\PassOptionsToPackage{dvipsnames,svgnames,table}{xcolor}
\documentclass[10pt,journal,compsoc]{IEEEtran}

\ifCLASSOPTIONcompsoc
  \usepackage[nocompress]{cite}
\else
  \usepackage{cite}
\fi

\ifCLASSINFOpdf
\else
\fi

\usepackage[utf8]{inputenc} 
\usepackage[T1]{fontenc}    
\usepackage{hyperref}       
\usepackage{url}            
\usepackage{booktabs}       
\usepackage{amsfonts}       
\usepackage{nicefrac}       
\usepackage{microtype}      
\usepackage{xcolor}         
\usepackage{graphicx}
\usepackage{subfigure}
\usepackage{amsmath,amssymb} 
\usepackage{caption} 

\usepackage{wrapfig}
\usepackage[american]{babel}

\usepackage{graphicx} 
\usepackage{algorithm}
\usepackage{algorithmic}
\usepackage{times}
\usepackage{soul}
\usepackage{url}
\usepackage[utf8]{inputenc}
\usepackage{amsmath}
\usepackage{amsthm}
\usepackage{booktabs}
\usepackage{algorithm}
\usepackage{algorithmic}
\usepackage{subfigure}
\usepackage{amsmath}
\usepackage{amssymb}
\usepackage{booktabs}
\usepackage{multirow}
\usepackage{paralist,algorithmic,algorithm}
\usepackage[american]{babel}
\usepackage{microtype}
\usepackage{lipsum}
\usepackage{bm}
\usepackage{overpic}
\usepackage[switch]{lineno}  %
 
\newcommand{\x}{{\bf x}}
\newcommand{\w}{{\bm w}}
\newcommand{\p}{{\bm p}}

\newcommand{\D}{\mathcal{D}}
\newcommand{\Y}{\mathcal{Y}}

\newcommand{\R}{\mathbb{R}}

\newcommand{\eg}{\emph{e.g.}}
\newcommand{\ie}{\emph{i.e.}}

\newtheorem{definition}{Definition}
\newcommand{\bfname}[1]{{\bf #1}}

\usepackage{threeparttable}
\usepackage{amssymb}
\usepackage{pifont}
\definecolor{DarkGreen}{RGB}{1,50,32}
\definecolor{dr}{RGB}{47,85,151}
\definecolor{dreg}{RGB}{112,48,160}
\definecolor{dn}{RGB}{255, 117, 143}
\definecolor{pr}{RGB}{228,197,111}
\definecolor{kd}{RGB}{251,133,0}
\definecolor{mr}{RGB}{84,130,53}
\definecolor{tbs}{RGB}{2,48,71}

\newcommand{\citep}[1]{\cite{#1}}

\usepackage{csquotes}
\begin{document}
%
\title{Class-Incremental Learning: A Survey}
%
%
%
%

\author{Da-Wei Zhou, Qi-Wei Wang, Zhi-Hong Qi, Han-Jia Ye, De-Chuan Zhan, Ziwei Liu
\thanks{
	This work is partially supported by National Science and Technology Major Project (2022ZD0114805), Fundamental Research Funds for the Central Universities (2024300373),
	NSFC (62376118, 62006112, 62250069, 61921006), Collaborative Innovation Center of Novel Software
	Technology and Industrialization, China Scholarship Council,
	Ministry of Education, Singapore, under its MOE AcRF Tier 2 (MOET2EP20221- 0012), NTU NAP, and under the RIE2020 Industry Alignment Fund – Industry Collaboration Projects (IAF-ICP) Funding Initiative. 	(Corresponding authors: H.-J. Ye and Z. Liu.)
	
	D.-W. Zhou, Q.-W. Wang, Z.-H. Qi, H.-J. Ye,  and D.-C. Zhan are with 
	School of Artificial Intelligence,  Nanjing University, and National Key Laboratory for Novel Software Technology, Nanjing University,  Nanjing, 210023, China;
	E-mail: \{zhoudw, wangqiwei, qizh, yehj, zhandc\}@lamda.nju.edu.cn
	
	Work done when D.-W. Zhou was a visiting scholar at NTU.
	
	Z. Liu is with S-Lab, College of Computing and Data Science, Nanyang Technological University, Singapore, 639798. E-mail: ziwei.liu@ntu.edu.sg
	}
}

\IEEEtitleabstractindextext{%
\begin{abstract}
	
	Deep models, \eg, CNNs and Vision Transformers, have achieved impressive achievements in many vision tasks in the closed world. However, novel classes emerge from time to time in our ever-changing world, requiring a learning system to acquire new knowledge continually. 
	Class-Incremental Learning (CIL) enables the learner to incorporate the knowledge of new classes incrementally and build a universal classifier among all seen classes. Correspondingly, when directly training the model with new class instances, a fatal problem occurs --- the model tends to {\em catastrophically forget} the characteristics of former ones, and its performance drastically degrades. There have been numerous efforts to tackle catastrophic forgetting in the machine learning community. 
	In this paper, we survey comprehensively recent advances in class-incremental learning and summarize these methods from several aspects. We also provide a rigorous and unified evaluation of 17 methods in benchmark image classification tasks to find out the characteristics of different algorithms empirically. Furthermore, we notice that the current comparison protocol ignores the influence of memory budget in model storage, which may result in unfair comparison and biased results. Hence, we advocate fair comparison by aligning the memory budget in evaluation, as well as several memory-agnostic performance measures. 
	The source code is available at \url{https://github.com/zhoudw-zdw/CIL_Survey/}.

\end{abstract}
}

\maketitle

\IEEEdisplaynontitleabstractindextext

%
\IEEEpeerreviewmaketitle

\section{Introduction} \label{sec:intro}
\IEEEPARstart{R}{ecent} years have witnessed the rapid progress of deep learning, where neural networks have achieved or even surpassed human-level performances in many fields~\cite{silver2016mastering,jumper2021highly,feng2023learning}. The typical training process of a deep network requires pre-collected datasets in advance, \eg, large-scale images or texts --- the network undergoes the training process of the pre-collected dataset multiple epochs. However, training data is often with stream format in the open world~\cite{gomes2017survey}. These streaming data cannot be held for long due to storage constraints~\cite{krempl2014open} or privacy issues~\cite{de2021survey}, requiring the model to be updated incrementally with only new class instances. Such requirements trigger the prosperity of the Class-Incremental Learning (CIL) field, aiming to continually build a holistic classifier among all seen classes. The fatal problem in CIL is called {\it catastrophic forgetting}, \ie, directly optimizing the network with new classes will erase the knowledge of former ones and result in irreversible performance degradation. Hence, how to effectively resist catastrophic forgetting becomes the core problem in building CIL models.

\begin{figure}[t]
	\vspace{-4mm}
	\begin{center}
		\includegraphics[width=1\columnwidth]{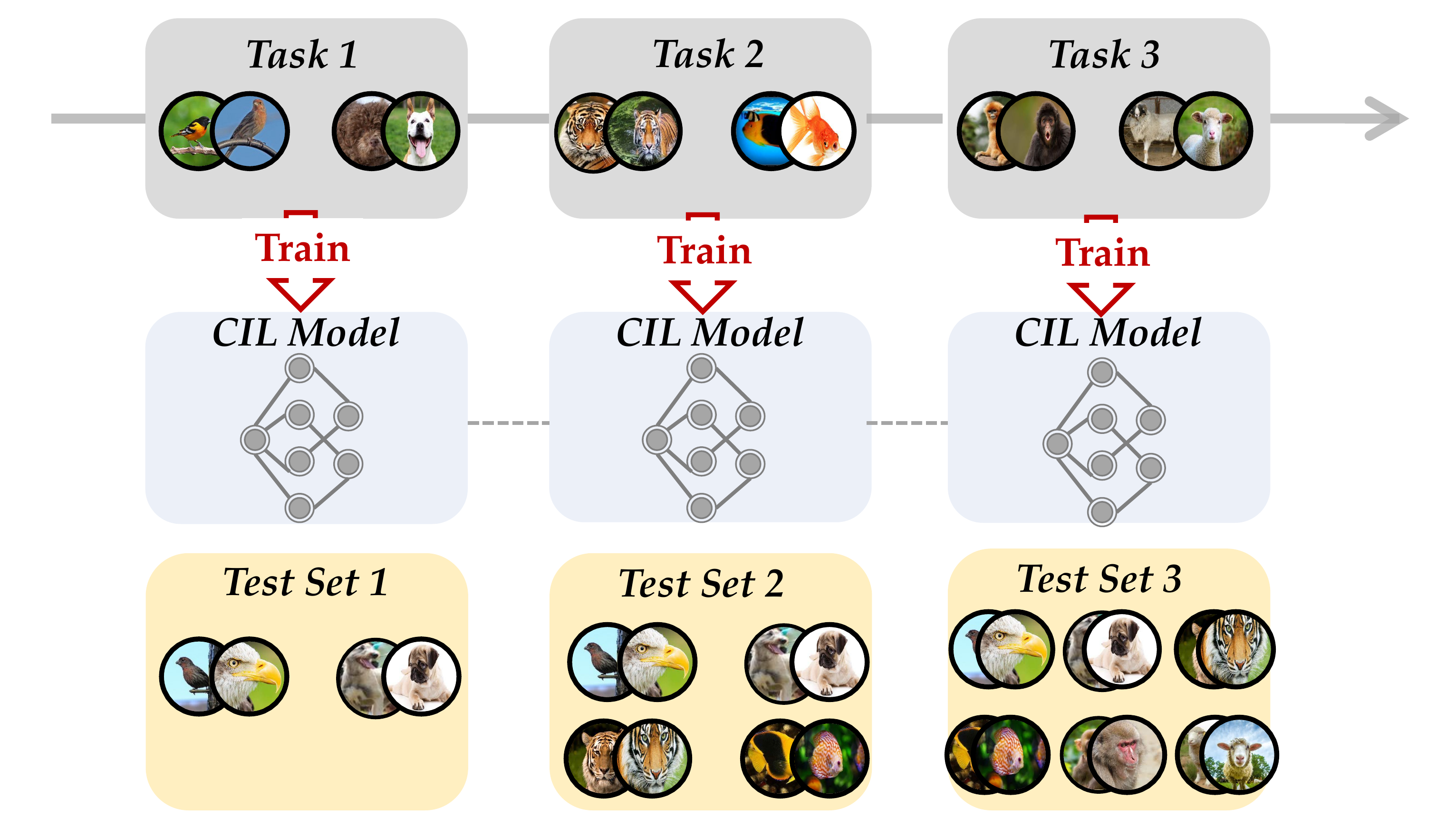}
	\end{center}
	\vspace{-5mm}
	\caption{ The setting of CIL. 
		Non-overlapping classes arrive sequentially, and the model needs to learn to classify all the classes incrementally. After learning each task, the model is evaluated among all seen classes. An ideal model should perform well in the newly learned classes and remember the former without forgetting.
	} \label{figure:intro}
	\vspace{-5mm}
\end{figure}

Figure~\ref{figure:intro} depicts the typical setting of CIL. Training data emerge sequentially in the stream format. In each timestamp, we can get a new training dataset (denoted as `task' in the figure) and need to update the model with the new classes. For example, the model learns `birds' and `dogs' in the first task, `tigers' and `fish' in the second task, `monkeys' and `sheep' in the third task, etc.
Afterward, the model is tested among all seen classes to evaluate whether it has discrimination for them. A good model should strike a balance between depicting the characteristics of new classes and preserving the pattern of formerly learned old classes. This trade-off is also known as the `stability-plasticity dilemma' in neural system~\cite{grossberg2012studies}, where stability denotes the ability to maintain former knowledge and plasticity represents the ability to adapt to new patterns.

\begin{figure*}[t]
	\vspace{-3mm}
	\begin{center}
		\includegraphics[width=1.9\columnwidth]{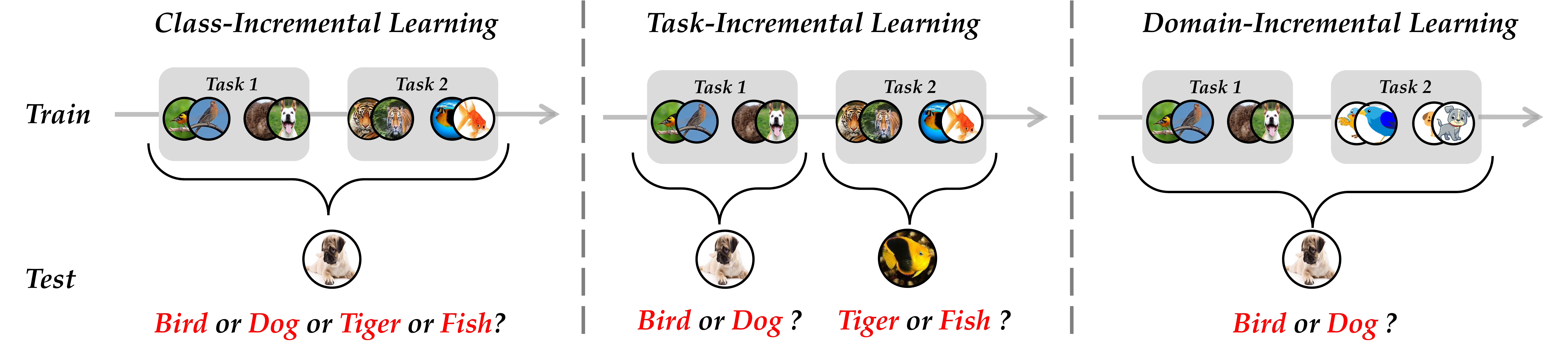}
	\end{center}
	\vspace{-4mm}
	\caption{ The setting of {\em Class-Incremental Learning} (CIL), {\em Task-Incremental Learning} (TIL), and {\em Domain-Incremental Learning} (DIL).
	CIL and TIL share the same training protocol, while TIL is much easier during inference, \ie, only requiring classifying among corresponding label spaces. DIL refers to the data stream with distribution change, where new tasks contain the same classes from different domains, \eg, cartoon and clip-art. 
	The distinction of these scenarios is proposed by~\cite{van2022three}.
	} \label{figure:class_task_domain}
	\vspace{-5mm}
\end{figure*}

Since the data stream comes continually and requires training a lifelong time, incremental learning is also known as `{\em continual learning}'~\cite{de2021survey} or `{\em lifelong learning}'~\cite{chen2018lifelong}. We interchangeably use these concepts in this paper.
Apart from class-incremental learning, there are other fine-grained settings addressing the incremental learning problem, \eg, Task-Incremental Learning (TIL) and Domain-Incremental Learning (DIL)~\cite{van2022three}. We show these three protocols in Figure~\ref{figure:class_task_domain}. TIL is a similar setting to CIL, and both of them observe incoming new classes in new tasks. However, the difference lies in the inference stage, where CIL requires the model to differentiate among all classes. By contrast, TIL only requires classifying the instance among the corresponding task space.
In other words, it does not require {\em cross-task} discrimination ability. Hence, TIL is {\em easier} than CIL, which can be seen as a particular case of CIL. On the other hand, DIL concentrates on the scenario with concept drift or distribution change~\cite{lu2018learning}, where new tasks contain instances from different domains but with the same label space. In this case, new domains correspond to the images in clip-art format. In this paper, we concentrate on the CIL setting, which is a more challenging scenario in the open world. 

There is also research about CIL before the prosperity of deep learning~\cite{zhou2002ensembling}. Typical methods try to solve the catastrophic forgetting problem with traditional machine learning models. However, most of them address the incremental learning within two tasks, \ie, the model is only updated with a single new stage~\cite{kuzborskij2013n,DaYZ14}. 
Furthermore, the rapid development of data collection and processing requires the model to grasp long-term and massive-scale data streams that traditional machine learning models cannot handle.
Correspondingly, deep neural networks with powerful representation ability well suit these requirements. As a result, deep learning-based CIL is becoming a hot topic in the machine learning and computer vision community. 

There are several related surveys discussing the incremental learning problem. For example, \cite{de2021survey} focuses on the task-incremental learning problem and provides a comprehensive survey. \cite{masana2022class} is a related survey on the class-incremental learning field, while it only discusses and evaluates the methods till 2020. However, with the rapid development of the CIL field, many great works are emerging day by day, which substantially boost the performance of benchmark settings~\cite{yan2021dynamically,wang2022foster,zhou2022model,douillard2022dytox}. On the other hand, with the prosperity of Vision Transformer (ViT)~\cite{dosovitskiy2020image} and pre-trained models, a heated discussion about ViT in CIL is attracting the attention of the community. 
Other surveys either focus on the specific field~\cite{mai2022online,biesialska2020continual} or lack the performance evolution among state-of-the-arts~\cite{van2022three,parisi2019continual,belouadah2021comprehensive}. Hence, it is urgent to provide an up-to-date survey containing popular methods to speed up the development of the CIL field.

In this paper, we aim for a comprehensive review of class-incremental learning methods and divide them into seven categories. We also provide a holistic comparison among different kinds of methods over benchmark datasets, \ie, CIFAR100~\cite{krizhevsky2009learning} and ImageNet100/1000~\cite{deng2009imagenet}. On the other hand, we highlight an important factor in CIL model evaluation, \ie, {\em memory budget}, and advocate fair comparison among different methods with an aligned budget. Correspondingly, we holistically evaluate the extensibility of CIL models with budget-agnostic measures.

In general, the contribution of this survey can be summarized as follows:
{\bf 1)}: We provide a comprehensive survey of CIL, including problem definitions, benchmark datasets, and different families of CIL methods. We organize these algorithms taxonomically (Table~\ref{table:taxonomy}) and chronologically (Figure~\ref{figure:timeline}) to give a holistic overview of state-of-the-art.
{\bf 2)}: We provide a rigorous and unified comparison among different methods on several publicly available datasets, including traditional CNN-backed and modern ViT-backed methods.	We also discuss the insights and summarize the common rules to inspire future research.
{\bf 3)}: To boost real-world applications, CIL models should be deployed not only on high-performance computers but also on edge devices. Therefore, we advocate evaluating different methods holistically by emphasizing the effect of memory budgets. Correspondingly, we provide a comprehensive evaluation of different methods given specific budgets as well as several new performance measures.

\section{Preliminaries} \label{sec:prelim}

\subsection{Problem Formulation}

{
\definition[] \label{def:cil} {\bf Class-Incremental Learning}
 aims to learn from an evolutive stream with new classes. Assume there is a sequence of $B$ training tasks\footnote{Also known as `session,' `phase,' and `stage.'} $\left\{\D^{1}, \D^{2}, \cdots, \D^{B}\right\}$ without overlapping classes, where $\D^{b}=\left\{\left(\x_{i}^{b}, y_{i}^{b}\right)\right\}_{i=1}^{n_b}$ is the $b$-th incremental step with $n_b$ training instances. $\x_i^b \in \R^D$ is an instance of class $y_i^b \in Y_b$, $Y_b$ is the label space of task $b$, where
 $Y_b  \cap Y_{b^\prime} = \varnothing$ for $b\neq b^\prime$. 
We can only access data from $\D^b$ when training task $b$. 
The ultimate goal of CIL is to continually build a classification model for all classes. In other words, the model should not only acquire the knowledge from the current task $\D^b$ but also preserve the knowledge from former tasks. 
After each task, the trained model is evaluated over all seen classes $\mathcal{Y}_b=Y_1 \cup \cdots Y_b$. Formally, CIL aims to fit a model $f(\x): X\rightarrow\mathcal{Y}_b$, which minimizes the expected risk:
\begin{equation} \label{eq:totalrisk}
	f^*=\underset{f \in \mathcal{H}}{\operatorname{argmin}} \enspace \mathbb{E}_{(\mathbf{x}, y) \sim \mathcal{D}_{t}^1\cup\cdots\mathcal{D}_{t}^b} \mathbb{I}(y \neq f(\mathbf{x})) \,,
\end{equation}
where $\mathcal{H}$ is the hypothesis space, $\mathbb{I}(\cdot)$ is the indicator function which outputs $1$ if the expression holds and $0$ otherwise. $\mathcal{D}_{t}^b$ denotes the data distribution of task $b$. A good CIL model satisfying Eq.~\ref{eq:totalrisk} has discriminability among all classes, which not only works well on new classes but also preserves the knowledge of former ones.
}

\noindent\textbf{Class Overlapping}: Typical CIL setting assumes $Y_b  \cap Y_{b^\prime} = \varnothing$ for $b\neq b^\prime$, \ie, there are no overlapping classes in different tasks. However, in the real world, it is common to observe the old classes emerging in new tasks. When $Y_b  \cap Y_{b^\prime} \neq \varnothing$, the setting is called blurry class-incremental learning (Blurry CIL)~\cite{bang2021rainbow}. It enables the model to revisit former instances in the later stage, which weakens the learning difficulty.

\noindent\textbf{Number of Instances}: Typical CIL setting assumes the training data of each class is balanced and many-shot (\eg, hundreds or thousands). However, the data collection may face challenges in the real world, \eg, we can only collect a limited number of training instances for rare birds. 
When the training instances of new classes are limited (\eg, 5-shot per class), the setting is called Few-Shot Class-Incremental Learning (FSCIL)~\cite{tao2020few}. When the training instances are highly imbalanced and long-tailed, the setting is called Long-Tailed Class-Incremental Learning (LTCIL)~\cite{liu2022long}. 

\noindent\textbf{Online CIL}: Although data comes with stream format, the model can conduct multi-epoch training with each task, \ie, offline training within each task. There are some works addressing fully online (one-pass) CIL, where each batch can be processed once and then dropped~\cite{mai2021supervised}. It is a specific case of the current setting, and we concentrate on the generalized CIL setting in this paper.

In the following discussions, we decompose the CIL model into the embedding module and linear layers, \ie, $f(\x)=W^{\top}\phi(\x)$,\footnote{We omit the bias term for ease of discussions.} where $\phi(\cdot):\mathbb{R}^{D} \rightarrow \mathbb{R}^{d}$, $W\in\mathbb{R}^{d\times |\mathcal{Y}_{b}|}$. The linear layer can be further decomposed into the combination of classifiers: $W=[\w_1,\w_2,\cdots,\w_{|\mathcal{Y}_{b}|}]$, where $\w_k\in\R^d$, $|\cdot|$ denotes the size of the set. The logits are then passed to the Softmax activation for further optimization, \ie, the output probability on class $k$ is denoted as:
\begin{align}\label{eq:softmax}
	\mathcal{S}_k(f(\x))=\frac{\exp^{\w_k^\top\phi(\x)/\tau}}
	{\sum_{j=1}^{|\mathcal{Y}_{b}|} \exp^{\w_j^\top\phi(\x)/\tau}} \,,
\end{align}
where $\tau$ is the temperature parameter.

\noindent\textbf{Backbones:} As defined in Eq.~\ref{eq:softmax}, the predictions are derived by feeding instance $\x$ into the embedding function and linear layer. The embedding function is designed to project input instances into the embedding space to reflect its semantic information. Hence, ideal CIL algorithms should work for any type of backbones, \eg, multi-layer perceptron (MLP)~\cite{kruse2022multi}, convolutional neural network (CNN)~\cite{lecun2010convolutional}, and Vision Transformer (ViT)~\cite{dosovitskiy2020image}. Specifically, we often treat the whole image as the input of MLP and CNN and utilize the final product as the embedding. By contrast, ViT transforms the image into a sequence of patches for patch features. These patches are then fed forward to the self-attention modules~\cite{vaswani2017attention} and MLP layer to get the contextualized information. Typical ViT appends an extra token  (\ie, $\texttt{[CLS]}$ token) to the set of patches and utilizes the final representation of $\texttt{[CLS]}$ token as the embedding.  Since ViT relies on the self-attention mechanism to relate and adjust patch-wise features, it is intuitive to influence the embedding context by adding task-specific tokens as the input. The basic difference between these backbones triggers different tuning algorithms in CIL, which will be further discussed in Section~\ref{sec:dynamic-networks}.

\noindent\textbf{Baseline in CIL}: In CIL, the typical baseline is sequential finetuning the model with the current dataset $\D^b$ (denoted as `Finetune'), whose loss function can be denoted as:
\begin{equation}\label{eq:finetune} \textstyle
	\mathcal{L}= \sum_{(\x,y)\in\mathcal{D}^b} \ell(f(\x),y) \,,
\end{equation}
where $\ell(\cdot,\cdot)$ measures the discrepancy between inputs, \eg, cross-entropy loss.
Finetune is known as the baseline method for CIL since it only concentrates on learning the new concepts in the current task. Consequently, it suffers severe forgetting since the model pays no attention to former ones.

\begin{table*}[t]
	\caption{ The taxonomy of CIL. The shade color in the last column denotes the subcategory, which is consistent with Figure~\ref{figure:timeline}.}
	\label{table:taxonomy}
	\vspace{-3mm}
	\centering
	\resizebox{0.88\textwidth}{!}{
		\begin{tabular}{@{}c|l|c|p{8cm}<{\centering}@{}}
			\toprule
			\multicolumn{2}{c|}{Algorithm Category} & Reference \\ 
			\midrule
			\midrule
			\multirow{3}{*}{\begin{tabular}[c]{@{}c@{}}\textcolor{red}{$\S~$}\ref{sec:data-replay}\\ Data Replay \end{tabular}} &
			\begin{tabular}[c]{@{}c@{}} {Direct Replay}\end{tabular}
			& 
			\cellcolor{dr!50}
			\cite{chaudhry2018riemannian,bang2021rainbow,aljundi2019gradient,isele2018selective,chaudhry2020using,de2021continual,iscen2020memory,zhao2021memory,liu2020mnemonics} \\ 
			\cmidrule(l){2-3} 
			\multicolumn{1}{c|}{}  &
			\begin{tabular}[c]{@{}c@{}} {Generative Replay}\end{tabular}
			& \cellcolor{dr!50}
			\cite{shin2017continual,he2018exemplar,hu2019overcoming,kemker2018fearnet,ostapenko2019learning,xiang2019incremental,wang2021ordisco,jiang2021ib,zhu2021prototype,petit2022fetril,jodelet2023class,gao2023ddgr} \\ 
\midrule
			\multicolumn{2}{c|}{\begin{tabular}[c]{@{}c@{}} {\textcolor{red}{$\S~$}\ref{sec:data-regularization}:~Data Regularization}\end{tabular}}   
			& \cellcolor{dreg!50} \cite{lopez2017gradient,chaudhry2018efficient,aljundi2019gradient,wang2021training,zeng2019continual,tang2021layerwise,riemer2018learning} \\ 
			\midrule
	\multirow{4}{*}{\begin{tabular}[c]{@{}c@{}}\textcolor{red}{$\S~$}\ref{sec:dynamic-networks}\\ Dynamic Networks\end{tabular}} &
			\begin{tabular}[c]{@{}c@{}} {Neuron Expansion}\end{tabular}
			& \cellcolor{dn!50}
			\cite{yoon2018lifelong,xu2018reinforced,li2019learn,ostapenko2019learning} \\ 
			\cmidrule(l){2-3} 
			\multicolumn{1}{c|}{}  &
			\begin{tabular}[c]{@{}c@{}} {Backbone Expansion}\end{tabular}
			& \cellcolor{dn!50} \cite{rusu2016progressive,aljundi2017expert,schwarz2018progress,zhao2021mgsvf,liu2021adaptive,pham2021dualnet,yan2021dynamically,wang2022foster,zhou2022model,wang2022beef} \\ 
			\cmidrule(l){2-3} 
			\multicolumn{1}{c|}{}  &
			\begin{tabular}[c]{@{}c@{}} {Prompt Expansion}\end{tabular}
			& \cellcolor{dn!50}
			\cite{douillard2022dytox,wang2022learning,wang2022dualprompt,seale2022coda,wang2022s} \\
\midrule
			\multicolumn{2}{c|}{\begin{tabular}[c]{@{}c@{}} {\textcolor{red}{$\S~$}\ref{sec:parameter-regularization}:~Parameter Regularization}\end{tabular}}         
			& \cellcolor{pr!50} \cite{kirkpatrick2017overcoming,zenke2017continual,chaudhry2018riemannian,aljundi2018memory,aljundi2019task,lee2017overcoming,yang2019,yang2021cost,lee2020continual} \\ 
			\midrule
			\multirow{4}{*}{\begin{tabular}[c]{@{}c@{}}\textcolor{red}{$\S~$}\ref{sec:knowledge-distillation}\\ Knowledge Distillation\end{tabular}} &
			\begin{tabular}[c]{@{}c@{}} {Logit Distillation}\end{tabular}
			& \cellcolor{kd!50}
			\cite{li2016learning,rebuffi2017icarl,wu2019large,hou2018lifelong,lee2019overcoming,zhang2020class,smith2021always,zhou2021co} \\ 
			\cmidrule(l){2-3} 
			\multicolumn{1}{c|}{}  &
			\begin{tabular}[c]{@{}c@{}} {Feature Distillation}\end{tabular}
			& \cellcolor{kd!50} \cite{hou2019learning,lu2022augmented,park2021class,dhar2019learning,kang2022class,douillard2020podnet,hu2021distilling,simon2021learning,jung2018less,li2019rilod} \\ 
			\cmidrule(l){2-3} 
			\multicolumn{1}{c|}{}  &
			\begin{tabular}[c]{@{}c@{}} {Relational Distillation}\end{tabular}
			& \cellcolor{kd!50} \cite{gao2022rdfcil,dong2021few,tao2020topology,tao2020few,liu2022model,asadi2023prototype} \\
			\midrule
			\multirow{4}{*}{\begin{tabular}[c]{@{}c@{}}\textcolor{red}{$\S~$}\ref{sec:model_rectify}\\ Model Rectify\end{tabular}} &
			\begin{tabular}[c]{@{}c@{}} {Feature Rectify}\end{tabular}
			& \cellcolor{mr!50}
			\cite{zhou2022forward,yu2020semantic,shi2022mimicking,jie2022alleviating,singh2020calibrating,singh2021rectification} \\ 
			\cmidrule(l){2-3} 
			\multicolumn{1}{c|}{}  &
			\begin{tabular}[c]{@{}c@{}} {Logit Rectify}\end{tabular} &
			\cellcolor{mr!50} \cite{hou2019learning,castro2018end,wu2019large,belouadah2019il2m} \\ 
			\cmidrule(l){2-3} 
			\multicolumn{1}{c|}{}  &
			\begin{tabular}[c]{@{}c@{}} {Weight Rectify}\end{tabular} &
			\cellcolor{mr!50} \cite{zhao2020maintaining,ahn2021ss,pernici2021class} \\
			\midrule
				\multicolumn{2}{c|}{\begin{tabular}[c]{@{}c@{}} {\textcolor{red}{$\S~$}\ref{sec:template-based}:~Template-Based Classification}\end{tabular}}         
			& \cellcolor{tbs!50} \cite{rebuffi2017icarl,de2021continual,zhang2021few,zhou2022forward,zhou2023revisiting,mcdonnell2023ranpac,goswami2023fecam,wang2023few,shi2023prototype,zhou2022few,wei2021incremental,van2021class} \\ 
			\bottomrule
		\end{tabular}
	}
	\vspace{-4mm}
\end{table*}

\subsection{Exemplars and Exemplar Set}

As defined in Definition~\ref{def:cil}, in each incremental task, the model can only access the current dataset $\D^{b}$. This helps to preserve user privacy and release the storage burden. However, this restriction is relaxed in some cases, and the model can keep a relatively small set, namely {\em exemplar set}, to reserve the representative instances from former tasks.

{
\definition[] {\bf Exemplar Set}\footnote{Also known as the `replay buffer' and `memory buffer.' } \label{def:exemplar_set}
is an extra collection of instances from former tasks $\mathcal{E}=\left\{\left(\x_{j}, y_{j}\right)\right\}_{j=1}^{M}$, $y_j \in \mathcal{Y}_{b-1}$.
 With the help of the exemplar set, the model can utilize $\mathcal{E}\cup\D^b$ for the update within each task. The model manages the exemplar set after the training process of each task.
}

\noindent\textbf{Exemplar Set Management}: Since the data stream is evolving, there are two main strategies to manage the exemplar set in CIL~\cite{hou2019learning}. The first way is to keep a fixed number of exemplars per class, \eg, $R$ per class. Under such circumstances, the size of the exemplar set will grow as the data stream evolves --- the model keeps $R|\mathcal{Y}_b|$ after the $b$-th task. This will result in a  linearly {\em growing} memory budget, which is inapplicable in real-world learning systems. To this end, another strategy advocates saving a fixed number of exemplars, \eg, $M$. The model keeps $[\frac{M}{|\mathcal{Y}_b|}]$ instances per class, where $[\cdot]$ denotes floor function. It helps to keep a fixed size of exemplars in the memory and release the storage burden.
 In this paper, we use the second strategy to organize the exemplar set.

\noindent\textbf{Exemplar Selection}: Exemplars are representative instances of each class, which need to be selected from the entire training set. An intuitive way to choose the exemplars is random sampling, which results in diverse instances. By contrast, a commonly used strategy is called herding~\cite{welling2009herding,rebuffi2017icarl}, aiming to select the most {\em representative} ones of each class. 
Given the instance set $X=\{\x_1,\x_2,\cdots,\x_n\}$ from class $y$, herding first calculates the class center with current embedding $\phi(\cdot)$: ${\bm \mu}_y \leftarrow \frac{1}{n} \sum_{i=1}^n \phi(\x_i)$.
Afterward, it iteratively appends instances into the exemplar set:
\begin{equation} \textstyle \label{eq:herding}
{\bm p}_k \leftarrow \underset{\x \in X}{\operatorname{argmin}}\left\|{\bm \mu}_y-\frac{1}{k}\left[\phi(\x)+\sum_{j=1}^{k-1} \phi\left({\bm p}_j\right)\right]\right\| \,,
\end{equation}
until $k$ reaches the memory bound of each class ($[\frac{M}{|\mathcal{Y}_b|}]$). The exemplar set of class $y$ is the concatenation of $\{{\bm p}_1, {\bm p}_2, \cdots {\bm p}_{[\frac{M}{|\mathcal{Y}_b|}]}\}$. Eq.~\ref{eq:herding} ensures that the average feature vector over all exemplars selected thus far is closest to the class mean.
Since the class center can be seen as the most representative pattern of each class, selecting exemplars near the center also enhances the representativeness. Herding is now a commonly used strategy to select exemplars in CIL, and we also adopt it in this paper. 

\section{Class-Incremental Learning: Taxonomy} \label{sec:taxonomy}

\begin{figure*}[t]
	\vspace{-3mm}
		\begin{overpic}[width=2\columnwidth]{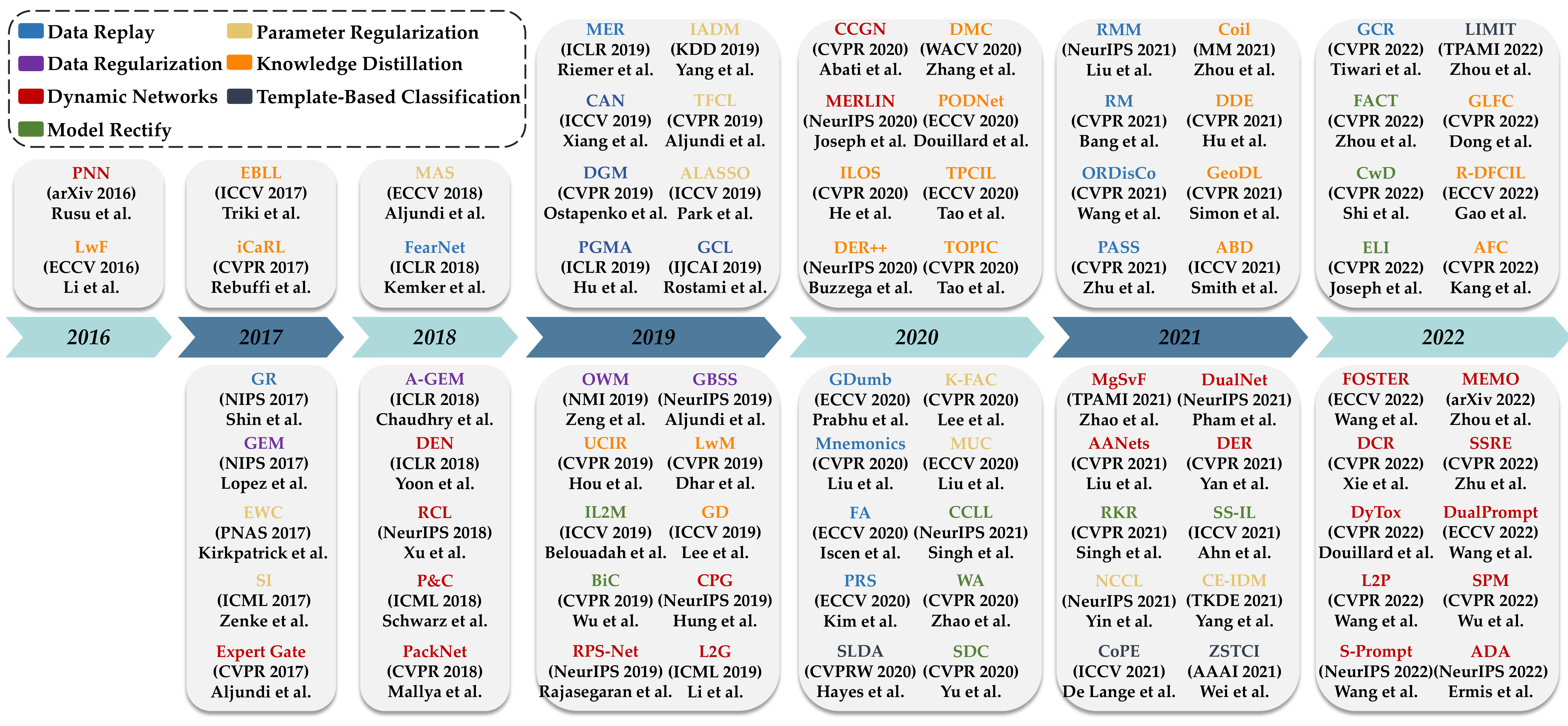}
			\put(7,35.0){\tiny {\cite{rusu2016progressive}}}
			\put(7,30.2){\tiny {\cite{li2016learning}}}
			
			\put(18.2,35.0){\tiny {\cite{rannen2017encoder}}}
			\put(18.2,30.2){\tiny {\cite{rebuffi2017icarl}}}
			\put(18.2,21.8){\tiny {\cite{shin2017continual}}}
			\put(18.2,17.8){\tiny {\cite{lopez2017gradient}}}
			\put(18.2,13.2){\tiny {\cite{kirkpatrick2017overcoming}}}
			\put(18.2,9.0){\tiny {\cite{zenke2017continual}}}
			\put(19.5,4.4){\tiny {\cite{aljundi2017expert}}}
			
			\put(29.2,35.0){\tiny {\cite{aljundi2018memory}}}
			\put(29.8,30.2){\tiny {\cite{kemker2018fearnet}}}
			\put(29.8,21.8){\tiny {\cite{chaudhry2018efficient}}}
			\put(29.2,17.8){\tiny {\cite{yoon2018lifelong}}}
			\put(29.2,13.4){\tiny {\cite{xu2018reinforced}}}
			\put(29.2,9.0){\tiny {\cite{schwarz2018progress}}}
			\put(30.0,4.4){\tiny {\cite{mallya2018packnet}}}
			
			\put(40.3,44.2){\tiny {\cite{riemer2018learning}}}
			\put(40.3,39.5){\tiny {\cite{xiang2019incremental}}}
			\put(40.3,35.0){\tiny {\cite{ostapenko2019learning}}}
			\put(40.3,30.2){\tiny {\cite{hu2019overcoming}}}
			\put(40.3,21.8){\tiny {\cite{zeng2019continual}}}
			\put(40.3,17.8){\tiny {\cite{hou2019learning}}}
			\put(40.3,13.2){\tiny {\cite{belouadah2019il2m}}}
			\put(40.3,9.0){\tiny {\cite{wu2019large}}}
			\put(40.8,4.4){\tiny {\cite{rajasegaran2019random}}}

			\put(47.1,44.2){\tiny {\cite{yang2019}}}
			\put(47.1,39.5){\tiny {\cite{aljundi2019task}}}
			\put(47.5,35.0){\tiny {\cite{park2019continual}}}
			\put(47.1,30.2){\tiny {\cite{rostami2019complementary}}}
			\put(47.1,21.8){\tiny {\cite{aljundi2019gradient}}}
			\put(47.1,17.8){\tiny {\cite{dhar2019learning}}}
			\put(47.1,13.2){\tiny {\cite{lee2019overcoming}}}
			\put(47.1,9.0){\tiny {\cite{hung2019compacting}}}
			\put(47.1,4.4){\tiny {\cite{li2019learn}}}
			
			\put(56.8,44.2){\tiny {\cite{abati2020conditional}}}
			\put(57.2,39.5){\tiny {\cite{joseph2020incremental}}}
			\put(56.8,35.0){\tiny {\cite{he2020incremental}}}
			\put(56.8,30.2){\tiny {\cite{buzzega2020dark}}}
			\put(56.6,21.8){\tiny {\cite{prabhu2020gdumb}}}
			\put(57.7,17.8){\tiny {\cite{liu2020mnemonics}}}%
			\put(56.6,13.2){\tiny {\cite{iscen2020memory}}}
			\put(56.6,9.0){\tiny {\cite{kim2020imbalanced}}}
			\put(56.6,4.4){\tiny {\cite{hayes2020lifelong}}}
			
			\put(63.8,44.2){\tiny {\cite{zhang2020class}}}
			\put(64.2,39.5){\tiny {\cite{douillard2020podnet}}}
			\put(63.8,35.0){\tiny {\cite{tao2020topology}}}
			\put(63.8,30.2){\tiny {\cite{tao2020few}}}
			\put(63.8,21.8){\tiny {\cite{lee2020continual}}}
			\put(63.6,17.8){\tiny {\cite{liu2020more}}}
			\put(63.6,13.2){\tiny {\cite{singh2020calibrating}}}
			\put(63.6,9.0){\tiny {\cite{zhao2020maintaining}}}
			\put(63.6,4.4){\tiny {\cite{yu2020semantic}}}
			
			\put(73.3,44.2){\tiny {\cite{liu2021rmm}}}
			\put(73.3,39.5){\tiny {\cite{bang2021rainbow}}}
			\put(73.9,35.0){\tiny {\cite{wang2021ordisco}}}
			\put(73.3,30.2){\tiny {\cite{zhu2021prototype}}}
			\put(73.3,21.8){\tiny {\cite{zhao2021mgsvf}}}
			\put(73.5,17.8){\tiny {\cite{liu2021adaptive}}}
			\put(73.3,13.2){\tiny {\cite{singh2021rectification}}}
			\put(73.3,9.0){\tiny {\cite{yin2021mitigating}}}
			\put(73.3,4.4){\tiny {\cite{de2021continual}}}
			
			\put(80.5,44.2){\tiny {\cite{zhou2021co}}}
			\put(80.5,39.5){\tiny {\cite{hu2021distilling}}}
			\put(80.5,35.0){\tiny {\cite{simon2021learning}}}
			\put(80.5,30.2){\tiny {\cite{smith2021always}}}
			\put(80.8,21.8){\tiny {\cite{pham2021dualnet}}}
			\put(80.8,17.8){\tiny {\cite{yan2021dynamically}}}
			\put(80.8,13.2){\tiny {\cite{ahn2021ss}}}
			\put(81.0,9.0){\tiny {\cite{yang2021cost}}}
			\put(80.8,4.4){\tiny {\cite{wei2021incremental}}}
			
			\put(89.5,44.2){\tiny {\cite{tiwari2022gcr}}}
			\put(89.5,39.5){\tiny {\cite{zhou2022forward}}}
			\put(89.5,35.0){\tiny {\cite{shi2022mimicking}}}
			\put(89.5,30.2){\tiny {\cite{joseph2022energy}}}
			\put(90.1,21.8){\tiny {\cite{wang2022foster}}}
			\put(89.5,17.8){\tiny {\cite{xie2022general}}}
			\put(89.5,13.2){\tiny {\cite{douillard2022dytox}}}
			\put(89.5,9.0){\tiny {\cite{wang2022learning}}}
			\put(90.3,4.4){\tiny {\cite{wang2022s}}}
			
			\put(97.0,44.2){\tiny {\cite{zhou2022few}}}
			\put(97.0,39.5){\tiny {\cite{dong2022federated}}}
			\put(97.5,35.0){\tiny {\cite{gao2022rdfcil}}}
			\put(97.0,30.2){\tiny {\cite{kang2022class}}}
			\put(97.0,21.8){\tiny {\cite{zhou2022model}}}
			\put(97.0,17.8){\tiny {\cite{zhu2022self}}}
			\put(98.0,13.2){\tiny {\cite{wang2022dualprompt}}}
			\put(97.0,9.0){\tiny {\cite{wu2022class}}}
			\put(97.0,4.4){\tiny {\cite{ermismemory}}}
		\end{overpic}
		\vspace{-3mm}
		\caption{ The roadmap of class-incremental learning. We organize representative methods chronologically to show the concentration at different stages. Different colors of these methods denote the sub-categories in Table~\ref{table:taxonomy}. Knowledge distillation and data replay dominated the research before 2021, while model rectify and dynamic networks became popular after 2021.
		} \label{figure:timeline} 
			\vspace{-5mm}
	\end{figure*}
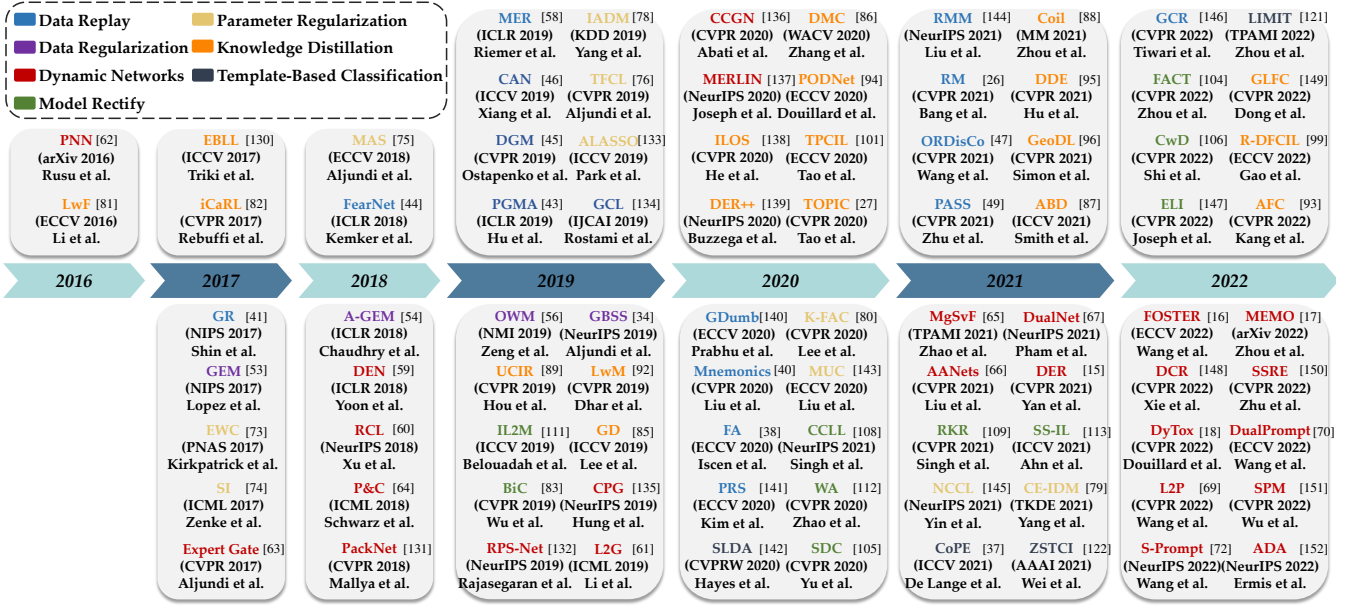

There are numerous works addressing CIL in recent years, raising a heated discussion among machine learning and computer vision society. 
We organize these methods taxonomically from {\em seven aspects}, as shown in Table~\ref{table:taxonomy}. 
Specifically, data replay and data regularization concentrate on solving CIL with exemplars, either by revisiting former instances or using them as indicators to regularize model updating. 
Dynamic networks expand the network structure for stronger representation ability, while parameter regularization-based methods regularize the model parameters to prevent them from drifting away to resist forgetting. 
Moreover, knowledge distillation builds the mapping between incremental models to resist forgetting, and model rectification aims to reduce the biased prediction of incremental learners. Template-based classification aims to transform inference into query-template matching.
 We list the representative methods chronologically in Figure~\ref{figure:timeline} to show the research focus of different periods. 
 
 Note that the classification rules of these categories are based on the `key point' (or the special focus) of each algorithm. With the rapid development of class-incremental learning, some techniques are becoming the well-acknowledged baseline to be shared among a variety of algorithms. Hence, these seven categories are not mutually exclusive, and there is no strict boundary between them. We organize them into several categories to enhance a holistic understanding of CIL from a specific perspective.
In the following sections, we will discuss CIL methods from these aspects.

\subsection{Data Replay}\label{sec:data-replay}

`Replay' is important in human cognition system~\cite{wilson1994reactivation,tambini2013persistence,barry2023neural} --- a student facing final exams shall go over the textbooks to recall former memory and knowledge. This phenomenon can also be extended to the network training process, where a network can overcome catastrophic forgetting by revisiting former exemplars. Correspondingly, an intuitive way~\cite{ratcliff1990connectionist,rebuffi2017icarl} is to save an extra exemplar set $\mathcal{E}$ (as defined in Definition~\ref{def:exemplar_set}) and include it into the model updating process:
\begin{equation}\label{eq:replay} \textstyle
	\mathcal{L}= \sum_{(\x,y)\in\left(\mathcal{D}^b\cup\mathcal{E}\right)} \ell(f(\x),y) \,.
\end{equation}
Comparing Eq.~\ref{eq:replay} to Eq.~\ref{eq:finetune}, we can find that exemplars are concatenated to the training set for updating, enabling the retrospective review of former knowledge when learning new concepts.

\noindent\textbf{Hot to construct the exemplar set?} Utilizing the exemplar set for rehearsal is simple yet effective, which leads to numerous following works. 
The exemplar sampling process is also similar to the active learning protocol, and some works propose corresponding sampling measures to select the {\em informative} ones. For example, \cite{chaudhry2018riemannian} suggests sampling exemplars with high prediction entropy and near the decision boundary. The model will achieve higher generalization ability by replaying these `hard' exemplars. \cite{bang2021rainbow} proposes estimating exemplars' uncertainty via data augmentation. They choose instances with large prediction diversity by aggregating the predictions of multiple augmented instances. Similarly, \cite{aljundi2019gradient} proposes to sample exemplars with a greedy strategy in online incremental learning.
It proves that exemplar selection is equivalent to maximizing the diversity of exemplars with parameters gradient as the feature. 
Without explicit task boundaries, \cite{isele2018selective} explores the reservoir sampling process to ensure the exemplars are i.i.d. sampled. 
\cite{chaudhry2020using} formulates the replay process into a bi-level optimization and keeps intact predictions on some anchor points of
past tasks. \cite{de2021continual} introduces data replay in prototypical network~\cite{snell2017prototypical}, and utilizes the exemplars as pseudo-prototypes for embedding evaluation.
Mnemonics~\cite{liu2020mnemonics} proposes a way to parameterize exemplars and optimize them in a meta-learning manner. The framework is trained through bi-level optimizations, \ie, model-level and exemplar-level, which can be combined with various CIL algorithms.

\noindent\textbf{Memory-efficient memory}: Since exemplars are raw images, directly saving a set of instances may consume enormous memory costs. To this end, several works are proposed to build a {\em memory-efficient} replay buffer~\cite{zhou2022model}. \cite{iscen2020memory} argues that extracted features have lower dimensions than raw images and proposes saving features in the exemplar set to release the burden. Similarly, \cite{zhao2021memory} proposes keeping low-fidelity images instead of raw ones. However, since the distributions of extracted features and low-fidelity images may differ from the raw images, an extra adaptation process is needed for these methods, adding to the algorithm's complexity.

\noindent\textbf{Generative replay}: The above-mentioned methods achieve competitive performance by replaying former instances in the memory. Apart from directly saving instances in the exemplar set, generative models show the potential to model the distribution and generate instances~\cite{goodfellow2014generative,kingma2013auto}, which have also been applied to class-incremental learning.
We denote the aforementioned methods directly saving instances for replay as `direct replay,' and the methods utilizing generative models as `generative replay.'

There often exist two models in generative replay-based CIL, \ie, the generative model for data generation and the classification model for prediction. GR~\cite{shin2017continual} firstly proposes to utilize the generative adversarial network (GAN)~\cite{goodfellow2014generative} in CIL. In each updating process, it utilizes the GAN to generate the instances from former classes and then updates GAN and classification model with both old and new classes. ESGR~\cite{he2018exemplar} extends GR by saving extra exemplars. It also proposes to train a separate GAN for each incremental task, which does not require updating GAN incrementally. \cite{hu2019overcoming} extends GR by introducing the dynamic parameter generator for model adaptation at test time. FearNet~\cite{kemker2018fearnet} explores brain-inspired CIL and 
 uses a dual-memory system in which new memories are consolidated from a network for recent memories. Recently, \cite{ostapenko2019learning,xiang2019incremental,wang2021ordisco} explore the application of conditional GAN~\cite{mirza2014conditional} in CIL, and~\cite{jiang2021ib} adopts variational auto-encoders (VAE)~\cite{kingma2013auto} to model data distribution. Similarly, \cite{zhu2021prototype,petit2022fetril} model each class into a Gaussian distribution and sample instances directly from the class center. With the prosperity of diffusion models~\cite{ho2020denoising}, recent works also consider using diffusion models as the data generator for high-quality samples. Correspondingly, 
 SDDR~\cite{jodelet2023class}  studies the use of a pre-trained diffusion model as a complementary source of data. Similarly, DDGR~\cite{gao2023ddgr} adopts a diffusion model as the generator and calculates an instruction operator through the classifier to instruct the generation of samples. There are also works~\cite{zajkac2023exploring,smith2023continual} on solving the catastrophic forgetting of these diffusion models.
 
  \noindent\textbf{Discussions}:
  Direct replay is a simple yet effective strategy, which has been widely applied to camera localization~\cite{korycki2021class}, semantic segmentation~\cite{maracani2021recall}, video classification~\cite{villa2022vclimb}, and action recognition~\cite{park2021class}. Since it directly optimizes the loss over old exemplars, it is found to help continual learners stay in the low-loss region of prior tasks during the optimization strategy~\cite{verwimp2021rehearsal}.  However, since the exemplar set only saves a tiny portion of the training set, data replay may suffer the overfitting problem and weaken the generalizability~\cite{lopez2017gradient,verwimp2021rehearsal}.
  Considering repeated optimization on a small pool
  of data inevitably leads to tight and unstable decision boundaries, several works tackle this problem by constraining the model's layer-wise Lipschitz constants with regard to exemplars~\cite{bonicelli2022effectiveness} or enlarging representational variations to alleviate representation collapse~\cite{yu2022continual}.
  Besides, the data-imbalance problem~\cite{zhang2021deep} also occurs due to the gap between the few-shot exemplars and the many-shot training set.
  Iteratively optimizing the imbalanced training set introduces extra bias in the classifier, and several works address this problem with balanced sampling~\cite{wang2022foster,castro2018end}. Finally, since direct replay requires saving exemplars of former classes, it will face privacy issues when the raw data contains face images or resource deficiency when the raw data contains images of rare animals~\cite{tao2020few}. Under such circumstances, algorithms should be designed to learn without exemplars~\cite{zhu2021prototype} or using privacy-friendly strategies like feature replay~\cite{petit2022fetril}.
  
  On the other hand, the performance of generative replay methods relies on the quality of generated data. They are found to work well on simple datasets~\cite{van2020brain,wang2021triple} while failing in complex, large-scale inputs~\cite{yoshihashi2019classification}.
  To tackle this problem, some works find generating features is much easier than generating raw images in terms of computational complexity and semantic information. Hence, they either utilize conditional GAN to generate features~\cite{liu2020generative} or VAE to model the internal representations~\cite{van2020brain}. 
  Additionally, recent advances in diffusion models reveal a promising way to generate instances with pre-trained diffusion models~\cite{jodelet2023class}, while utilizing pre-trained models leads to an unfair comparison to other methods without extra information.
  Furthermore, when sequentially updating the generative model, the catastrophic forgetting phenomena can also be observed on these generative models~\cite{zajkac2023exploring,smith2023continual}. As a result, algorithms should be designed to address catastrophic forgetting in two aspects (\ie, classifier aspect and generative model aspect) when using generative replay.

\subsection{Data Regularization}\label{sec:data-regularization}

Apart from directly replaying former data, another group of works tries to regularize the model with former data and control the optimization direction. Since learning new classes will result in the catastrophic forgetting of old ones, the intuitive idea is to ensure that optimizing the model for new classes will {\em not hurt} former ones. GEM~\cite{lopez2017gradient} aims to find the model that satisfies:
\begin{align} \label{eq:gem}
	&f^*= \underset{f \in \mathcal{H}}{\operatorname{argmin}} \sum_{(\x,y)\in\mathcal{D}^b}
	\ell\left(f(\x),y \right) \quad\\ \notag
	& \text { s.t. } \sum_{(\x_j,y_j)\in\mathcal{E}}
	\ell\left(f(\x_j),y_j \right) \leq \sum_{(\x_j,y_j)\in\mathcal{E}}
	\ell\left(f^{b-1}(\x_j),y_j \right) \,,
\end{align}
where $f^{b-1}$ stands for the incremental model after training the last task $\D^{b-1}$. 
Eq.~\ref{eq:gem} optimizes the model with a restriction, which requires the loss calculated with the exemplar set not to exceed the former model. Since exemplars are representative instances from former classes, GEM strikes a balance between learning new classes and preserving former knowledge. Furthermore, it transforms the constraints in Eq.~\ref{eq:gem} into:
\begin{equation} \label{eq:gem2}
	\left\langle g, g_{old}\right\rangle:=\left\langle\frac{\partial \ell\left(f(\x), y\right)}{\partial \theta}, \frac{\partial \ell\left(f, \mathcal{E}\right)}{\partial \theta}\right\rangle \geq 0 \,,
\end{equation}
where $g, g_{old}$ denotes the gradients of the current updating step and exemplar set, respectively.  Eq.~\ref{eq:gem2} requires the angle between gradients to be {\em acute}. If all the inequality constraints are satisfied, then the proposed parameter update is unlikely to increase the loss of previous tasks. However, if violations occur, GEM proposes to project the gradients $g$ to the closest gradient $\tilde{g}$ satisfying the constraints. GEM further transforms the optimization into a Quadratic Program (QP) problem. However, since the regularization in Eq.~\ref{eq:gem2} is defined among all exemplars, it requires calculating loss among exemplar sets and solving the QP problem in each optimization step. Hence, optimizing GEM is very time-consuming. To this end, A-GEM~\cite{chaudhry2018efficient} is proposed to speed up the optimization by relaxing the constraints in Eq.~\ref{eq:gem2} into a random batch. A similar idea is also adopted in~\cite{aljundi2019gradient}.

There are other methods to address the regularization problem with exemplars. For example, Adam-NSCL~\cite{wang2021training} proposes to sequentially optimize network parameters by projecting the candidate parameter update into the approximate null space of all previous tasks.
OWM~\cite{zeng2019continual} only allows weight modification in the direction orthogonal to the subspace spanned by all previously learned inputs. LOGD~\cite{tang2021layerwise} further decomposes the gradients into shared and task-specific ones. In model updating, the gradient should be close to the gradient of the new task, consistent with the gradients shared by all old tasks, and orthogonal to the space spanned by the gradients specific to the old tasks.

\noindent\textbf{Discussions}: Data regularization methods utilize the exemplar set in another manner, \ie,  treating the loss of them as the {\em indicator} of forgetting. 
They assume that the loss on exemplars is consistent with the prior tasks and align the parameter updates with the direction of the exemplar set. Hence, previous knowledge can be preserved for these exemplars due to the aligned gradient direction. However, these assumptions may not stand in some cases~\cite{chaudhry2018efficient}, which results in poor performance. To this end, \cite{guo2022adaptive} gets rid of the requirement of exemplars and manually projects the gradient direction to be orthogonal with previous ones. On the other hand, some works assume the updating rule can be meta-learned~\cite{finn2017model,chao2020revisiting} from a series of related tasks. MER~\cite{riemer2018learning} regularizes the objective of data replay so that gradients on incoming examples are more likely to have transferability and less likely to have interference with respect to past examples.
iTAML~\cite{rajasegaran2020itaml} separates the generic feature extraction module from the task-specific classifier, thereby minimizing interference and promoting a shared feature space among tasks. 
\cite{wang2022anti} extends data replay with adversarial attack and meta-learns an adaptive fusion module to help allocate capacity to the knowledge of different difficulties. 

Moreover, since data regularization and data replay both need to save previous data in the memory, similar problems will also occur in data regularization-based methods, \eg, overfitting, generalization issues~\cite{lopez2017gradient,verwimp2021rehearsal}, and privacy concerns~\cite{dong2022federated}. Consequently, it would be interesting to design privacy-friendly algorithms to build the regularization term with intermediate products for real-world applications.

\subsection{Dynamic Networks}\label{sec:dynamic-networks}
Deep neural networks are proven to produce task-specific features~\cite{goodfellow2016deep}. For example, when the training dataset contains `cars,' the model tends to depict the wheels and windows. However, if the model is updated with new classes containing `cats,' the features would be adapted for beards and stripes. Since the capacity of a model is limited, adapting to new features will result in the {\em overwriting} of old ones and forgetting~\cite{yoon2018lifelong}. Hence, utilizing the extracted features for beards and strides is inefficient for recognizing a car. To this end, dynamic networks are designed to dynamically adjust the model's representation ability to fit the evolving data stream.
There are several ways to expand representation ability, and we divide them into three sub-groups, \ie, {\em neuron expansion}, {\em backbone expansion}, and {\em prompt expansion}.

\subsubsection{Neuron Expansion}
Early works focus on {\em neuron expansion} adding neurons when the representation ability is insufficient to capture new classes. 
DEN~\cite{yoon2018lifelong} formulates the adjusting process into selection, expansion, duplication, and elimination. Facing a new incremental task, the model first selectively retrains the neurons that are relevant to this task. If the retrained loss is still above some threshold, DEN considers expanding new neurons top-down and eliminating the useless ones with group-sparsity regularization. Afterward, it calculates the neuron-wise drift and duplicates neurons that drift too much from the original values. Apart from heuristically expanding and shrinking the network structure, RCL~\cite{xu2018reinforced} formulates the network expansion process into a reinforcement learning problem and searches for the best neural architecture for each incoming task. Similarly, Neural Architecture Search (NAS)~\cite{elsken2019neural} is also adopted to find the optimal structure for each of the sequential tasks~\cite{li2019learn}.

\begin{figure}[t]
	\vspace{-3mm}
	\begin{center}
\includegraphics[width=1\columnwidth]{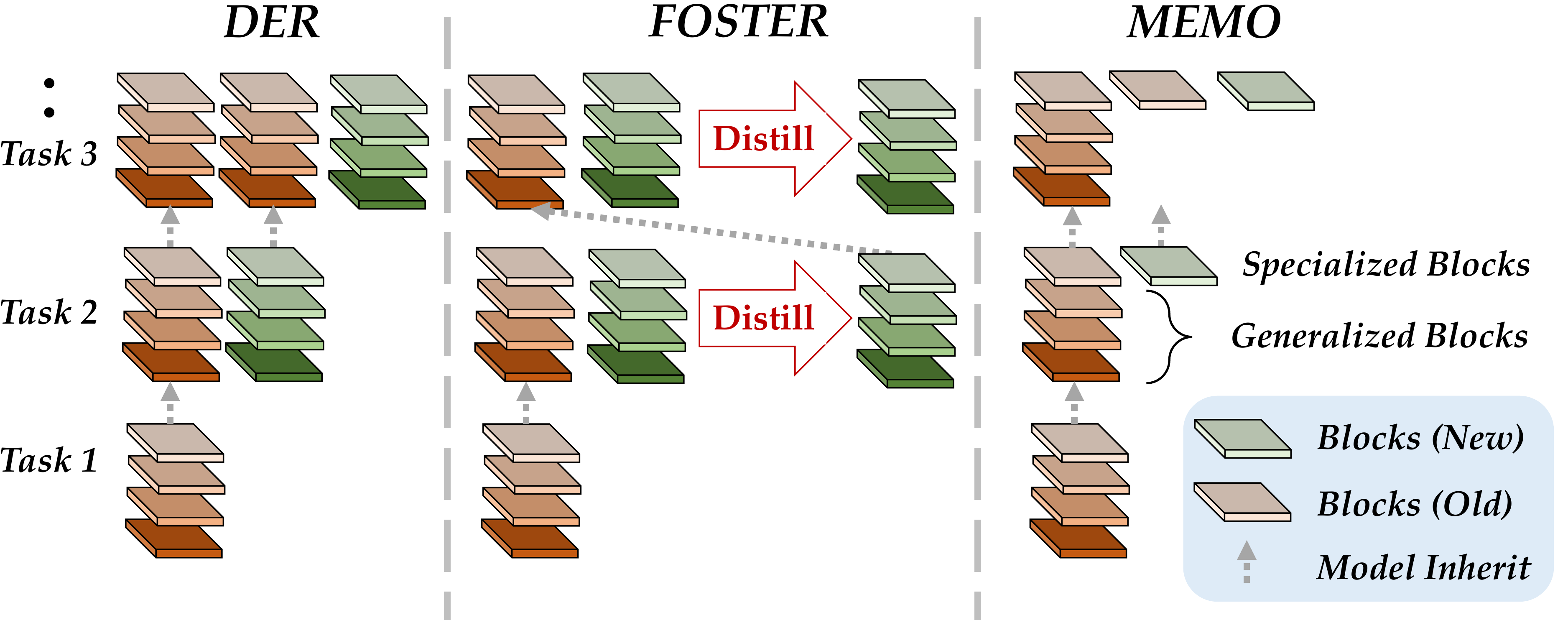}
	\end{center}
	\vspace{-5mm}
	\caption{ Illustration of network structure evolving in backbone expansion. \textbf{Left:} DER expands a new backbone per incremental task. \textbf{Middle:} FOSTER adds an extra model compression stage, which maintains limited model storage. \textbf{Right:} MEMO decouples the network structure and only expands specialized blocks.
	}
	\vspace{-5mm}
	\label{figure:dynamic_networks}
\end{figure}

\subsubsection{Backbone Expansion}
Expanding neurons shows competitive results with expandable representation. Correspondingly, several works try to duplicate the backbone network for stronger representation ability. PNN~\cite{rusu2016progressive} proposes learning a new backbone for each new task and fixing the former in incremental learning. It also adds layer-wise connections between old and new models to reuse former features. Expert Gate~\cite{aljundi2017expert} also expands the backbone per incremental task while it requires learning an extra gate to map the instance to the most suitable pathway during inference. To release the expansion cost, P\&C~\cite{schwarz2018progress} suggests a progression-compression protocol --- it first expands the network to learn representative representations. Afterward, a compression process is conducted to control the total budget. AANets~\cite{liu2021adaptive} partially expands stable and plastic blocks and aggregates their predictions to enhance model's representation ability. 
\cite{zhao2021mgsvf,pham2021dualnet} maintain a dual-branch network for class-incremental learning, one for fast adaptation and one for slow adaption. 
Recently, DER~\cite{yan2021dynamically} has been proposed to address the CIL problem. Similar to PNN, it expands a new backbone when facing new tasks and aggregates the features with a larger FC layer. Take the second incremental task for an example, where the model output is aggregated as:
\begin{equation}
	f(\x)=W_{new}^\top [\phi_{old}(\x),\phi_{new}(\x)] \,,
\end{equation}
where $\phi_{old}$ is the former backbone, and  $\phi_{new}$ is the newly initialized backbone. $[ \cdot,\cdot]$ denotes feature aggregation, and $W_{new}\in \mathbb{R}^{2d\times |\mathcal{Y}_{b}|}$ is the newly initialized FC layer. In the model updating process, the old backbone is frozen to maintain former knowledge:
\begin{align} \label{eq:der} 
	\mathcal{L}=	 \sum_{k=1}^{|\mathcal{Y}_b|}-\mathbb{I}(y=k) \log \mathcal{S}_k(W_{new}^\top [\bar{\phi}_{old}(\x),\phi_{new}(\x)]) \,,
\end{align}
where $\bar{\phi}_{old}(\x)$ denotes the old backbone is frozen. DER also adopts an auxiliary loss to differentiate between old and new classes. Eq.~\ref{eq:der} depicts a way to continually expand the model with new features. Under such circumstances, if the old backbone is optimized with `cars,' the features extracted by $\phi_{old}$ are then representative of wheels and windows. The new backbone trained with `cats' is responsible for extracting beards and stripes.
Since the old backbone is frozen in later stages, learning new classes will not overwrite the features of old ones, and forgetting is alleviated. Figure~\ref{figure:dynamic_networks} (left) depicts the model evolution of DER.

However, saving a backbone per task requires numerous memory size in DER, and many works are proposed to obtain expandable features with a {\em limited} memory budget.
 FOSTER~\cite{wang2022foster} formulates the learning process in Eq.~\ref{eq:der} as a feature-boosting~\cite{zhou2012ensemble} problem. It argues that not all expanded features are needed for incremental learning and need to be integrated to reduce redundancy. For example, suppose old classes contain `tigers,' and new classes contain `zebras.' 
 In that case, the stripe will be a useful feature that both old and new backbones could extract.
 Under such circumstances, forcing the new backbone to extract the same features is less effective for recognition. Hence, FOSTER adds an extra model compression process by knowledge distillation~\cite{hinton2015distilling}:
\begin{align} \label{eq:foster}
	\min_{f_s(\x)} \operatorname{KL}
	\left(
	\mathcal{S}\left(
	f_{t}(\x)
	\right) 
	\|
	\mathcal{S}\left(
	f_{s}(\x)
	\right)
	\right) \,.
\end{align}
Eq.~\ref{eq:foster} aims to find the student model $f_{s}$ that has the same discrimination ability as the teacher model $f_{t}$ by
 minimizing the discrepancy between them. The teacher is the {\em frozen} expanded model with two backbones: $f_{t}(\x)=W_{new}^\top [\phi_{old}(\x),\phi_{new}(\x)]$, and the student is the newly initialized model $
f_{s}(\x)=W^\top\phi(\x)$. Hence, the number of backbones is consistently {\em limited} to a single one, and the memory budget will not suffer catastrophic expansion. Figure~\ref{figure:dynamic_networks} (middle) depicts the model evolution of FOSTER. 

MEMO~\cite{zhou2022model} addresses the memory problem in CIL, aiming to enable model expansion with the {\em least} budget cost. It finds that in CIL, shallow layers of different models are similar, while deep layers are diverse. In other words, shallow layers are more generalizable, while deep layers are specific to the task, making expanding shallow layers less memory-efficient for CIL.
 Hence, MEMO proposes to decouple the backbone at middle layers: $\phi(\x)=\phi_{s}(\phi_{g}(\x))$, where specialized block $\phi_{s}$ corresponds to the deep layers in the network, while generalized block $\phi_{g}$ corresponds to the rest shallow layers. Compared to DER, MEMO only expands specialized blocks $\phi_s$, and transforms Eq.~\ref{eq:der} into:
 \begin{align} \label{eq:memo} \notag
	 \sum_{k=1}^{|\mathcal{Y}_b|}-\mathbb{I}(y=k) \log \mathcal{S}_k(W_{new}^\top [{\phi_s}_{old}(\phi_g(\x)),{\phi_s}_{new}(\phi_g(\x))]) \,,
\end{align}
which indicates that task-specific deep layers can be built for each task upon the shared shallow layers $\phi_g(\x)$. Figure~\ref{figure:dynamic_networks} (right) depicts the model evolution of MEMO.

\subsubsection{Prompt Expansion}
Recently, Vision Transformer (ViT)~\cite{dosovitskiy2020image} has attracted the attention of the computer vision community, and many works tend to design CIL learners using ViT as the backbone. 
DyTox~\cite{douillard2022dytox} is the first work to explore ViT in CIL, which finds that model expansion in ViT is much easier than in convolutional networks. In DyTox, only task tokens are expanded for each new task, which requires much less memory than saving the whole backbone. Similarly, L2P~\cite{wang2022learning} and DualPrompt~\cite{wang2022dualprompt} explore how to build CIL learners with pre-trained ViT. They borrow ideas from Visual Prompt Tuning (VPT)~\cite{jia2022visual} to incrementally finetune the model with prompts. In L2P, the pre-trained ViT is frozen during the learning process, and the model only optimizes the prompts to fit new patterns. The prompt pool is defined as: $	\mathbf{P}=\left\{P_1, P_2, \cdots, P_M\right\}$,
where $M$ is the total number of prompts,  $P_i\in\R^{L_p\times d}$ is a single prompt with token length $L_p$ and the same embedding size $d$ as the instance embedding $\phi(\x)$.
The prompts are organized as key-value pairs --- each instance selects the most similar prompts in the prompt pool via KNN search. It obtains instance-specific predictions by adapting the input embeddings as: $	{\x}_p=\left[P_{s_1} ; \cdots ; P_{s_N} ; \phi(\x)\right], \quad 1 \leq N \leq M$, where
$P_{s_j}$ are the selected instance-specific prompts.
The adapted embeddings are then fed into the self-attention layers~\cite{vaswani2017attention} to obtain instance-specific representations.
 CODA-Prompt~\cite{seale2022coda} extends the prompt search with the attention mechanism. Apart from pre-trained ViT, S-Prompt~\cite{wang2022s} utilizes the pre-trained language-vision model CLIP~\cite{radford2021learning} for CIL, which simultaneously learns language prompts and visual prompts to boost representative embeddings. Apart from expanding prompts, other lightweight modules can also be dynamically expanded in CIL~\cite{zhou2024continual,zhou2024expandable,zhou2023learning}

\noindent\textbf{Discussions}: 
Learning dynamic networks, especially backbone expansion methods, has achieved state-of-the-art performance in recent years~\cite{zhou2022model,yan2021dynamically,wang2022foster,wang2022beef}. However, it often requires expandable memory budgets, which is unsuitable for incremental learning on edge devices. To tackle this problem, further model compression~\cite{wang2022foster}, decoupling~\cite{zhou2022model}, and pruning can be adopted to alleviate the memory budget. Additionally, training DER requires an individual backbone for each task and aggregates all historical backbones as the feature extractor. It implicitly results in an unfair comparison to other methods with a single backbone~\cite{zhou2022model}. In this paper, we systematically investigate the fair comparison protocol between these dynamic networks and others in Section~\ref{sec:fair_exp}, \ref{sec:auc_exp}. Additionally, expanding backbones ignores the semantic information across tasks, \eg, when the old task contains `tigers' and the new task contains `zebras,' features like `stripes' will be extracted by multiple backbones, resulting in feature redundancy. Hence, analyzing the semantic relationship across tasks~\cite{zhou2021co} can help detect the feature redundancy, and contrastive learning~\cite{cha2021co2l} can be adopted for generalizable features.

On the other hand, most prompt expansion methods rely on the pre-trained models as initialization. Without such generalizable backbones, lightweight model updating with prompts often fails~\cite{tang2023prompt}. However, a pre-trained model is not always available for some specific downstream tasks, \eg, face recognition and speech recognition. Therefore, how to get rid of the dependence on pre-trained models is essential for these methods in real-world applications. Additionally, these prompt expansion-based methods tend to select instance-specific prompts based on a batch of instances. This requirement also needs to be satisfied during inference for accurate prompt retrieval~\cite{wang2022learning}, which implicitly results in an unfair comparison. Since a batch of instances is utilized to get the prompt, the context among instances becomes available, which is against the common sense of i.i.d. testing in machine learning.

Apart from these groups, there are works addressing network masks to divide a large network into sub-networks for each task~\cite{hung2019compacting,rajasegaran2019random,abati2020conditional}. However, deciding the activation of a specific sub-network requires the task identifier or learning extra task classifiers. On the other hand, several works~\cite{ostapenko2021continual,ermismemory} propose to design specific modules for incremental new tasks, \eg, adapters~\cite{rebuffi2017learning}. However, manually handcrafting these modules requires heuristic designs or task-specific priors.

\subsection{Parameter Regularization}\label{sec:parameter-regularization}

Dynamic networks seek to adjust model capacity with data evolves. However, if the model structure is fixed and unchangeable, how can we adjust the plasticity to resist catastrophic forgetting? Parameter regularization methods consider that the contribution of each parameter to the task is not equal. Hence, they seek to evaluate each parameter's {\em importance}  to the network and keep the important ones static to maintain former knowledge.

Typical works estimate a distribution over the model parameters and use it as the prior when learning new tasks. Due to the large amounts of parameters, the estimation process often assumes them to be independent. EWC~\cite{kirkpatrick2017overcoming} is the first work addressing parameter regularization. It maintains an importance matrix with the same scale of the network, \ie, $\Omega$. Denote the $k$-th model parameters as $\theta_k$, the importance of $\theta_k$ is represented by $\Omega_k\ge 0$ (the larger $\Omega_k$ indicates $\theta_k$ is more important).
Apart from the training loss in Eq.~\ref{eq:finetune} to learn new classes, EWC builds an additional regularization term to remember old ones:
\begin{equation}\label{eq:ewc}
	 \mathcal{L}= \ell(f(\x),y)+ \frac{1}{2}\lambda\sum_{k}\Omega_k(\theta_k^{b-1}-\theta_k)^2 \,.
\end{equation}
The parameter-wise regularization term is calculated based on two parts. $\theta_k^{b-1}$ denotes the $k$-th parameter after learning last task $\D^{b-1}$. Hence,  $(\theta_k^{b-1}-\theta_k)^2$ represents the parameter drift from the last stage, and $\Omega_k$ weighs it to ensure important parameters do not shift away from the last stage. Since the model at the last stage represents the `old' knowledge, consolidating important parameters can prevent the knowledge from being forgotten.

Eq.~\ref{eq:ewc} depicts a way to penalize essential parameters, and there are different ways to calculate the importance matrix $\Omega$. In EWC, Fisher information matrix~\cite{le2012asymptotic} is adopted to estimate $\Omega$. However, the importance calculation in EWC is conducted at the end of each task, which ignores the optimization dynamics along the model training trajectory. To this end, SI~\cite{zenke2017continual} proposes to estimate $\Omega$ in an online manner and weigh the importance via its contribution to loss decay. RWalk~\cite{chaudhry2018riemannian} combines these importance estimation techniques. \cite{aljundi2018memory,aljundi2019task} resort to an extra unlabeled dataset for online evaluation. IMM~\cite{lee2017overcoming} finds a maximum of the
mixture of Gaussian posteriors with the estimated Fisher information matrix. IADM~\cite{yang2019} and CE-IDM~\cite{yang2021cost} analyze the capacity and sustainability of different layers and find that different layers have different characteristics in CIL. In detail, shallow layers converge faster but have limited representation ability. By contrast, deep layers converge slowly while having powerful discrimination abilities. Hence, IADM augments EWC with an ensemble of different layers and learns layer-wise importance matrix in an online manner. K-FAC~\cite{lee2020continual} extends the Fisher information matrix approximation with the Kronecker factorization technique.

\noindent\textbf{Discussions}:
Although parameter and data regularization (Section~\ref{sec:data-regularization}) both exert regularization terms to resist forgetting, their basic idea differs substantially. Specifically, data regularization relies on the exemplar set to direct the optimization direction, while parameter regularization is based on the parameter-wise importance to construct the regularization term.

 As shown in Figure~\ref{figure:timeline}, parameter regularization methods have attracted the attention of the community in the early years. 
\cite{benzing2022unifying} shows that despite stemming from very different motivations, both SI~\cite{zenke2017continual} and MAS~\cite{aljundi2018memory} approximate the square root of
the Fisher Information, with the Fisher being the theoretically justified basis of EWC. 
However, estimating parameter importance requires saving the matrix with the same scale as the backbone. It faces the same risk as dynamic networks in that the memory budget is linearly increasing when learning more and more tasks. 

On the other hand, the importance matrix may conflict at different incremental stages~\cite{zhou2021co}, making it hard to optimize the model and achieve poor performance on new tasks. Hence, although these works achieve competitive results in task-incremental learning, many works~\cite{van2022three} find that parameter regularization-based methods perform poorly in the class-incremental learning scenario. To this end, some works try to alleviate the intransigence by learning a new backbone and consolidating them into a single one~\cite{lee2017overcoming,schwarz2018progress}. In that case, the learning of new tasks will not be affected by the regularization term, and the parameter importance will only be considered during the consolidation process, enabling the model to be fully fitted to new tasks.

\subsection{Knowledge Distillation}\label{sec:knowledge-distillation}

Training data is evolving in the learning process, requiring tuning the model sequentially. We can denote the model after the previous stage $f^{b-1}$ as the `old model' and the current updating model $f$ a s the `new model.' Assuming the old model is a good classifier for all the seen classes in $\Y_{b-1}$, how can we utilize it to resist forgetting in the new model?
To enable the old model to assist the new model, an intuitive way utilizes the concept proposed in~\cite{hinton2015distilling}, \ie, knowledge distillation (KD). KD enables the knowledge transfer from a teacher model to the student model, with which we can teach the new model not to forget.
There are several ways to build the distillation relationship, and we divide these KD-based methods into three subgroups, \ie, logit distillation, feature distillation, and relational distillation.

LwF~\cite{li2016learning} is the first success to apply knowledge distillation into CIL. Similar to Eq.~\ref{eq:ewc}, it builds the regularization term via knowledge distillation to resist forgetting:
{\small
	\begin{align}\label{eq:lwf}
	\mathcal{L}= \underbrace{
		\vphantom{\sum_{k=1}^{|\mathcal{Y}_{b-1}|}}
		\ell(f(\x),y)}_{\text{Learning New Classes}}+ \underbrace{
	\sum_{k=1}^{|\mathcal{Y}_{b-1}|}-
	\mathcal{S}_k({f}^{b-1}(\x))
	\log \mathcal{S}_k(f(\x))}_{{\text{Remembering Old Classes}}} \,,
\end{align}}where the old model ${f}^{b-1}$ is frozen during updating. The regularization term builds the mapping between the old and new models by forcing the predicted probability among old classes to be the same. Given a specific input $\x$, the output probability of the $k$-th class reveals the semantic similarity of the input to this class. Hence, Eq.~\ref{eq:lwf} forces the semantic relationship of the old and new models to be the same and resists forgetting. iCaRL~\cite{rebuffi2017icarl} extends LwF with the exemplar set, which helps to further recall former knowledge during incremental learning. Additionally, it drops the fully-connected layers and follows~\cite{mensink2013distance} to utilize nearest-mean-of-exemplars during inference. Eq.~\ref{eq:lwf} strikes a trade-off between old and new classes, where the former part aims to learn new classes and the latter one maintains old knowledge. Since the number of old and new classes may differ in different incremental stages, BiC~\cite{wu2019large} extends Eq~\ref{eq:lwf} by introducing a {\em dynamic} trade-off term:
\begin{equation}\label{eq:icarl_tradeoff} \notag
	\mathcal{L}= (1-\lambda)\ell(f(\x),y)+ \lambda \sum_{k=1}^{|\mathcal{Y}_{b-1}|}-
	\mathcal{S}_k({f}^{b-1}(\x))
	\log \mathcal{S}_k(f(\x))\,,  
\end{equation}
where $\lambda=\frac{\lvert\mathcal{Y}_{b-1}\rvert}{\lvert\mathcal{Y}_{b}\rvert}$ denotes the proportion of old classes among all classes. It increases as incremental tasks evolve, indicating that the model pays more attention to old ones. 

LwF inspires the community to build the mapping between models, making knowledge distillation a useful tool in CIL. 
D+R~\cite{hou2018lifelong} suggests changing the first part in Eq.~\ref{eq:lwf} into a distillation loss by training an extra expert model.
GD~\cite{lee2019overcoming} proposes to select wild data for model distillation and designs a confidence-based sampling method to effectively leverage external data.
Similarly, DMC~\cite{zhang2020class} proposes to train a new model in each incremental stage and then compress them into a single model via an extra unlabeled dataset. Finally, if no additional data is available for knowledge distillation,  ABD~\cite{smith2021always} proposes distilling synthetic data for incremental learning. 
These methods only concentrate on utilizing the old model to help resist forgetting in the new model.  
However, COIL~\cite{zhou2021co} suggests conducting {\em bidirectional} distillation with co-transport, where semantic relationships between old and new models are both utilized.

\begin{figure}[t]
	\begin{center}
		\includegraphics[width=1\columnwidth]{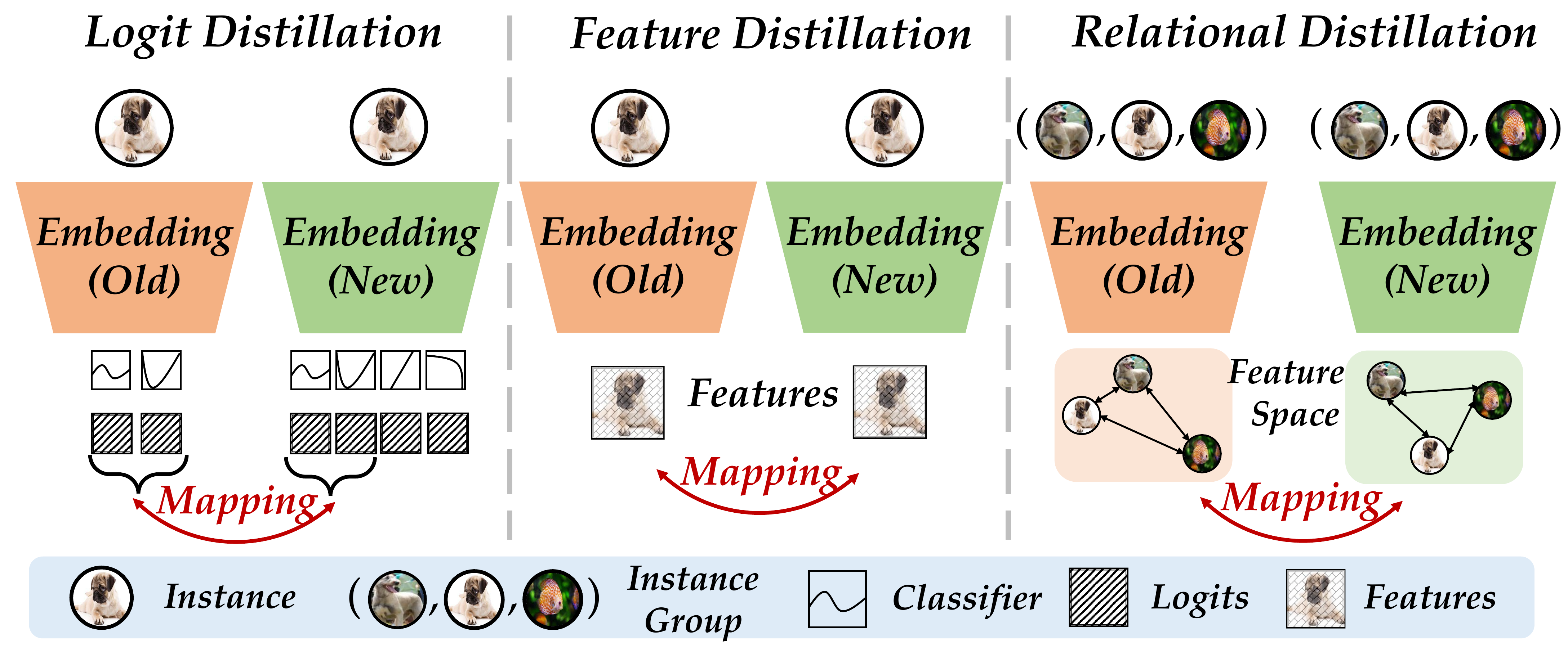}
	\end{center}
	\vspace{-5mm}
	\caption{ Illustration of knowledge distillation in CIL. \textbf{Left:} Logit distillation aligns the model outputs to make the old and new models share the same semantic relationship. 
		\textbf{Middle:} Feature distillation aligns the features produced by the old and new models to ensure the new model does not forget old features.
		\textbf{Right:} Relational distillation resorts to structural inputs, \eg, triples, and aligns the input relationship of the old and new model.
	} \label{figure:knowledge_distillation}
	\vspace{-5mm}
\end{figure}

Apart from distilling the logits, 
some works propose to distill the intermediate product of deep models, \eg, extracted features. UCIR~\cite{hou2019learning} replaces the regularization term in Eq.~\ref{eq:lwf} into:
\begin{align}\label{eq:ucir}
	\mathcal{L}= \ell(f(\x),y)+ (1- 
	\langle
	\frac{\phi^{b-1}(\x)}{\lVert\phi^{b-1}(\x)	\rVert},
	\frac{\phi(\x)}{\lVert\phi(\x)	\rVert} \rangle 	)
	\,.
\end{align}
Eq.~\ref{eq:ucir} forces the features extracted by the new embedding module to be the same as the old one, which is a stronger regularization than Eq.~\ref{eq:lwf}. Several works follow it to utilize feature distillation in CIL~\cite{lu2022augmented,park2021class,jung2018less,li2019rilod}, while others address distilling other products. 
LwM~\cite{dhar2019learning} suggests penalizing the changes in classifiers' attention maps to resist forgetting. AFC~\cite{kang2022class} conducts the distillation considering the importance of different feature maps.
PODNet~\cite{douillard2020podnet}  minimizes the difference of the pooled intermediate features in the height and width directions instead of
performing element-wise distillations. CVIC~\cite{zhao2021video} decouples the distillation term into spatial and temporal features for video classification.
DDE~\cite{hu2021distilling} distills the causal effect from the old training to preserve the old knowledge.
GeoDL~\cite{simon2021learning} conducts distillation based on the projection of two sets of features from old and new models.

However, both logit and feature distillation address the instance-wise mapping between old and new models. To reveal the structural information in model distillation, several works suggest conducting {\em relational} knowledge distillation~\cite{park2019relational}. The differences between these groups of knowledge distillation are shown in Figure~\ref{figure:knowledge_distillation}.

To conduct relational distillation, a group of instances needs to be extracted, \eg, triplets. We denote the extracted triplets as $\{\x_i,\x_j,\x_k\}$, where $\x_i$ is called the anchor. In a triplet, the target neighbor $\x_j$ is similar to the anchor $\x_i$ with the same class, while the impostor $\x_k$ is dissimilar to $\x_i$ (usually from different classes). R-DFCIL~\cite{gao2022rdfcil} suggests mapping the angle among triplets:
\begin{equation} \label{eq:rkd}
	 \sum_{\{\x_i, \x_j, \x_k\} \in \D^{b}}\left\|\cos \angle \mathbf{t}_i \mathbf{t}_j \mathbf{t}_k-\cos \angle \mathbf{s}_i \mathbf{s}_j \mathbf{s}_k\right\| \,,
\end{equation}
where $\mathbf{t}_m=\phi^{b-1}(\x_m)$ is the representation in the old model's embedding space, and $\mathbf{s}_m=\phi(\x_m)$ denotes the representation in the current model. The cosine value is calculated in the corresponding embedding space. Eq.~\ref{eq:rkd} provides a way to encode the old model's structural information into the new model and align the feature space softly. ERL~\cite{dong2021few} extends this regularization into few-shot CIL scenarios. TPCIL~\cite{tao2020topology} models the relationship with the elastic Hebbian graph and penalizes the changing of the topological relations between vertices. TOPIC~\cite{tao2020few} further explores neural gas network to model the class-wise relationship. 
Apart from the triplet relationship, MBP~\cite{liu2022model} extends the regularization to the instance neighborhood, requiring the old and new models to have the same distance ranking in the neighborhood.

\noindent\textbf{Discussions}: 
Knowledge distillation is a general idea to build the mapping between a set of methods, which has been widely adopted in class-incremental learning with many formats (\eg, logits, features, relationships). Due to their flexibility, knowledge distillation-based methods have also been widely applied to various incremental learning tasks, \eg, semantic segmentation~\cite{yang2022uncertainty}, person re-identification~\cite{pu2021lifelong}, human action recognition~\cite{park2021class}, and federated learning~\cite{dong2022federated}. Since a set of models exists in CIL, it is intuitive to build the student-teacher mapping in CIL, making knowledge distillation an essential solution for most works.  

However, since the knowledge distillation term aims to strike a balance between learning the new and remembering the old, it is hard to control the precise trade-off term between plasticity and stability. Specifically, giving more importance to the knowledge distillation term will harm the plasticity of learning new tasks, while giving lower importance will cause catastrophic forgetting or feature overwriting.
Compared to dynamic networks, knowledge distillation-based methods lack the ability to learn more informative features as data evolves.  \cite{zhou2022model} compares knowledge distillation-based and dynamic network-based methods and finds that knowledge distillation-based methods and dynamic networks have their advantages given different memory budgets. Specifically, knowledge distillation methods show stronger performance given limited memory, while dynamic networks require an adequate memory budget to perform competitively. Besides, feature/relational distillation only regularizes the extracted features to be similar, thus regularizing the embedding function to resist forgetting. However, due to the data characteristics of CIL, the classifier layer also gets biased and forgets former knowledge and knowledge distillation-based methods cannot handle such challenges.

\subsection{Model Rectify} \label{sec:model_rectify}

Assuming we can get all the training datasets at once and shuffle them for training with multiple epochs, the model will not suffer any forgetting and will perform well among all classes. Such a protocol is known as the upper bound of class-incremental learning, denoted as `Oracle.' However, since the models trained with incremental data suffer catastrophic forgetting, several methods try to find the abnormal behaviors in CIL models and rectify them like the oracle model. These abnormal behaviors include the output logits, classifier weights, and feature embedding.

The first method addressing the bias of the CIL model is UCIR~\cite{hou2019learning}. It finds that the weight norm of new classes is significantly larger than old ones, and the model tends to predict instances as the new classes with larger weights. Hence, UCIR proposes to utilize a cosine classifier to avoid the influence of biased classifiers: $
f(\x)=\frac{W}{\lVert W\rVert}^\top\frac{\phi(\x)}{\lVert\phi(\x)\rVert}$. Thus, the weight norm will not influence model predictions in incremental learning. WA~\cite{zhao2020maintaining} further normalizes the weight after every optimization step. It also introduces weight clipping to ensure the predicted probability is proportionate to classifier weights.
SS-IL~\cite{ahn2021ss} explains the reason for weight drifting, which is caused by the imbalance phenomena between old and new instances. 
Since the number of new class instances is much more than that of old ones, optimizing the model with cross-entropy loss will increase the weight of new classes and decrease old ones. Hence, SS-IL suggests separated softmax operation and task-wise knowledge distillation to alleviate the influence of imbalanced data. RPC~\cite{pernici2021class} claims that the classifiers of all classes can be pre-allocated and fixed. This makes it impossible for classifiers to be biased toward new classes.

On the other hand, several works find that the predicted logits of new classes are much larger than old ones. E2E~\cite{castro2018end} proposes to finetune the fully-connected layers with a balanced dataset after each stage. Furthermore, BiC~\cite{wu2019large} proposes to attach an extra rectification layer to adjust the predictions. The extra layer only have two parameters, \ie, re-scale parameter $\alpha$ and bias parameter $\beta$, and the rectified output for the $k$-th class is denoted as:
\begin{align} \label{eq:bic}
	\hat{f}(\x)_k= \begin{cases}
		\alpha \w_k^\top\phi(\x)+\beta ,& k\in{Y}_b\\
		\w_k^\top\phi(\x) ,& otherwise
	\end{cases} \,.
\end{align}
Only the logits for new classes ($k\in{Y}_b$) are rectified after each incremental task. BiC separates an extra validation set from the exemplar set, \ie, $\mathcal{E}=\mathcal{E}_{train}\cup\mathcal{E}_{val}$, and uses the validation set to tune the rectification layer. On the other hand, IL2M~\cite{belouadah2019il2m} suggests re-scaling the outputs with historical statistics. Suppose an instance is predicted as a new class. In that case, the logits will be re-scaled to ensure the predictions of old and new classes follow the same distribution. 

Lastly, since the embedding module is sequentially updated in CIL, some works try to rectify the biased representations of incremental models. For example, SDC~\cite{yu2020semantic} utilizes a nearest-mean-of-exemplars classifier, which calculates the class centers and assigns instances to the nearest class center. However, since the embedding is incrementally updated, the class centers calculated in the former stage may suffer a drift in the next stage, making the classification results unreliable. Since old class instances are not available in the current stage, SDC aims to calibrate the class centers of old classes with the drift of new classes. CwD~\cite{shi2022mimicking} analyzes the differences of embeddings among the CIL model and the oracle model and finds that the embeddings of the oracle model scatter more uniformly. It aims to make the CIL model similar to the oracle by enforcing the eigenvalues to be close. 
ConFiT~\cite{jie2022alleviating} relieves the feature drift from the middle layers. MRFA~\cite{zheng2024multi} finds the all-layer margin of replay samples shrinks as data evolves.
Other works address the model weight rectification.
 CCLL~\cite{singh2020calibrating} aims to calibrate the activation maps of old models during incremental learning. RKR~\cite{singh2021rectification} proposes to rectify the convolutional weights of old models when learning new tasks. FACT~\cite{zhou2022forward} depicts a new training paradigm for CIL, namely {\em forward compatible training}. Since the embedding space is endlessly adjusted for new classes, FACT proposes to pre-assign the embedding space for new classes to relieve the burden of embedding tuning.

\noindent\textbf{Discussions}:
Model rectify-based methods aim to reduce the inductive bias in the CIL model and align it to the oracle model. This line of work helps to understand the inherent factor of catastrophic forgetting. Apart from the rectification methods listed in this section, \cite{ahn2021ss} addresses that the bias of the CIL model is from the imbalanced data stream. \cite{cha2021co2l} finds that the embedding trained with contrastive loss suffers less forgetting than cross-entropy loss. 
\cite{pham2021continual} finds that the batch normalization layers~\cite{ioffe2015batch} are biased in CIL and proposes re-normalizing the layer outputs. \cite{zheng2023preserving} finds that vision transformers gradually lose the locality information when incrementally updated and proposes to insert the prior information about locality into the self-attention process. 
These works often treat the oracle model as the example, and design training techniques to reflect oracle model's characteristics.

Apart from mimicking the oracle model, there are also works addressing the forward compatibility~\cite{zhou2022forward} and flat loss landscape. Since the ultimate goal of CIL is to find a flat minimum in loss landscape among all tasks, several works aim to achieve this goal during the first stage~\cite{shi2021overcoming,mirzadeh2020understanding}.
It is worth exploring other factors in catastrophic forgetting and corresponding solutions in the future. On the other hand, the oracle model is obtained via joint training of all data via supervised loss. Other task-agnostic features could also help build a holistic classifier, \eg, via contrastive learning~\cite{cha2021co2l}, which the oracle model does not possess.

\subsection{Template-Based Classification} \label{sec:template-based}
Lastly, we discuss template-based classification, which is widely adopted in CIL. If we can build a `template' for each class, the classification can be done by matching the query instance to the most similar template.
A popular approach is to utilize class prototypes~\cite{snell2017prototypical} as the template, which is rooted in cognitive science~\cite{nosofsky1986attention}. The prototype in deep neural networks is often defined as the average vector in the embedding space. For example, we can utilize the current embedding function $\phi(\cdot)$ to extract the prototype of the $i$-th class:
\begin{equation}  \label{eq:prototype} \textstyle
	{\bm p}_{i}= \frac{1}{N} \sum_{j=1}^{|\mathcal{{D}}^b|}\mathbb{I}(y_j=i)\phi(\x_j) \,,
\end{equation} 
where $N$ is the instance number of class $i$. In Eq.~\ref{eq:prototype}, the class prototype is calculated via the class center in the embedding space. Hence, we can make inference without relying on the fully connected layer by matching an instance to the nearest prototype:
\begin{equation} \label{eq:icarl}
	y^*=\underset{y=1, \ldots, |\mathcal{Y}_b| }{\operatorname{argmin}}\left\|\phi(\x)-\p_y\right\| \,.
\end{equation}
As discussed in model rectification-based methods, sequentially updating the incremental model will result in bias in the fully-connected layer. Correspondingly, iCaRL~\cite{rebuffi2017icarl} suggests conducting inferences via Eq.~\ref{eq:icarl}, which is also known as nearest-class-mean classifier~\cite{mensink2013distance}. Since inference is conducted by instance-prototype matching in the same embedding space, the bias among different stages can be alleviated. 
However, utilizing prototype-based inference also faces another challenge, \ie, the embedding mismatch between stages. Since the embedding function keeps changing among different stages, the prototypes of former stages may be incompatible with the query embedding of later stages. This phenomenon is also known as the semantic drift~\cite{yu2020semantic}. To fill in this gap, a naive solution is to utilize the set of exemplars and re-calculate class prototypes after each stage~\cite{rebuffi2017icarl,de2021continual}.
 Since the exemplar set $\mathcal{E}$ contains instances of former classes, re-calculating all class prototypes after each incremental stage ensures the compatibility between prototypes and the latest embedding function. 
However, when exemplars are unavailable, specific algorithms need to be designed to compensate for the semantic drift.

\noindent\textbf{Prototype-based Inference without Exemplars}: When exemplars are unavailable, there are two main solutions to maintain a prototype classifier. A naive solution is to freeze the embedding function after the first incremental stage~\cite{zhou2022forward,zhou2023revisiting}, which forces the prototypes of different stages to be compatible. Such a learning process assumes the embedding function trained with first-stage data is generalizable enough for future tasks, relying on the vast number of training instances in $\D^1$. As a result, researchers tend to design suitable training techniques with $\D^1$ to obtain a generalizable feature space for future tasks. \cite{shi2023prototype,zhu2021prototype} draw inspiration from contrastive learning and design pre-text tasks to enhance the representation learned by the first stage. 
CEC~\cite{zhang2021few} meta-learns a graph model to adjust new class prototypes with known ones, which propagates context information between classifiers for adaptation. 
LIMIT~\cite{zhou2022few} finds that using prototypes extracted by the first stage backbone tends to predict instances into classes of the first stage. It proposes calibrating the prediction logits by meta-learning a transformer block between old and new classes. 
TEEN~\cite{wang2023few} systematically analyzes the performance gap between old and new classes and finds that the prototypical network forces the model to predict new classes into the most similar old class. Hence, it suggests pushing the prototypes of new classes to old classes for a calibrated decision boundary.
Moreover, \cite{zhou2022forward}  aims to enhance the model's forward compatibility by reserving the embedding space for new classes so that new classes can be inserted into the embedding space without harming existing ones. It allocates `virtual prototypes' for new classes and explicitly reserves the embedding space for them using a bimodal target label and manifold mixup~\cite{verma2019manifold} to generate new class instances.

When we have a pre-trained embedding function as initialization, ADAM~\cite{zhou2023revisiting} finds that using a prototype-based classifier easily beats the state-of-the-art prompt-based methods~\cite{wang2022dualprompt,wang2022learning}. However, although the pre-trained model possesses generalizable features, it still lacks task-specific information on incremental datasets. Hence, it designs the `adapt and merge' protocol to unify the generalizability of the pre-trained model and the adaptivity of downstream tasks. With the pre-trained embedding, it finetunes it with the first stage dataset $\D^1$, and concatenates the pre-trained and adapted embeddings for prototype extraction. Based on ADAM, RanPAC~\cite{mcdonnell2023ranpac} designs random projection to project the concatenated features into a high-dimensional space, within which classes are separated more clearly. It also incrementally updates a Mahalanobis distance-based classifier, which shows stronger performance than cosine distance between prototypes and query embedding. FeCAM~\cite{goswami2023fecam} also finds the inadequacy of the cosine classifier and proposes using a Bayesian classifier instead. 

Apart from freezing the embedding, some works try to compensate for the semantic drift among different stages. Since prototypes will drift with the ever-changing embedding functions, they aim to estimate such drift to estimate the prototypes in the latest embedding space. SDC~\cite{yu2020semantic} aims to measure the prototype drift between different stages and utilizes the weighted combination of current stage data for reference. ZSTCI~\cite{wei2021incremental} achieves this goal by mapping prototypes of different stages into the same embedding space and designing a prototype alignment loss across stages.

Finally, another line of work considers generative classification~\cite{van2021class}. Different from estimating the class prototype as a template, the template for each class is a generative model. Hence, the inference process can be measured via the likelihood of the query instance under such a generative model. However, it requires more calculation budgets for generative templates during inference than prototype-based methods. 

\noindent\textbf{Discussions}: There are two advantages of template-based classification. Firstly, when the exemplar set is available, using template-based classification enables query-prototype matching in the embedding space. Since the classifier will be biased after incremental learning, utilizing such a matching target alleviates the inductive bias during inference~\cite{rebuffi2017icarl}. Secondly, when the pre-trained model (or model trained with large base classes) is available as initialization, the feature representations are generalizable and can be transferred to downstream tasks. Hence, freezing the embedding and using template-based classification can take full use of the generalizable features, and such a non-incremental learner will not suffer forgetting due to the frozen backbone~\cite{zhou2023revisiting,mcdonnell2023ranpac,goswami2023fecam}.

 However, there are also some drawbacks. When the exemplar set is not available, re-calculating the prototypes is impossible, and it requires a complex adjusting process to overcome the semantic drift~\cite{yu2020semantic,wei2021incremental}. Secondly, when freezing the backbone and taking a template-based classifier, the model sacrifices its adaptivity for downstream tasks. When there are significant domain gaps between the pre-trained model and the downstream data~\cite{hendrycks2021natural,alfassy2022feta}, template-based classification shall fail due to the incapability to extract generalizable features. 
  In that case, continually adjusting the backbone could be more suitable to extract task-specific features.
 Finally, we can use an energy-based model~\cite{lecun2006tutorial} to compute for each class an energy value rather than a likelihood~\cite{li2022energy} to alleviate the high cost of the generative model.

\section{Experimental Evaluation}
In this section, we conduct comprehensive experiments to evaluate the performance of different kinds of CIL methods with benchmark datasets. We first introduce the benchmark experimental setting, dataset split, evaluation protocol, and implementation details. Afterward, we aim to compare these methods from three aspects:
\begin{itemize}
	\item How do these methods perform on {\em benchmark} datasets?
	\item Are they {\em fairly} compared? How to fairly compare them?
	\item How to evaluate them with {\em memory-agnostic} measure?
\end{itemize}
Specifically, Sections~\ref{sec:cifar_exp} and \ref{sec:imagenet_exp} answer the first question, and Section~\ref{sec:fair_exp} answers the second question. We provide the holistic performance measures in Section~\ref{sec:auc_exp} to answer the third question and summarize the results in Section~\ref{sec:comparison_results}. 
We report more results, measures, and visualizations in the supplementary material.

\begin{figure*}[t]
	\vspace{-2mm}
	\begin{center}

		\includegraphics[width=1.9\columnwidth]{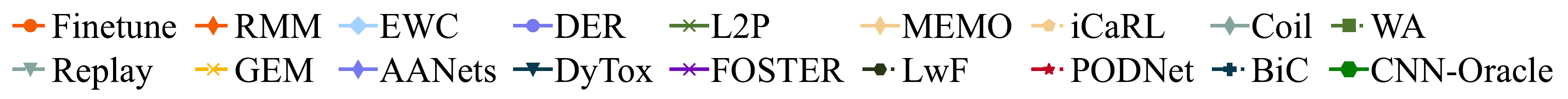}\\
		\vspace{-2mm}
		\subfigure[CIFAR100 Base0 Inc5 ]
		{	\includegraphics[width=.475\columnwidth]{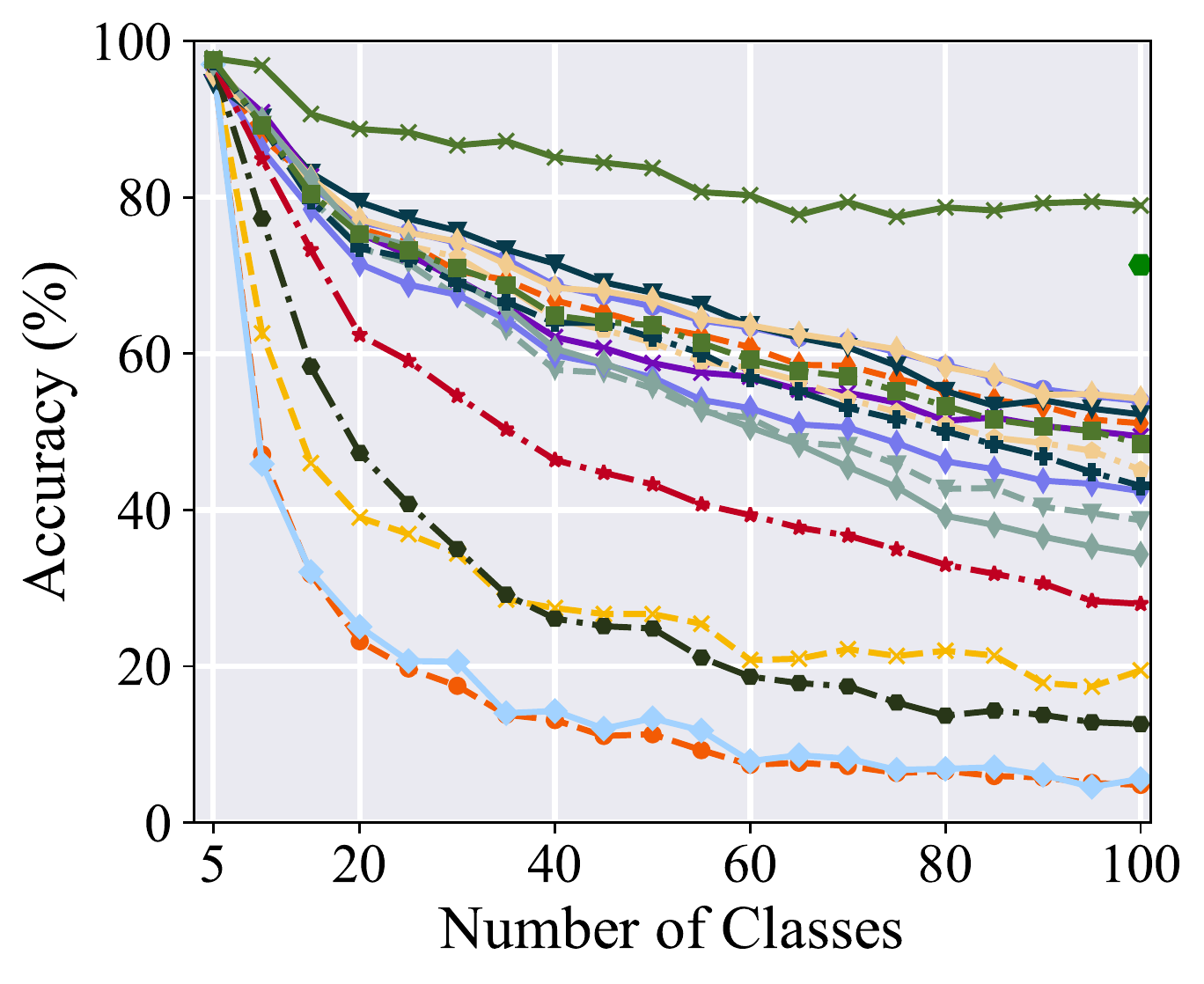}
		}
		\hfill
		\subfigure[CIFAR100 Base0 Inc10]
		{	\includegraphics[width=.475\columnwidth]{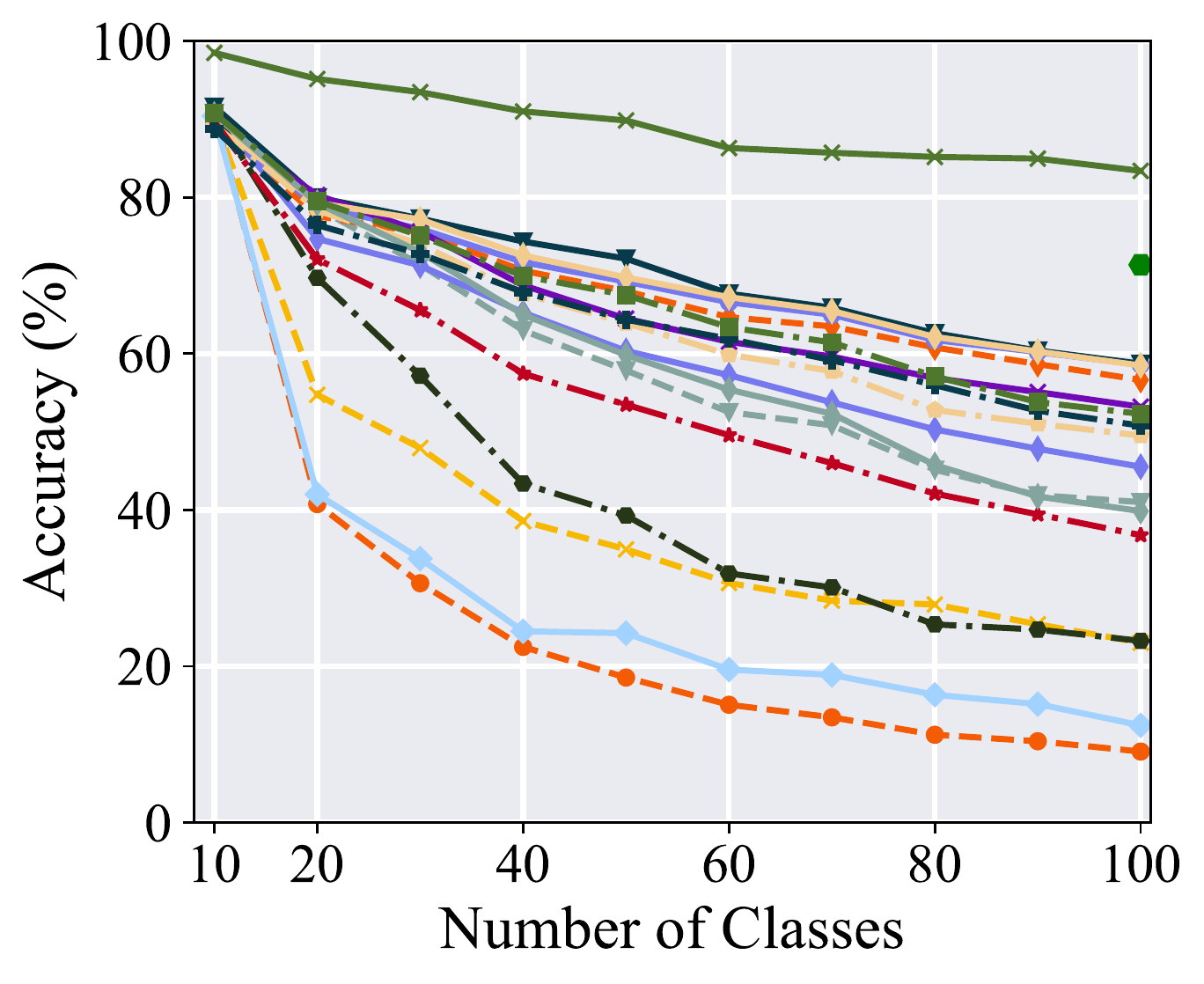}		
			\label{figure:cifar100b}
		}
		\hfill
		\subfigure[CIFAR100 Base50 Inc10]
		{	\includegraphics[width=.475\columnwidth]{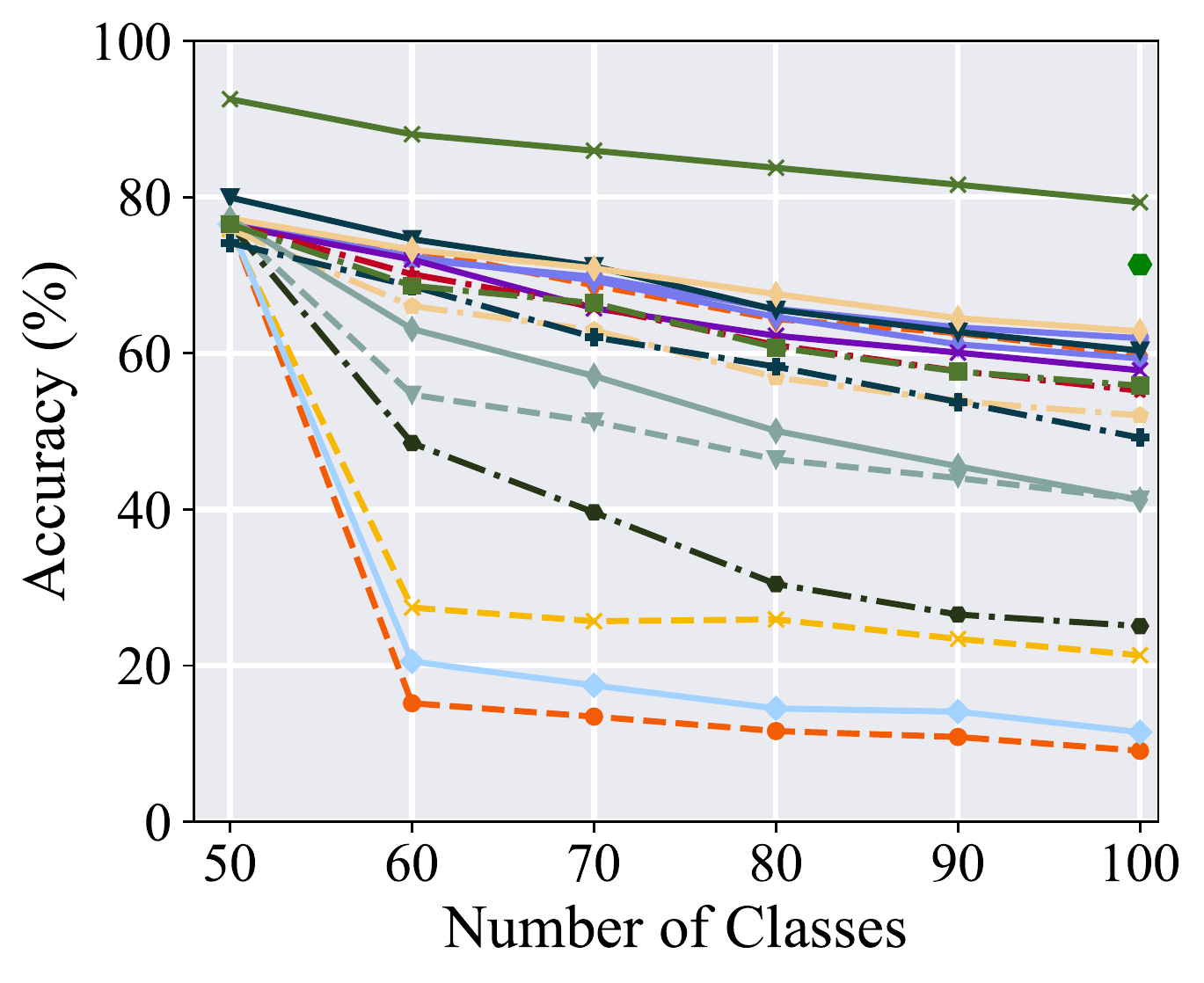}
		}
		\hfill
		\subfigure[CIFAR100 Base50 Inc25]
		{	\includegraphics[width=.475\columnwidth]{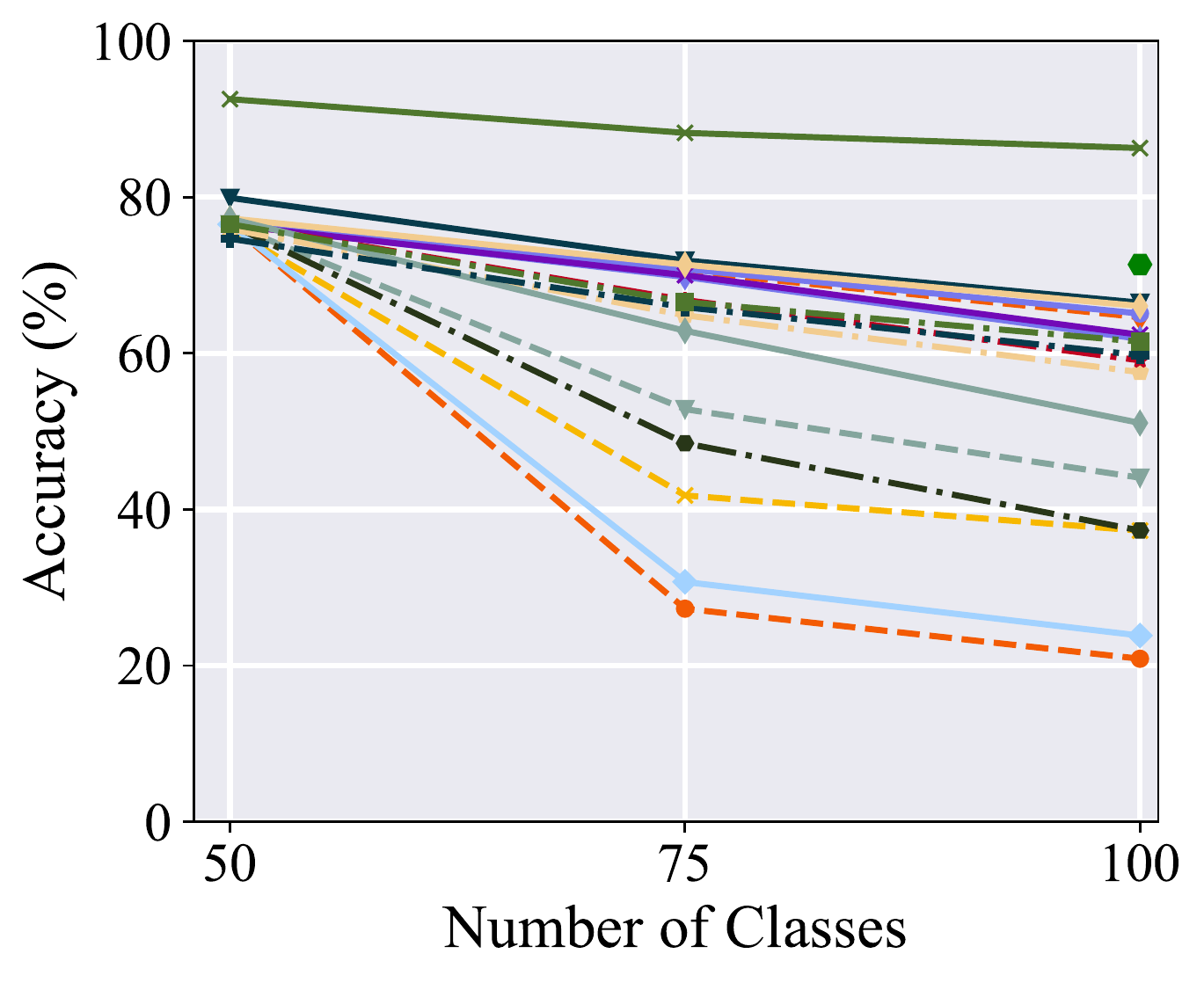}
		}\\
		\vspace{-3mm}
		\subfigure[ImageNet100 Base0 Inc5 ]
		{	\includegraphics[width=.475\columnwidth]{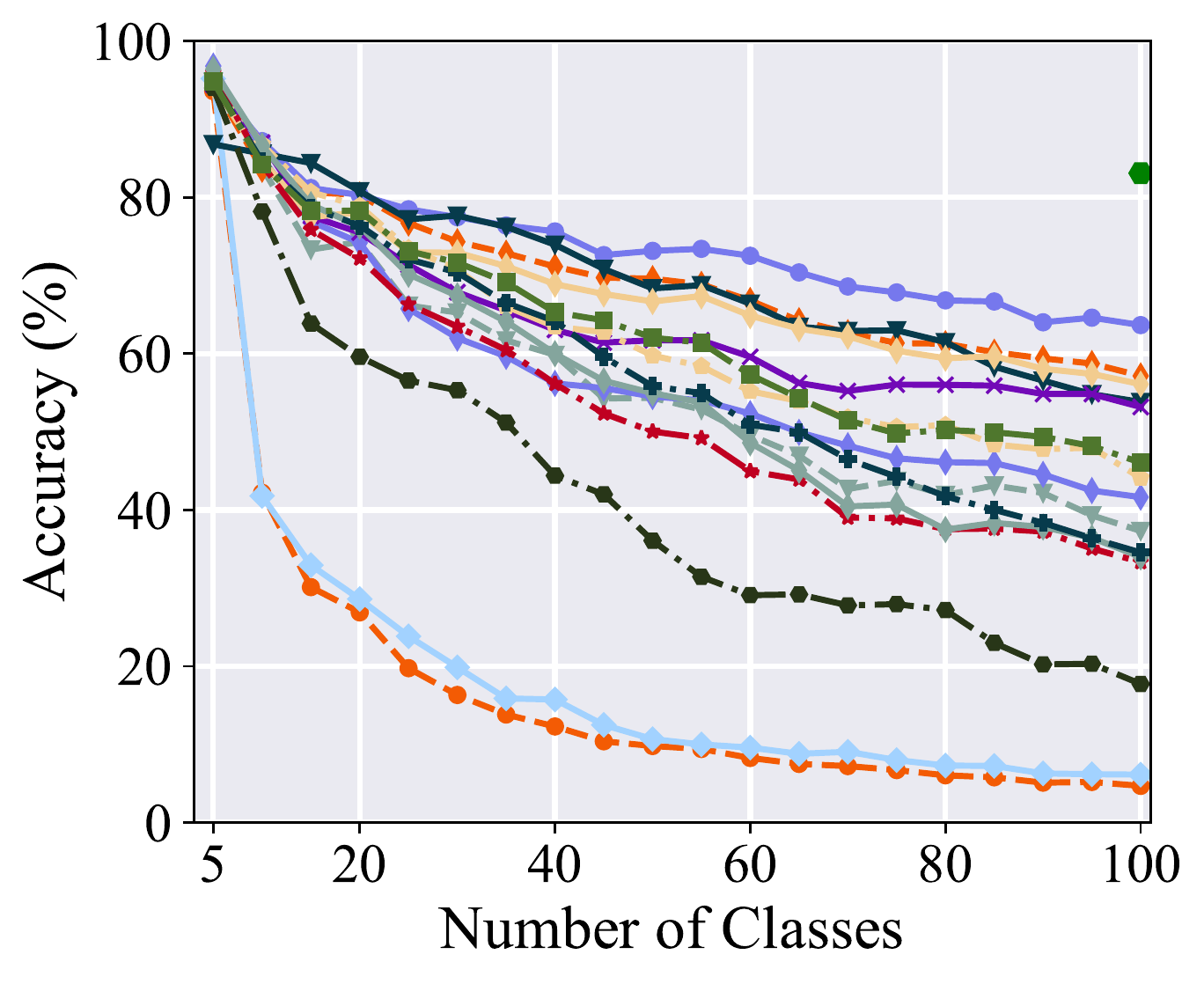}
		}
		\hfill
		\subfigure[ImageNet100 Base50 Inc10]
		{	\includegraphics[width=.475\columnwidth]{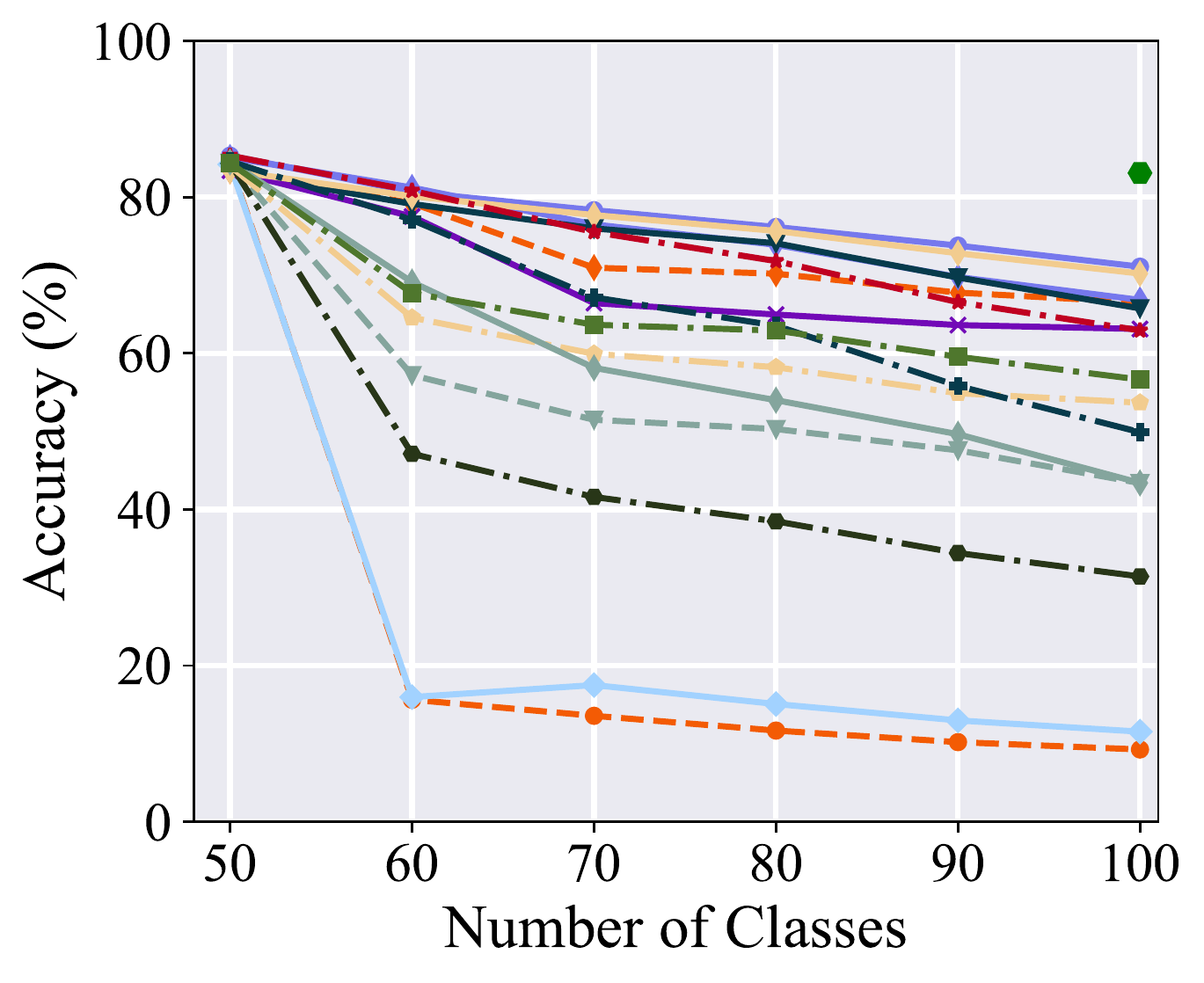}
		}
		\hfill 
		\subfigure[ImageNet1000 Base0 Inc100 ]
		{	\includegraphics[width=.47\columnwidth]{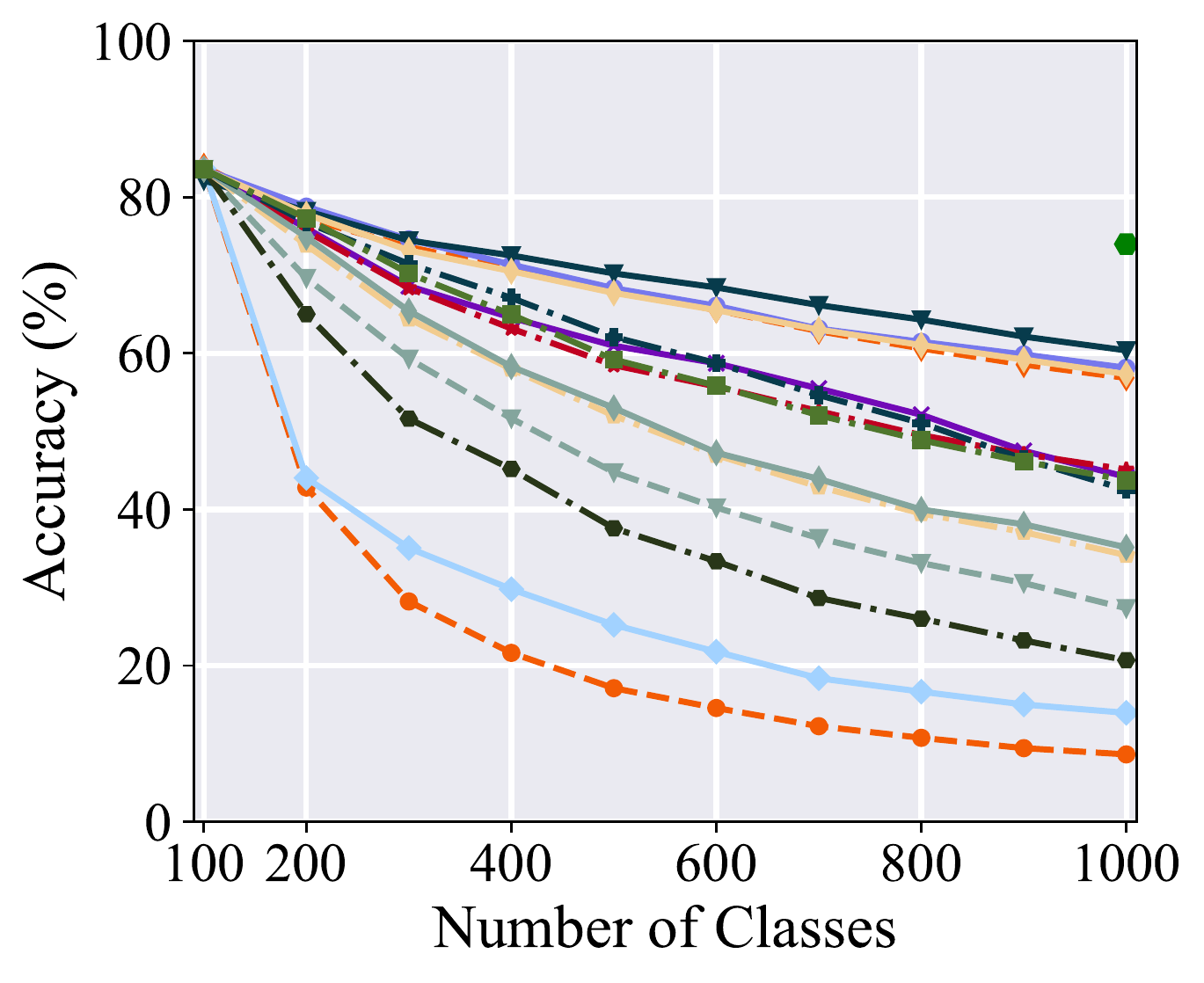}
		}
		\hfill 
		\subfigure[ImageNet1000 Base500 Inc100]
		{	\includegraphics[width=.47\columnwidth]{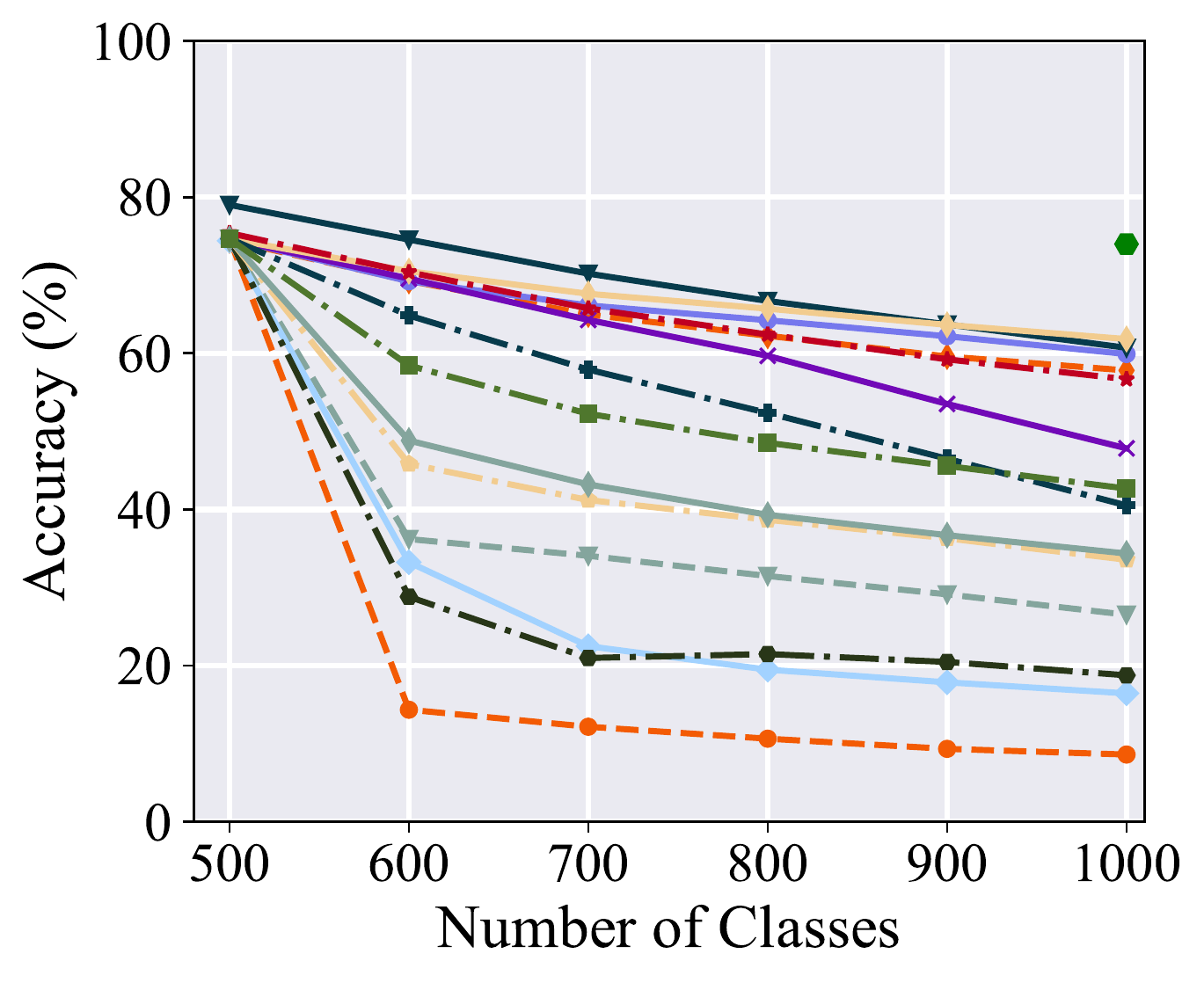}
		}
	\end{center}
	\vspace{-5mm}
	\caption{ Incremental performance of different methods on CIFAR100 (a-d), ImageNet100 (e-f), and ImageNet1000 (g-h). Legends are shown at the top of this figure, and we report the results of more settings in the supplementary.
	} \label{figure:imagenet100_top1}
	\vspace{-5mm}
\end{figure*}

\subsection{Experimental Settings}

\subsubsection{Benchmark Datasets}

iCaRL~\cite{rebuffi2017icarl} firstly formulates the comparison protocol of class-incremental learning, which was widely followed and compared in other works. It suggests using CIFAR100~\cite{krizhevsky2009learning} and ImageNet100/1000~\cite{russakovsky2015imagenet} for evaluation. 	 \bfname{CIFAR100} contains 100 classes with 60,000 images, in which 50,000 are training instances, and 10,000 are testing ones, with 100 images per class. Each image is represented by 32$\times$32 pixels. 
\bfname{ImageNet1000}  is a large-scale dataset with 1,000 classes, with about 1.28 million images for training and 50,000 for validation. 
\bfname{ImageNet100}  is the subset of ImageNet1000 containing 100 classes~\cite{rebuffi2017icarl}. These classes are selected from the first 100 classes after a random shuffle.
Some works~\cite{kirkpatrick2017overcoming,wang2022s,ahn2021ss,tao2020few} utilize
MNIST~\cite{lecun2010mnist}, CUB200~\cite{WahCUB2002011} and {\it mini}ImageNet~\cite{russakovsky2015imagenet} for evaluation, while the aforementioned datasets are the most widely adopted in the current CIL community, and we choose them for model evaluation.

\begin{table}[t]
	\caption{ Average and last accuracy performance comparison on CIFAR100.  `\#P' represents the number of parameters (million).
	}\label{tab:average_acc_cifar}
	\vspace{-2mm}
	\centering
		\begin{tabular}{@{}lccccccccc ccc}
			\toprule
			\multicolumn{1}{c}{\multirow{2}{*}{Method}} & 
			\multicolumn{3}{c}{Base0 Inc5} & \multicolumn{3}{c}{Base0 Inc10}  \\
			& {\#P} & {$\bar{\mathcal{A}}$} & ${\mathcal{A}_B}$ & {\#P} & {$\bar{\mathcal{A}}$} & ${\mathcal{A}_B}$ 
			\\
			\midrule
			Finetune   & 0.46 &17.59 &4.83 & 0.46 &26.25 & 9.09  \\
			EWC    & 0.46 & 18.42 & 5.58 & 0.46 &29.73 & 12.44 \\
			LwF   & 0.46 &30.93 & 12.60 & 0.46 &43.56 & 23.25  \\
			\midrule
			GEM    & 0.46 & 31.73 & 19.48 & 0.46 & 40.18 & 23.03  \\
			Replay & 0.46 & 58.20 & 38.69 & 0.46 & 59.31 & 41.01 \\
			RMM   & 0.46 &65.72 & 51.10 & 0.46 &68.54 & 56.64  \\
			iCaRL    & 0.46 & 63.51 & 45.12 & 0.46 &64.42 & 49.52  \\
			PODNet   & 0.46 & 47.88 & 27.99 & 0.46 & 55.22 & 36.78  \\
			Coil    & 0.46 & 57.68 & 34.33 & 0.46 &60.27 & 39.85 \\
			WA & 0.46 &64.65 & 48.46 & 0.46 & 67.09 & 52.30  \\
			BiC   & 0.46 &62.38 & 43.08 & 0.46 & 65.08 & 50.79 \\
			FOSTER   & 0.46 &63.38 & 49.42 & 0.46 & 66.49 & 53.21  \\
			AANets & 0.99 & 59.34 & 42.42 & 0.99 &61.73 &45.53\\
			DER    & 9.27 &  67.99 & 53.95 & 4.60 & 69.74 & \bf 58.59 \\
			MEMO    & 7.14 &\bf 68.10 & \bf 54.23 & 3.62 & \bf 70.20 & 58.49 \\
			\midrule
			DyTox & 10.7 & 68.06 & 52.23 & 10.7 & 71.07 & 58.72 \\
			L2P & 85.7 & 84.00 & 78.96 & 85.7 & 89.35 & 83.39 \\
			\bottomrule
		\end{tabular}
	\vspace{-4mm}
\end{table}

\noindent\textbf{Dataset Split}: Following the protocol defined in~\cite{rebuffi2017icarl}, all classes are first shuffled by Numpy random seed $1993$. Afterward, there are two different ways to split the classes into incremental stages:
$\bullet$ \bfname{Train from scratch (TFS)}: splits all classes equally into each incremental stage. For example, if there are $B$ stages and $C$ classes in total, each incremental task contains ${C}/{B}$ classes for training.\\
$\bullet$ {  \bfname{Train from half (TFH)}: splits half of the total classes as the first incremental task and equally assigns the rest classes into the following stages. Specifically, it assigns $C/2$ classes to the first task and $C/2(B-1)$ classes to the rest tasks.

Both of these settings are widely adopted in the current CIL community~\cite{hou2019learning,wu2019large}. Hence, we unify these settings as `Base-$m$, Inc-$n$', where $m$ stands for the number of classes in the first stage, and $n$ stands for the number of classes in each incremental task. $m=0$ stands for the TFS protocol.
We use the \emph{same} training splits for every compared method for a fair comparison. 
The testing set is the same as the original one for holistic evaluation.

\subsubsection{Evaluation Metrics} 
There are several metrics to evaluate the CIL model. We denote the Top-1 accuracy after the $b$-th task as $\mathcal{A}_b$, and higher $\mathcal{A}_b$ indicates a better prediction accuracy. Since the CIL model is continually updated, the accuracy often decays with more tasks incorporated. Hence, the accuracy after the last stage ($\mathcal{A}_B$) is a proper metric for measuring the overall accuracy among all classes. 

However, only comparing the final accuracy ignores the performance evolution along the learning trajectory. Hence, another metric denoted as `average accuracy' considers the performance of all incremental stages: $\bar{\mathcal{A}}=\frac{1}{B}\sum_{b=1}^B\mathcal{A}_b$. A higher average accuracy denotes a stronger performance along the incremental stages.
Apart from these measures, we also consider forgetting and intransigence~\cite{chaudhry2018riemannian} measures in the supplementary.

\begin{figure*}[t]
	\vspace{-3mm}
	\begin{center}
		\subfigure[CIFAR100 Base0 Inc5 ]
		{	\includegraphics[width=.48\columnwidth]{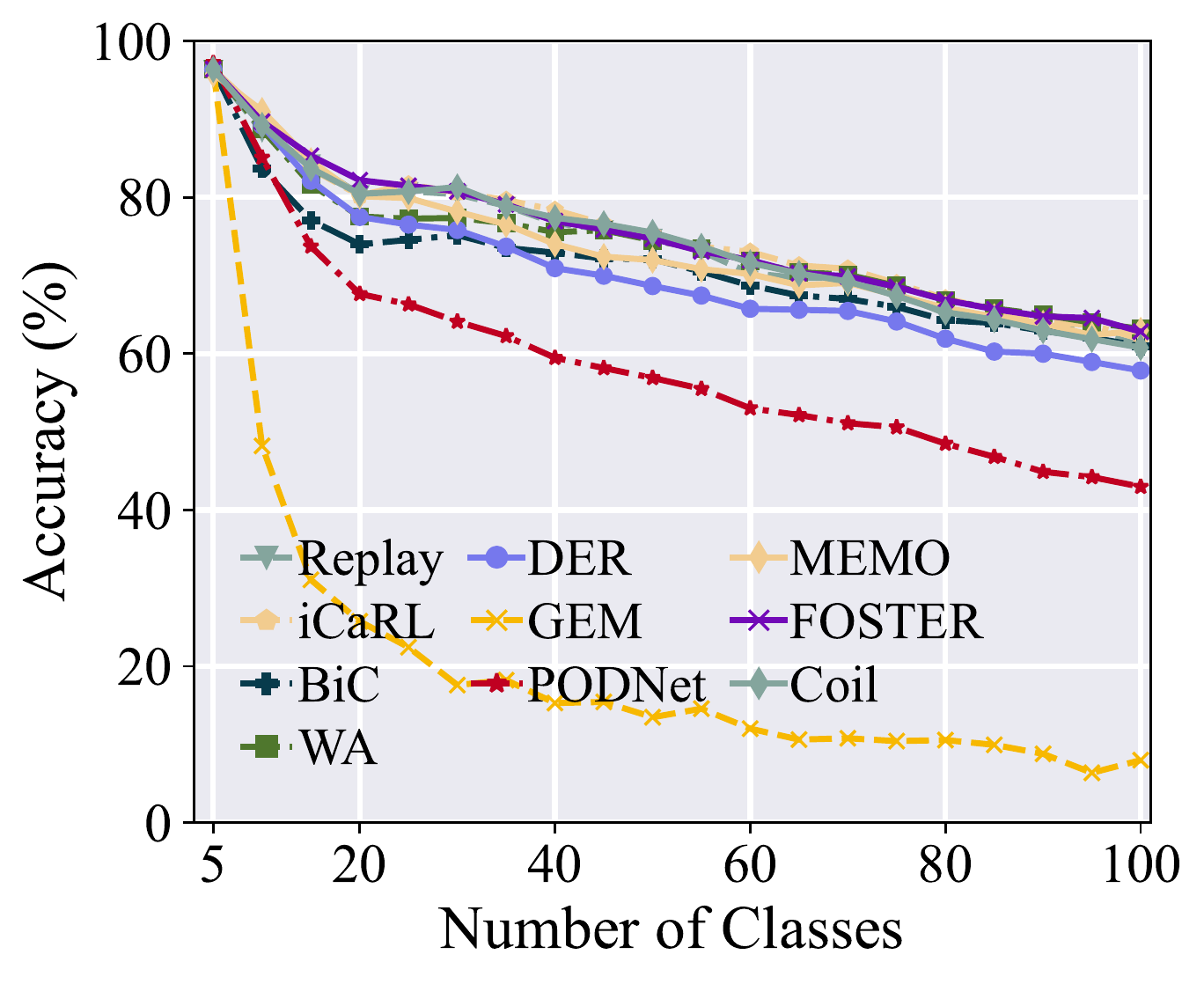}
		}
		\hfill
		\subfigure[CIFAR100 Base0 Inc10]
		{	\includegraphics[width=.48\columnwidth]{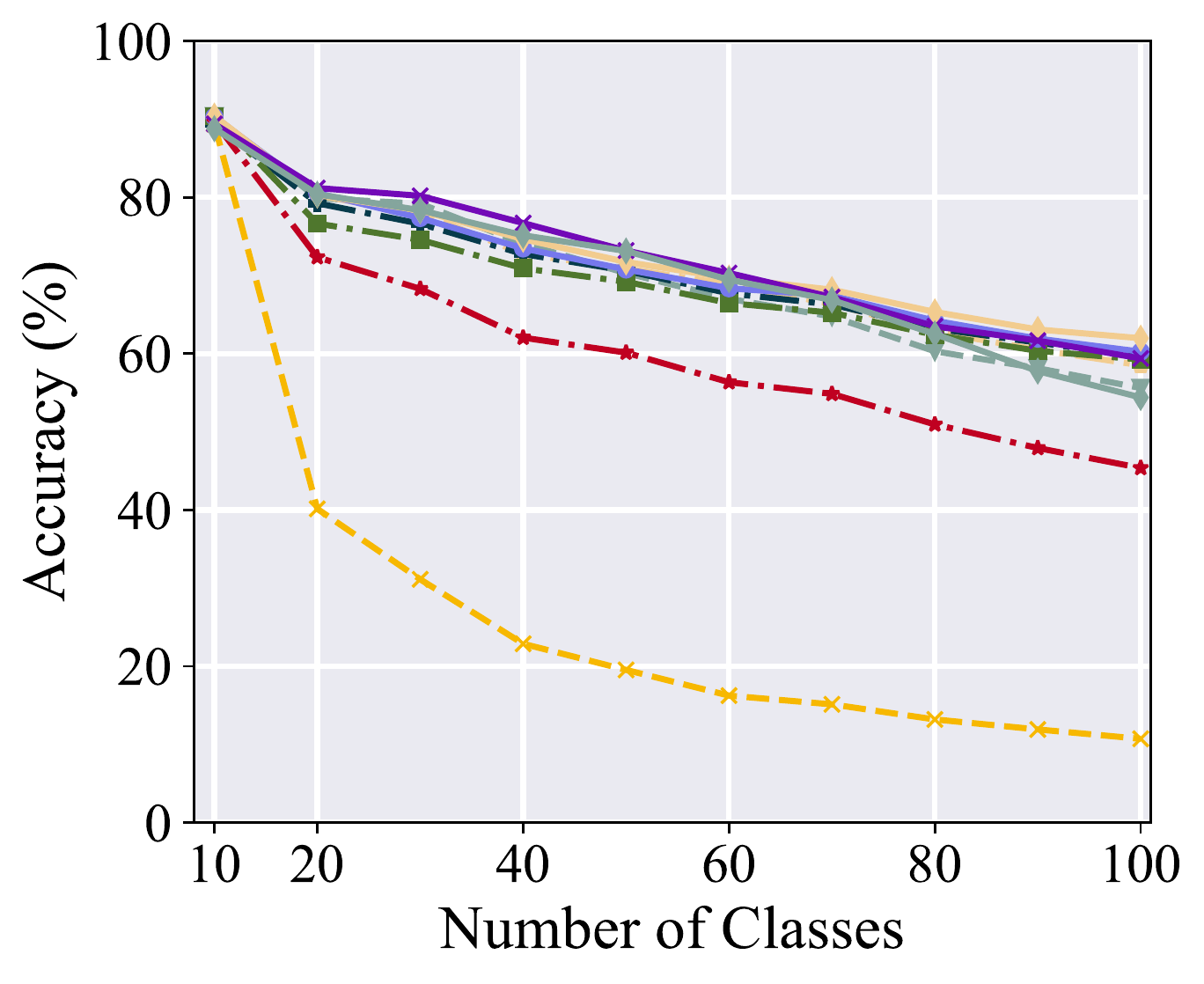}		
			\label{figure:fair_comparisonb}
		}
		\hfill
		\subfigure[ImageNet100 Base50 Inc5]
		{	\includegraphics[width=.48\columnwidth]{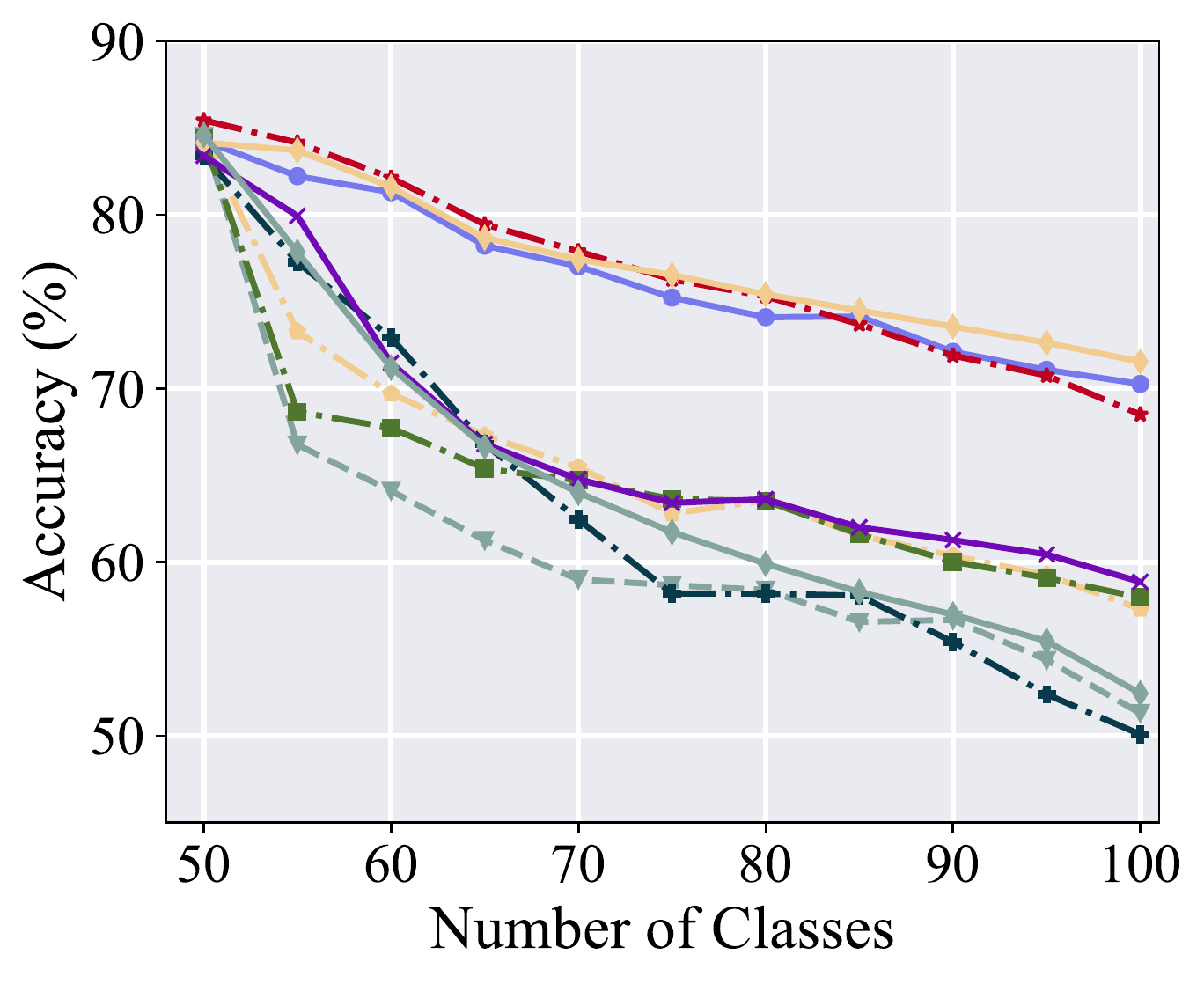}
		}
		\hfill
		\subfigure[ImageNet100 Base0 Inc10]
		{	\includegraphics[width=.48\columnwidth]{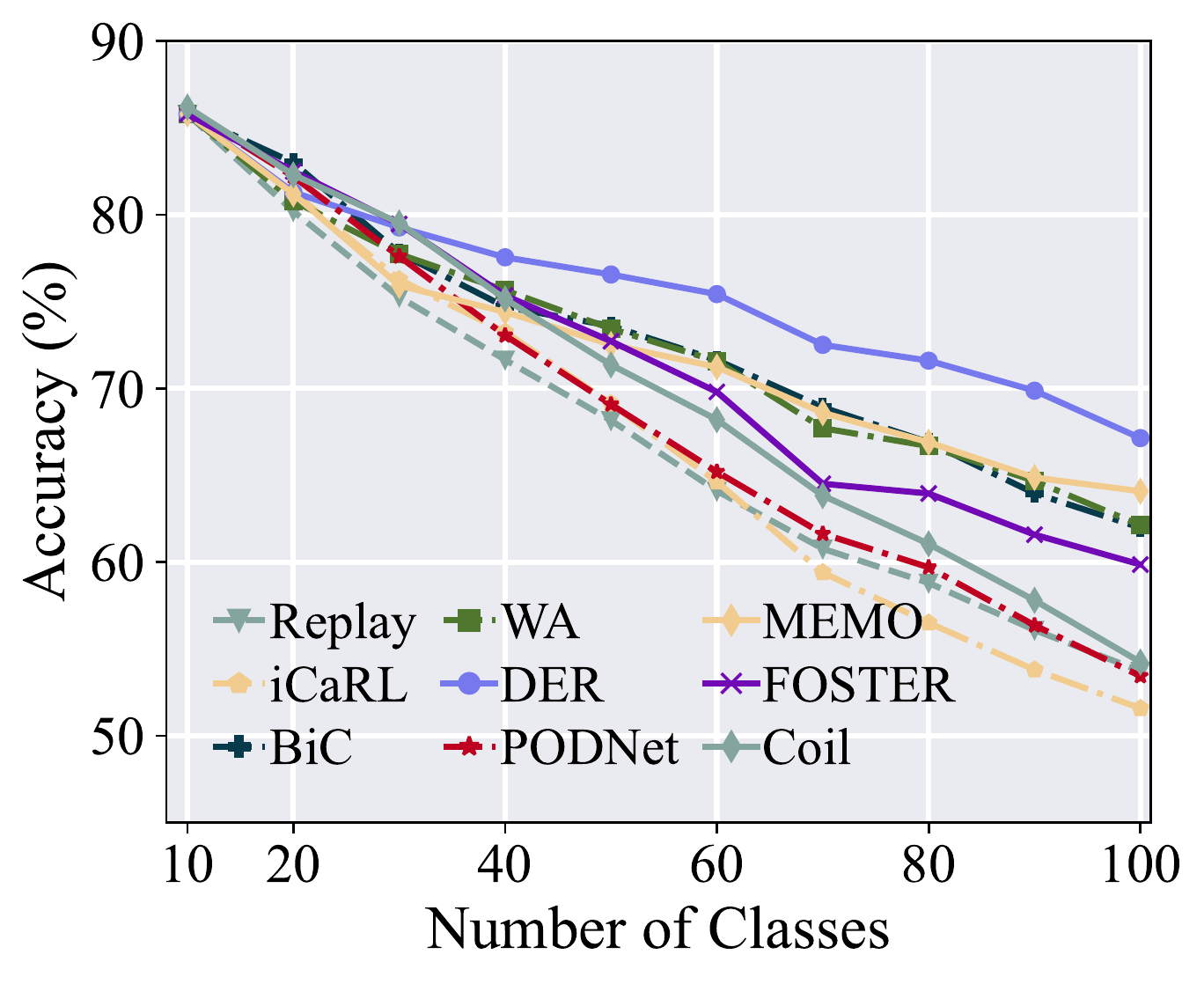}
		}
	\end{center}
	\vspace{-5mm}
	\caption{ Incremental accuracy of different methods with {\bf aligned memory cost}. Legends are shown in (a) and (d). 
	} \label{figure:fair_comparison}
	\vspace{-5mm}
\end{figure*}

\subsubsection{Implementation Details}

\noindent {\bf Selected methods:} In the comparison, we aim to contain all kinds of methods in Table~\ref{table:taxonomy}. We systematically choose 17 methods, including: Replay~\cite{ratcliff1990connectionist}, RMM~\cite{liu2021rmm} ({\it data replay}), GEM~\cite{lopez2017gradient} ({\it data regularization}), EWC~\cite{kirkpatrick2017overcoming} ({\it parameter regularization}), AANets~\cite{liu2021adaptive}, FOSTER~\cite{wang2022foster}, MEMO~\cite{zhou2022model}, DER~\cite{yan2021dynamically}, DyTox~\cite{douillard2022dytox}, L2P~\cite{wang2022learning} ({\it dynamic networks}), 
LwF~\cite{li2016learning}, iCaRL~\cite{rebuffi2017icarl}, PODNET~\cite{douillard2020podnet}, Coil~\cite{zhou2021co} ({\it knowledge distillation}),
 WA~\cite{zhao2020maintaining}, BiC~\cite{wu2019large} ({\it model rectify}).
 We also report the baseline method `Finetune,' which updates the model with Eq.~\ref{eq:finetune}. 

 The choice of these methods follows the development timeline of  CIL, which is also the way we introduce these works.  Additionally, it not only contains all seven aspects of CIL algorithms in our taxonomy but also includes early works (\eg, EWC, LwF, iCaRL) and recent state-of-the-art (DER, MEMO, L2P). The choice also gives consideration to CNN-based methods,  ViT-based methods (DyTox), and even pre-trained ViT-based methods (L2P). 
 Specifically, RMM is a specific technique to efficiently organize the memory budget, which can be orthogonally combined with other methods, and we combine it with FOSTER, denoted as FOSTER+RMM. Similarly, we combine AANets with LUCIR~\cite{hou2019learning}, denoted as LUCIR+AANets.
 L2P requires pre-trained ViT as the backbone model, while others are trained from scratch.

\noindent {\bf Training details:} We implement the above methods with PyTorch~\cite{paszke2019pytorch} and PyCIL~\cite{zhou2021pycil}. 
Specifically, we use the \emph{same} network backbone for \emph{all} CNN-based compared methods, \ie, ResNet32~\cite{he2015residual} for CIFAR100 and ResNet18 for ImageNet. 
  We use SGD with an initial learning rate of 0.1 and momentum of 0.9. The training epoch is set to 170 for all datasets with a batch size of 128. The learning rate suffers a decay of $0.1$ at 80 and 120 epochs. 
  For ViT-based methods like DyTox and L2P, we follow the original implementation and use ConViT~\cite{d2021convit} for DyTox and pre-trained ViT-B/16~\cite{dosovitskiy2020image} for L2P. The optimization parameters of them are set according to the original paper since ViT has a different optimization preference to CNN.
  We follow the original paper to set the algorithm-specific parameters, \eg, splitting 10\% exemplars from the exemplar set as validation for BiC, setting the temperature $\tau$ to 5 and using a 10 epochs warm-up for DER, using $\ell_2$ norm to normalize the fully-connected layers in WA. For EWC, the $\lambda$ parameter is done via a grid search among $\{1,10^1,10^2,10^3,10^4\}$, and we find $10^3$ leads to its best performance.
  
  Research has shown that a good starting point (\ie, the first stage accuracy $\mathcal{A}_1$) implies better transferability and will suffer less forgetting~\cite{castro2018end}.
  It must be noted that the performance gap of the first stage should be {\em eliminated} so as not to affect the forgetting evaluation. Hence, to make the methods share the same starting point, we utilize the same training strategy, data augmentation, and hyperparameters. We use basic data augmentation, \eg, random crop, horizontal flip, and color jitter for CIFAR100 and ImageNet.

  It must be noted that Finetune, EWC, LwF, and L2P are exemplar-free methods, and we do not use any exemplar set for them. For other methods, we follow the benchmark setting to set the number of exemplars to 2,000 for CIFAR100 and ImageNet100 and 20,000 for ImageNet1000.
  These exemplars are equally sampled from each seen class via the herding~\cite{welling2009herding} algorithm in Definition~\ref{def:exemplar_set}.

\subsection{Comparison on Small-Scale Datasets} \label{sec:cifar_exp}

This section includes the comparison on the small-scale dataset, \ie, CIFAR100. We compare these methods under TFS and TFH settings with four data splits and report the incremental performance in Figure~\ref{figure:imagenet100_top1} (top). We summarize the average and final performance, the number of parameters in Table~\ref{tab:average_acc_cifar}.

As we can infer from these figures, finetune shows the worst performance among all settings, verifying the fact that the model will suffer forgetting when sequentially learning new concepts. Regularizing the parameters, \ie, EWC, shows a negligible improvement over finetune. By contrast, LwF adds knowledge distillation loss to resist forgetting, which substantially improves the performance. When the exemplar set is available, directly replaying them during model updating can further enhance the performance by a substantial margin. Coil and iCaRL combine the exemplar replay and knowledge distillation and further obtain a performance boost than the vanilla replay. Model rectify methods, \ie, BiC and WA, rectify the bias in iCaRL and further improve the accuracy. However, recent methods based on dynamic networks (\ie, DER, FOSTER, MEMO, and DyTox) show competitive results with the help of multiple backbones. It indicates that saving more backbones can substantially help the model overcome forgetting. 
When it comes to pre-trained models, L2P obtains the best performance among all methods. However, other methods are trained from scratch, while L2P relies on the ViT pre-trained on ImageNet-21K, making it unfair to directly compare these two lines of methods.

On the other hand, we can infer from different settings that TFH requires more {\em stability} than TFS. Since the evaluation is based on the accuracy among all classes, remembering old classes becomes more critical when there is a large group of base classes.

\subsection{Comparison on Large-Scale Datasets}\label{sec:imagenet_exp}

In this section, we evaluate different methods on the large-scale dataset, \ie, ImageNet100 and ImageNet1000. We compare these methods under the TFS and TFH settings and report the incremental performance (top-1 accuracy) of different methods in Figure~\ref{figure:imagenet100_top1} (bottom). 
We report the top-5 accuracy and summarize the average and final performance, the number of parameters in the supplementary. Since GEM requires saving a large-scale matrix for solving the QP problem, it cannot be conducted with the ImageNet dataset. On the other hand, since the ViT in L2P is pre-trained on ImageNet-21K, incrementally training it on ImageNet is meaningless, and we do not report its results.

As we can infer from these figures, most methods share the same trend as CIFAR100. Replay acts as a strong baseline in both small-scale and large-scale inputs, verifying the effectiveness of exemplars in incremental learning. Dynamic networks consistently show the best performance among all settings, outperforming other methods by a substantial margin in the benchmark comparison.

\begin{figure}[t]
	\begin{center}
		\subfigure[Memory size of Figure~\ref{figure:cifar100b}]
		{	\includegraphics[width=.45\columnwidth]{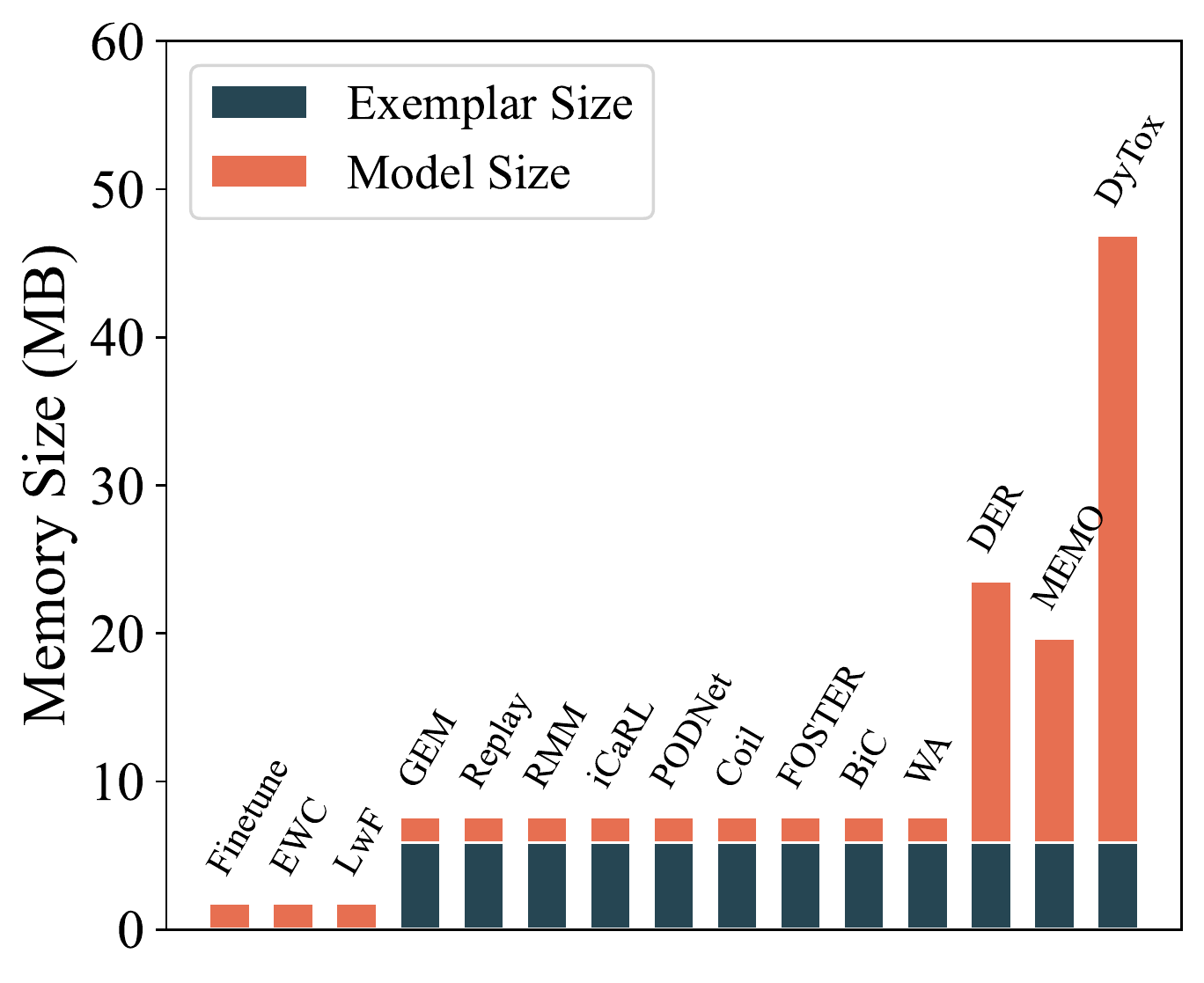}
			\label{fig:memory_budgeta}
		}
		\subfigure[Memory size of Figure~\ref{figure:fair_comparisonb} ]
		{	\includegraphics[width=.45\columnwidth]{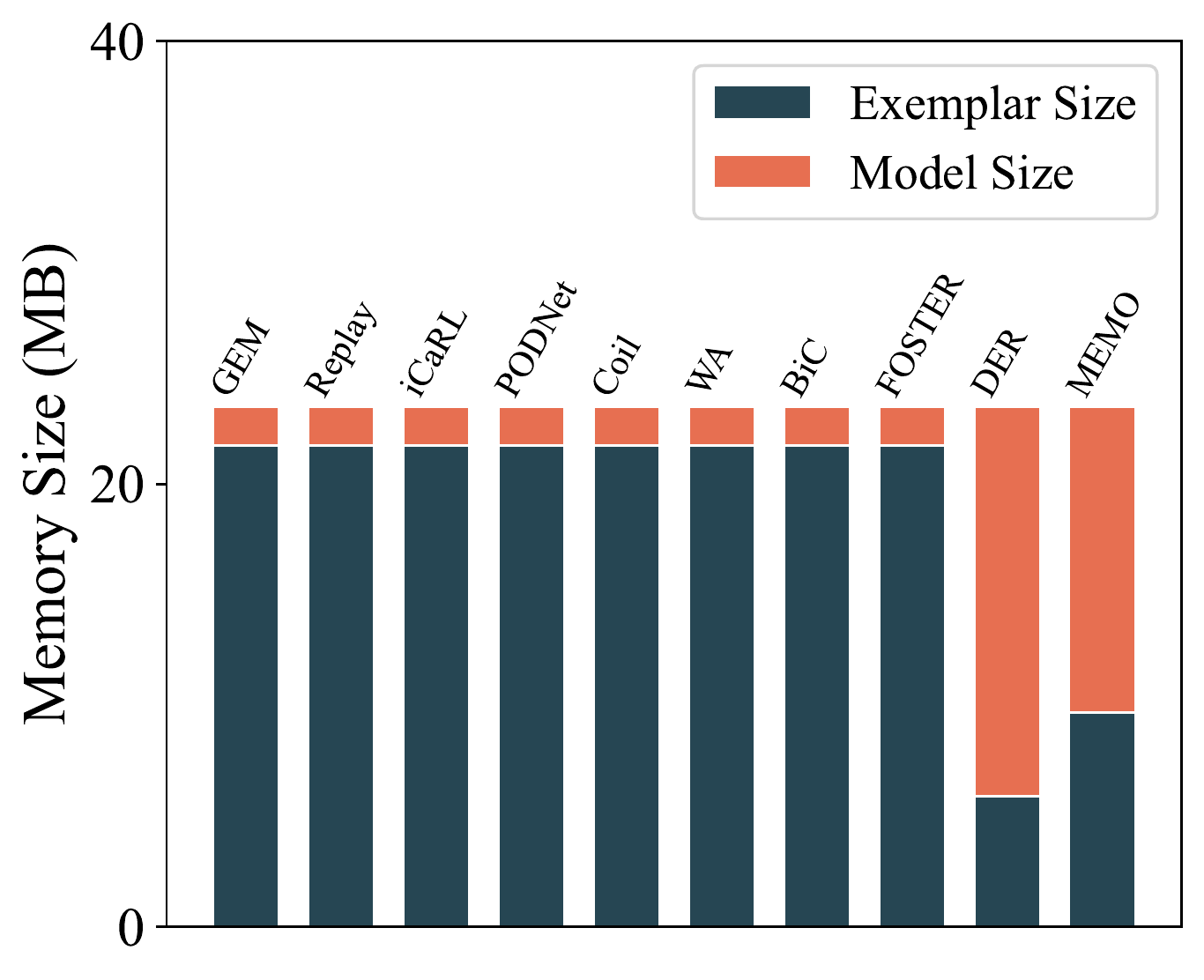}		
			\label{fig:memory_budgetb}
		}
	\end{center}
	\vspace{-5mm}
	\caption{ Memory size of different comparison protocols. Dark bars denote the budget for exemplars, and red bars represent the budget for keeping the model. Different methods should be aligned to the same budget for a fair comparison, as shown in (b).
	} 
	\vspace{-1mm}
	\label{figure:memory_budget}
\end{figure}

\begin{table}[t]
	\caption{ Performance comparison on CIFAR100 with aligned memory cost.
		`\#P' represents the number of parameters (million). 
		`\#$\mathcal{E}$' denotes the number of exemplars, and `MS' denotes the memory size (MB). 
	}\label{tab:fair_comparison}
	\vspace{-3mm}
	\centering
	\begin{tabular}{@{}lccccccccccccccc}
		\toprule
		\multicolumn{1}{c}{\multirow{2}{*}{Method}} & 
		\multicolumn{5}{c}{CIFAR100 Base0 Inc10}  \\
		& {\#P} & {\#{$\mathcal{E}$}} & MS& {$\bar{\mathcal{A}}$} & ${\mathcal{A}_B}$ 
		\\
		\midrule
		GEM &0.46 & 7431 &23.5 & 27.03 & 10.72   \\
		Replay &0.46 & 7431 &23.5 & 69.97 & 55.61\\
		iCaRL  &0.46 & 7431 &23.5 & 70.94 &58.52 \\
		PODNet &0.46 & 7431 &23.5 & 60.80 & 45.38\\
		Coil  &0.46 & 7431 &23.5 & 70.69 & 54.40\\
		WA &0.46 & 7431 &23.5 & 69.55 & 59.26\\
		BiC&0.46 & 7431 &23.5 &70.69& 59.60\\
		FOSTER  &0.46 & 7431 &23.5 &72.28 & 59.39\\
		DER &4.60 & 2000 &23.5&71.47&60.26 \\
		MEMO &3.62 & 3312 &23.5 & \bf 72.37& \bf 61.98\\
		\bottomrule
	\end{tabular}
	\vspace{-5mm}
\end{table}

\subsection{Memory-Aligned Comparison}\label{sec:fair_exp}

Former experimental evaluations indicate dynamic networks
show the best performance among all methods. However, are these methods {\em fairly} compared? We argue that these dynamic networks implicitly introduce an extra memory budget, namely {\it model buffer} for keeping old models. The additional
buffer results in an unfair comparison to those methods without storing models. Taking Table~\ref{tab:average_acc_cifar} for an example, the number of parameters in DER is ten times that of iCaRL in the CIFAR100 Base0 Inc10 setting. We visualize the memory cost of different methods in Figure~\ref{fig:memory_budgeta} and find that dynamic networks obtain better performance at the expense of more memory budgets. It makes directly comparing different kinds of methods {\em unfair}.

In this section, we follow~\cite{zhou2022model} for a fair comparison among different methods with different memory budgets. For those
methods with different memory costs, we need to {\em align} the performance measure at the same memory scale for a fair comparison. For example, a ResNet32 model costs $463,504$ parameters (float), while a CIFAR image requires $3\times32\times32$ integer numbers (int). Hence, the budget for saving a backbone is equal to saving 
$463,504$ floats $\times4$ bytes/float $\div (3\times32\times32)$ bytes/image $\approx 603$ instances for CIFAR. A fair comparison between different methods can be made by equipping the other methods with more exemplars.
Since DER requires saving all the backbones from history, we align the memory cost of other methods to DER with extra exemplars.
 We visualize the memory budget of different methods under the current protocol in Figure~\ref{fig:memory_budgetb}, where the total budgets of different methods are aligned to the same scale.

We report the results under the current protocol in Figure~\ref{figure:fair_comparison} and the detailed performance in Table~\ref{tab:fair_comparison}.
Since Finetune, LwF, and EWC cannot be combined with the exemplar set, we do not report the results of these methods. Similarly, DyTox and L2P utilize different kinds of backbones, and we do not report their results.
As we can infer from these results, the gap between dynamic networks and other methods is no longer large under fair comparison. For example, DER outperforms iCaRL by $9.07$\% in terms of the final accuracy in the benchmark comparison of CIFAR100 Base0 Inc10, while the gap decreases to $1.74$\% under the current protocol.

\begin{table}[t]
	\caption{ Memory-agnostic performance measures for CIL. AUC depicts the dynamic ability with the change of memory size. }\label{tab:auc}
	\centering
	\begin{tabular}{@{}lccccccccccccccc}
		\toprule
		\multicolumn{1}{c}{\multirow{2}{*}{Method}} & 
		\multicolumn{2}{c}{CIFAR100 Base0 Inc10}   & 
		\multicolumn{2}{c}{ImageNet100 Base50 Inc5} \\
		& AUC-A & AUC-L & AUC-A & AUC-L
		\\
		\midrule
		GEM    &4.31  & 1.70& - & -\\
		Replay & 10.49  & 8.02& 553.6&470.1\\
		iCaRL    & 10.81  & 8.64& 607.1 & 527.5 \\
		PODNet   &9.42& 6.80 & 701.8 & 624.9\\
		Coil    &10.60 &7.82& 601.9 &486.5 \\
		WA & 10.80  &8.92& 666.0 & 581.7 \\
		BiC  & 10.73&8.30 & 592.7 &474.2\\
		FOSTER   &\bf 11.12 &\bf 9.03 & 638.7 & 566.3	\\
		DER    & 10.74  & 8.95& 699.0 &639.1\\
		MEMO    & 10.85  &\bf 9.03& \bf 713.0 & \bf 654.6\\
		\bottomrule
	\end{tabular}
	\vspace{-3mm}
\end{table}

\begin{figure}[t]
	\begin{center}
		\subfigure[CIFAR100 Base0 Inc10 ]
		{	\includegraphics[width=.473\columnwidth]{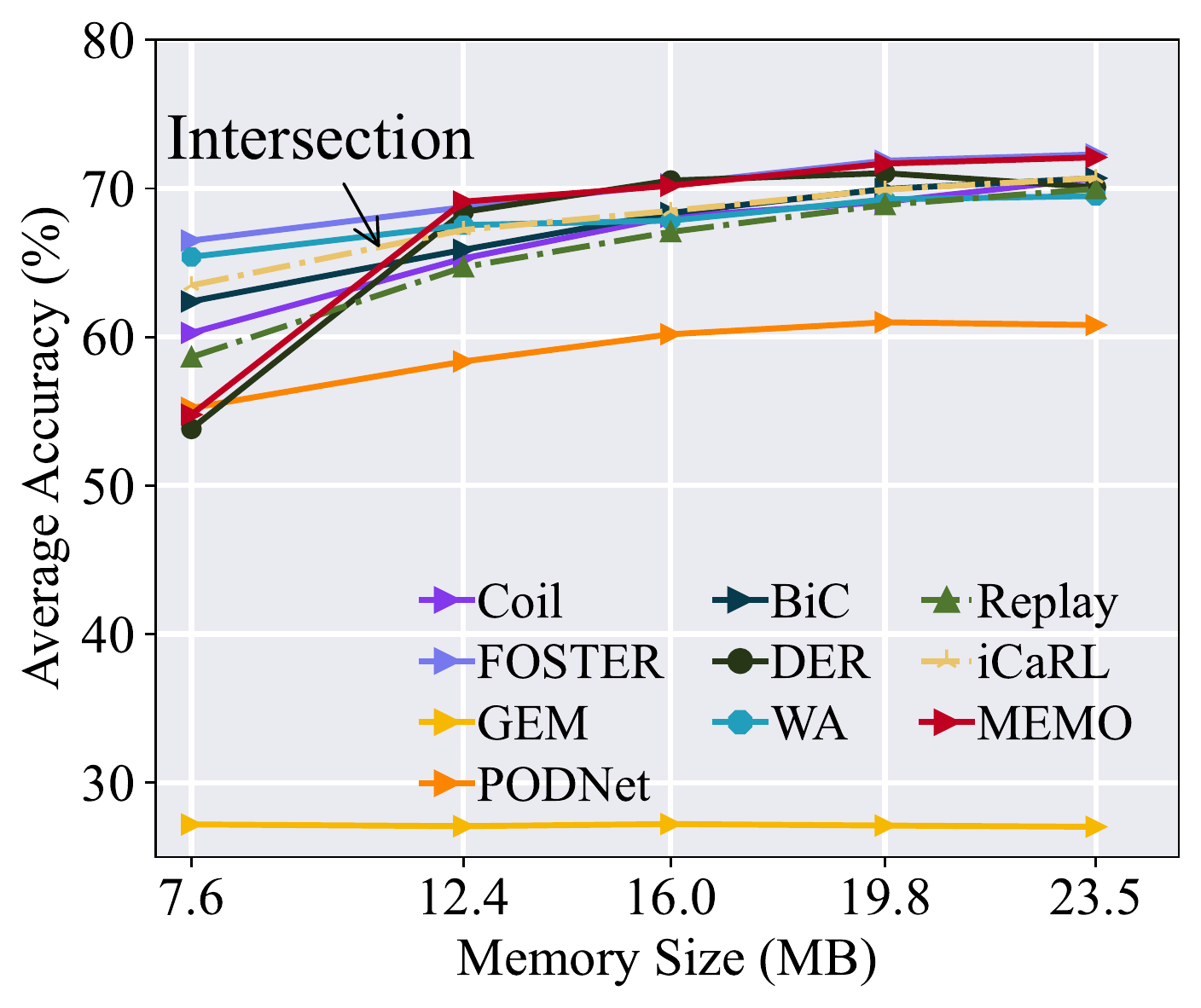}
		}
		\hfill
		\subfigure[ImageNet100 Base50 Inc5]
		{	\includegraphics[width=.473\columnwidth]{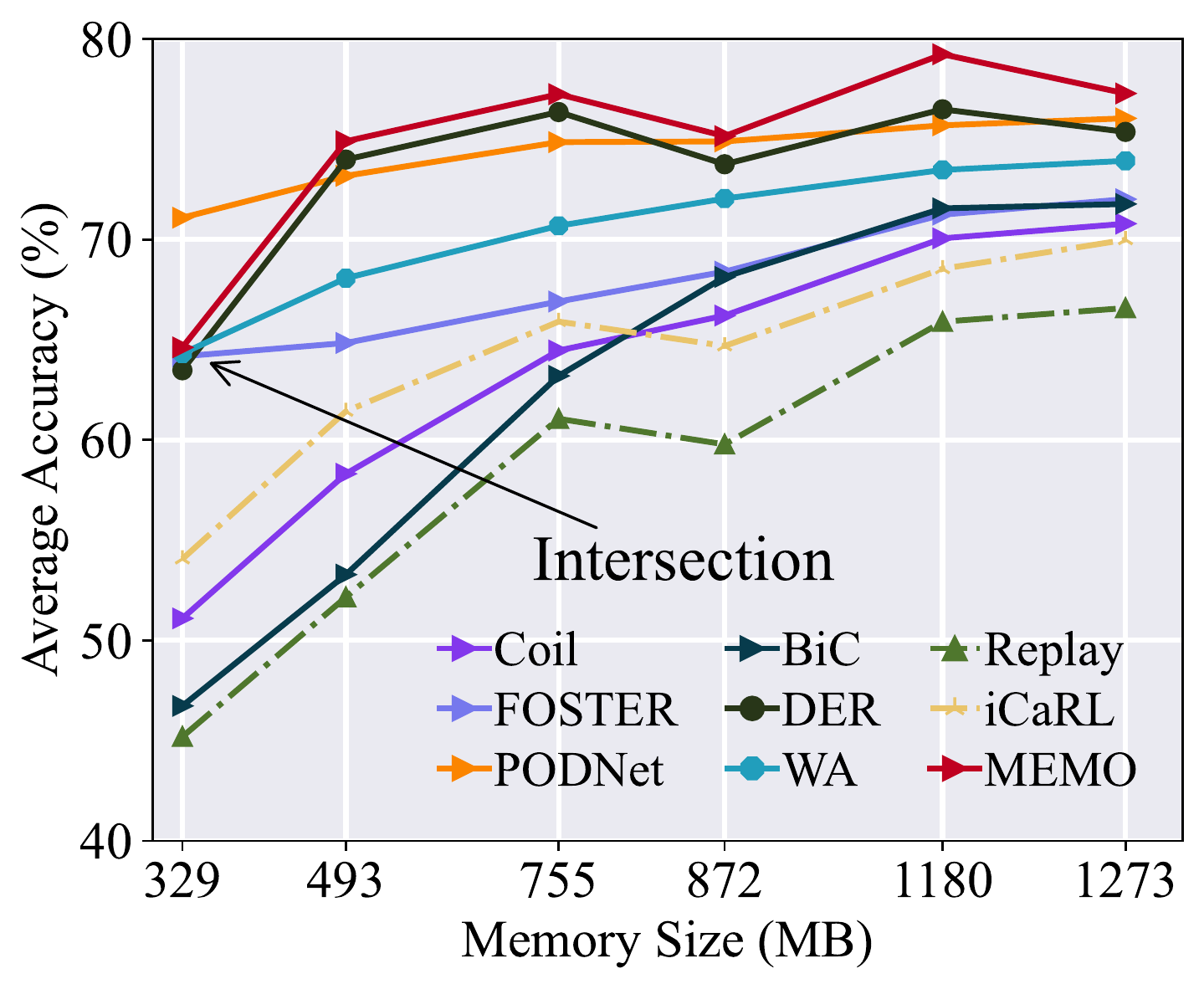}		
		}
		\vspace{-3mm}
	\end{center}
	\caption{ Average performance-memory curve of different methods with different datasets. Dynamic networks perform better with large budgets, while other methods dominate small ones.
	} \label{figure:auc_measure}
	\vspace{-5mm}
\end{figure}

\subsection{Memory-Agnostic Measure}\label{sec:auc_exp}

Section~\ref{sec:fair_exp} enables fair comparison among different methods by aligning the memory cost. However, the comparison is made by aligning the memory cost of other methods to DER. In contrast, open-world applications require conducting class-incremental learning in various scenarios, \ie, high-performance computers and edge devices are both essential. Hence, it requires a {\em memory-agnostic measure} for CIL to measure the model's extendability given {\em any} memory budget.

To this end, we can set several `comparison budgets' and align the memory cost of different methods to them. The budget list starts from a small value and incrementally enlarges, containing the requirement of different scale models. In this setting, we set the budget of the start point to a single backbone and gradually increase it to the budget cost of DER. 
For those algorithms without extra model storage, we equip more exemplars to meet the selected budget. However, when the selected budget is smaller than the required size, dynamic networks (\ie, DER and MEMO) cannot be deployed with the benchmark backbone. Therefore, we choose smaller backbones with fewer parameters for alignment. Hence, we can measure the performance of different methods at different memory scales and draw the performance-memory curve, as shown in Figure~\ref{figure:auc_measure}. The X-coordinate corresponds to the memory cost, and the Y-coordinate indicates the average performance $\bar{\mathcal{A}}$ (or last performance ${\mathcal{A}_B}$). 
We suggest the area under the performance-memory curve
(AUC) since the curve of each method indicates the dynamic ability with the change of model size. We calculate {\bf AUC-A} and {\bf AUC-L}, standing for the AUC under the average performance-memory and last performance-memory curves.

As we can infer from Table~\ref{tab:auc}, there are two main conclusions. Firstly, there exists an {\em intersection} between dynamic networks and other methods, where other methods perform better given a small memory size. In contrast, dynamic networks perform better given a large memory size.
In other words, {\em there is no free lunch in CIL}, and different kinds of methods have their dominant domains. Secondly, the AUC measure provides a holistic way to measure the {\em extendability} of different methods given various memory budgets. 
FOSTER and MEMO show competitive results regarding the AUC-A/L measure, implying their stronger extendability.
It would be interesting to introduce it as the performance measure and design CIL methods for real-world applications. We report implementation details and the performance of each budget point in the supplementary.

\subsection{Discussions about the Comparison} \label{sec:comparison_results}

With the above experimental evaluations from three aspects, we have the following conclusions: {\bf 1)} Equipping the model with exemplars is a simple and effective way to resist forgetting in CIL models. 
		{\bf 2)} Knowledge distillation performs better than parameter regularization in resisting forgetting with the same cost.
		{\bf 3)} Model rectification can further boost the performance of other CIL models in a plug-and-play manner.
		{\bf 4)} Pre-trained models can ease the burden of incremental learning and show very strong performance. However, since the features of pre-trained models are already available and do not need to be incrementally learned, comparing pre-trained models to other methods may be unfair.
		{\bf 5)} Dynamic networks show the best performance in the evaluation at the cost of extra memory budgets. However, when changing the extra budgets into equal size of exemplars, the performance gap becomes smaller.
		{\bf 6)} AUC-A and AUC-L provide a way to evaluate different CIL methods in a {\em memory-agnostic} manner, which can help select the method with {\em extendability} given any budget.

\section{Future Directions}

In this section, we discuss the possible future directions of the class-incremental learning field.

\noindent\textbf{CIL with Complex Inputs}: In the real world, data is often with complex format, \eg, few-shot~\cite{tao2020few}, imbalanced~\cite{liu2022long}, weak supervised~\cite
{cha2021co2l,zhong2023weakly},
multi modal~\cite{radford2021learning,zhou2023learning}, concept drift~\cite{lu2018learning}, novel classes~\cite{zhou2021learning} etc. CIL methods should be able to handle these real-world scenarios for better generalizability. 
Specifically, \cite{tao2020few,zhou2022forward,zhou2022few,zhang2021few,zhuang2023gkeal} address the few-shot CIL problem, where the model is required to be adapted to incoming few-shot classes. \cite{kim2020imbalanced,liu2022long} propose to tackle the long-tailed CIL problem, where the head classes are easy to collect with adequate instances while tail classes are scarce. Recently, with the prosperity of CLIP~\cite{radford2021learning}, incrementally training the language vision models to handle multi-modal data streams~\cite{yan2022generative} is becoming popular. Since the labeling cost is always expensive in the real world, there are some works addressing training CIL models in a semi-supervised or unsupervised manner~\cite{cha2021co2l}. 
\cite{xie2022general} proposes a unified framework to handle CIL with concept drift. If the test dataset contains instances from unknown classes, open-set recognition~\cite{zhou2021learning} and novel class discovery~\cite{rios2022incdfm} can equip the model with the detection ability, which refers to the open-world recognition problem~\cite{bendale2015towards}.  
Lastly, real-world data may emerge hierarchically, and the labels could evolve from coarse-grained to fine-grained. CIL with refined concepts~\cite{abdelsalam2021iirc} is also vital to building real-world learning systems.

\noindent\textbf{CIL with General Data Stream}: Current CIL methods have a set of restrictions on the data stream, \eg, saving exemplars for rehearsal, undergoing multi-epoch offline training within an incremental task, etc. Future CIL algorithms should be able to conduct fully online training~\cite{guo2022online} without requiring the task boundaries~\cite{pourcel2022online}. On the other hand, saving exemplars from the history may violate user privacy in some cases. To this end, exemplar-free CIL~\cite{petit2022fetril,zhu2022self,zhu2021class,zhuang2022acil,Zhuang_DSAL_AAAI2024} should be conducted to enable the model to be adapted without the help of exemplars. Lastly, most CIL methods rely on the number of base classes to define hyper-parameters in model optimization, where more base classes require larger stability and fewer base classes require larger plasticity. Therefore, designing the algorithm to handle CIL problems given any base classes~\cite{liu2023Online} is also essential to the real world.

\noindent\textbf{CIL with Any Memory/Computational Budgets}: Dynamic networks have obtained impressive performance in recent years~\cite{wang2022foster}. However, most of them require an extra memory budget to save the external backbones. In the real world, a good CIL algorithm should handle different budget restrictions. For example, an algorithm should be able to learn with high-performance computers or with edge devices (e.g., smartphones), and both scenarios are essential. Hence, deploying and comparing CIL models in the real world should take the memory budgets into consideration. The performance measures of AUC-A/L~\cite{zhou2022model} are proper solutions to holistically compare different methods given any memory budgets. On the other hand, future CIL methods are also encouraged to handle specific learning scenarios, \eg, \cite{wang2022sparcl} addresses training CIL systems under resource-limited scenarios. 
Another important characteristic is the computational budget~\cite{prabhu2023computationally}. 
For realistic scenarios with high-throughput streams,  computational bottlenecks impose implicit constraints on learning from past samples that can be too many to be revisited during training. In that case, we can only update the model with limited iterations and cannot tackle all the data. Hence, designing computational-efficient algorithms in real-world applications remains a promising direction for relating CIL to realistic scenarios.

\noindent\textbf{CIL with Pre-trained Models}: Recently, pre-trained models have shown to work competitively with their strong transferability, especially for ViT-based methods~\cite{wang2022dualprompt,wang2022learning,seale2022coda,wang2022s}. The pre-trained models provide generalizable features for the downstream tasks, enabling the incremental model to adjust with minimal cost~\cite{jia2022visual}. CIL with pre-trained models is a proper way to handle real-world incremental applications from an excellent starting point. However, since the final target for incremental learning is to build a generalizable feature continually, some will argue that pre-trained models weaken the difficulty of incremental learning. From this perspective, designing proper algorithms to train the CIL model from scratch is more challenging.
On the other hand, pre-trained language-vision models~\cite{radford2021learning} have shown powerful generalizability in recent years, and exploring the ensemble of various pre-trained models is also an exciting topic.

\noindent\textbf{CIL with Bidirectional Compatibility}:
Compatibility is a design characteristic in software engineering community~\cite{gheorghioiu2003interprocedural,nagarakatte2009softbound,varma2009backward,xu2004efficient}, which was introduced to the machine learning community in~\cite{bansal2019updates,srivastava2020empirical}. Backward compatibility allows for interoperability with an older legacy system. In contrast, forward compatibility allows a system to accept input intended for a later version of itself. In the incremental learning field, compatibility is also a core problem, where the new model is required to understand the features produced by the old model to classify old classes.
Most methods aim to enhance backward compatibility by making the new model similar to the old one. However, FACT~\cite{zhou2022forward} addresses the forward compatibility in CIL, where the model should be prepared for future updates, acting like the pre-assigned interface. Furthermore, it will be interesting to address bidirectional compatibility~\cite{zhou2021co} in the future. 

\noindent\textbf{Analyzing the Reason Behind Catastrophic Forgetting}: Model rectification-based methods aim to reduce the inductive bias in the CIL model. Recently, more works have tried to analyze the reason for forgetting in CIL. \cite{pham2021continual} suggests that batch normalization layers are biased when sequentially trained, which triggers the different activation of old and new classes. \cite{zhu2021class} finds that representations with large eigenvalues transfer better and suffer less forgetting, and~\cite{guo2022online} shows that eigenvalues can be enlarged by maximizing the mutual information between old and new features. \cite{kim2022theoretical} theoretically decomposes the CIL problem into within-task and task-id predictions. It further proves that good within-task and task-id predictions are necessary and sufficient for good CIL performances.
\cite{caccia2022new} finds that applying
data replay causes the newly added classes’ representations to overlap significantly with
the previous classes, leading to highly disruptive parameter updates.  \cite{mirzadeh2020understanding} empirically shows that
the amount of forgetting correlates with the geometrical properties of the convergent points. It would be interesting to explore more reasons for catastrophic forgetting theoretically and empirically in the future.

\section{Conclusion}

Real-world applications often face streaming data, with which the model should be incrementally updated without catastrophic forgetting. In this paper, we provide a comprehensive survey about class-incremental learning by dividing them into seven categories taxonomically and chronologically.
Additionally, we provide a holistic comparison among different methods on several publicly available datasets. With these results, we discuss the insights and summarize the common rules to inspire future research. Finally, we highlight an important factor in CIL comparison, namely memory budget, and advocate evaluating different methods holistically by emphasizing the effect of memory budgets. We provide a comprehensive evaluation of different methods given specific budgets as well as some new performance measures. We expect this survey to provide an effective way to understand current state-of-the-art and speed up the development of the CIL field.

\ifCLASSOPTIONcompsoc
{
	\small
\bibliographystyle{unsrt}
\bibliography{paper}
}

\ifCLASSOPTIONcaptionsoff
  \newpage
\fi

\appendices


\begin{center}
	\textbf{\large Supplementary Material }
\end{center}

The supplementary material mainly contains additional experiments that cannot be reported due to page limit, which is organized as follows:
\begin{itemize}
	\item Section~\ref{sec:supp_benchmark} provides extra results in benchmark comparison, including detailed results on CIFAR100, ImageNet100, and Top-5 accuracy on ImageNet.
	\item Section~\ref{sec:supp_vis} provides visualization results on the confusion matrix and weight norm.
	\item Section~\ref{sec:supp_memory_aligned} provides extra results on other settings in the memory-aligned comparison.
	\item Section~\ref{sec:supp_AUC} reports the detailed implementations of the AUC-A/L measure, including the per-point setting of memory,  backbone, and incremental performance.
	\item Section~\ref{sec:supp_detailed_perf} reports the incremental learning performance (per-stage accuracy) of different methods in the benchmark comparison.
	\item Section~\ref{sec:supp_forgetting} reports the extra evaluation measures, \ie, forgetting and intransigence among all compared methods.
	\item Section~\ref{sec:supp_replay} conducts experiments on the type of replay strategies, including direct replay, feature replay, and generative replay.
	\item Section~\ref{sec:compared_methods} gives a detailed introduction about selected compared methods in the main paper.
	\item Section~\ref{sec:supp_detection_seg} discusses the taxonomy in the main paper and introduces the related topics about incremental object detection and semantic segmentation.
	\item Section~\ref{sec:supp:further} explores further analysis in class-incremental learning, including the influence of exemplars and computational efficiency.
\end{itemize}

\begin{figure*}[t]
	\begin{center}
		\subfigure[CIFAR100 Base0 Inc5 ]
		{	\includegraphics[width=.64\columnwidth]{pics/cifarb0inc5}
		}
		\hfill
		\subfigure[CIFAR100 Base0 Inc10]
		{	\includegraphics[width=.64\columnwidth]{pics/cifarb0inc10}		
		}
		\hfill
		\subfigure[CIFAR100 Base0 Inc20]
		{	\includegraphics[width=.64\columnwidth]{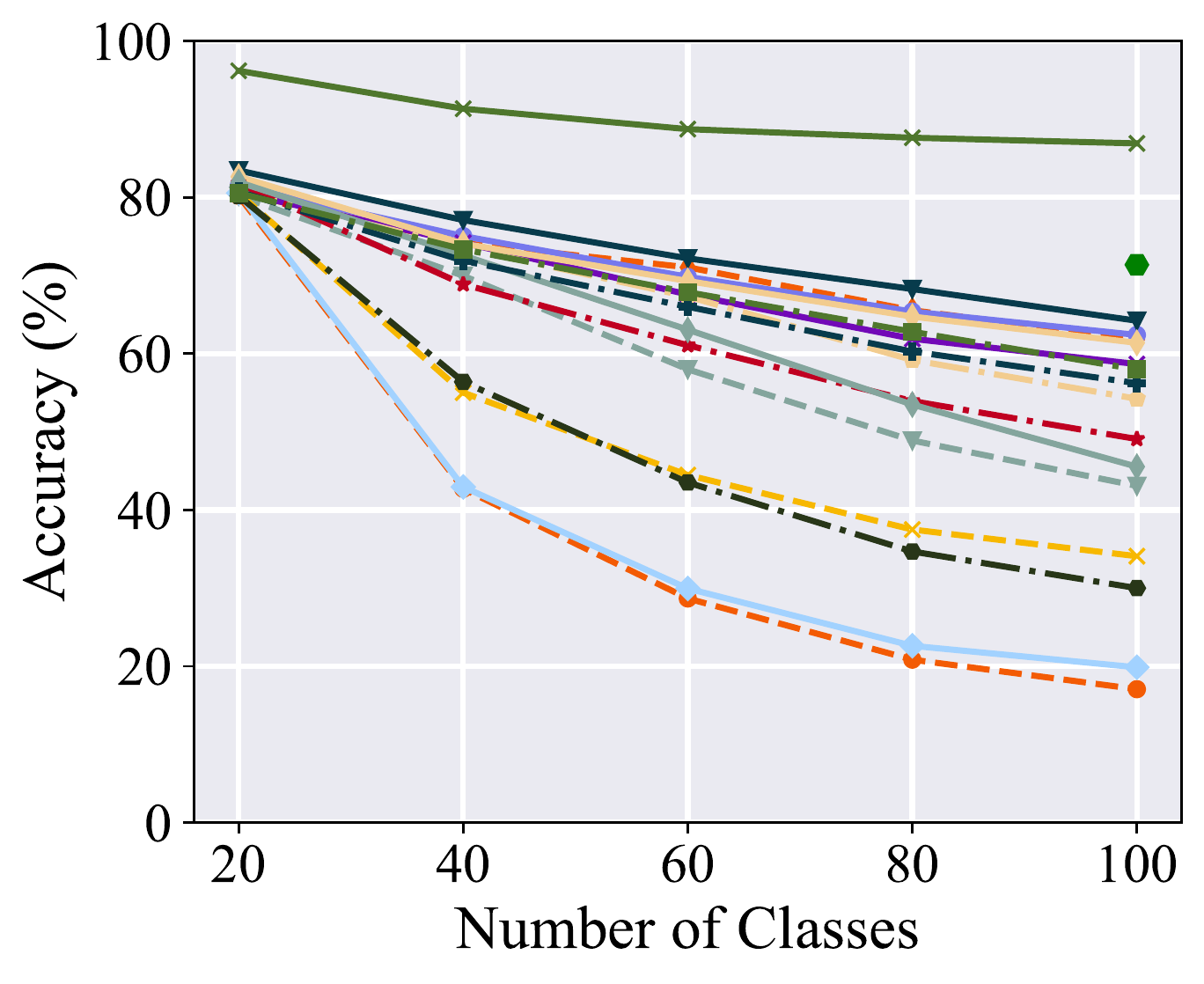}	
		}
		\\
		\includegraphics[width=1.9\columnwidth]{pics/cifar_legend}
		\subfigure[CIFAR100 Base50 Inc50]
		{	\includegraphics[width=.64\columnwidth]{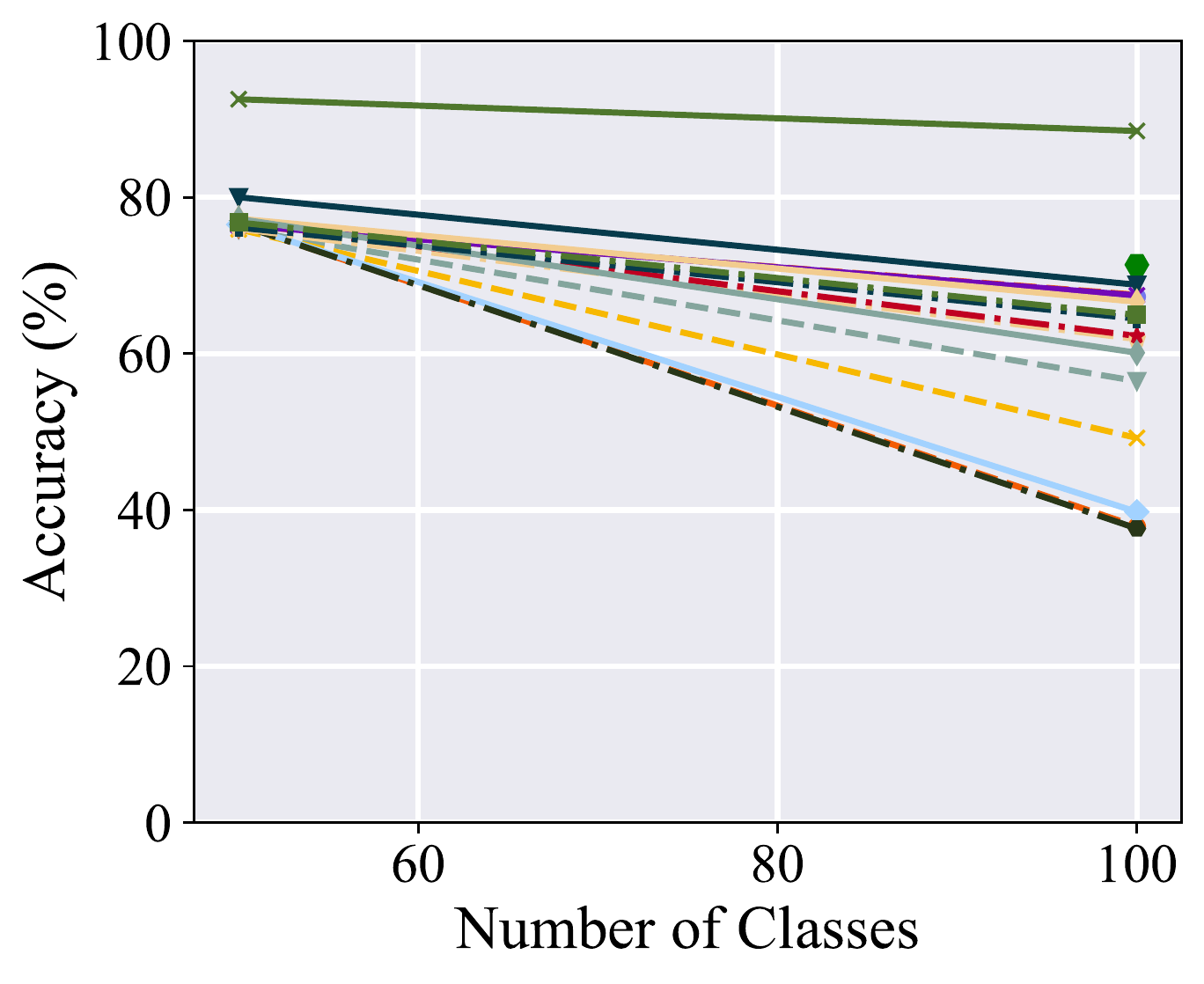}
		}
		\hfill
		\subfigure[CIFAR100 Base50 Inc10]
		{	\includegraphics[width=.64\columnwidth]{pics/cifarb50inc10}
		}
		\hfill
		\subfigure[CIFAR100 Base50 Inc25]
		{	\includegraphics[width=.64\columnwidth]{pics/cifarb50inc25}
		}
	\end{center}
	\vspace{-4mm}
	\caption{	Incremental accuracy of different methods on CIFAR100.
	} \label{figure:cifar100_full}
	\vspace{-4mm}
\end{figure*}

\section{Supplied Results in Benchmark Comparison} \label{sec:supp_benchmark}
In this section, we report the full results of the benchmark comparison. With a bit of redundancy, we list the incremental performance of all dataset splits. Specifically, we first report the detailed results of CIFAR100 and ImageNet100 and then report the top-5 accuracy of the ImageNet comparison.

\subsection{Detailed Results of CIFAR100 }

We first report the results of CIFAR100 where we select six dataset splits (\ie, Base0 Inc5, Base0 Inc10, Base0 Inc20, Base50 Inc50, Base50 Inc10, Base50 Inc25). The results are shown in Figure~\ref{figure:cifar100_full}, and we list the number of parameters, the average accuracy, and the last accuracy in Table~\ref{tab:average_acc_cifar_full}.

As we can infer from these figures, L2P consistently obtains the best performance among all settings. However, it must be noted that L2P relies on the ImageNet-21K pre-trained Vision Transformer with 85M parameters. At the same time, typical methods train ResNet32 with 0.46M parameters from scratch. Hence, comparing pre-trained model-based methods to typical methods is unfair. We also notice that dynamic networks, \eg, DER and MEMO obtain better performance at the cost of more parameters. Therefore, we report Table~\ref{tab:average_acc_cifar_full} to point out that parameter cost is sometimes ignored in the benchmark comparison, which shall lead to unfair results. These comparison details also motivate us to design {\em fair} comparison protocols for class-incremental learning.

\begin{table*}[t]
	\caption{ Average and last accuracy performance comparison on CIFAR100.  `\#P' represents the number of parameters (million). }\label{tab:average_acc_cifar_full}
	\centering
		\begin{tabular}{@{}lccccccccc ccc}
			\toprule
			\multicolumn{1}{c}{\multirow{2}{*}{Method}} & 
			\multicolumn{3}{c}{Base0 Inc5} & \multicolumn{3}{c}{Base0 Inc10} & \multicolumn{3}{c}{Base0 Inc20} & \multicolumn{3}{c}{Base50 Inc10}\\
			& {\#P} & {$\bar{\mathcal{A}}$} & ${\mathcal{A}_B}$ & {\#P} & {$\bar{\mathcal{A}}$} & ${\mathcal{A}_B}$ & {\#P} & {$\bar{\mathcal{A}}$} & ${\mathcal{A}_B}$
			& {\#P} & {$\bar{\mathcal{A}}$} & ${\mathcal{A}_B}$\\
	\midrule
Finetune   & 0.46 &17.59 &4.83 & 0.46 &26.25 & 9.09 & 0.46 & 37.90 & 17.07 & 0.46 & 22.79 & 9.09 \\
EWC    & 0.46 & 18.42 & 5.58 & 0.46 &29.73 & 12.44 & 0.46 & 39.19 & 19.87 & 0.46  &25.77 & 11.47\\
LwF   & 0.46 &30.93 & 12.60 & 0.46 &43.56 & 23.25 & 0.46 & 48.96 & 30.00 & 0.46 &41.12 & 25.06 \\
\midrule
GEM    & 0.46 & 31.73 & 19.48 & 0.46 & 40.18 & 23.03 & 0.46 & 50.33 & 34.09 & 0.46&33.28 & 21.33  \\
Replay & 0.46 & 58.20 & 38.69 & 0.46 & 59.31 & 41.01 & 0.46 &  60.03 & 43.08 & 0.46&52.37 & 41.26  \\
RMM   & 0.46 &65.72 & 51.10 & 0.46 &68.54 & 56.64 & 0.46 &  70.64 & 61.81 & 0.46 & 67.53 & 59.75 \\
iCaRL    & 0.46 & 63.51 & 45.12 & 0.46 &64.42 & 49.52 & 0.46 & 67.00 & 54.23 & 0.46 & 61.29 & 52.04 \\
PODNet   & 0.46 & 47.88 & 27.99 & 0.46 & 55.22 & 36.78 & 0.46 & 62.96 & 49.08 & 0.46 & 64.45 & 55.21 \\
Coil    & 0.46 & 57.68 & 34.33 & 0.46 &60.27 & 39.85 & 0.46 & 63.33 & 45.54 & 0.46 &55.71 & 41.24 \\
WA & 0.46 &64.65 & 48.46 & 0.46 & 67.09 & 52.30 & 0.46 & 68.51 & 57.97 & 0.46&64.32 & 55.85  \\
BiC   & 0.46 &62.38 & 43.08 & 0.46 & 65.08 & 50.79 & 0.46 & 67.03 & 56.22 & 0.46  &61.01 & 49.19\\
FOSTER   & 0.46 &63.38 & 49.42 & 0.46 & 66.49 & 53.21 & 0.46 & 68.60 & 58.67 & 0.46 &65.73 & 57.82 \\
DER    & 9.27 &  67.99 & 53.95 & 4.60 & 69.74 & \bf 58.59 & 2.30 & \bf 70.82 & \bf 62.40 &2.76 &68.24 & 61.94\\
MEMO    & 7.14 &\bf 68.10 & \bf 54.23 & 3.62 & \bf 70.20 & 58.49 & 1.87 & 70.43 & 61.39 &2.22 &\bf 69.39 & \bf 62.83\\
\midrule
DyTox & 10.7 &68.06 & 52.23 & 10.7 & 71.07 & 58.72 & 10.7 & 73.05 & 64.22 & 10.7 & 69.07 & 60.35\\
L2P & 85.7 & 84.00 & 78.96 & 85.7 & 89.35 & 83.39 & 85.7 & 90.17 & 86.91 & 85.7 & 85.21 & 79.34\\
\bottomrule
		\end{tabular}
\end{table*}

\subsection{Detailed Results of ImageNet100 }

We report the results of ImageNet100 where we select six dataset splits (\ie, Base0 Inc5, Base0 Inc10, Base0 Inc20, Base50 Inc50, Base50 Inc10, Base50 Inc25). The results are shown in Figure~\ref{figure:imagenet100_top1_full}, and we list the number of parameters, the average accuracy, and the last accuracy in Table~\ref{tab:average_acc_imagenet100}. The conclusions are consistent with former small-scale datasets, and dynamic networks perform better at the cost of more memory budget.

It must be noted that there are two methods not included in the ImageNet comparison, \ie, GEM and L2P. GEM requires solving the QP problem and saving a large-scale matrix, which cannot be deployed with ImageNet. Besides, L2P utilizes an ImageNet-21K pre-trained backbone for class-incremental learning~\cite{wang2022learning}, which overlaps with the dataset to be learned. Hence, these two methods are not included in the comparison.

\begin{figure*}[t]
	\begin{center}
			
			\subfigure[ImageNet100 Base0 Inc5 ]
			{	\includegraphics[width=.64\columnwidth]{pics/imagenet100b0inc5}
				}
			\hfill
			\subfigure[ImageNet100 Base0 Inc10]
			{	\includegraphics[width=.64\columnwidth]{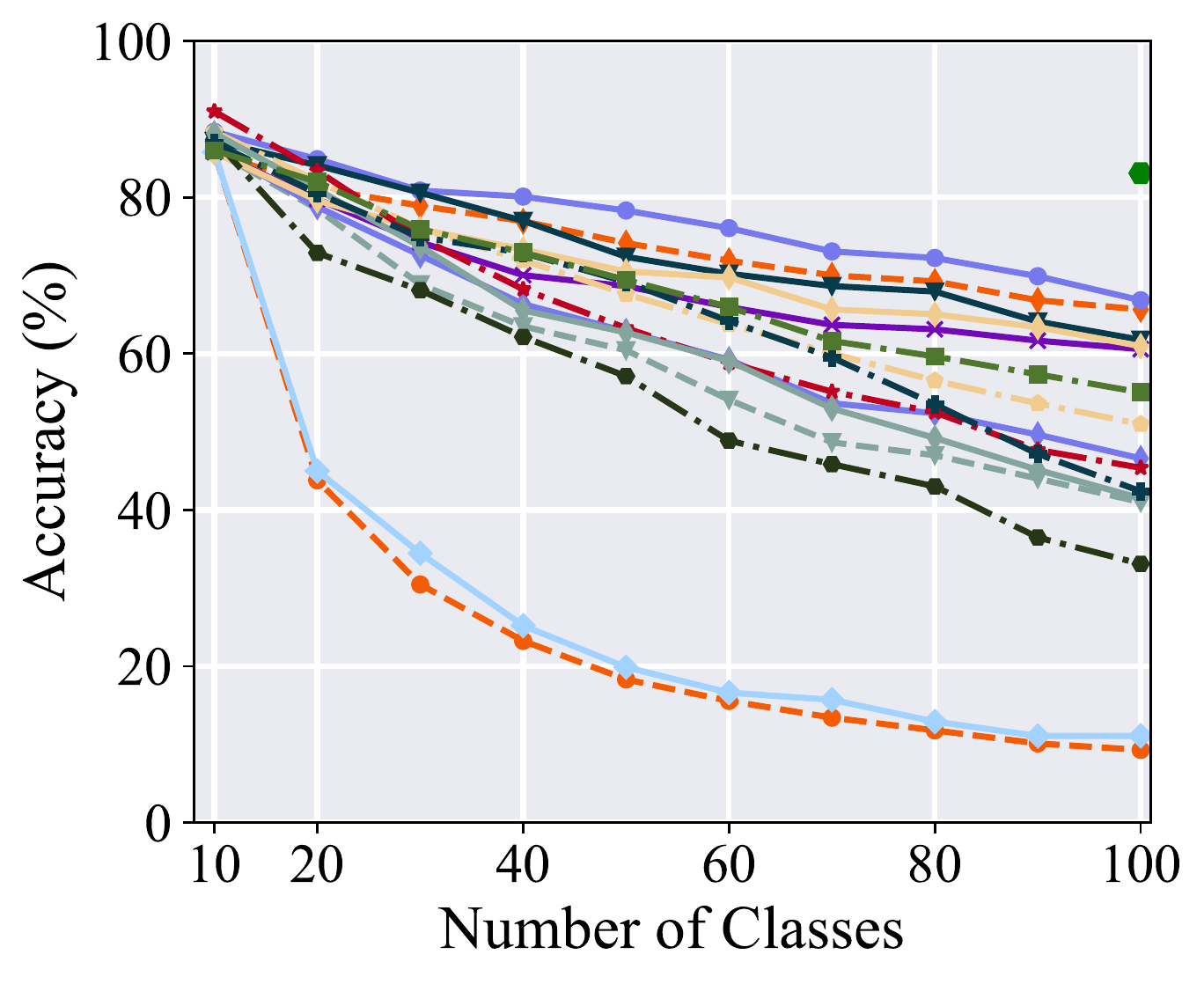}		
				}
			\hfill
			\subfigure[ImageNet100 Base0 Inc20]
			{	\includegraphics[width=.64\columnwidth]{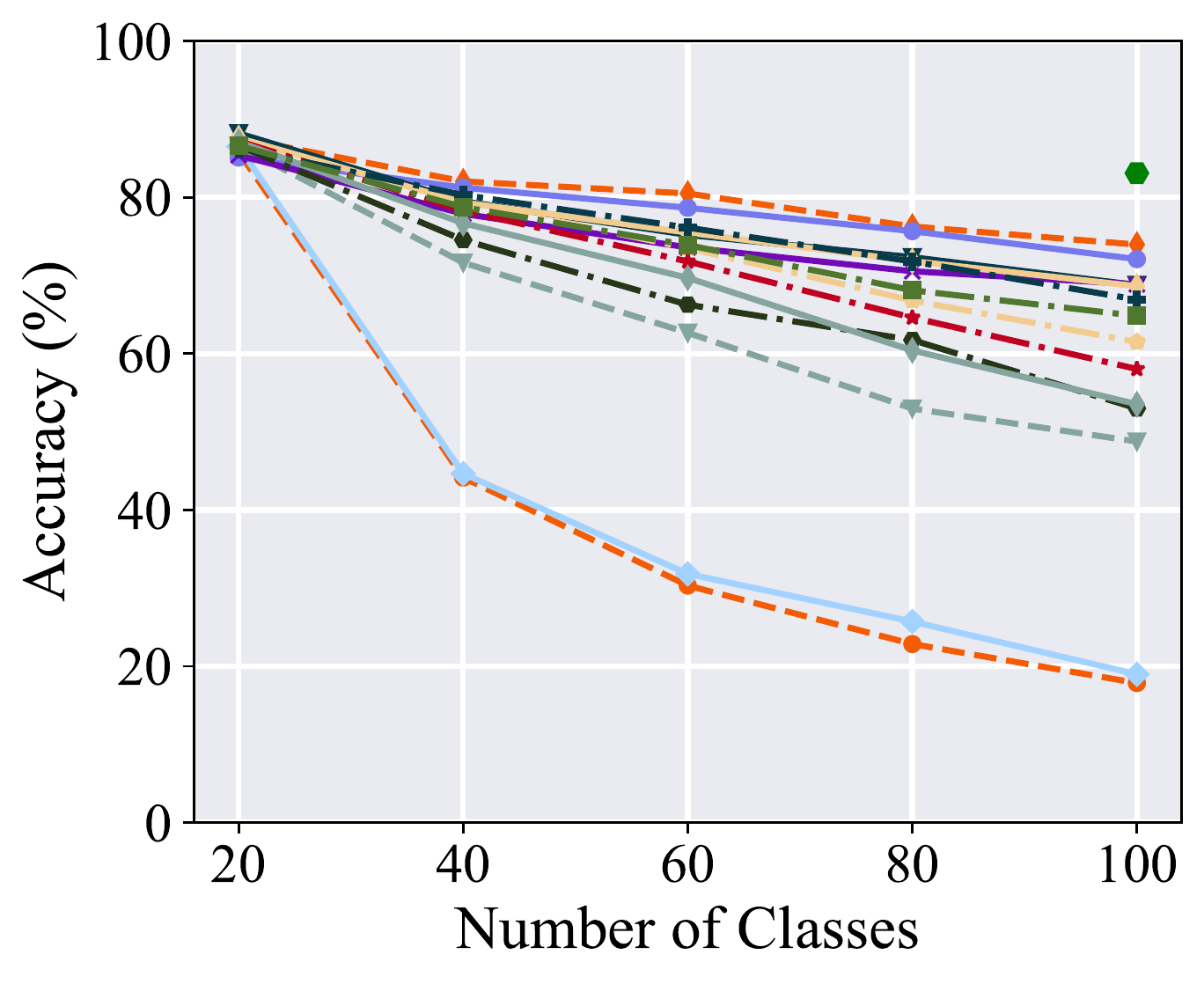}	
				}
			\\
				\includegraphics[width=1.9\columnwidth]{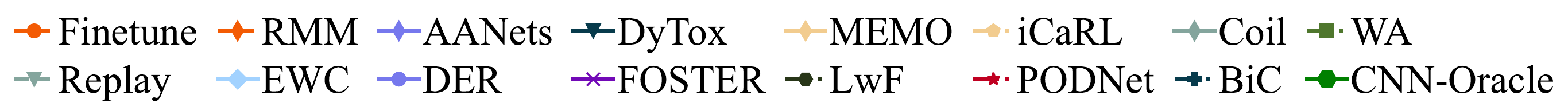}
			\subfigure[ImageNet100 Base50 Inc50]
			{	\includegraphics[width=.64\columnwidth]{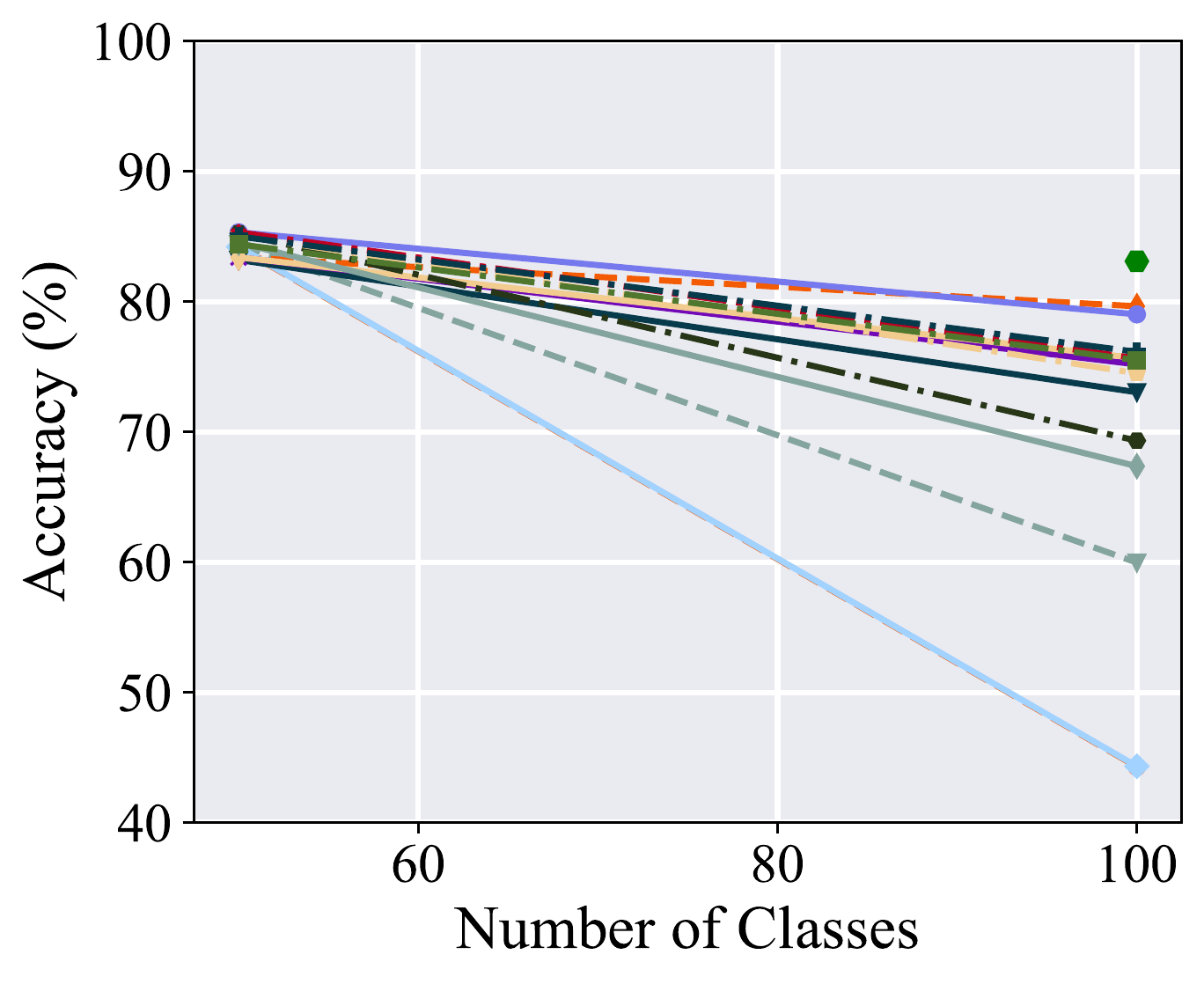}
				}
			\hfill
			\subfigure[ImageNet100 Base50 Inc10]
			{	\includegraphics[width=.64\columnwidth]{pics/imagenet100b50inc10}
				}
			\hfill
			\subfigure[ImageNet100 Base50 Inc25]
			{	\includegraphics[width=.64\columnwidth]{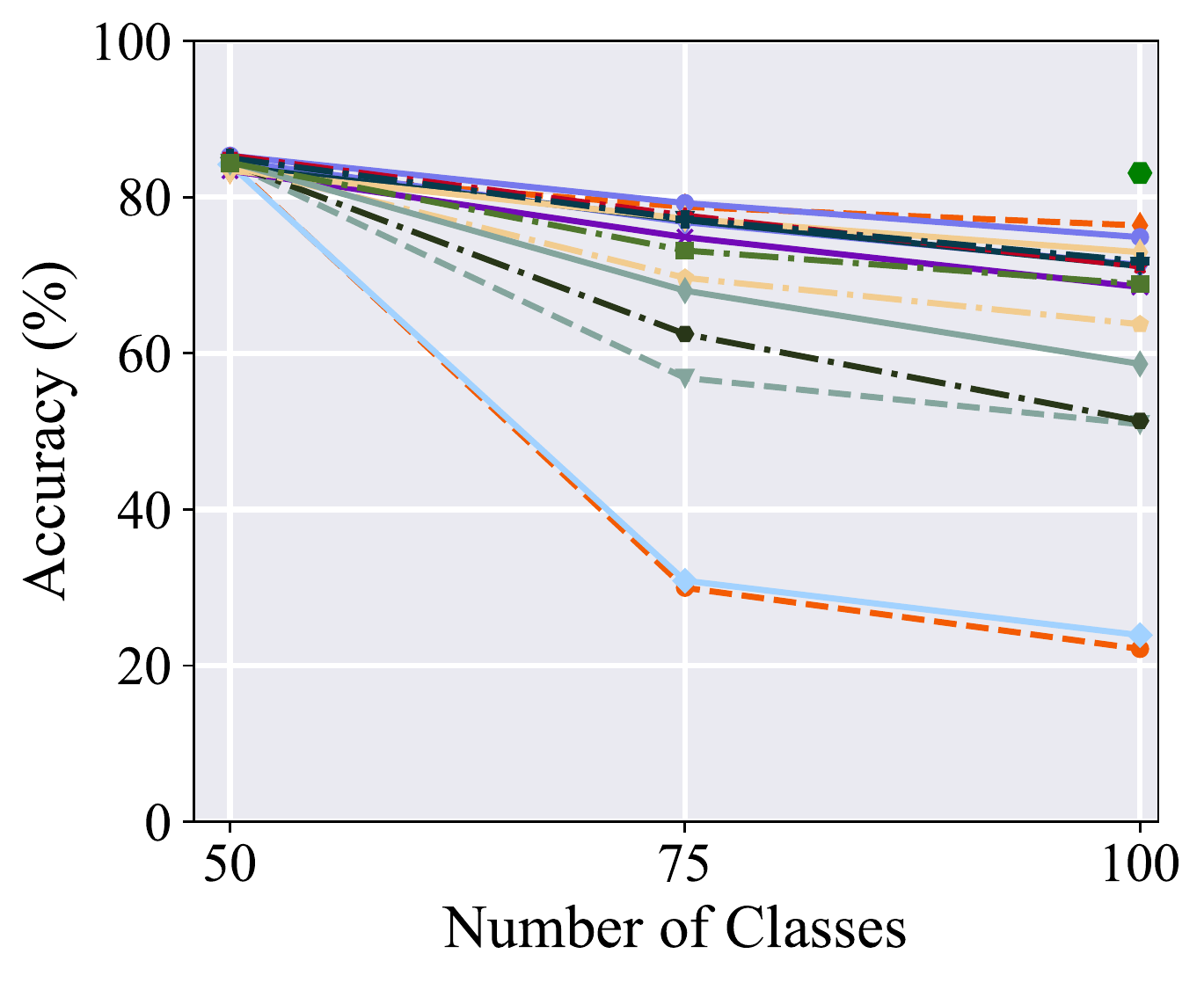}
				}
		\end{center}
\vspace{-4mm}
	\caption{Incremental top-1 accuracy of different methods on ImageNet100. 
		} \label{figure:imagenet100_top1_full}
\end{figure*}

\begin{table*}[t]
	\caption{  Average and last top-1 accuracy performance comparison on ImageNet100.  `\#P' represents the number of parameters (million). }\label{tab:average_acc_imagenet100}
	
	\centering
	\begin{tabular}{@{}lccccccccc ccc}
		\toprule
		\multicolumn{1}{c}{\multirow{2}{*}{Method}} & 
		\multicolumn{3}{c}{Base0 Inc5} & \multicolumn{3}{c}{Base0 Inc10} & \multicolumn{3}{c}{Base0 Inc20} & \multicolumn{3}{c}{Base50 Inc10}\\
		& {\#P} & {$\bar{\mathcal{A}}$} & ${\mathcal{A}_B}$ & {\#P} & {$\bar{\mathcal{A}}$} & ${\mathcal{A}_B}$ & {\#P} & {$\bar{\mathcal{A}}$} & ${\mathcal{A}_B}$
		& {\#P} & {$\bar{\mathcal{A}}$} & ${\mathcal{A}_B}$\\
		\midrule
		Finetune   &11.17 &17.06 & 4.70 & 11.17 & 26.19 & 9.30 & 11.17 & 40.20 & 17.86 & 11.17 &24.12 & 9.26 \\
		EWC     &11.17 & 18.78 & 6.14 & 11.17 &27.78 & 11.10 & 11.17 & 41.54 & 18.98 & 11.17 &26.21 & 11.54\\
		LwF    &11.17 & 41.76 & 17.74 & 11.17 & 55.50 & 33.10 & 11.17 & 68.43 & 53.00 & 11.17 & 46.24 & 31.42 \\
		\midrule
		Replay  &11.17 &  56.37 & 37.32 & 11.17 &59.21 & 41.00 & 11.17 & 64.53 & 48.76 & 11.17 & 55.73 & 43.38 \\
		RMM    &11.17 & 69.70 & 57.16 & 11.17 & 74.07 & 65.66 & 11.17 & \bf  80.08 & \bf 73.96 & 11.17 &73.02 & 66.52 \\
		iCaRL     &11.17 &  62.36 & 44.10 & 11.17 & 67.11 & 50.98 & 11.17 & 73.57 & 61.50 & 11.17 &62.56 & 53.68 \\
		PODNet   &11.17 &  53.70 & 33.34 & 11.17 &64.03 & 45.40 & 11.17 & 71.99 & 58.04 & 11.17 & 73.83 & 62.94 \\
		Coil    &11.17 & 56.21 & 34.00 & 11.17 & 61.91 & 41.50 & 11.17 & 69.49 & 53.54 & 11.17& 59.80 & 43.40\\
		WA  &11.17 & 62.96 & 46.06 & 11.17 & 68.60 & 55.04 & 11.17 & 74.44 & 64.84 & 11.17 &65.81 & 56.64\\
		BiC   &11.17 & 58.03 & 34.56 & 11.17 & 65.13 & 42.40 & 11.17 & 76.29 & 66.92 & 11.17 & 66.36 & 49.90 \\
		FOSTER    &11.17 & 64.45 & 53.18 & 11.17 & 69.36 & 60.58 & 11.17 & 75.27 & 68.88 & 11.17 &69.85 & 63.12 \\
		DER  &  223.4 & \bf 73.79 & \bf 63.66 &111.7 &\bf 77.08 &\bf 66.84 &55.85 & 78.56 & 72.10 &67.02 &\bf 77.57 & \bf 71.10\\
		MEMO  & 170.6 & 68.19 & 56.10 &86.72 & 71.00 & 60.96 &44.75 & 76.59 & 68.64 &53.14 &76.66 & 70.22 \\
		\midrule
		DyTox & 11.00 & 69.57 & 53.82 & 11.00 &73.40 & 61.78 & 11.00 &76.81 & 68.78 & 11.00 & 74.65 & 65.76 \\
		\bottomrule
	\end{tabular}
\end{table*}

\subsection{ImageNet Top-5 Results}

In this section, we report the top-5 accuracy of different methods in the ImageNet100 comparison. As shown in Figure~\ref{figure:imagenet100_top5}, the ranking of different methods remains the same as in the top-1 accuracy comparison.
We also report the top-5 accuracy in the ImageNet1000 comparison in Figure~\ref{figure:imagenet1000_top5}.

\begin{figure*}[t]
	\begin{center}
		
		\subfigure[ImageNet100 B0 Inc5 ]
		{	\includegraphics[width=.64\columnwidth]{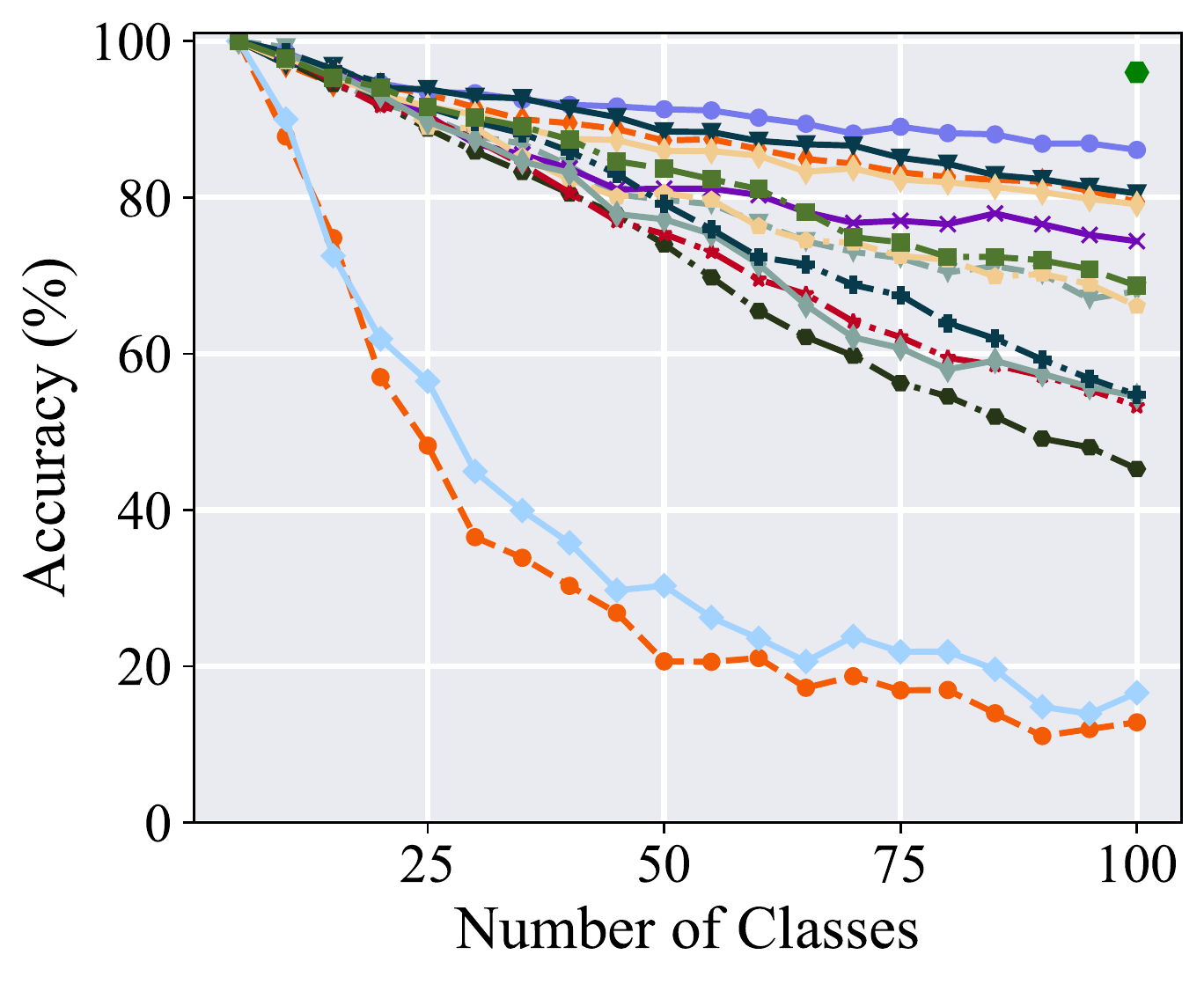}
		}
		\hfill
		\subfigure[ImageNet100 B0 Inc10]
		{	\includegraphics[width=.64\columnwidth]{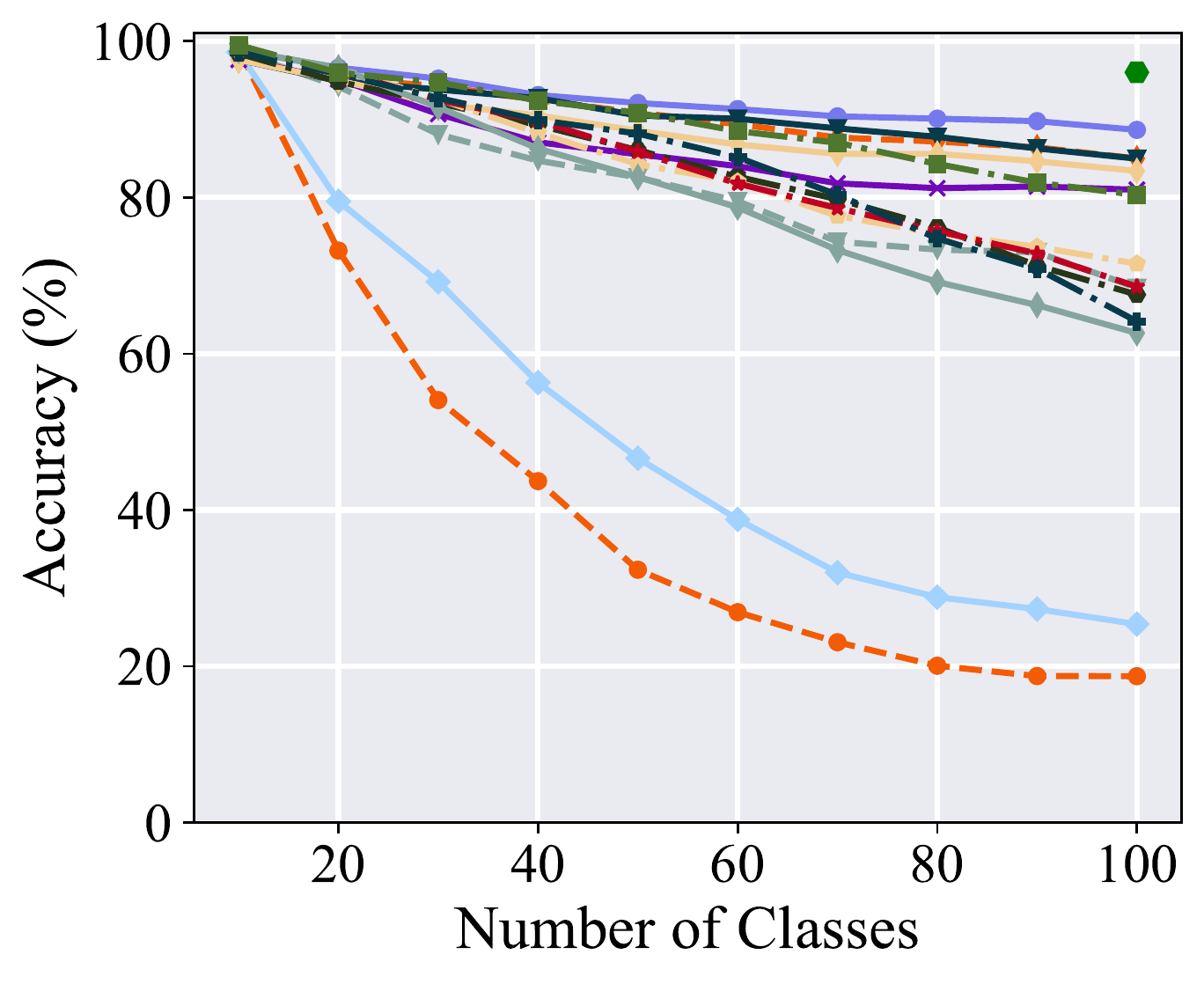}		
		}
		\hfill
		\subfigure[ImageNet100 B0 Inc20]
		{	\includegraphics[width=.64\columnwidth]{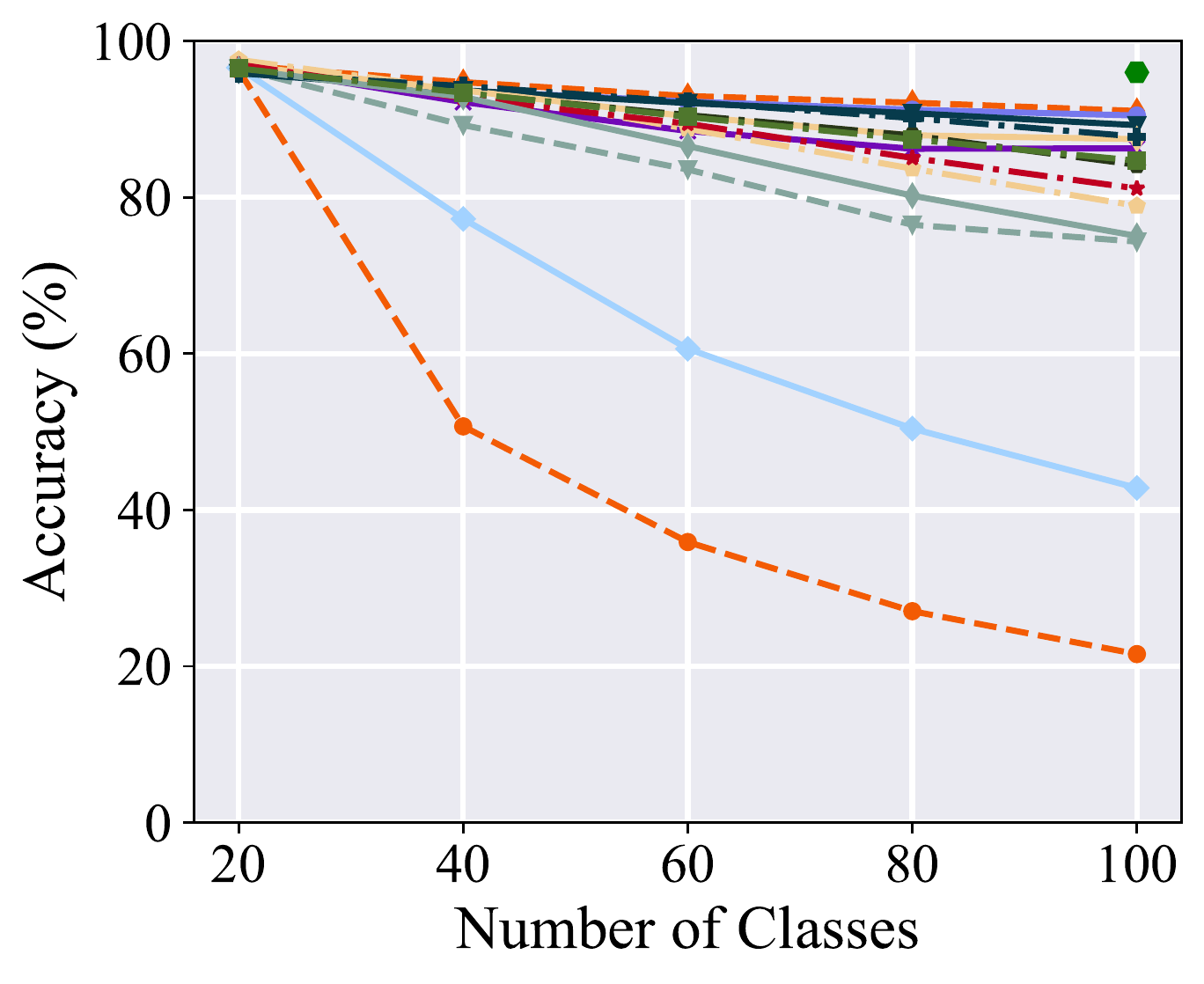}	
		}
		\\
			\includegraphics[width=1.9\columnwidth]{pics/imagenet_legend}
		\subfigure[ImageNet100 B50 Inc50]
		{	\includegraphics[width=.64\columnwidth]{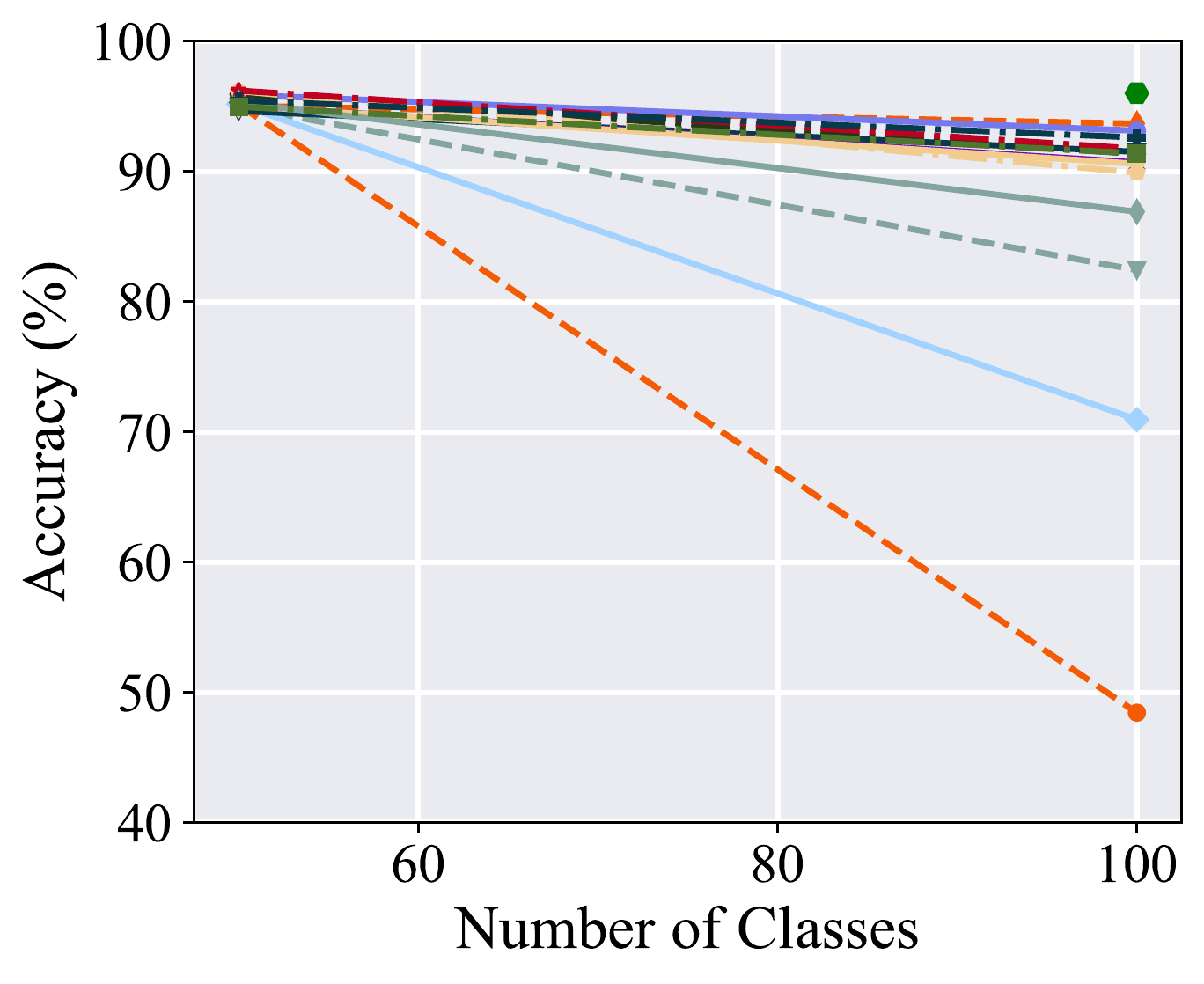}
		}
		\hfill
		\subfigure[ImageNet100 B50 Inc10]
		{	\includegraphics[width=.64\columnwidth]{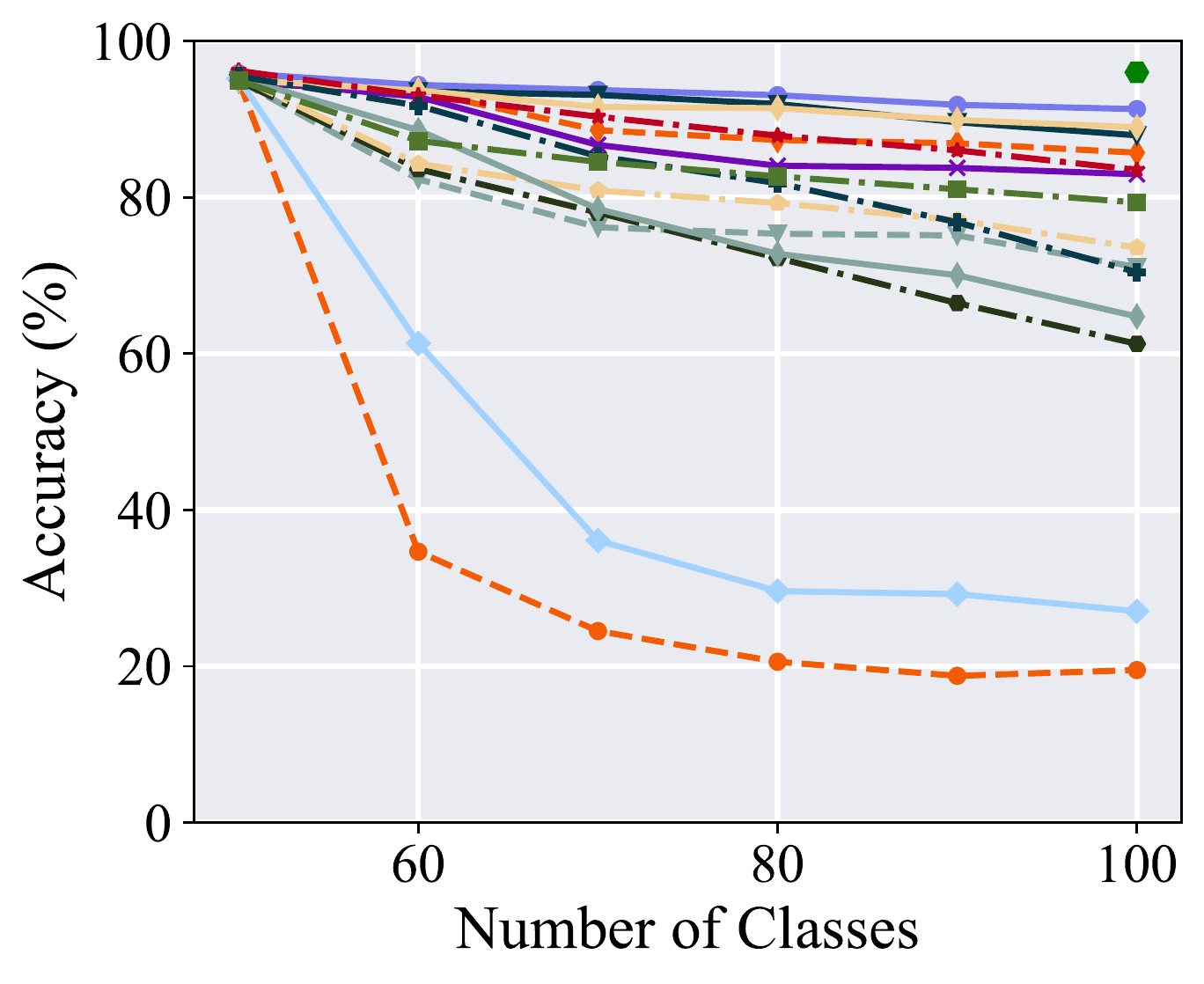}
		}
		\hfill
		\subfigure[ImageNet100 B50 Inc25]
		{	\includegraphics[width=.64\columnwidth]{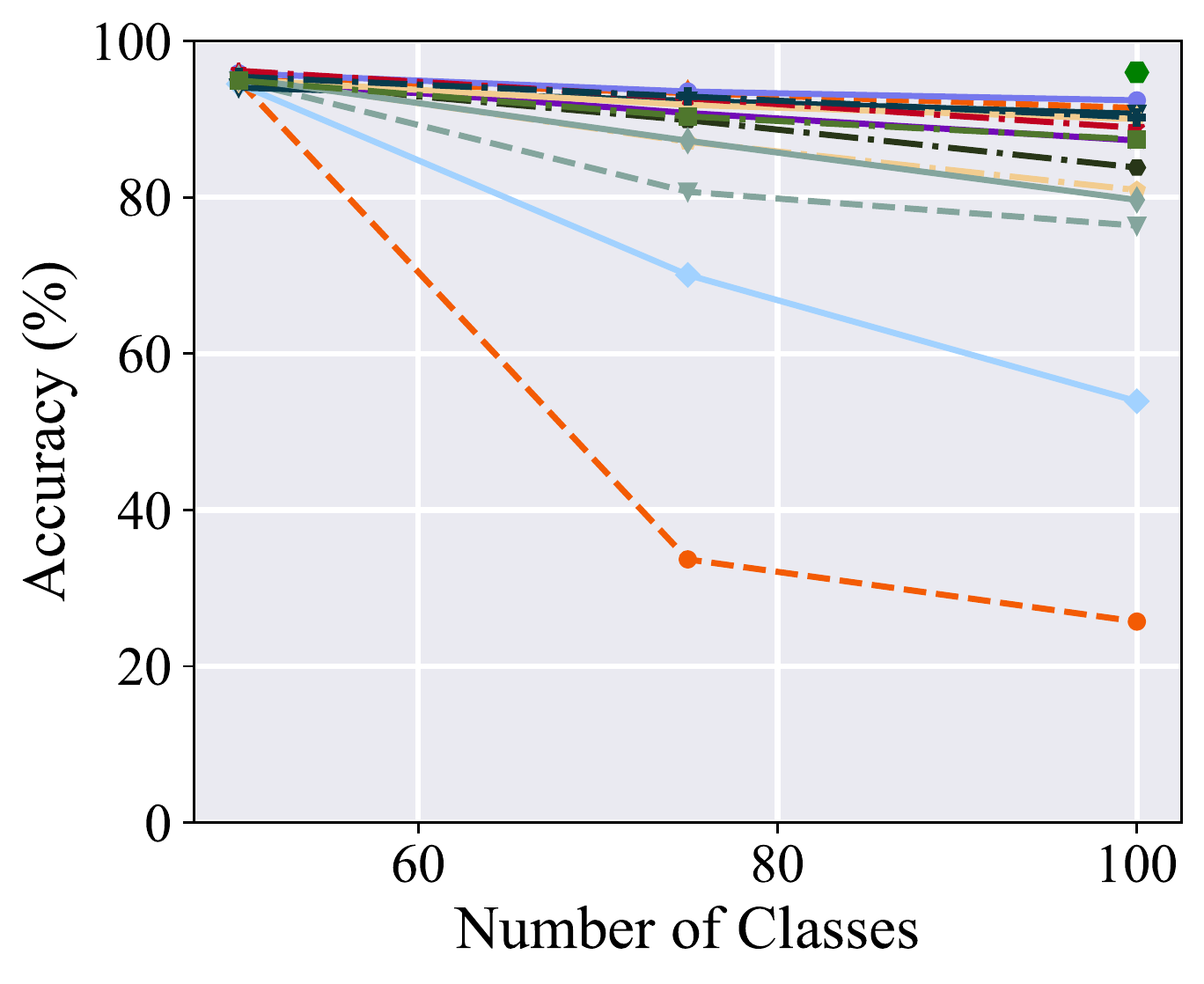}
		}
	\end{center}
	\vspace{-4mm}
	\caption{Incremental top-5 accuracy of different methods on ImageNet100. 
	} \label{figure:imagenet100_top5}
\end{figure*}

\begin{figure*}[t]
	\begin{center}
			\includegraphics[width=1.9\columnwidth]{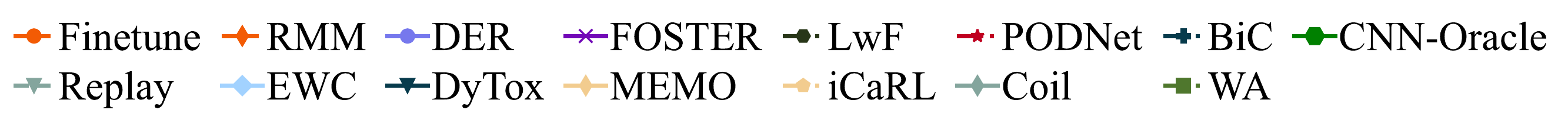}
		\subfigure[ImageNet1000 B0 Inc100 ]
		{	\includegraphics[width=.97\columnwidth]{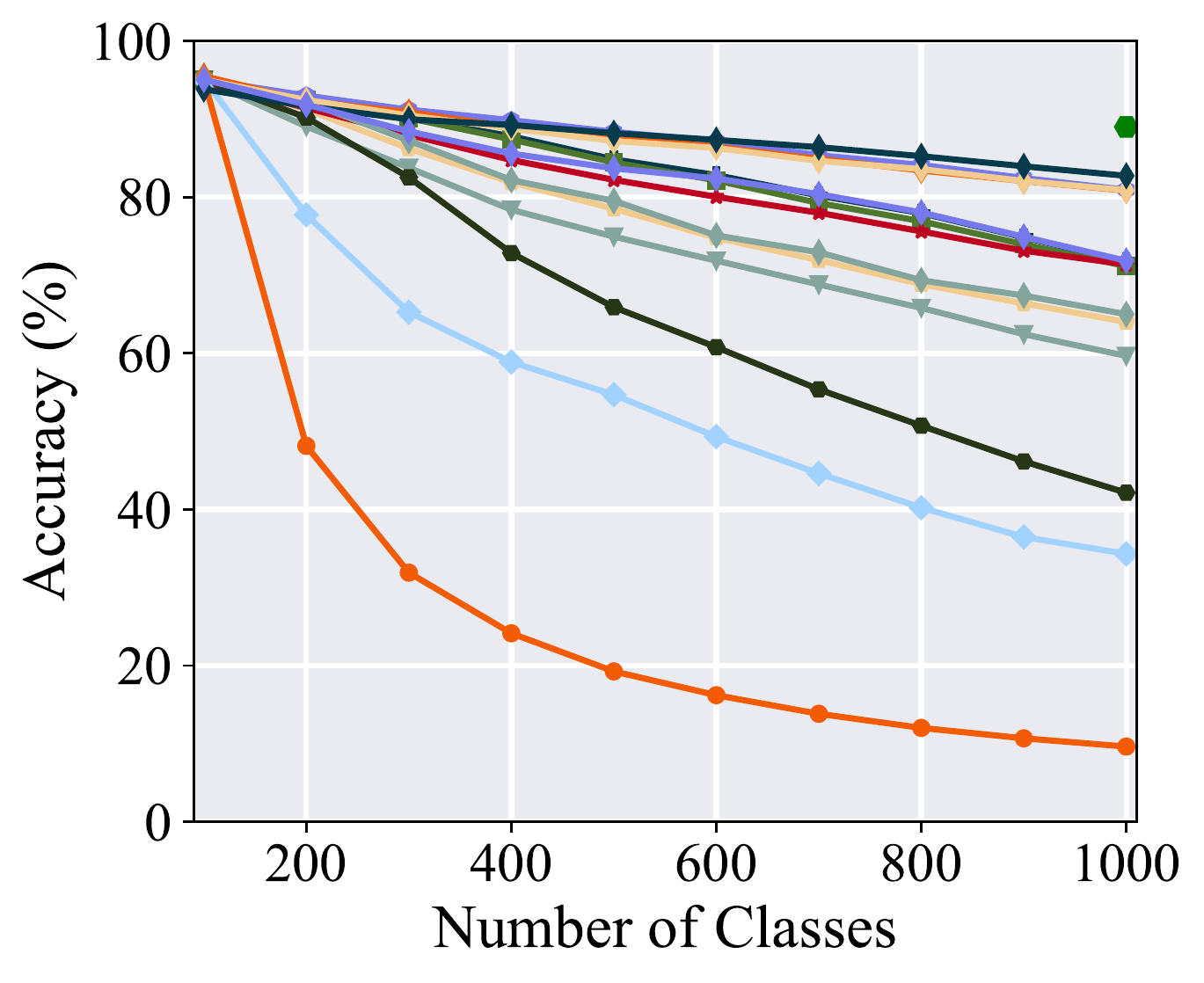}
		}
		\subfigure[ImageNet1000 B500 Inc100]
		{	\includegraphics[width=.97\columnwidth]{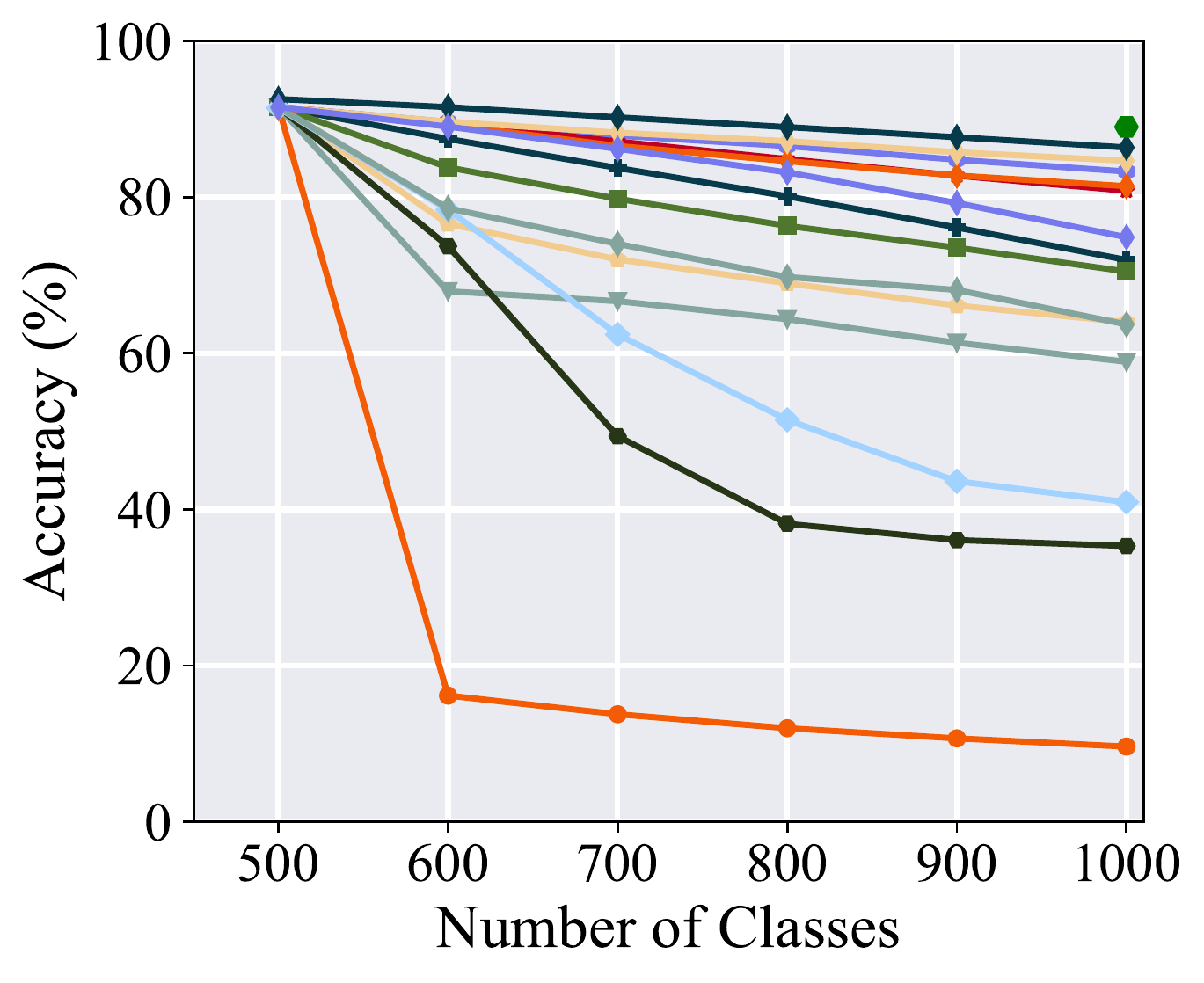}		
		}
	\end{center}
		\vspace{-4mm}
	\caption{Incremental top-5 accuracy of different methods on ImageNet1000.  
	} \label{figure:imagenet1000_top5}
		\vspace{-4mm}
\end{figure*}

\begin{figure*}[t]
	\begin{center}
		\subfigure[Finetune ]
		{	\includegraphics[width=.36\columnwidth]{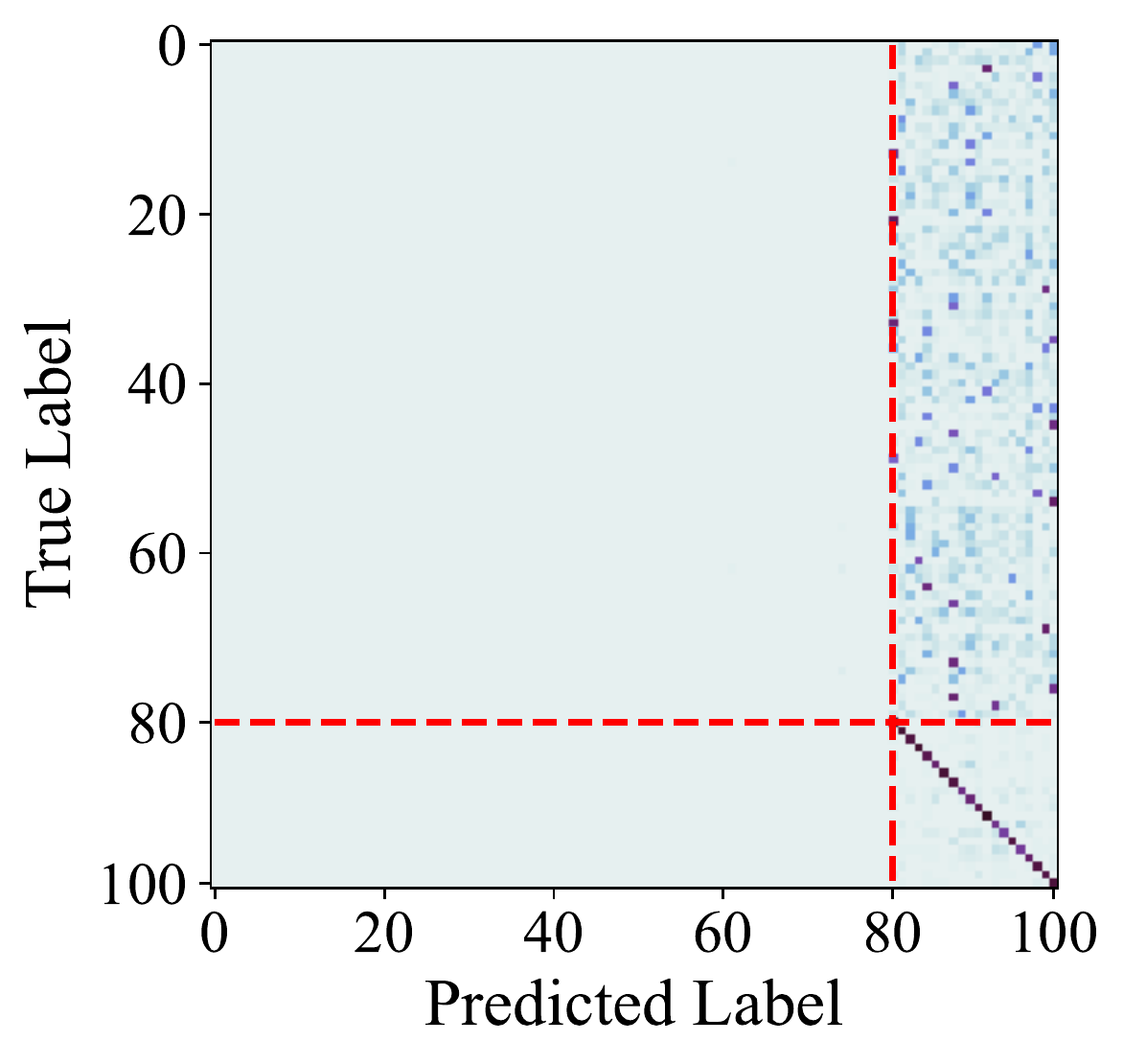}
		}
		\hfill
		\subfigure[EWC]
		{	\includegraphics[width=.36\columnwidth]{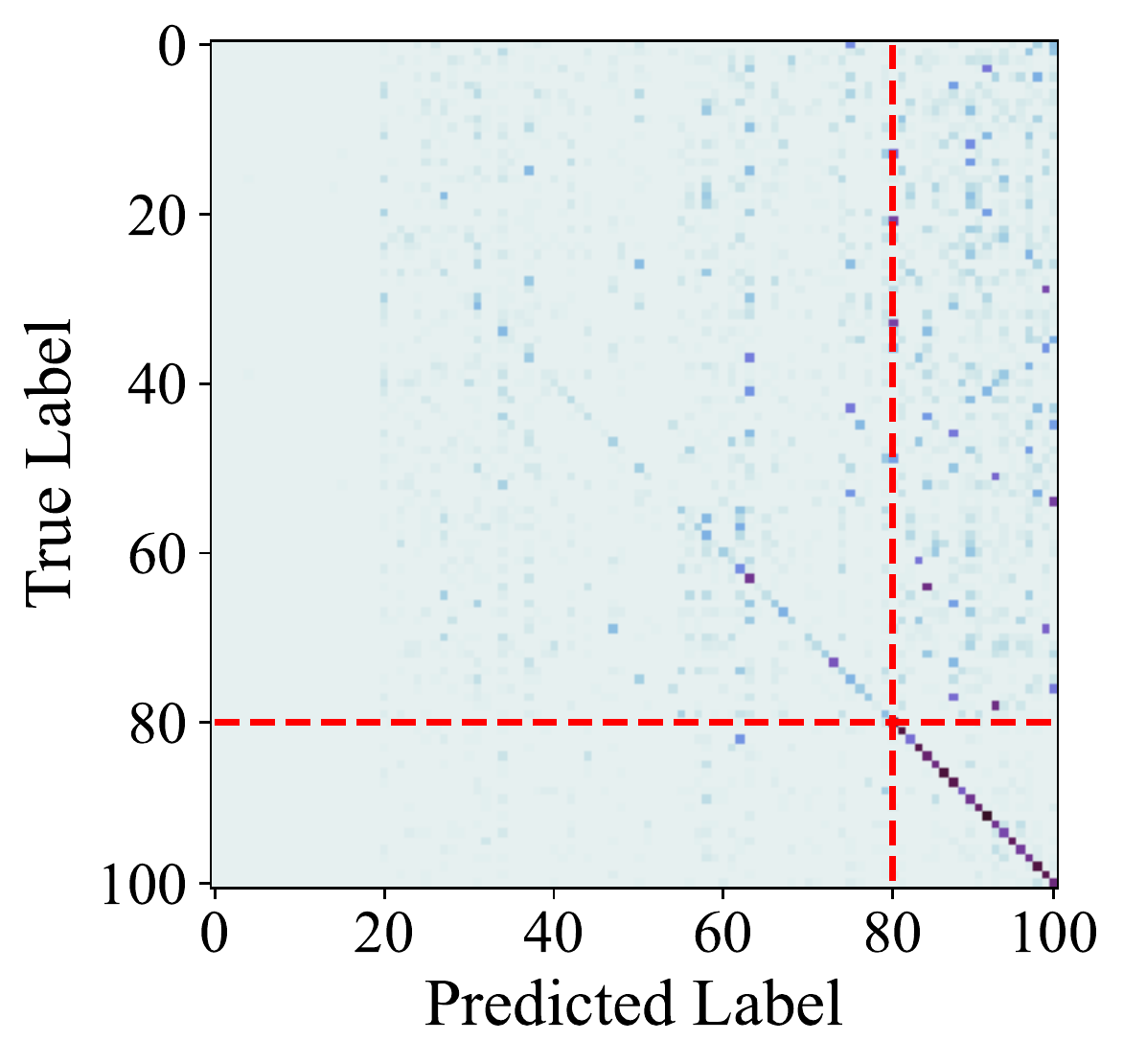}		
		}
		\hfill
		\subfigure[LwF]
		{	\includegraphics[width=.36\columnwidth]{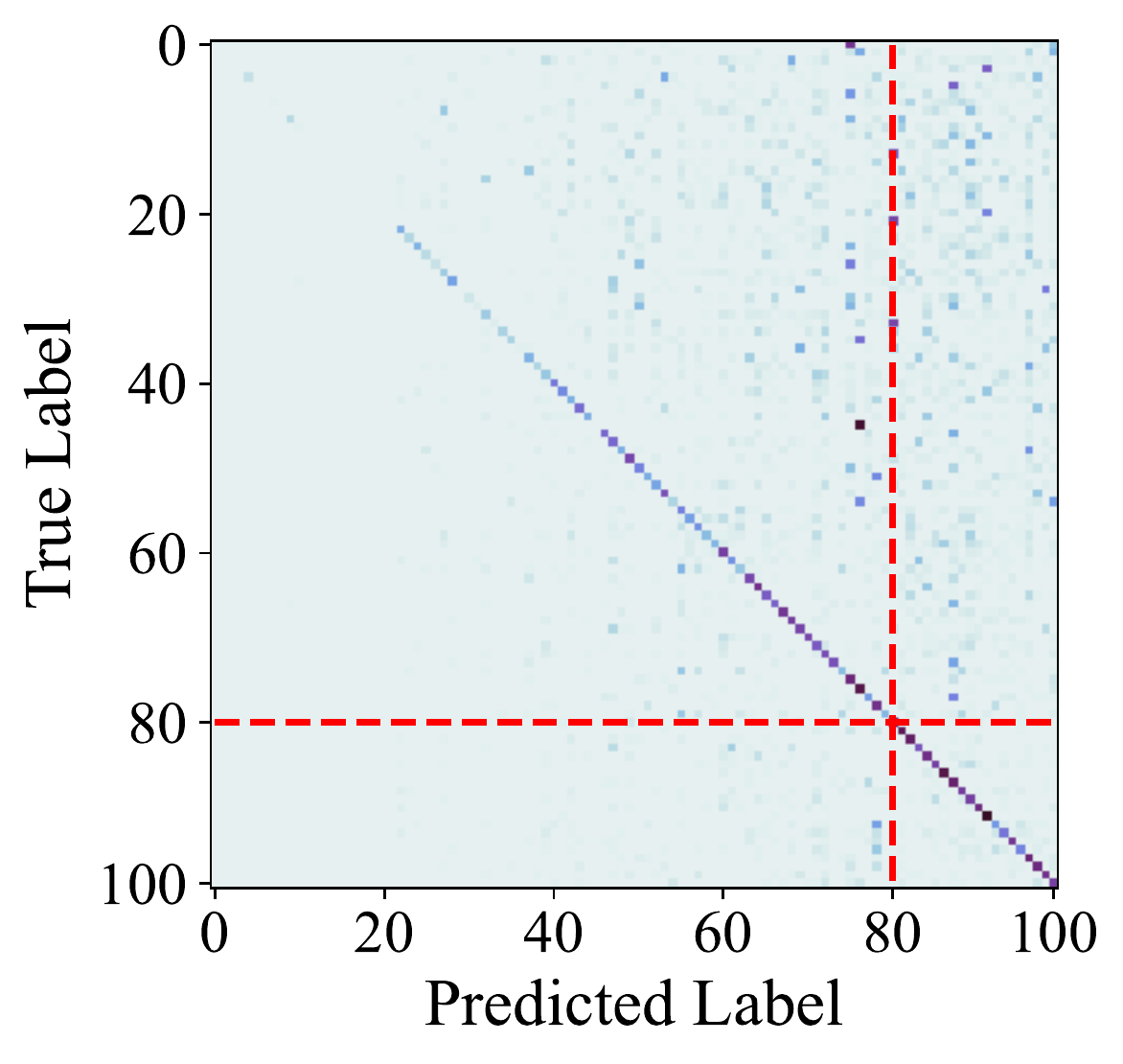}	
		}
		\hfill
		\subfigure[Replay]
		{	\includegraphics[width=.36\columnwidth]{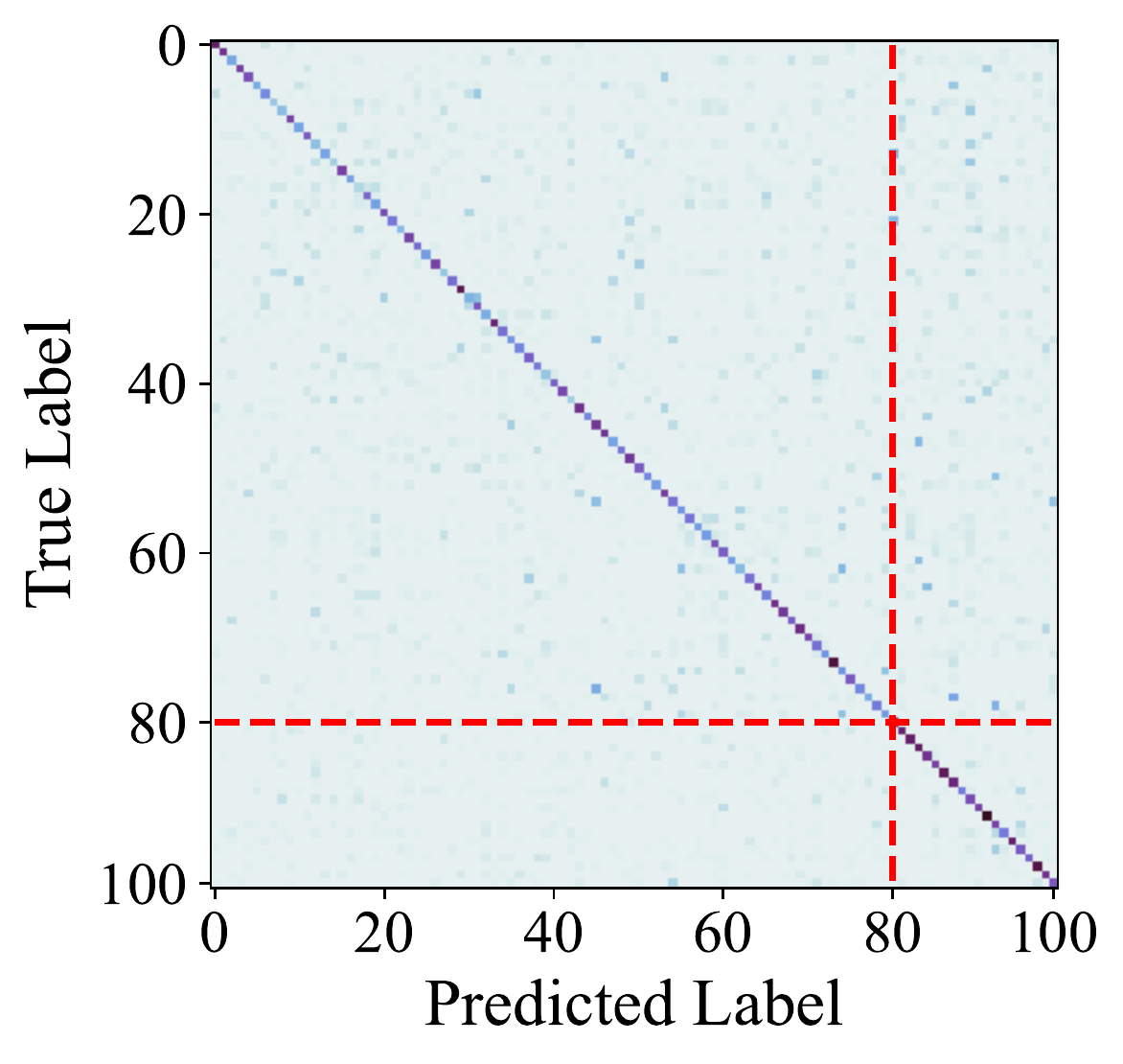}
		}
		\hfill
		\subfigure[GEM]
		{	\includegraphics[width=.36\columnwidth]{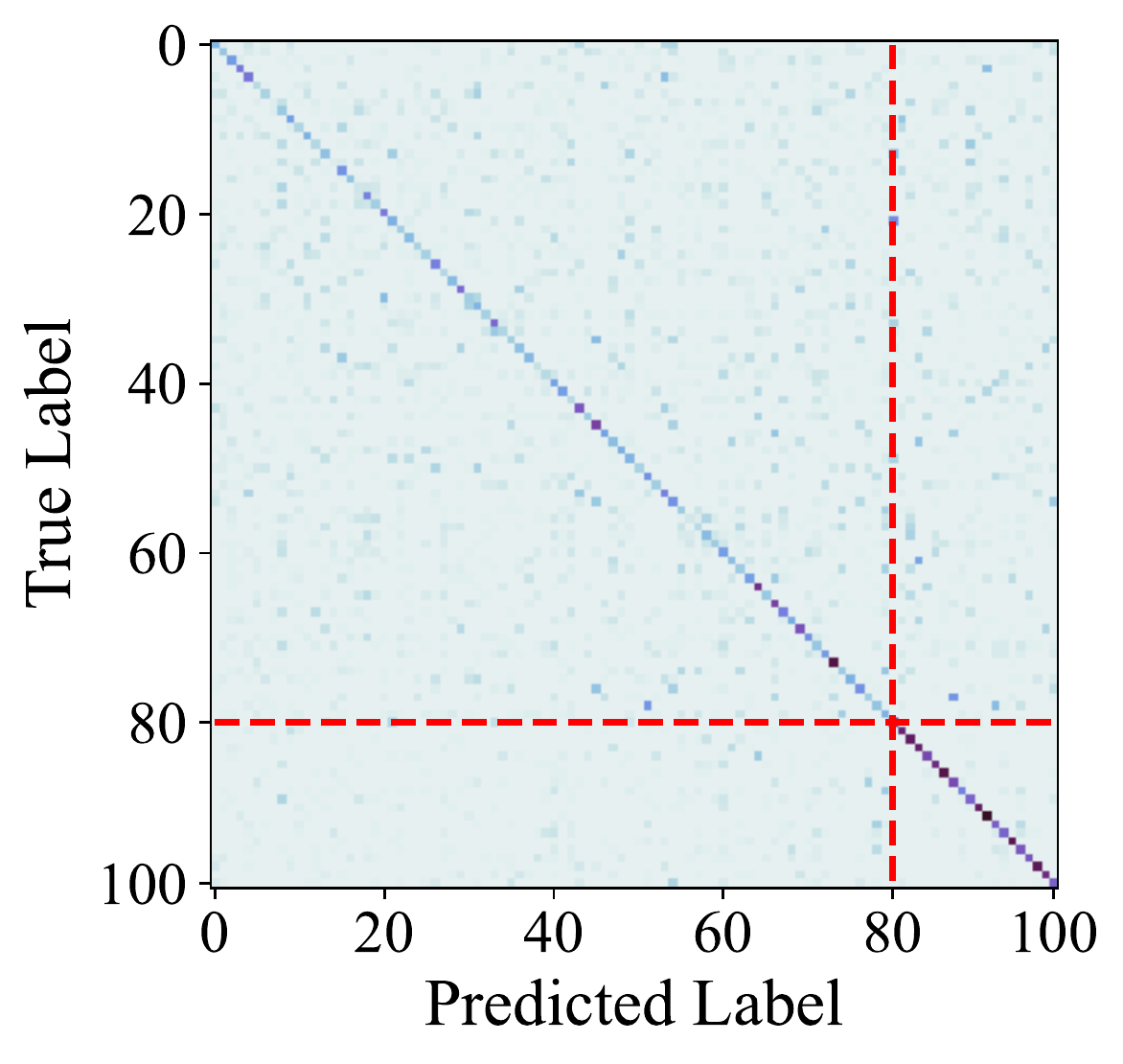}
		} \\
		\subfigure[iCaRL]
		{	\includegraphics[width=.36\columnwidth]{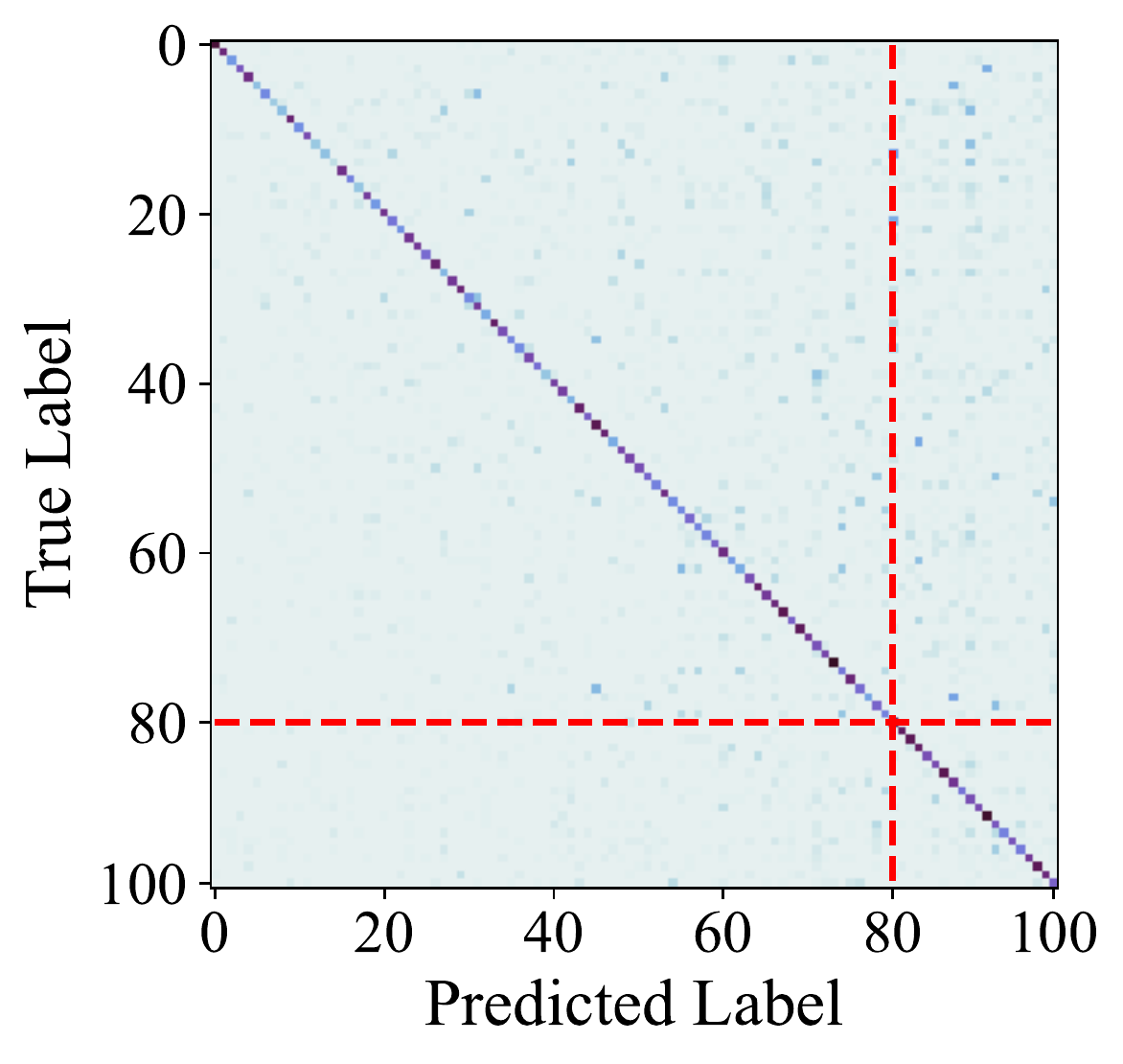}
		}
		\hfill
		\subfigure[BiC ]
		{	\includegraphics[width=.36\columnwidth]{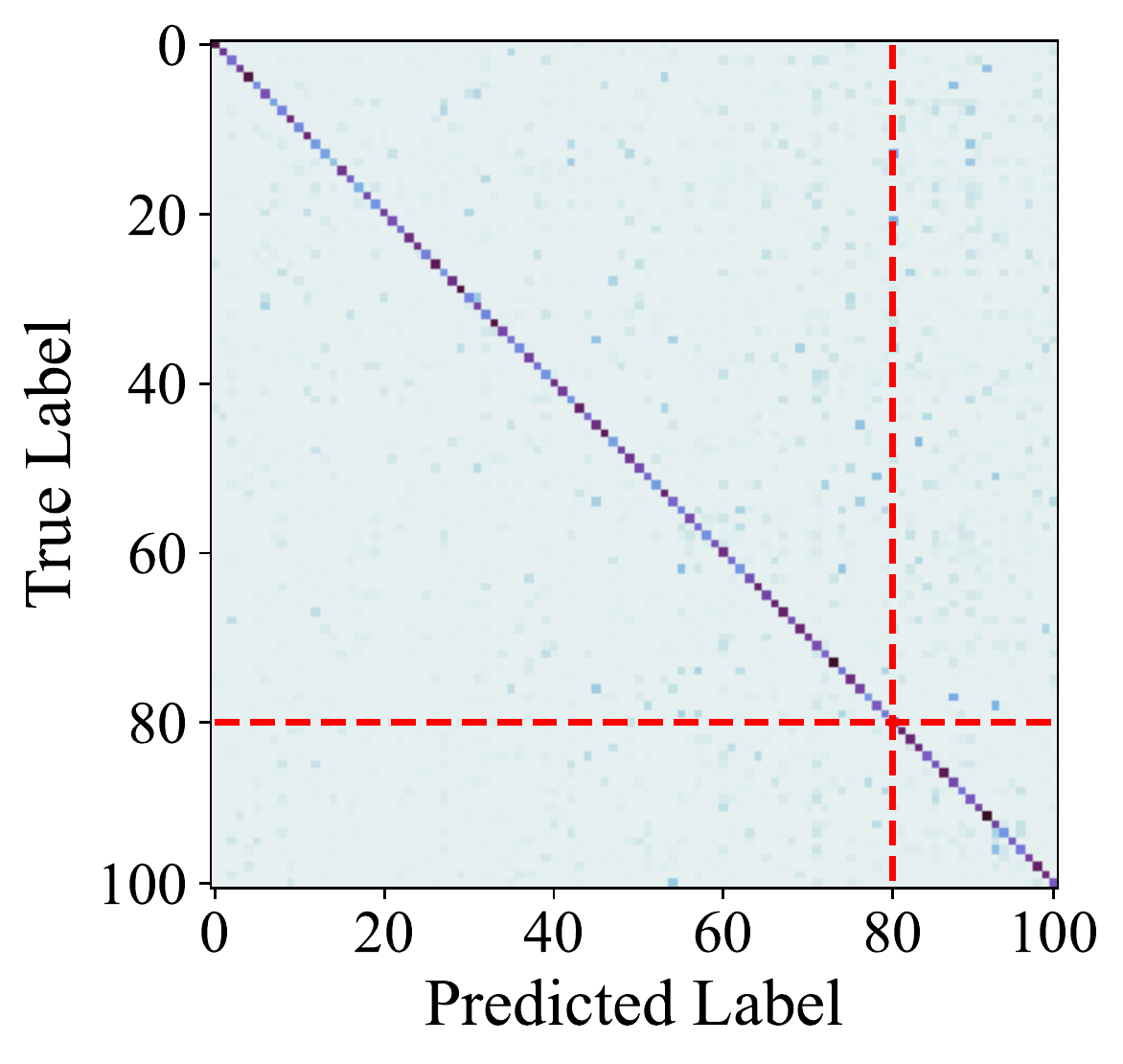}
		}
		\hfill
		\subfigure[WA]
		{	\includegraphics[width=.36\columnwidth]{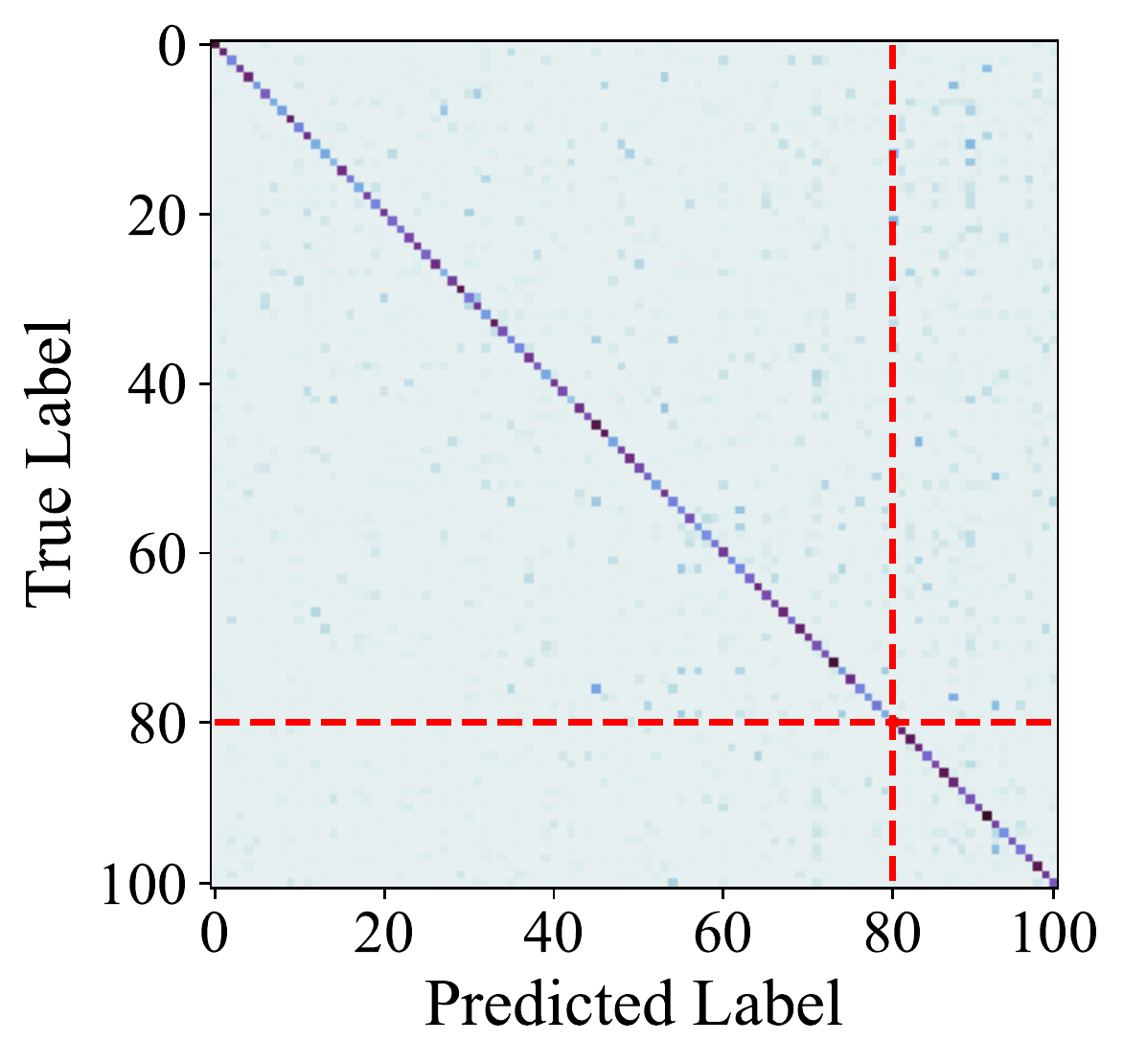}		
		}
		\hfill
		\subfigure[PODNet]
		{	\includegraphics[width=.36\columnwidth]{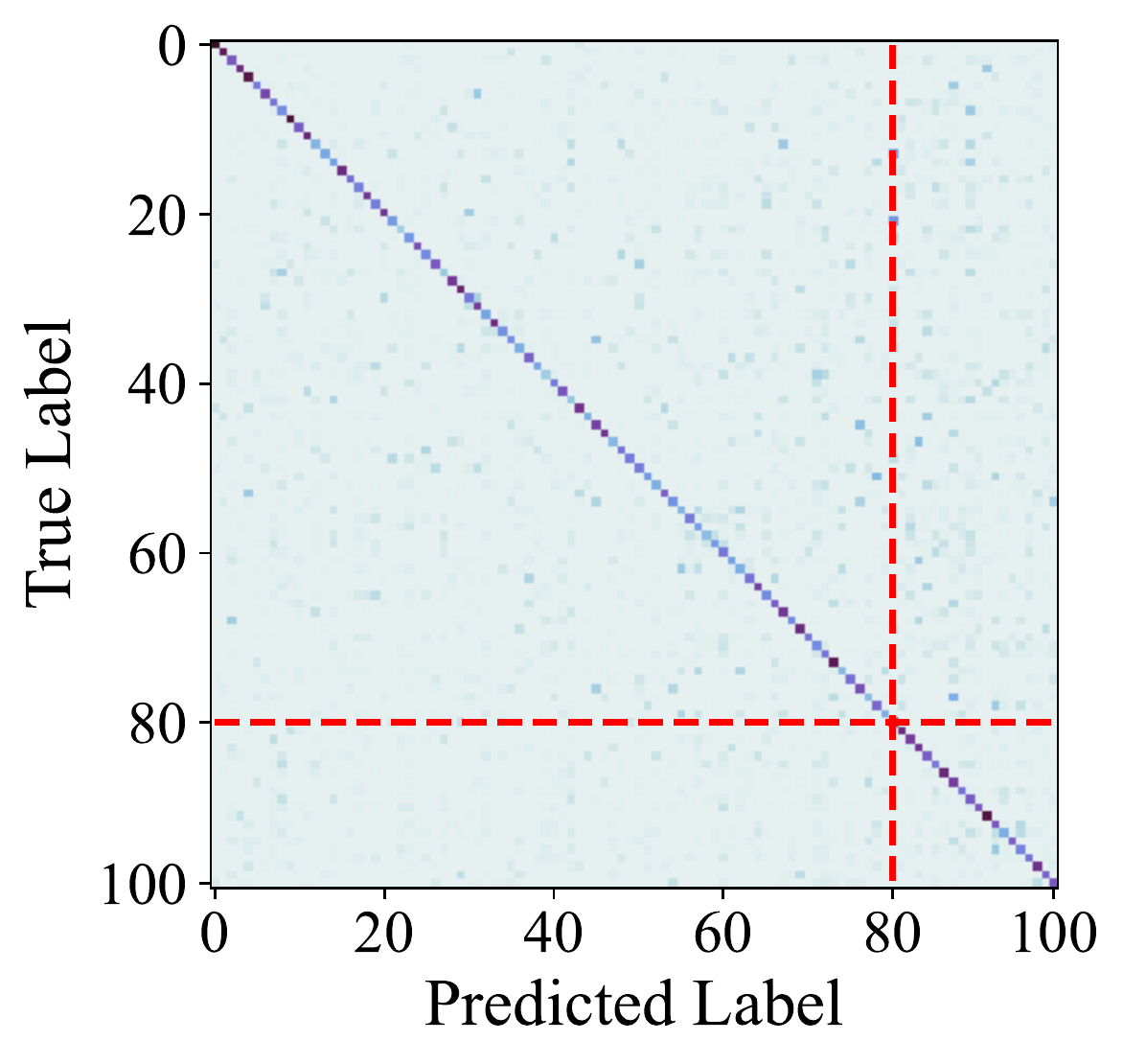}	
		}
		\hfill
		\subfigure[DER]
		{	\includegraphics[width=.36\columnwidth]{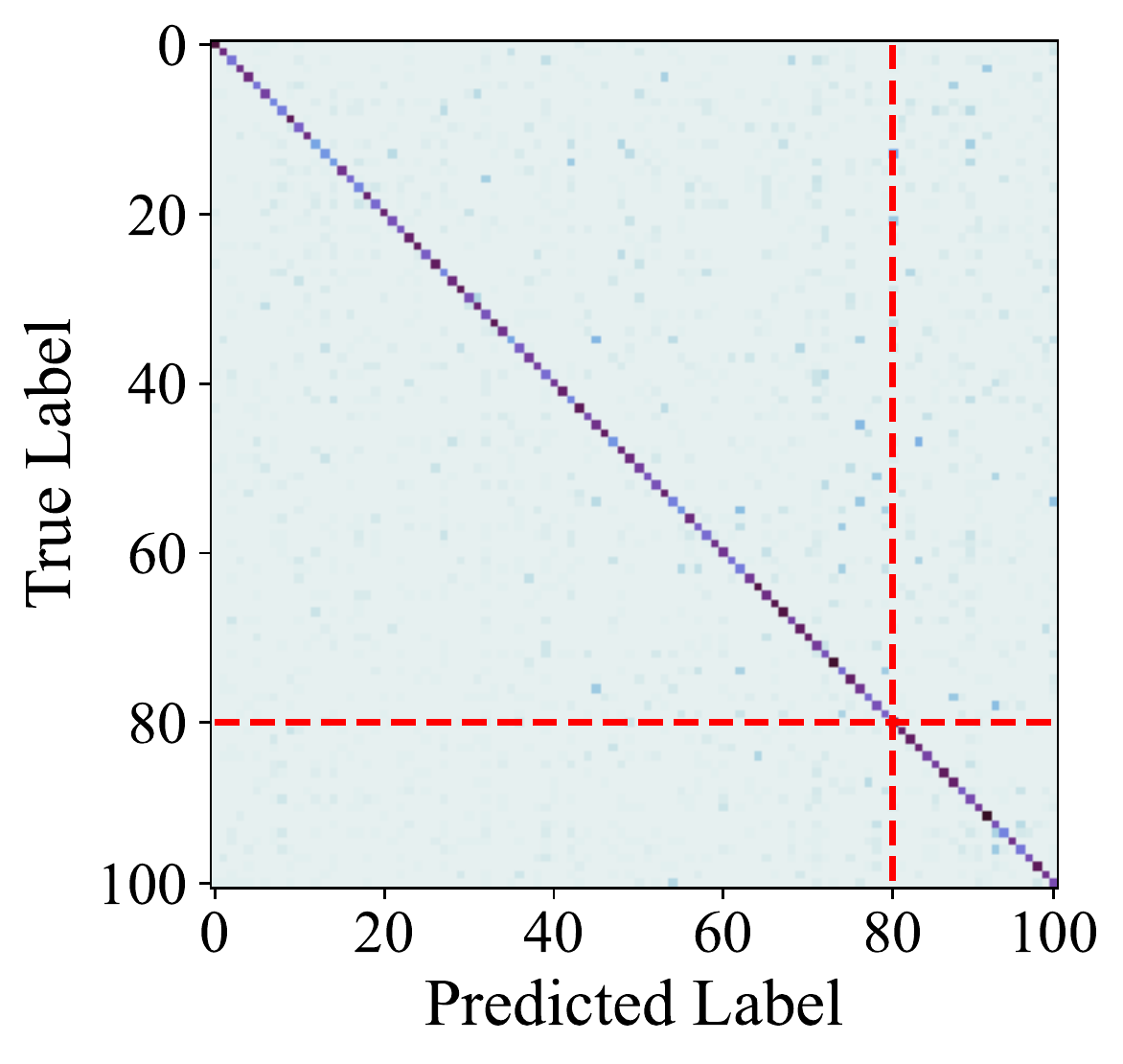}	
		}
		\\
		\subfigure[RMM+FOSTER]
		{	\includegraphics[width=.36\columnwidth]{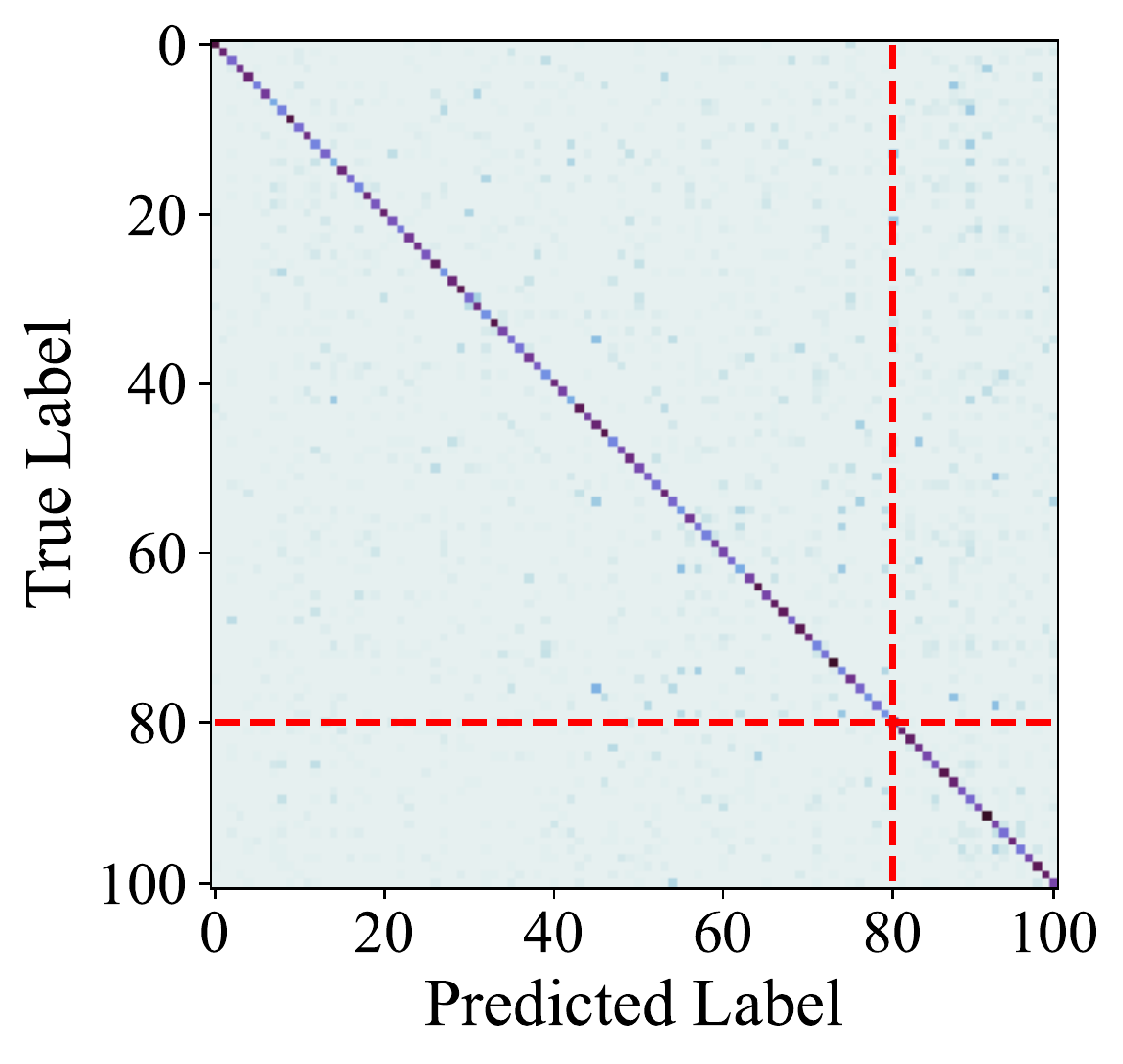}
		}
		\hfill
		\subfigure[Coil ]
		{	\includegraphics[width=.36\columnwidth]{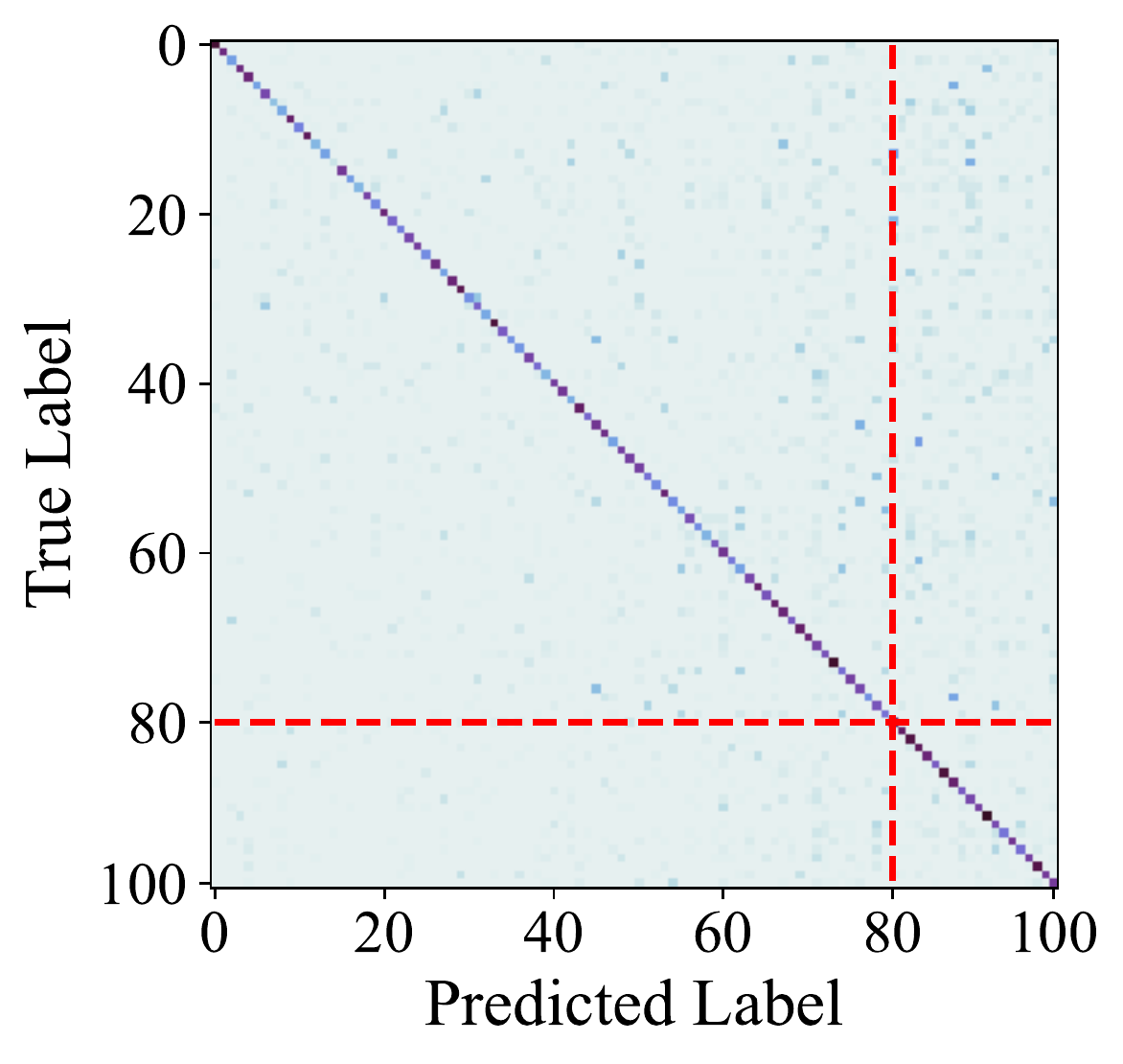}
		}
		\hfill
		\subfigure[FOSTER]
		{	\includegraphics[width=.36\columnwidth]{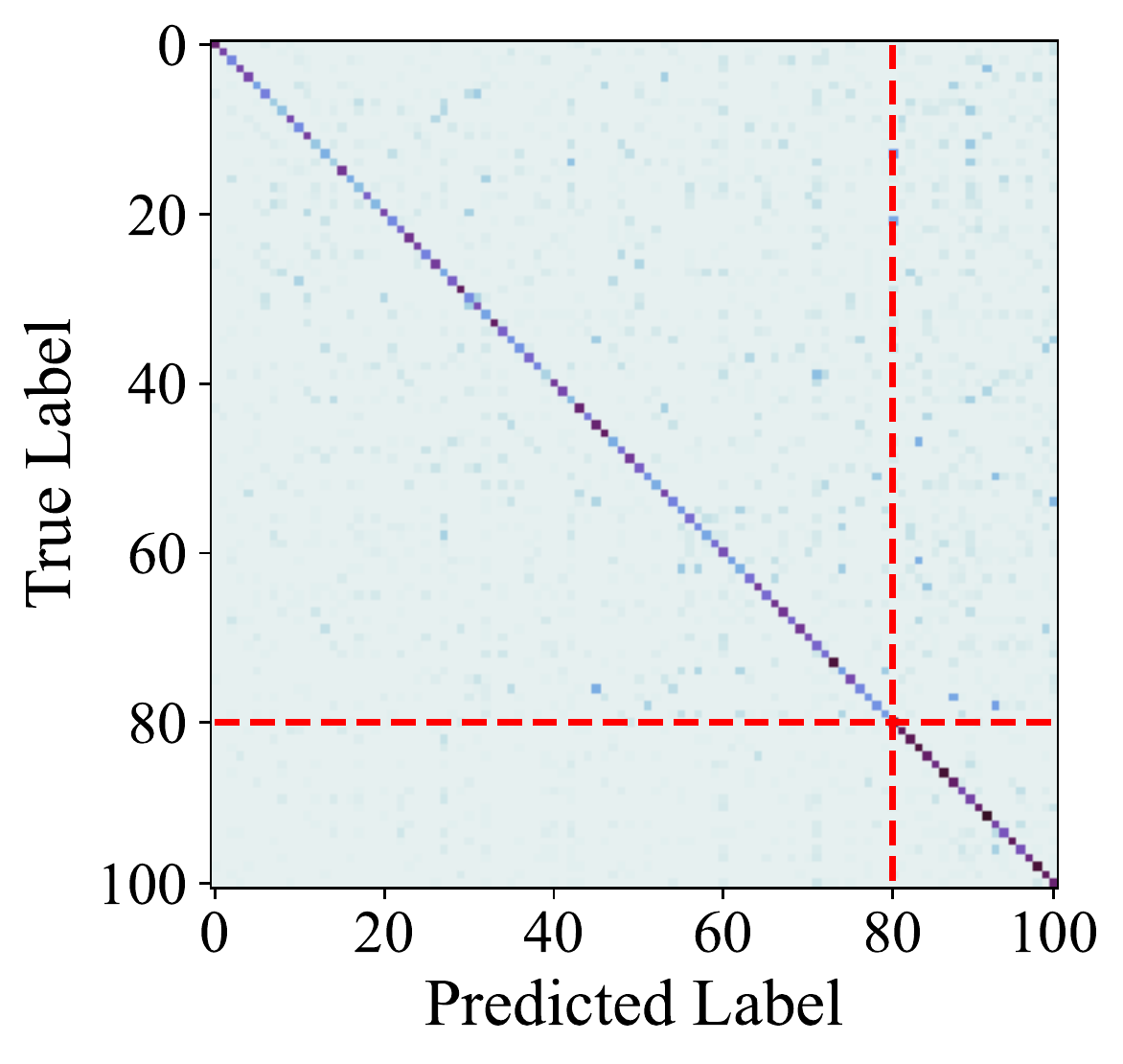}		
		}
		\hfill
		\subfigure[MEMO]
		{	\includegraphics[width=.36\columnwidth]{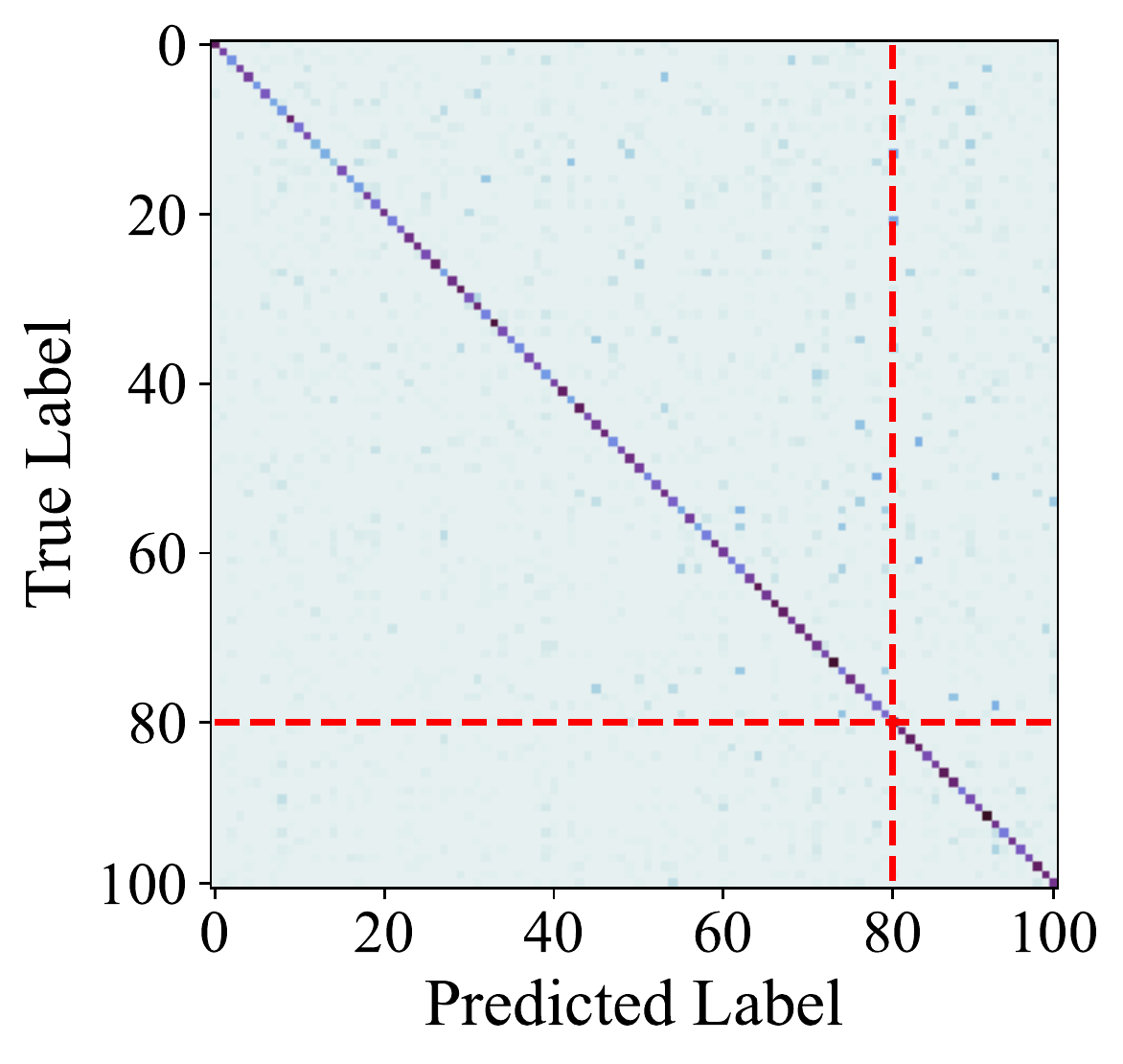}	
		}
		\hfill
		\subfigure[Dytox]
		{	\includegraphics[width=.36\columnwidth]{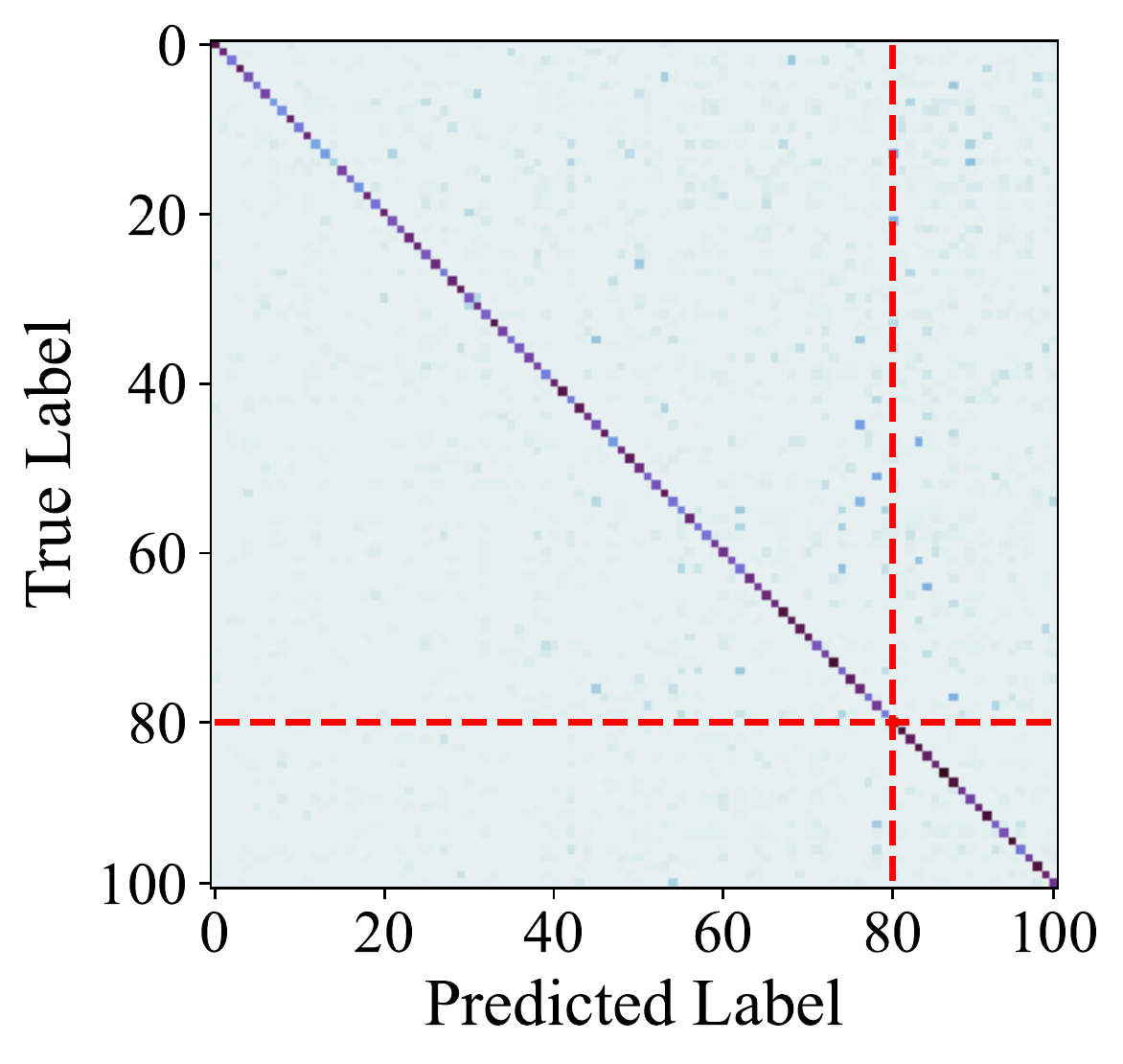}	
		}
	\end{center}
	\caption{Confusion matrix of different methods on CIFAR100 Base0 Inc20 setting after the last incremental stage.
	} \label{figure:confusion_matrix_full}
\end{figure*}

\begin{figure*}[t]
	\begin{center}
		
		\subfigure[Finetune ]
		{	\includegraphics[width=.36\columnwidth]{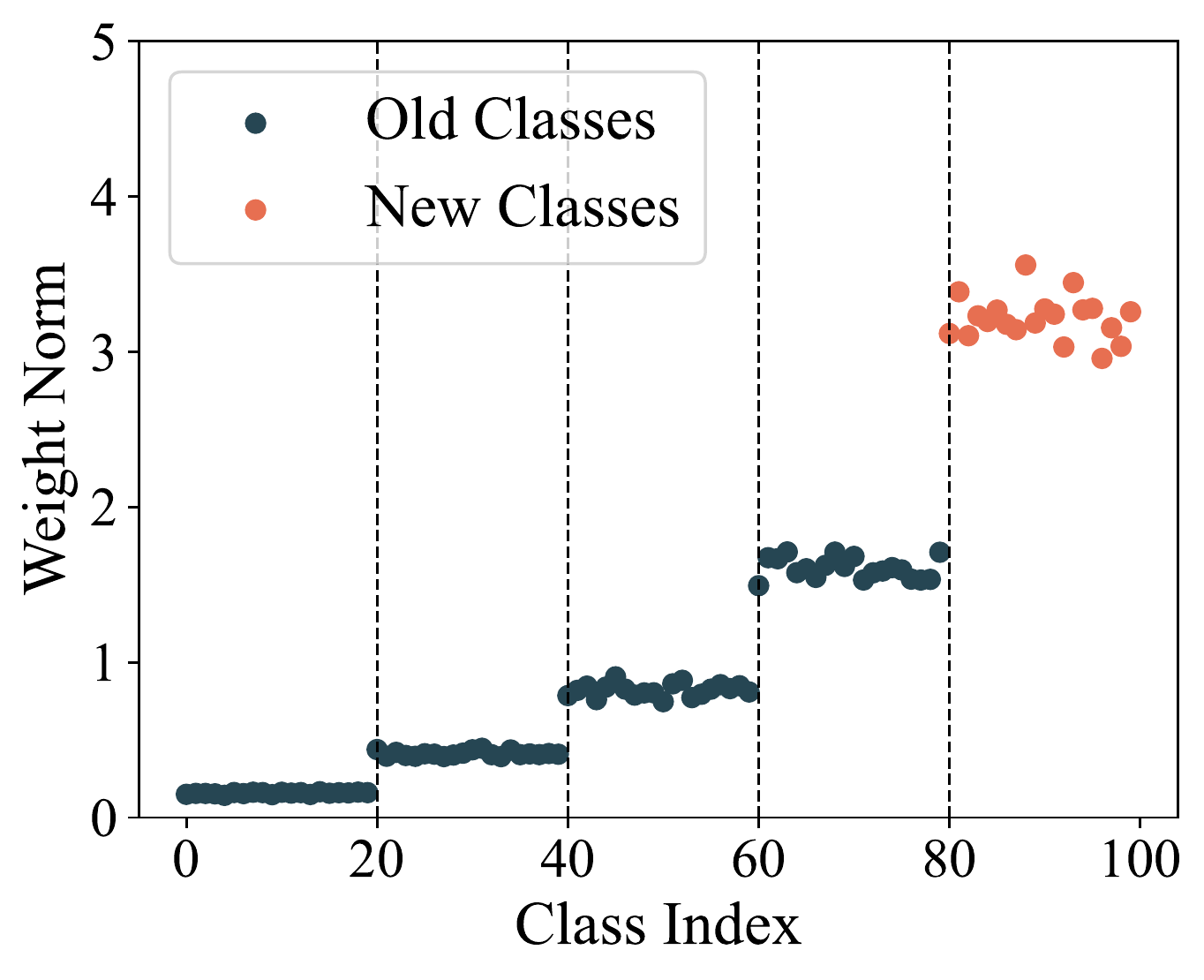}
						\label{fig:fc-ft}
		}
		\hfill
		\subfigure[EWC]
		{	\includegraphics[width=.36\columnwidth]{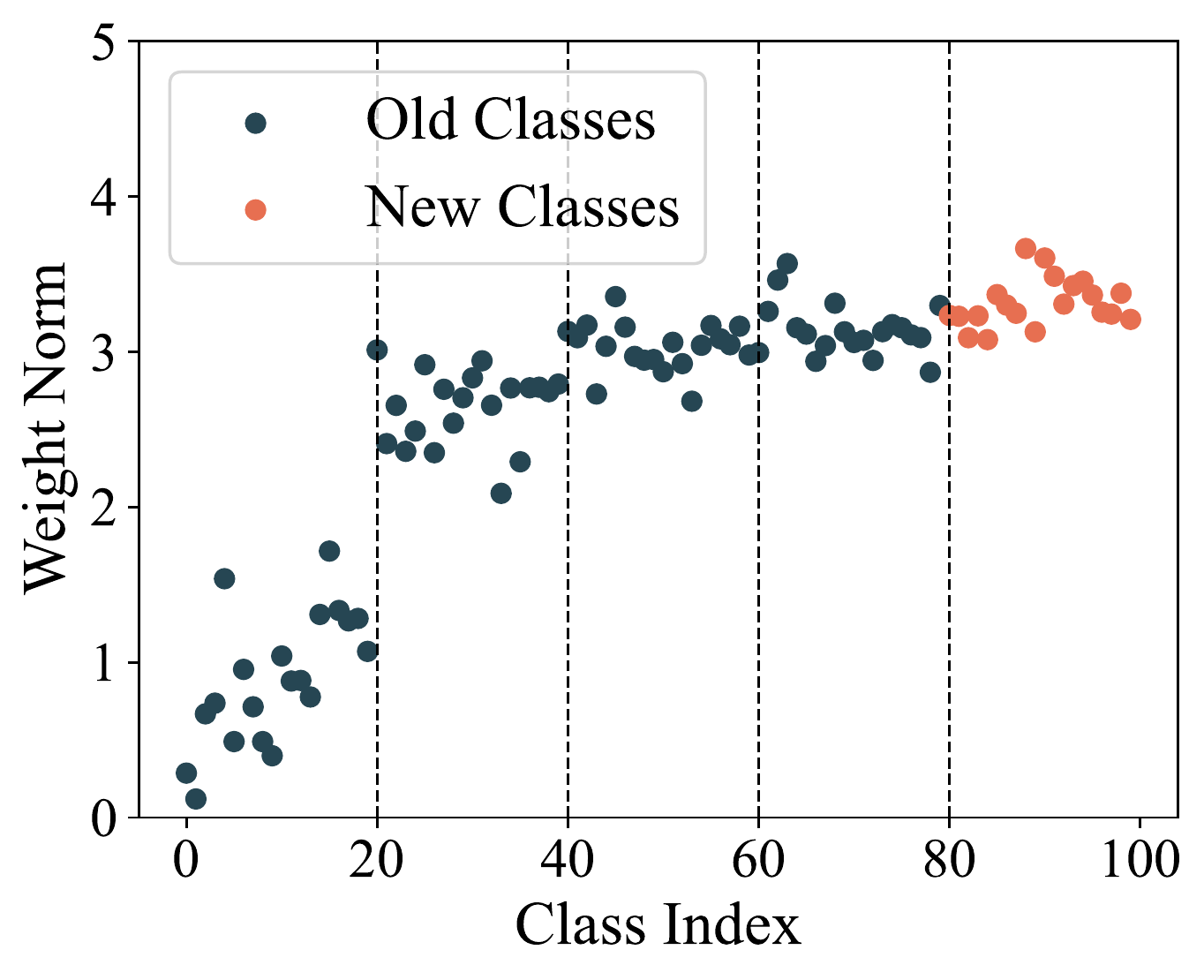}		
		}
		\hfill
		\subfigure[LwF]
		{	\includegraphics[width=.36\columnwidth]{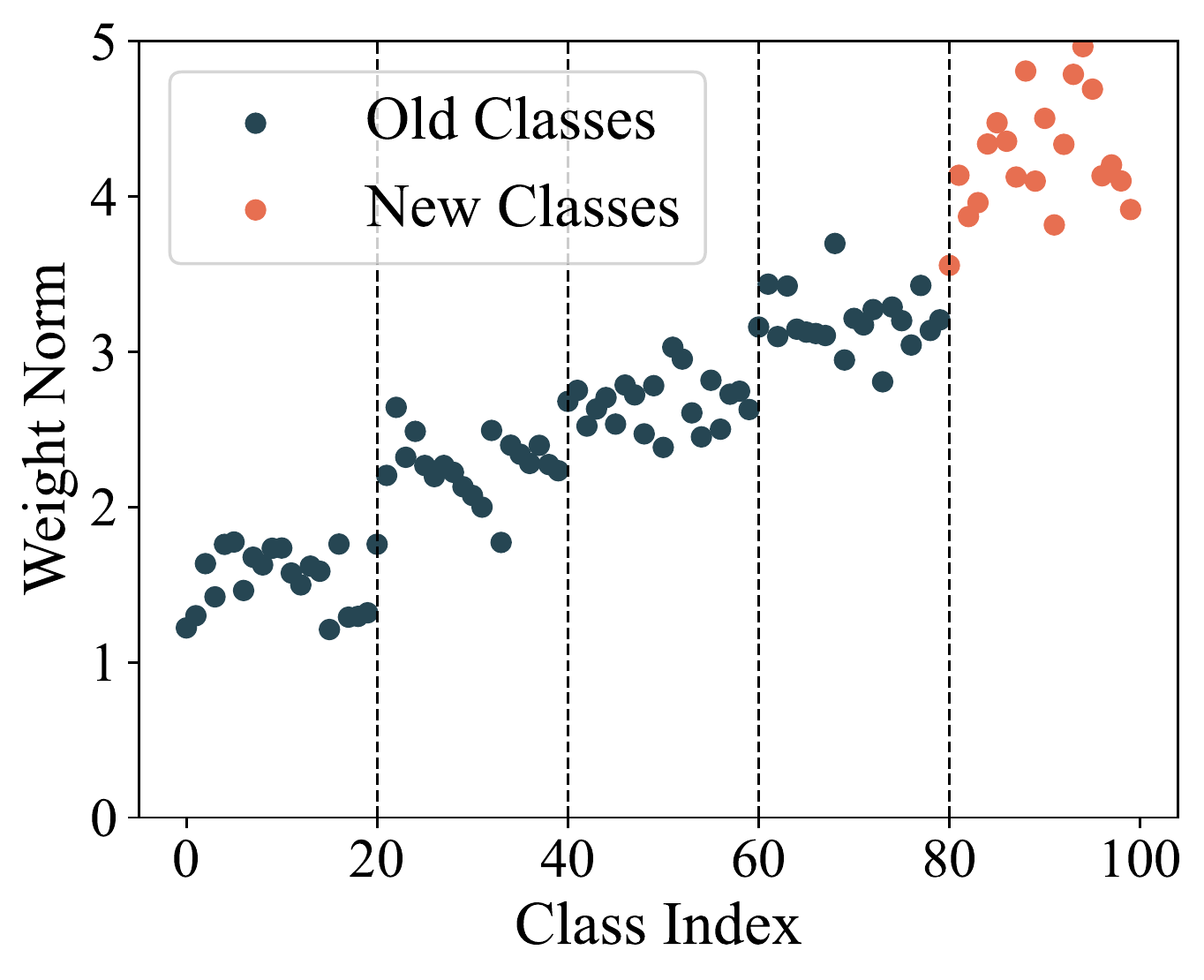}	
		}
		\hfill
		\subfigure[Replay]
		{	\includegraphics[width=.36\columnwidth]{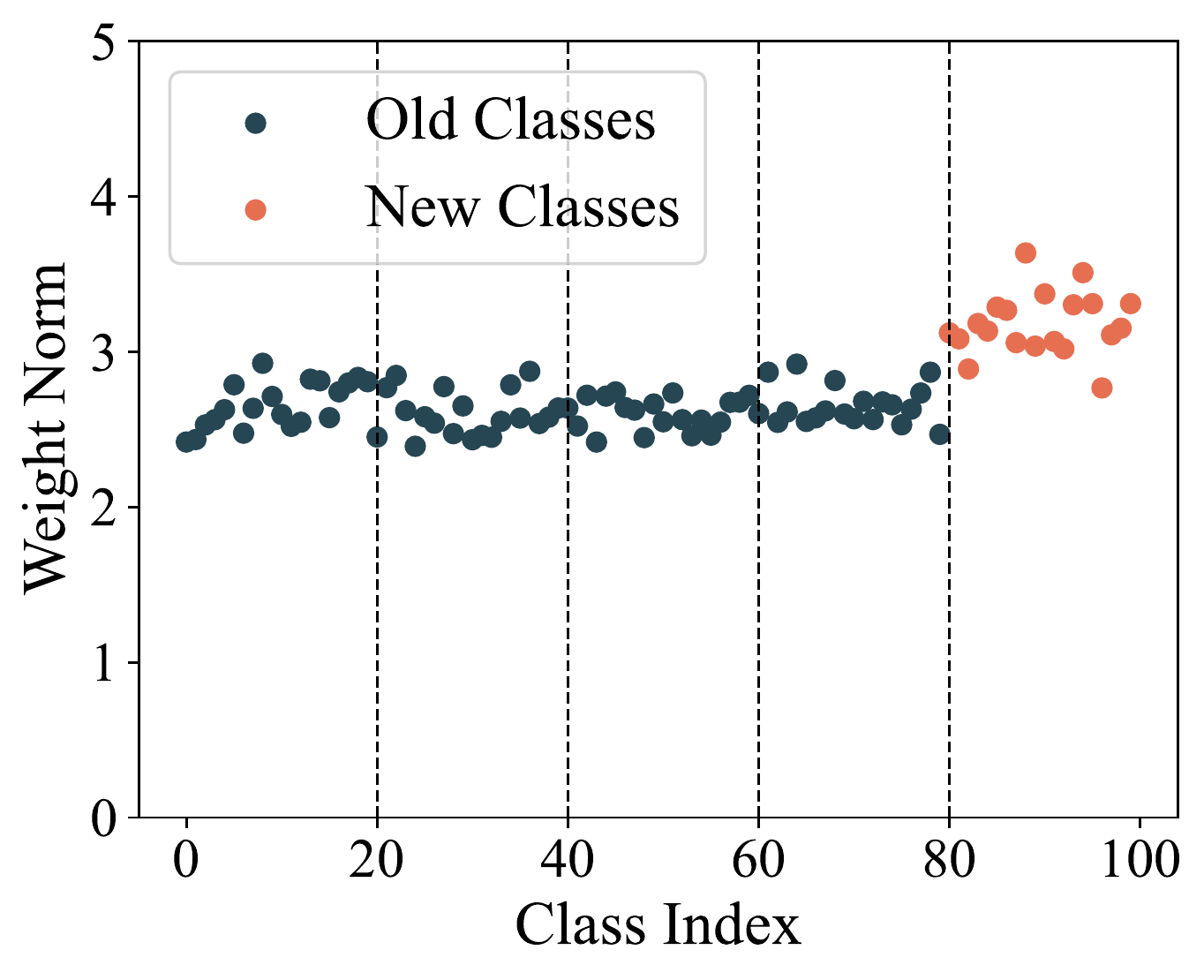}
		}
		\hfill
		\subfigure[GEM]
		{	\includegraphics[width=.36\columnwidth]{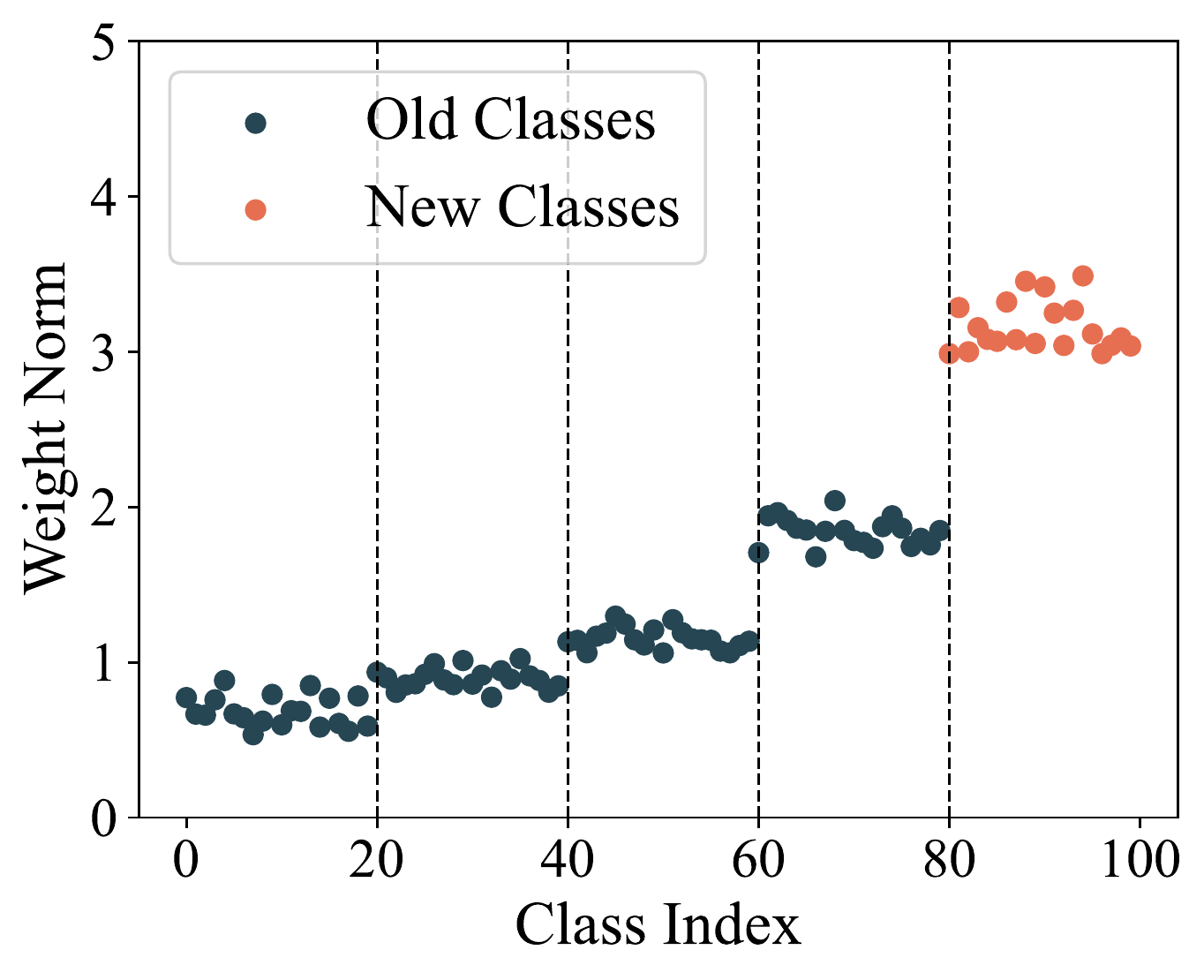}
		} \\
		\subfigure[iCaRL]
		{	\includegraphics[width=.36\columnwidth]{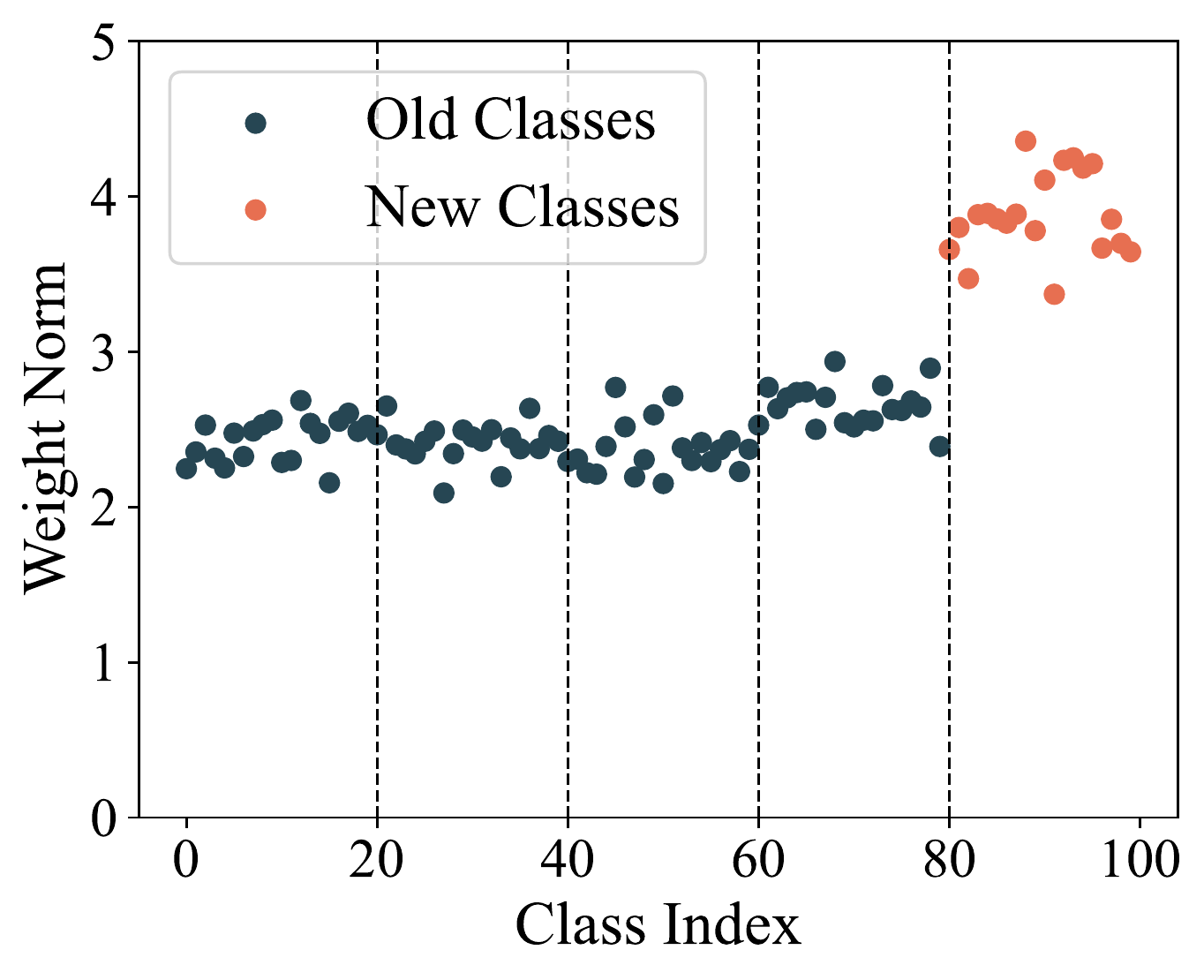}
		}
		\hfill
		\subfigure[BiC ]
		{	\includegraphics[width=.36\columnwidth]{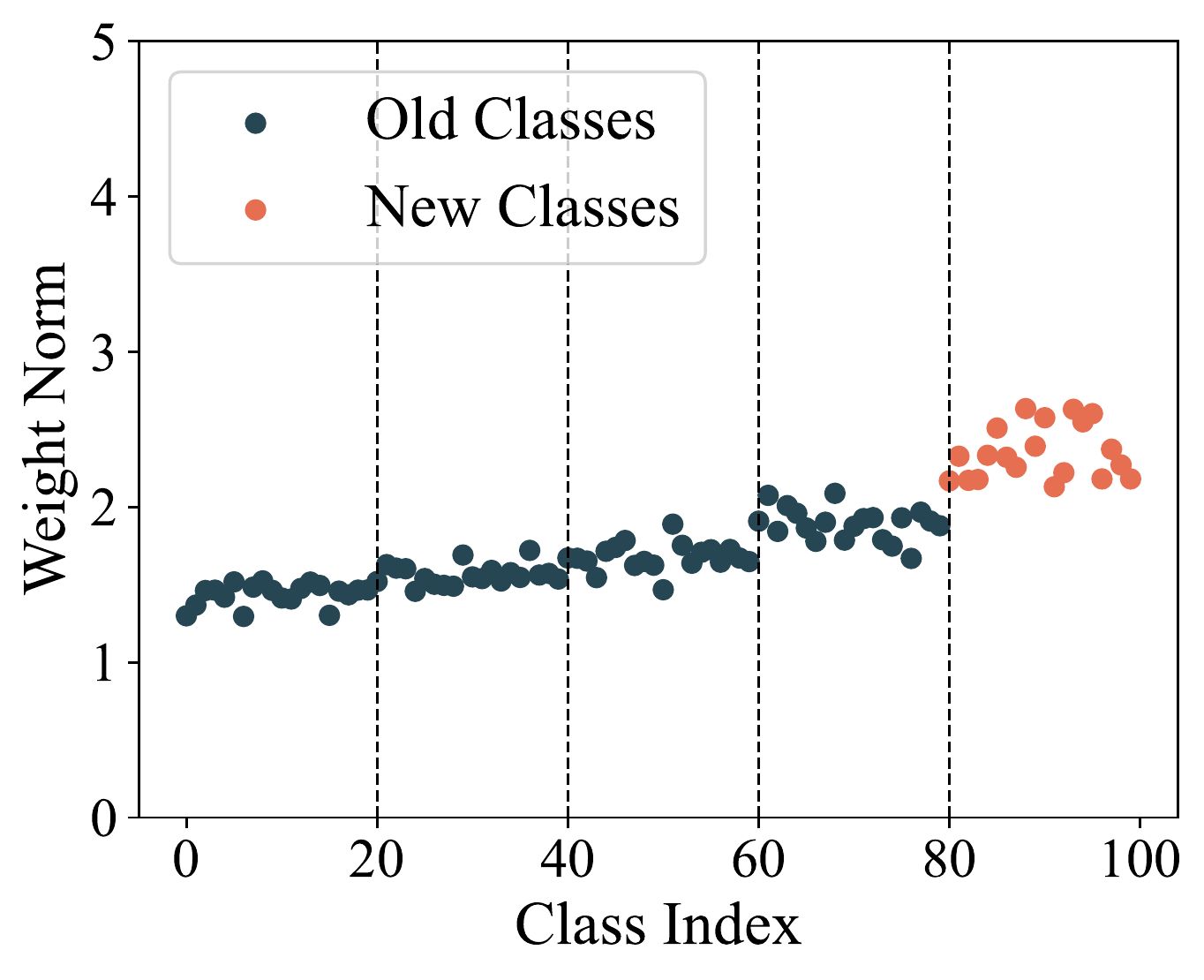}
		}
		\hfill
		\subfigure[WA]
		{	\includegraphics[width=.36\columnwidth]{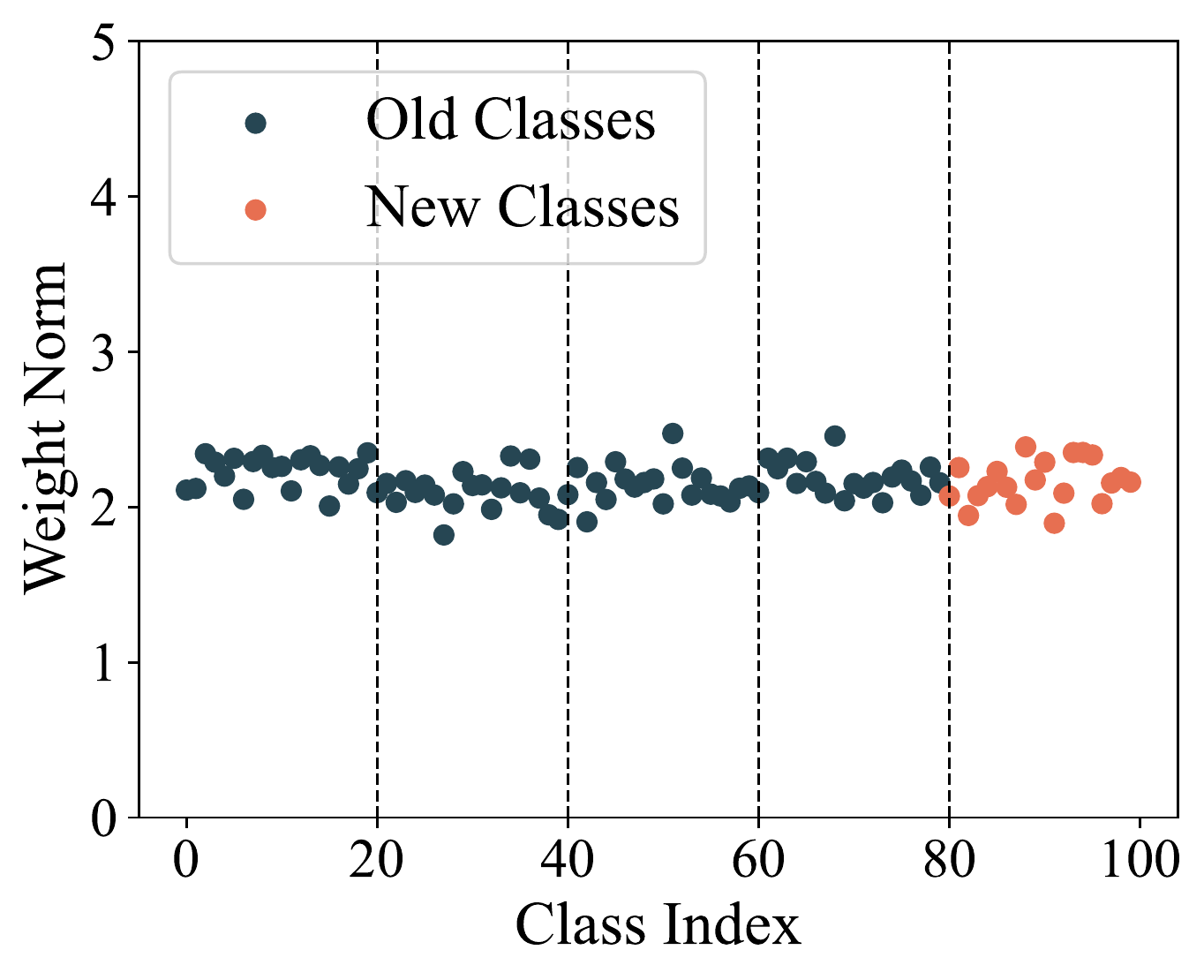}		
		}
		\hfill
		\subfigure[PODNet]
		{	\includegraphics[width=.36\columnwidth]{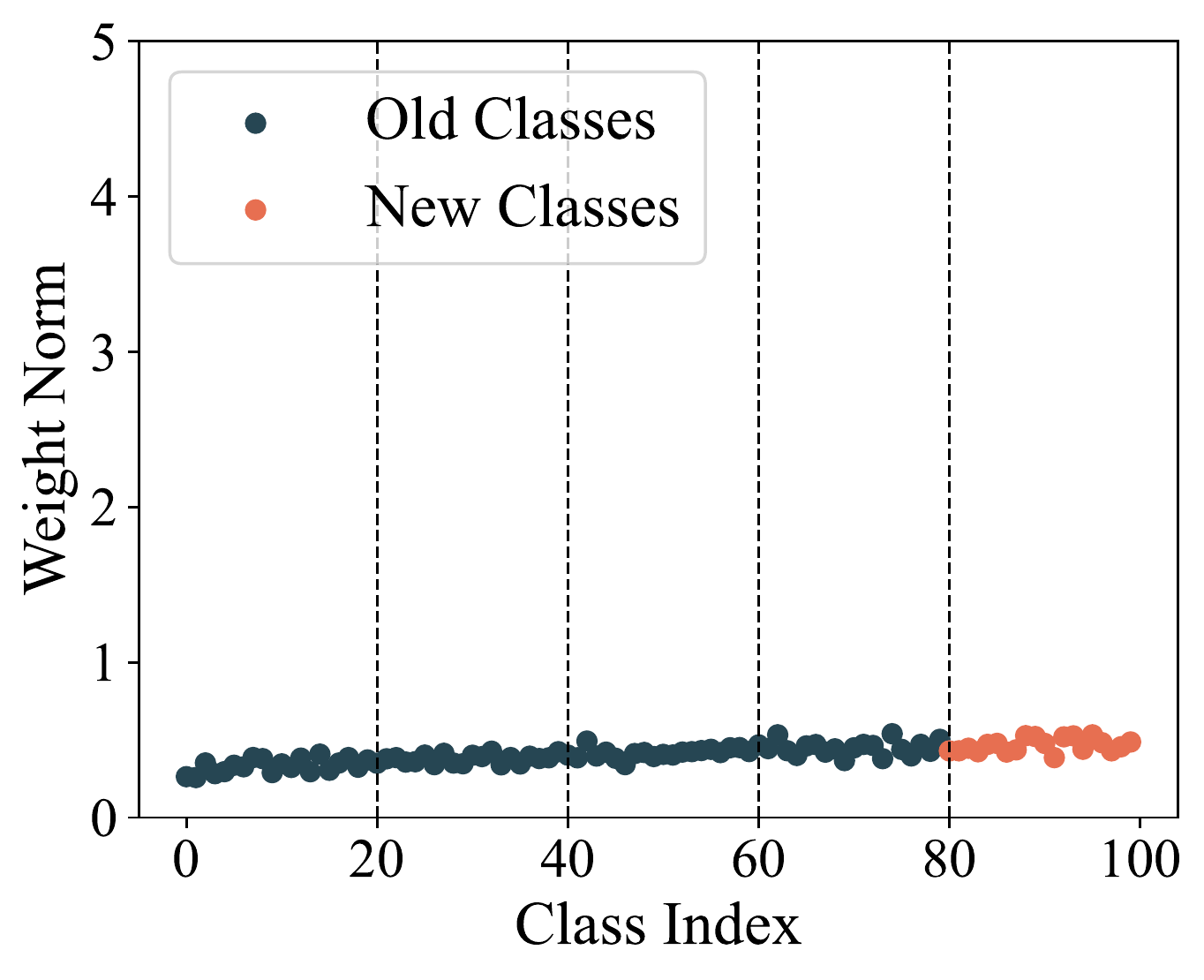}	
		}
		\hfill
		\subfigure[DER]
		{	\includegraphics[width=.36\columnwidth]{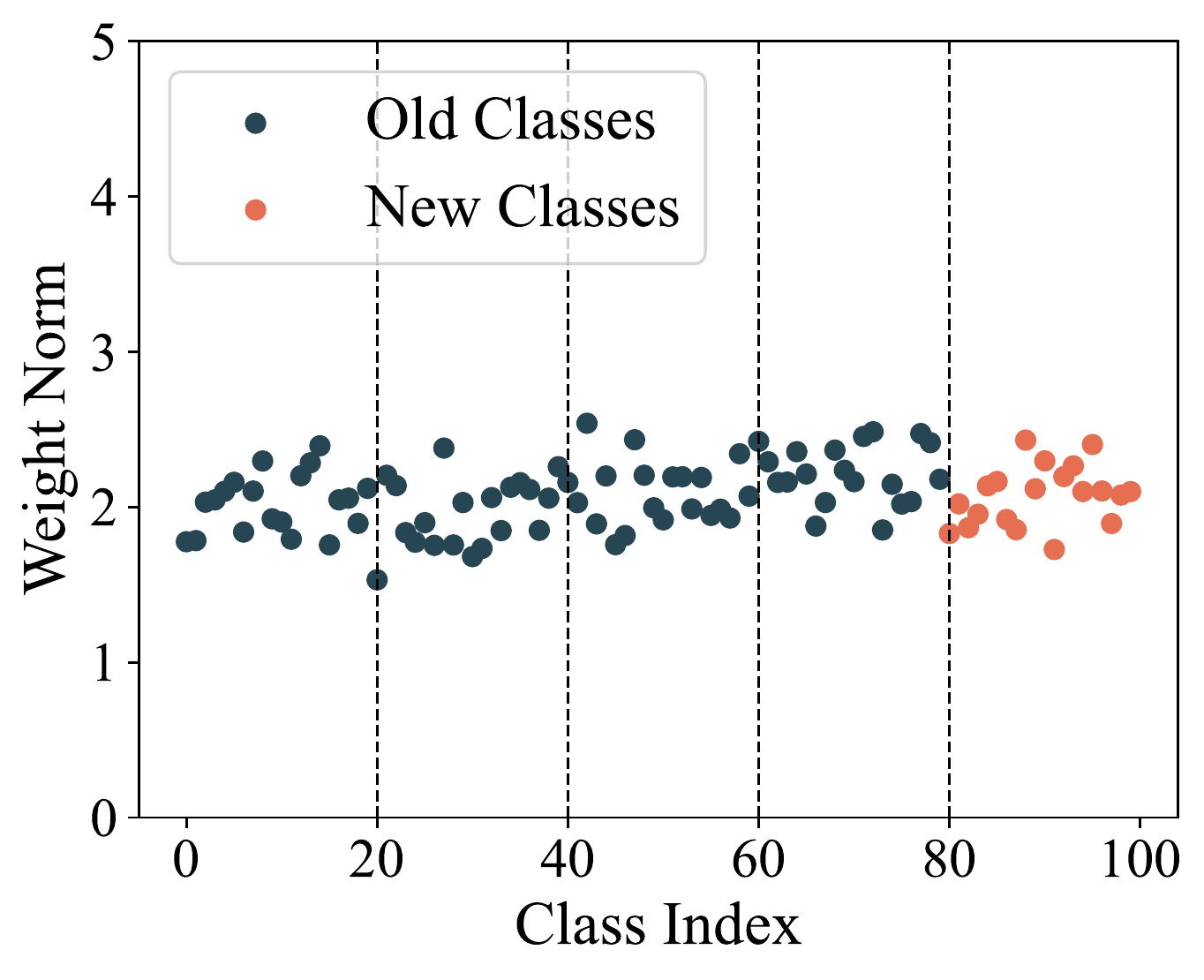}	
		}
		\\
		\subfigure[RMM+FOSTER]
		{	\includegraphics[width=.36\columnwidth]{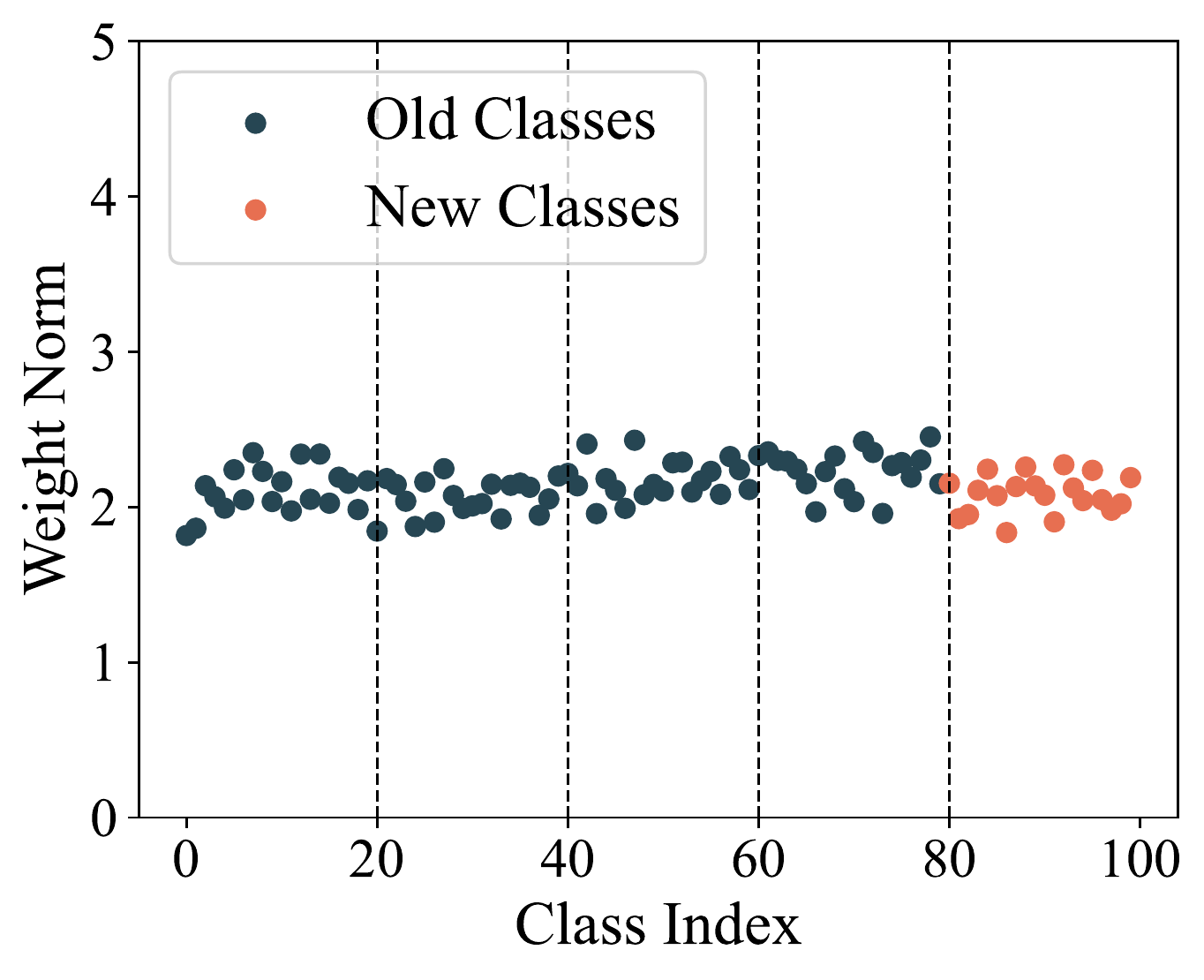}
		}
		\hfill
		\subfigure[Coil ]
		{	\includegraphics[width=.36\columnwidth]{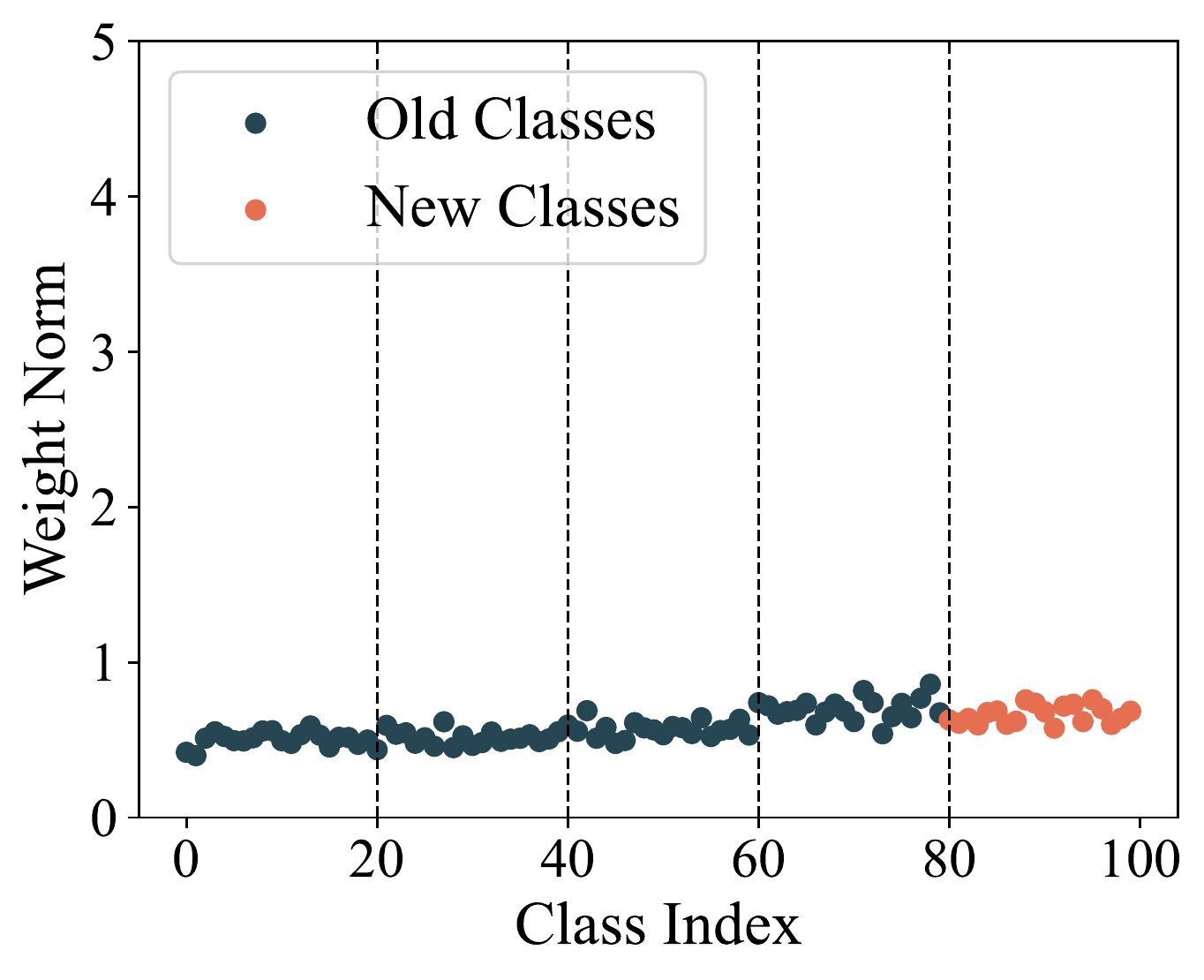}
		}
		\hfill
		\subfigure[FOSTER]
		{	\includegraphics[width=.36\columnwidth]{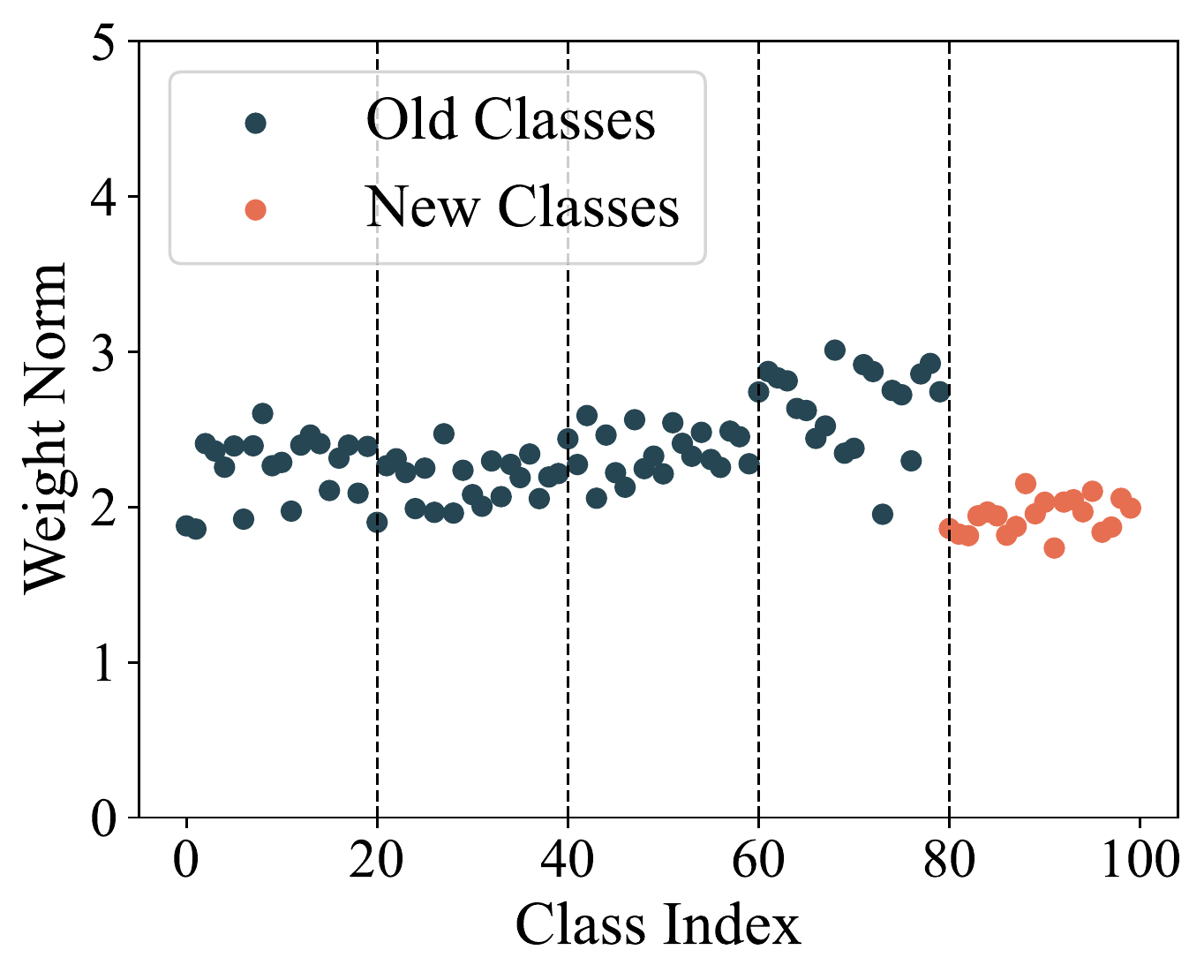}		
		}
		\hfill
		\subfigure[MEMO]
		{	\includegraphics[width=.36\columnwidth]{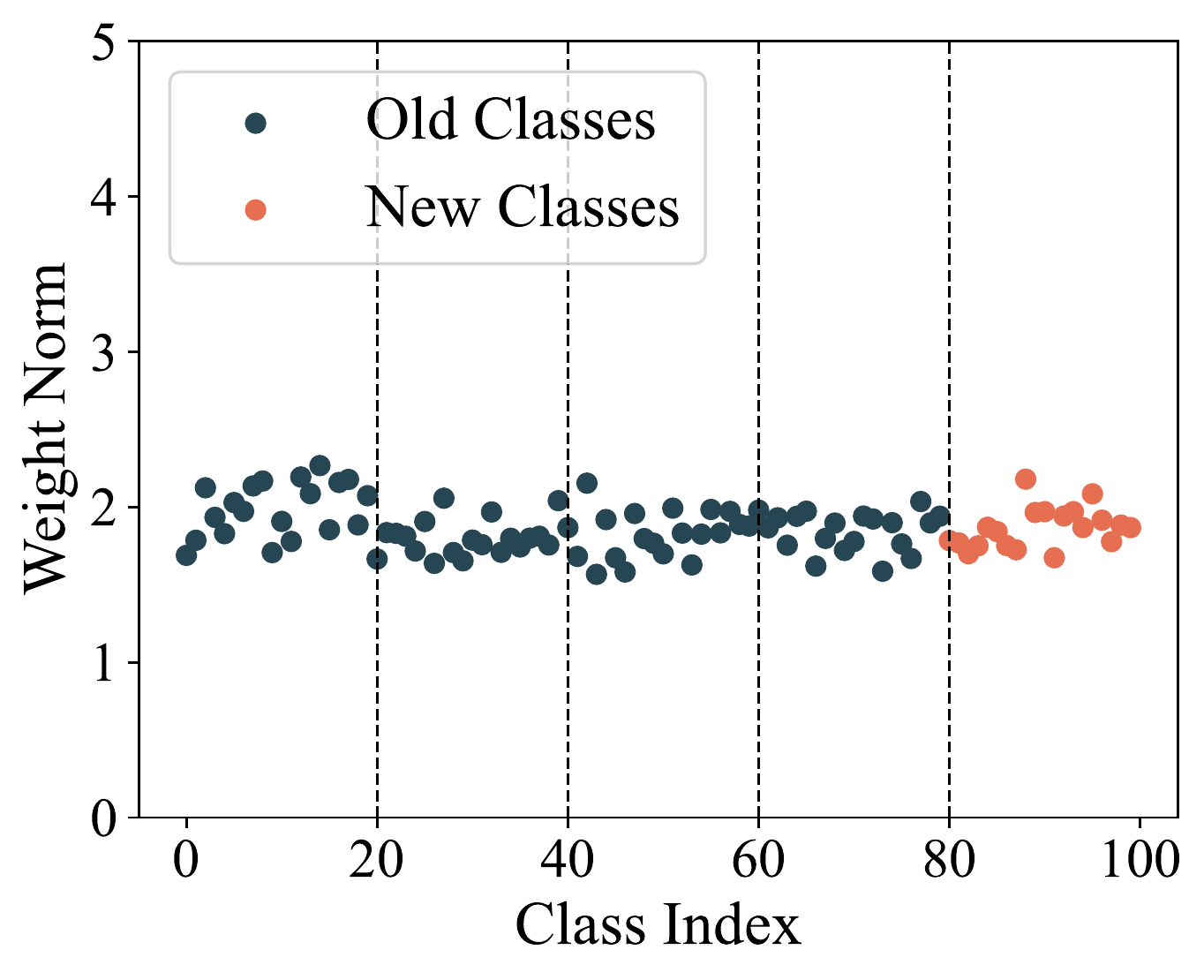}	
		}
		\hfill
		\subfigure[Dytox]
		{	\includegraphics[width=.36\columnwidth]{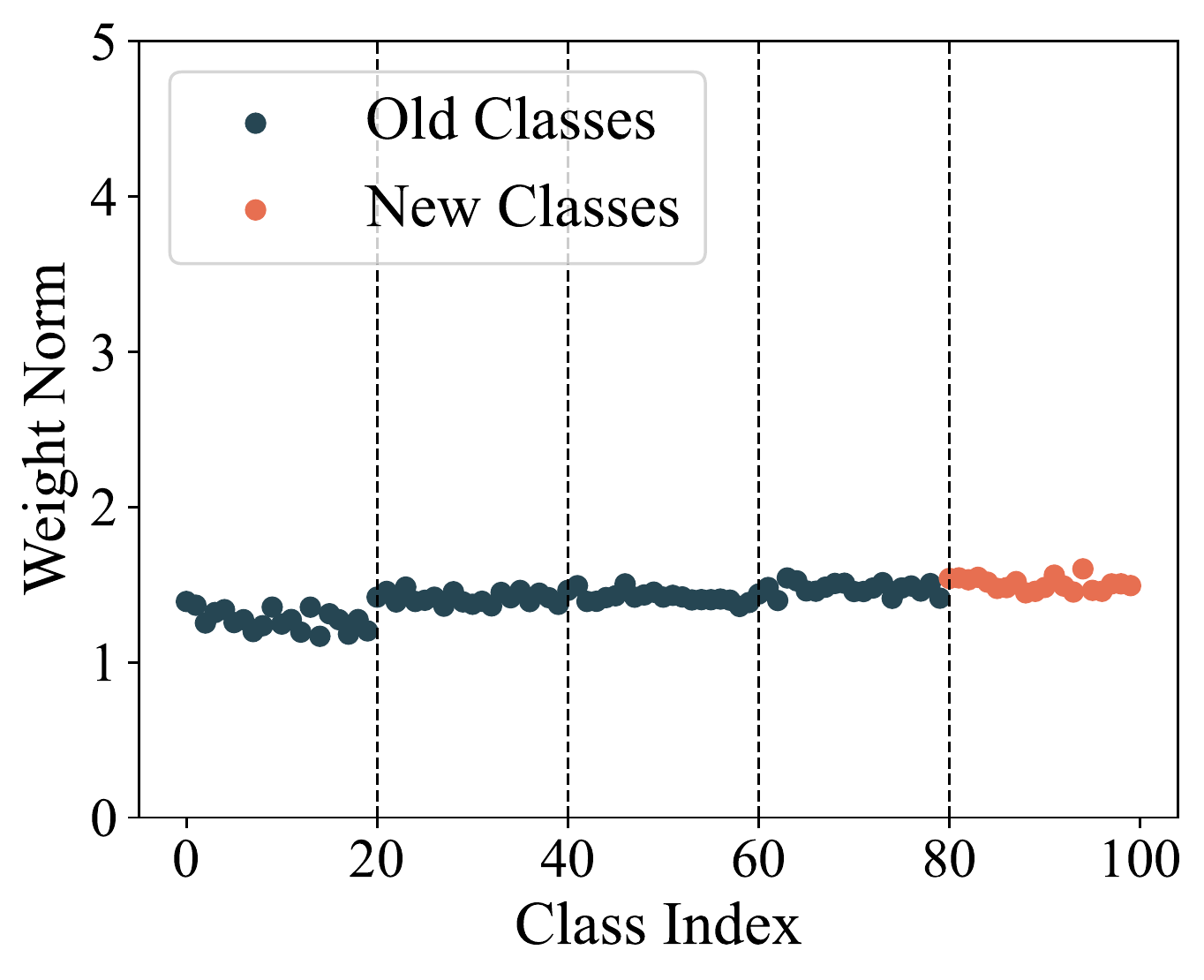}	
		}
	\end{center}
	\caption{Weight norm of different methods on CIFAR100 Base0 Inc20 setting. We visualize the norm after the last incremental stage.
	} \label{figure:weight_norm}
\end{figure*}

\section{Visualization Results} \label{sec:supp_vis}

In this section, we report the full visualization results in the CIFAR100 comparison. Specifically, we first show the confusion matrix of 15 methods in Figure~\ref{figure:confusion_matrix_full}. As we can infer from these figures, Finetune and EWC forget most knowledge from former classes, while LwF shows less forgetting with knowledge distillation. When comparing LwF to EWC, we find that knowledge distillation is a more suitable solution to resist catastrophic forgetting than parameter regularization without saving exemplars.
 When equipping the model with exemplars, the diagonal elements become brighter, implying that forgetting is alleviated.
The performance of Replay shows more steady improvement than finetune, indicating the critical role of exemplars in CIL.
However, when comparing iCaRL to GEM, we can find that data regularization shows inferior performance than knowledge distillation in the efficiency of using exemplars. Moreover, BiC and WA improve the performance of iCaRL by rectifying the bias in the model, indicating that these methods can be orthogonally equipped to other CIL methods for rectification. Lastly, dynamic networks (\ie, DER,
FOSTER, MEMO, and DyTox) still show the most competitive performance in these visualizations, indicating that the former backbones help the model retrieve old features and resist forgetting.

Additionally, we visualize the weight norm of the fully-connected layers in Figure~\ref{figure:weight_norm} to show the bias introduced by incremental learning. As we can infer from these figures, when sequentially finetuning the model, the weight norm of new classes will dominate the learning process due to data imbalance. As shown in Figure~\ref{fig:fc-ft}, the weight norm of finetune has a stair-step shape. Since the features are passed through the ReLU layer to avoid negative ones, the larger weight norm of new classes will result in an imbalance in the prediction results. These results are consistent with the confusion matrix in Figure~\ref{figure:confusion_matrix_full}, and we find that most CIL methods can relieve the imbalance in the fully connected layers.

\begin{table*}[t]
	\caption{ Average and last accuracy performance comparison with aligned memory cost.
		`\#P' represents the number of parameters (million). 
		`\#$\mathcal{E}$' denotes the number of exemplars, and `MS' denotes the memory size (MB).
	}\label{tab:fair_comparison_full}
	\centering
	\begin{tabular}{@{}lccccccccccccccc}
		\toprule
		\multicolumn{1}{c}{\multirow{2}{*}{Method}} & 
		\multicolumn{5}{c}{CIFAR100 Base0 Inc5} &
		\multicolumn{5}{c}{CIFAR100 Base0 Inc10} & \multicolumn{5}{c}{ImageNet100 Base0 Inc10} \\
		& {\#P} & {\#{$\mathcal{E}$}} & MS& {$\bar{\mathcal{A}}$} & ${\mathcal{A}_B}$ 
		& {\#P} & {\#{$\mathcal{E}$}} & MS& {$\bar{\mathcal{A}}$} & ${\mathcal{A}_B}$& {\#P} & {\#{$\mathcal{E}$}} & MS & {$\bar{\mathcal{A}}$} & ${\mathcal{A}_B}$
		\\
		\midrule
		GEM  & 0.46 & 13466& 41.22&20.30&7.99 &0.46 & 7431 &23.5 & 27.03 & 10.72 & -& -& -& - & -  \\
		Replay&0.46 & 13466& 41.22&74.25& 61.09 &0.46 & 7431 &23.5 & 69.97 & 55.61 & 11.17 & 4671 & 713.2&67.46 & 53.72\\
		iCaRL &0.46 & 13466& 41.22&\bf  75.17&61.88  &0.46 & 7431 &23.5 & 70.94 &58.52& 11.17 & 4671 & 713.2&67.16 &51.58 \\
		PODNet&0.46 & 13466& 41.22&59.03&42.98 &0.46 & 7431 &23.5 & 60.80 & 45.38& 11.17 & 4671 & 713.2&68.43&53.44\\
		Coil &  0.46 & 13466& 41.22&74.39&60.78 &0.46 & 7431 &23.5 & 70.69 & 54.40& 11.17 & 4671 & 713.2&69.97&54.24\\
		WA &0.46 & 13466& 41.22&74.07&\bf  63.18&0.46 & 7431 &23.5 & 69.55 & 59.26& 11.17 & 4671 & 713.2&72.62&62.12\\
		BiC &0.46 & 13466& 41.22&71.27&60.98 &0.46 & 7431 &23.5 &70.69& 59.60& 11.17 & 4671 & 713.2&72.83&61.98\\
		FOSTER& 0.46 & 13466& 41.22&75.06&62.82  &0.46 & 7431 &23.5 &72.28 & 59.39& 11.17 &4671 & 713.2&71.55&59.86\\
		DER& 9.27&2000 &41.22&70.41&57.85  &4.60 & 2000 &23.5&71.47&60.26& 111.7 & 2000 & 713.2&\bf 75.70&\bf 67.14 \\
		MEMO & 7.14& 4771& 41.22&73.57&62.97  &3.62 & 3312 &23.5 & \bf 72.37& \bf 61.98& 86.72 & 2652 & 713.2&72.55&64.08\\
		\bottomrule
	\end{tabular}
\end{table*}

\section{Supplied Results in Memory-Aligned Comparison}  \label{sec:supp_memory_aligned}
In the main paper, we report the incremental performance curve with aligned memory to DER. In this section, we report the detailed setting of those figures, including the number of exemplars, the number of parameters, the total memory size, and the detailed performance in Table~\ref{tab:fair_comparison_full}. Specifically, since DER and MEMO require an extra model budget compared to other methods, we equip more exemplars to other methods to make sure the total memory size is aligned to the same scale.

 As we can infer from the table, the gap between different methods becomes smaller when the total budget is aligned to the same scale. Specifically, the typical baseline method iCaRL shows the best average accuracy in the CIFAR100 Base0 Inc5 setting. In other words, saving exemplars can be more memory-efficient than saving backbones when fairly compared.

\begin{figure*}[t]
	\begin{center}
		
		\subfigure[Average accuracy curve of CIFAR100 Base0 Inc10 ]
		{	\includegraphics[width=.96\columnwidth]{pics/avgAUC_CIFAR} \label{figure:avg_auc_a}
		}
		\subfigure[Average accuracy curve of ImageNet100 Base50 Inc5]
		{	\includegraphics[width=.96\columnwidth]{pics/avgAUC_ImageNet}		
		}
		\subfigure[Last accuracy curve of CIFAR100 Base0 Inc10]
		{	\includegraphics[width=.96\columnwidth]{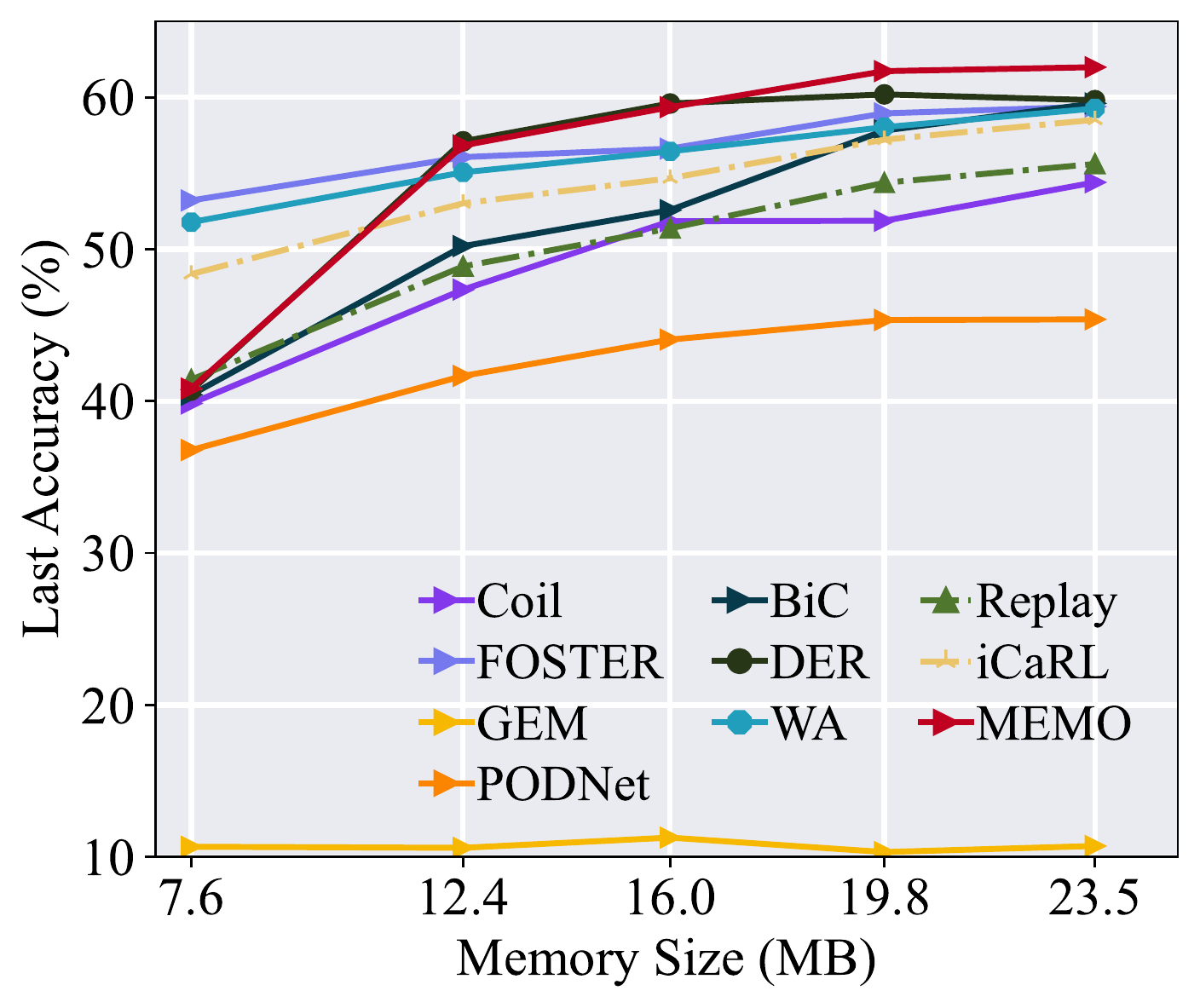}
		}
		\subfigure[Last accuracy curve of ImageNet100 Base50 Inc5]
		{	\includegraphics[width=.96\columnwidth]{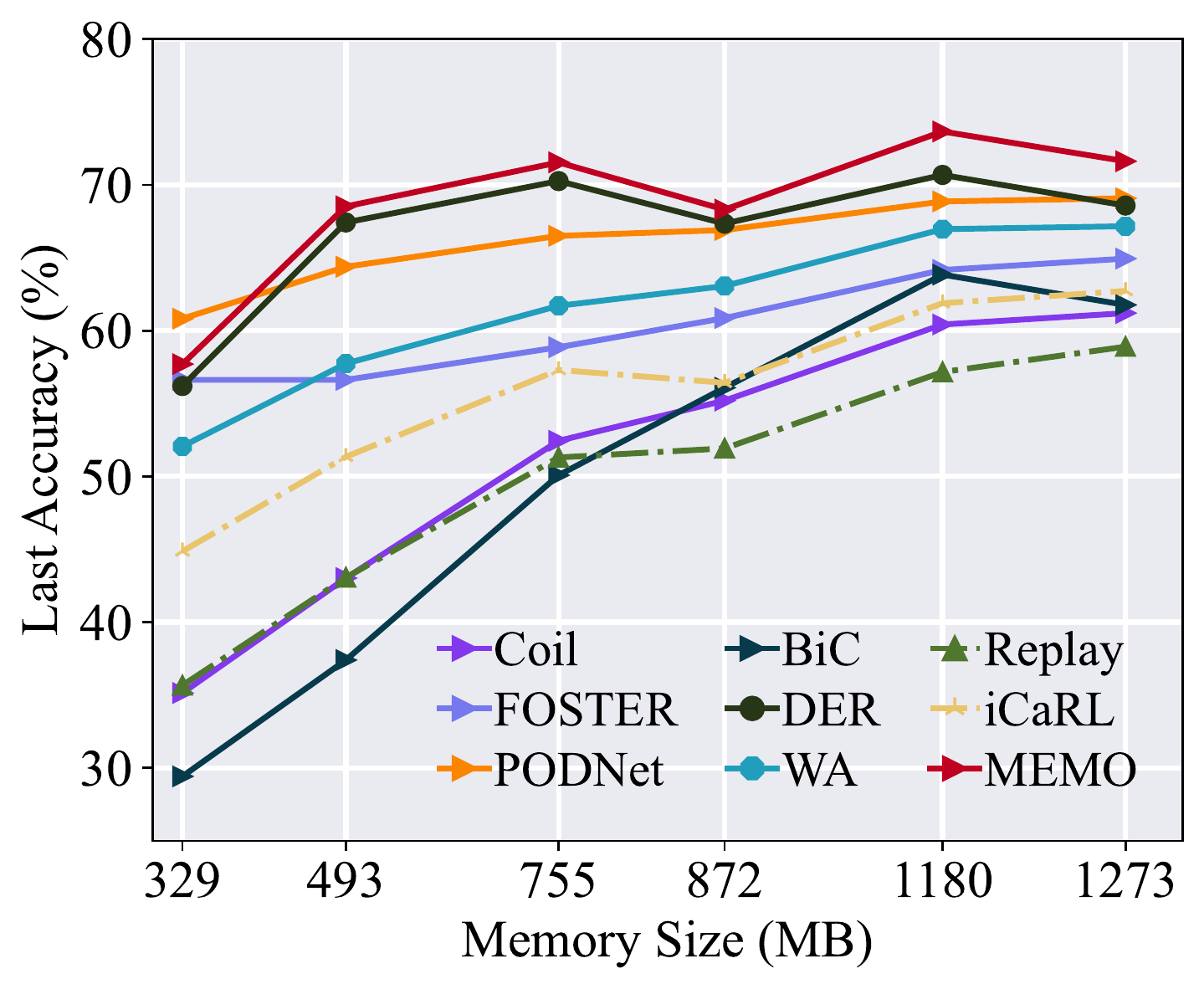}		
		}
		
	\end{center}
	\caption{Performance-memory curve of different methods with different datasets.
	} \label{figure:auc_measure_last}
\end{figure*}

\section{Details About the AUC-A/L Measure} \label{sec:supp_AUC}
In this section, we report the details in the memory-agnostic measure, \ie, AUC-A/L for class-incremental learning. We first highlight the importance of this measure and then give the implementation details of the curve, including the per-node performance and memory alignment details. We will start with CIFAR100~\cite{krizhevsky2009learning} and then discuss ImageNet100~\cite{deng2009imagenet}.

\subsection{Necessity of AUC in CIL} 

We report the full performance-memory curve in Figure~\ref{figure:auc_measure_last}, including the average and last accuracy-memory curves.
In each figure, the X-coordinate indicates the memory budget of the specific model, and the Y-coordinate represents the corresponding performance of the model.
To highlight the importance of the AUC measure, we begin with a simple question:
\begin{displayquote} 
	Given \emph{the same memory budget}, which algorithm should I choose for class-incremental learning? iCaRL or DER?
\end{displayquote}
It is a simple question that aims to explore a better algorithm with the same memory budget. However, when we compare the performance of iCaRL and DER in Figure~\ref{figure:avg_auc_a}, we find an intersection among these methods. Specifically, we find iCaRL has better accuracy when the memory size is 7.4 MB, while DER works better when the memory size is 23.5 MB. This results in a {\em contradictory} conclusion, and we cannot find a model that suits all memory budgets.

To this end, calculating AUC does not rely on the specific budget, and we can tell from the AUC table that DER has a better AUC performance, which means it has better expandability. In other words, accuracy can only measure the performance given a specific X-coordinate. In contrast, AUC measures the area under the incremental performance curve holistically.

\subsection{Implementation Details}
We give the implementation details in plotting the performance-memory curve in this section.
In the following discussions, we use $\mathcal{E}$ to represent the exemplar set. \#$\mathcal{E}$ denotes the number of exemplars, and $S(\mathcal{E})$ represents the memory size (in MB) that saving these exemplars consume. 
Following the benchmark implementation in the main paper, 2,000 exemplars are saved for every method for CIFAR100 and ImageNet100. Hence, the exemplar size of each method is denoted as \#$\mathcal{E}$$=2000+E$, where $E$ corresponds to the extra exemplars exchanged from the model size, as discussed in the main paper. We use `\#P' to represent the number of parameters and `MS' to represent the memory budget (in MB) it costs to save this model in memory.  
The total memory size (\ie, numbers on the X coordinate) is the summation of exemplars and the models:
\begin{align}
	\text{Memory Size}=\text{Model Size}+\text{Exemplar Size}\,,
\end{align}
which should be aligned for fair comparison as we advocated.

Specifically, {\it we can divide the selected methods into two groups}. The first group contains GEM~\cite{lopez2017gradient}, iCaRL~\cite{rebuffi2017icarl}, Replay~\cite{ratcliff1990connectionist}, WA~\cite{zhao2020maintaining}, PODNet~\cite{douillard2020podnet}, Coil~\cite{zhou2021co}, BiC~\cite{wu2019large} and FOSTER~\cite{wang2022foster}, whose use a single backbone for incremental learning in the inference stage. 
 Hence, they use the same network backbone with the same model size and equal memory sizes, and we denote them as {\bf SingleNet}.
  The second group contains DER~\cite{yan2021dynamically} and MEMO~\cite{zhou2022model}, which require more memory budget to save the extra model during inference. Specifically, DER sacrifices the memory size to store the backbone from history, which consumes the largest model size. On the other hand, compared to DER, MEMO does not keep the duplicated generalized blocks from history and saves much memory size to change into exemplars. 
  
  \noindent\textbf{Discussion about selected methods:} In this part, we choose ten methods for the comparison, \ie, Finetune, EWC, LwF, RMM, DyTox, and L2P are not included in the evaluation. Firstly, non-exemplar-based methods (Finetune, EWC, LwF) are unsuitable for the evaluation since the extendability relies on adding extra exemplars. Secondly, RMM advocates another evaluation protocol in terms of the {\em memory}, which is incompatible with ours. Lastly, DyTox and L2P rely on the vision transformer as the backbone, which has a much larger memory scale than benchmark backbones. Directly comparing ResNet to ViT may be unfair since these backbones have different characteristics in model optimization.

\noindent\textbf{How to choose the budget list?} As we can infer from Figure~\ref{figure:auc_measure_last}, there are several selected memory sizes in the figures, which formulate the X-coordinate. Firstly, the start point corresponds to the memory size of the first group, \ie, a single backbone with $2,000$ exemplars. It is a relatively small budget, which can be seen as the budget for edge devices. Correspondingly, since DER has the largest model size among all compared methods, we set the endpoint to the memory size of DER. After selecting the start and end points, we choose several intermediate budgets to formulate the budget list. The intermediate budgets are set according to the parameter size of DER and MEMO, as shown below.

\noindent\textbf{How to set the exemplar and model size?} For SingleNet, we can easily extend them by adding the number of exemplars. Since the model size is fixed for them, adding exemplars enables these methods to extend to a larger scale. However, we cannot use the same backbone as SingleNet for DER and MEMO when the memory size is small. Hence, we divide the model parameters into ten equal parts (since there are ten incremental tasks in the CIFAR100 Base0 Inc10 setting) and look for a backbone with similar parameter numbers. For example, we separately use
ConvNet, ResNet14, ResNet20, and ResNet26 as the backbone for these methods to match different memory scales. 
Specifically, we use the same backbone for DER and MEMO and another same backbone for SingleNet to keep the total memory budget at the same scale. We annotate the backbone type in the figures of configurations, \eg, `ResNet32/ResNet14' means we use ResNet32 for SingleNet and ResNet14 for DER and MEMO.

In the following discussions, we give the memory figures and tables to illustrate the implementation of different methods and report their incremental performance.  We start with CIFAR100 and then discuss ImageNet100.

\subsubsection{CIFAR100 Implementations}
There are five X coordinates in the curve of CIFAR100, \eg, $7.6$, $12.4$, $16.0$, $19.8$, and $23.5$ MB. Following, we show the detailed implementation of different methods at these scales.

\begin{figure}[H]
	\begin{center}
		\subfigure[Memory Usage]
		{	\includegraphics[width=.47\columnwidth]{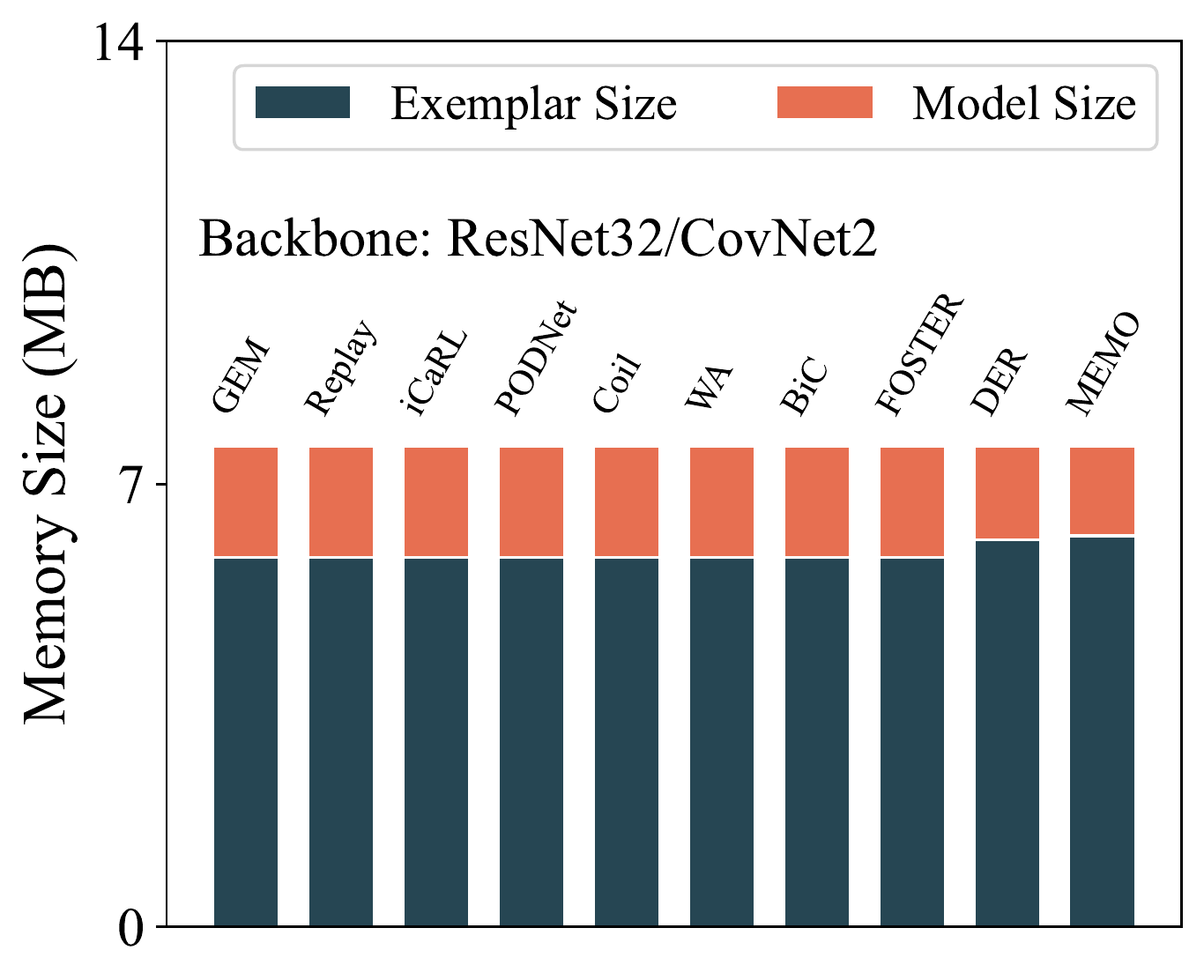}	
		}
		\subfigure[Performance Curve]
		{	\includegraphics[width=.47\columnwidth]{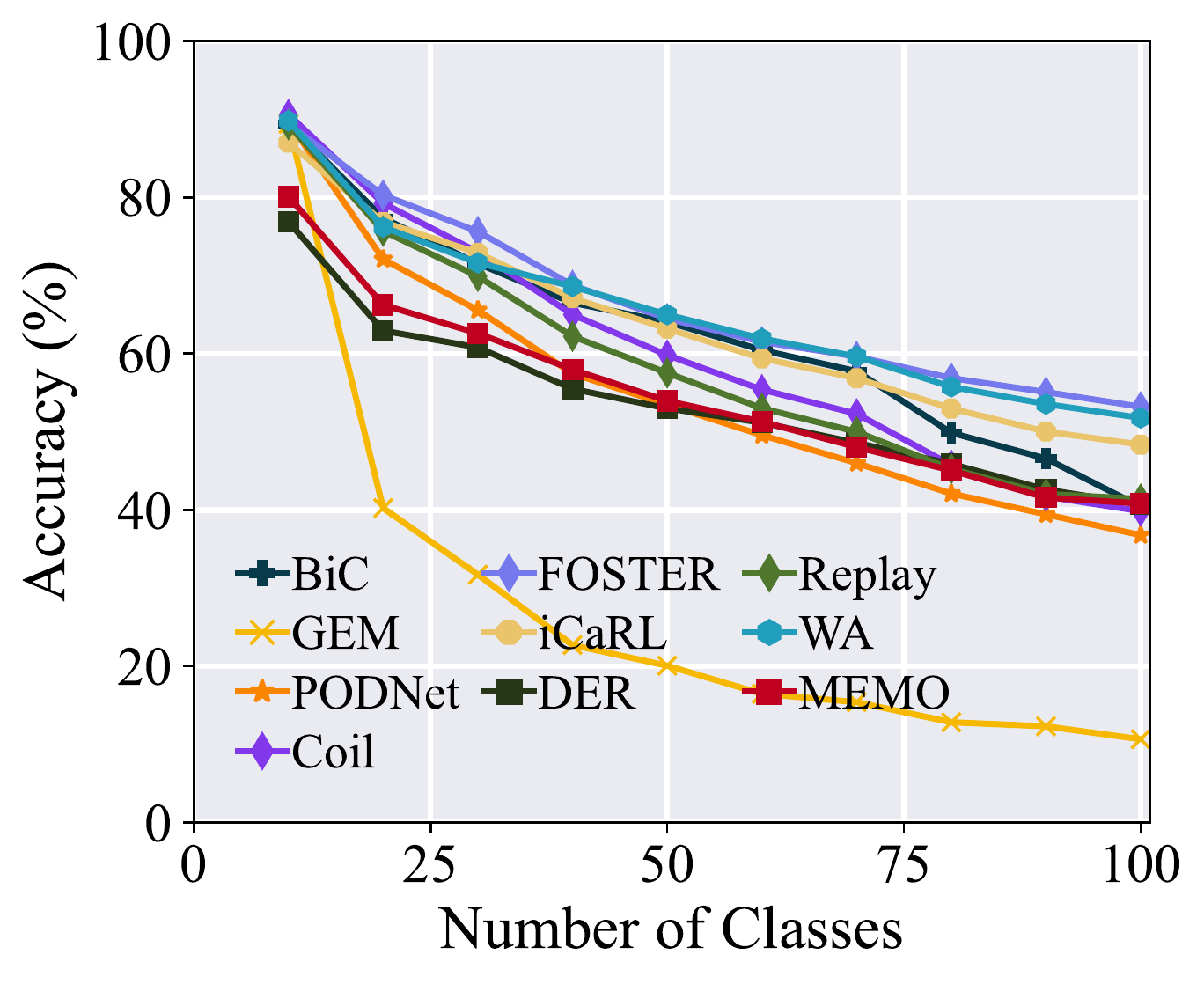}		
		}
	\end{center}
	\caption{Implementation details and performance curve of CIFAR100 when memory size=7.6 MB.
	} \label{fig:per-cif-7.6M}
\end{figure}

\begin{table}[H] 
	\caption{Numerical details when memory size=7.6 MB.	}
	\centering
	\begin{tabular}{@{\;}l@{\;}|@{\;}c@{\;}c@{\;}c@{\;}
			c@{\;} c@{\;}}
		\addlinespace
		\toprule
		7.6MB  & \#$\mathcal{E}$ & $S(\mathcal{E})$ & Model Type & \#P & Model Size\\
		\midrule
	SingleNet   &2000  & 5.85MB & ResNet32& 0.46M & 1.76MB \\
		DER   &2096  & 6.14MB & ConvNet2& 0.38M & 1.48MB \\
		MEMO &2118  & 6.20MB & ConvNet2& 0.37M & 1.42MB \\
		\bottomrule
	\end{tabular}\label{tab:per-cif-7.6M}
\end{table}

\noindent\textbf{CIFAR100 with 7.6 MB Memory Size:} The implementations are shown in Figure~\ref{fig:per-cif-7.6M}, and the memory size ($7.6$ MB) is relatively small. Since we need to align the total budget of these methods, we are only able to use small backbones for DER and MEMO. These small backbones, \ie, {\em ConvNet with two convolutional layers}, have much fewer parameters than ResNet32 and saving 10 ConvNets matches the memory size of a single ResNet32 (1.48MB versus 1.76MB).
We can infer that DER and MEMO are restricted by the learning ability of these small backbones, which perform poorly in the base session. These results are consistent with the conclusions in Figure~\ref{figure:auc_measure_last} that dynamic networks are inferior to SingleNet given a small memory budget.

We report the implementation details, including the number of exemplars, size of exemplars, type of backbones, number of parameters, and size of models in Table~\ref{tab:per-cif-7.6M}. The total memory budget, \ie, 7.6MB, is the summation of the exemplar size and model size.

\begin{figure}[H]
	\begin{center}
		\subfigure[Memory Usage]
		{	\includegraphics[width=.47\columnwidth]{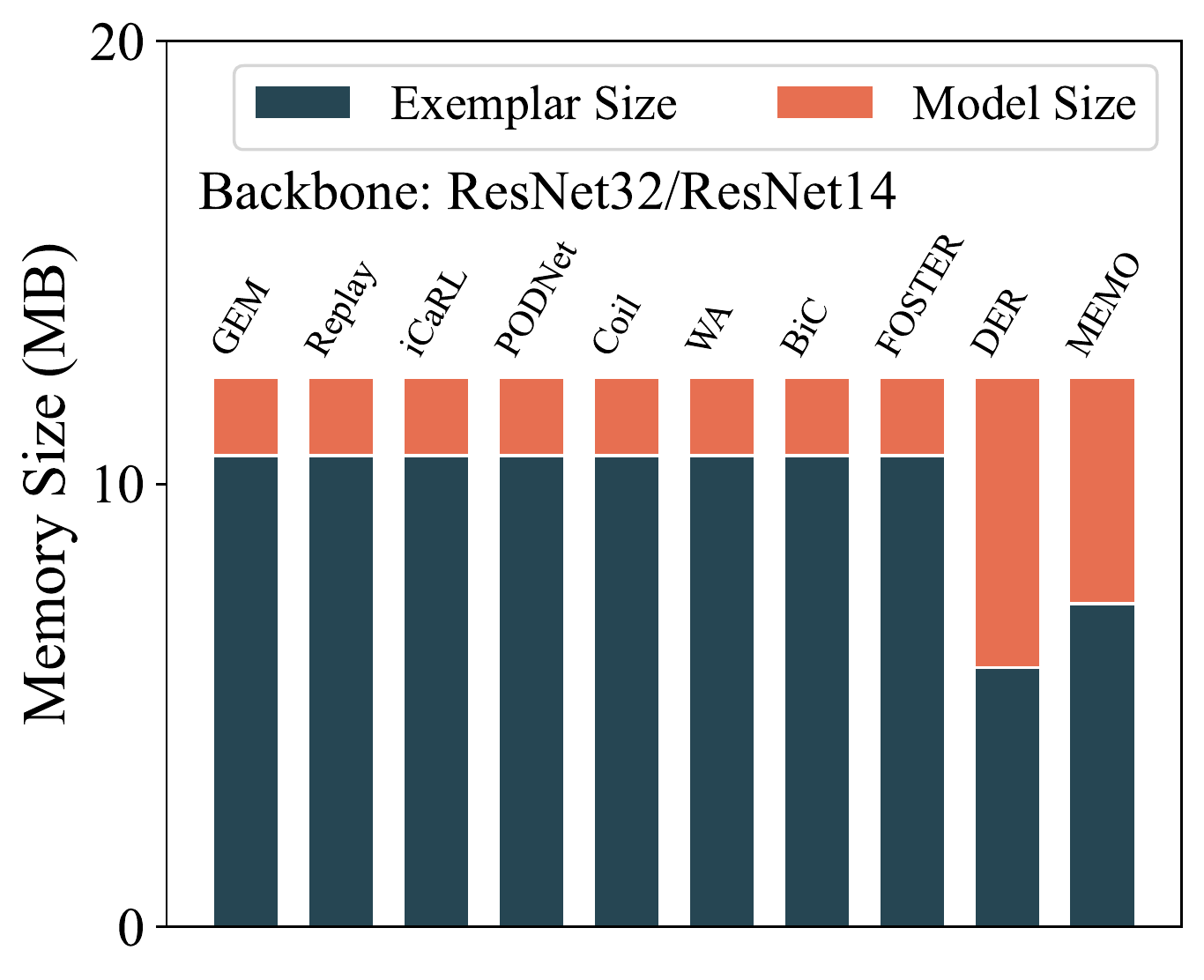}
		}
		\subfigure[Performance Curve]
		{	\includegraphics[width=.47\columnwidth]{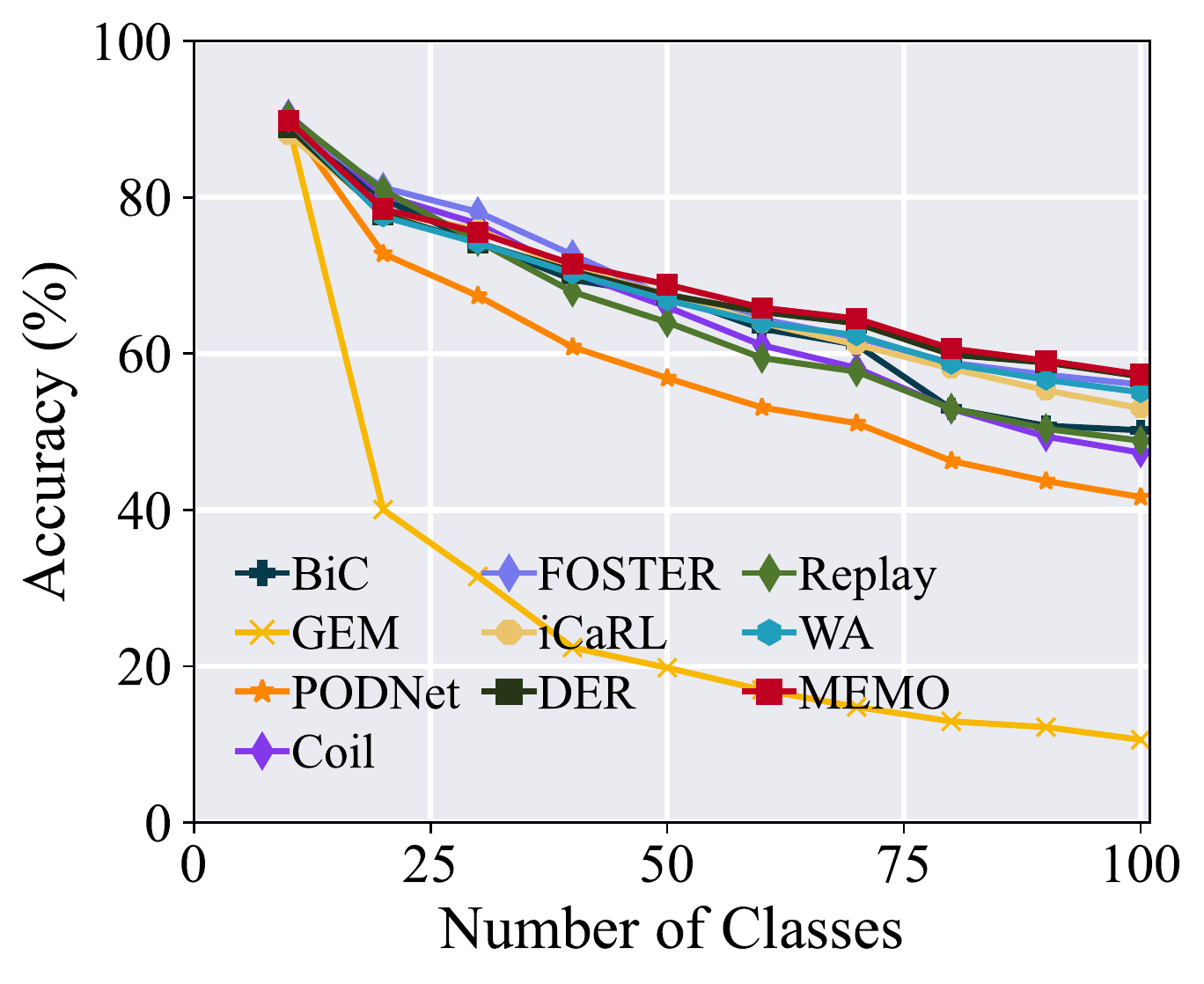}
		}
	\end{center}
	\caption{Implementation details and performance curve of CIFAR100 when memory size=12.4 MB.
	} \label{fig:cif-12.4M}
\end{figure}
\begin{table}[h]
	\caption{Numerical details when memory size=12.4 MB.	}
	\centering
	\begin{tabular}{@{\;}l@{\;}|@{\;}c@{\;}c@{\;}c@{\;}
			c@{\;} c@{\;}}
		\addlinespace
		\toprule
		12.4MB  & \#$\mathcal{E}$ & $S(\mathcal{E})$ & Model Type & \#P & Model Size\\
		\midrule
	SingleNet    & 3634  & 10.64MB & ResNet32& 0.46M & 1.76MB \\
	DER    &2000  & 5.85MB & ResNet14& 1.70M & 6.55MB \\
	MEMO &2495  & 7.32MB & ResNet14& 1.33M & 5.10MB \\
		\bottomrule
	\end{tabular}
\end{table}

\noindent\textbf{CIFAR100 with 12.4 MB Memory Size:} The implementations are shown in Figure~\ref{fig:cif-12.4M}.
By raising the total memory cost to $12.4$ MB, SingleNet can utilize the extra memory size to exchange $1634$ exemplars. At the same time, dynamic networks can switch to more powerful backbones, \ie, ResNet14, to get better representation ability. 
We can infer that DER and MEMO show competitive results with stronger backbones and outperform other methods in this setting. These results are consistent with the conclusions in Figure~\ref{figure:auc_measure_last} that the \emph{intersection} between these two groups of methods exists near the start point.

\begin{figure}[H]
	\begin{center}
		\subfigure[Memory Usage]
		{	\includegraphics[width=.47\columnwidth]{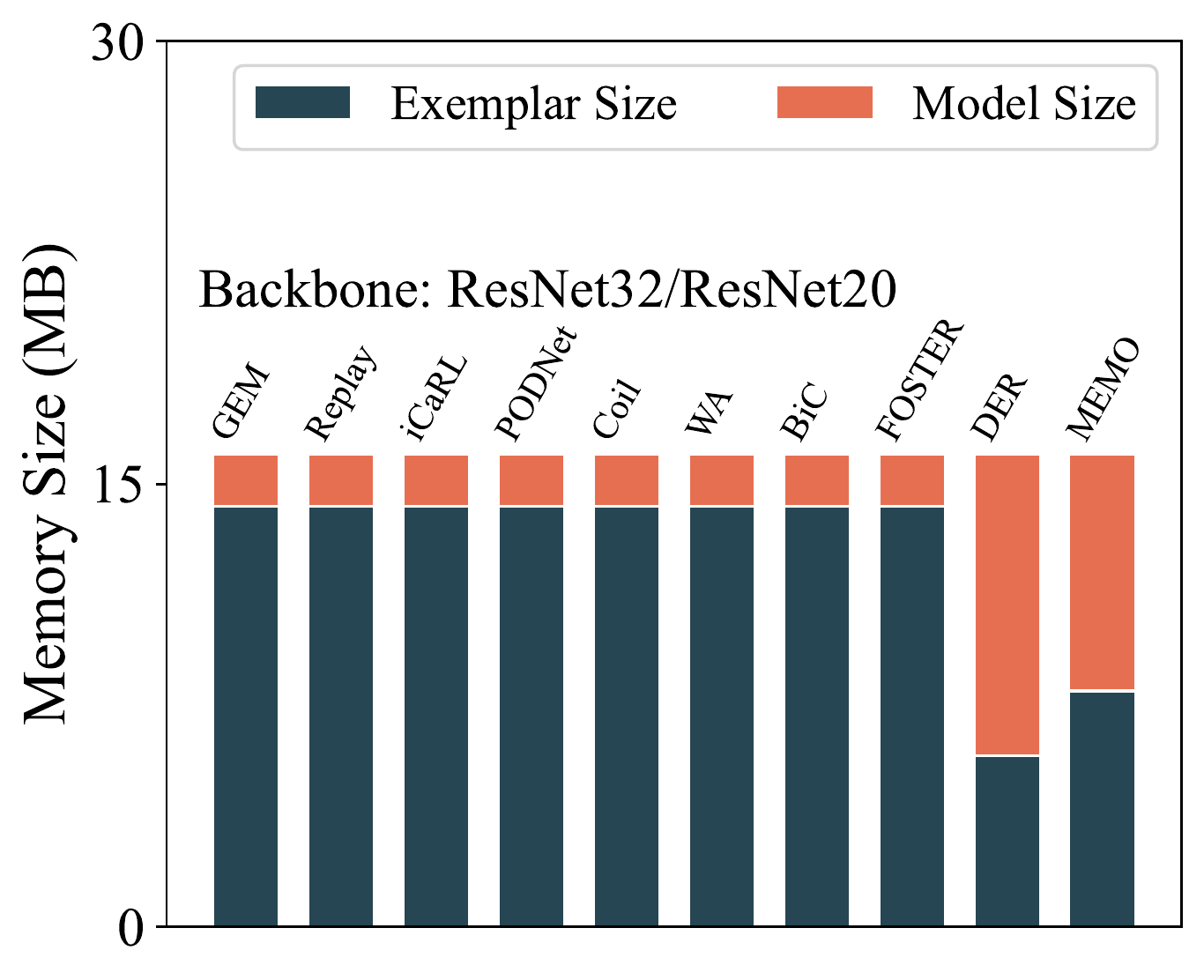}
		}
		\subfigure[Performance Curve]
		{	\includegraphics[width=.47\columnwidth]{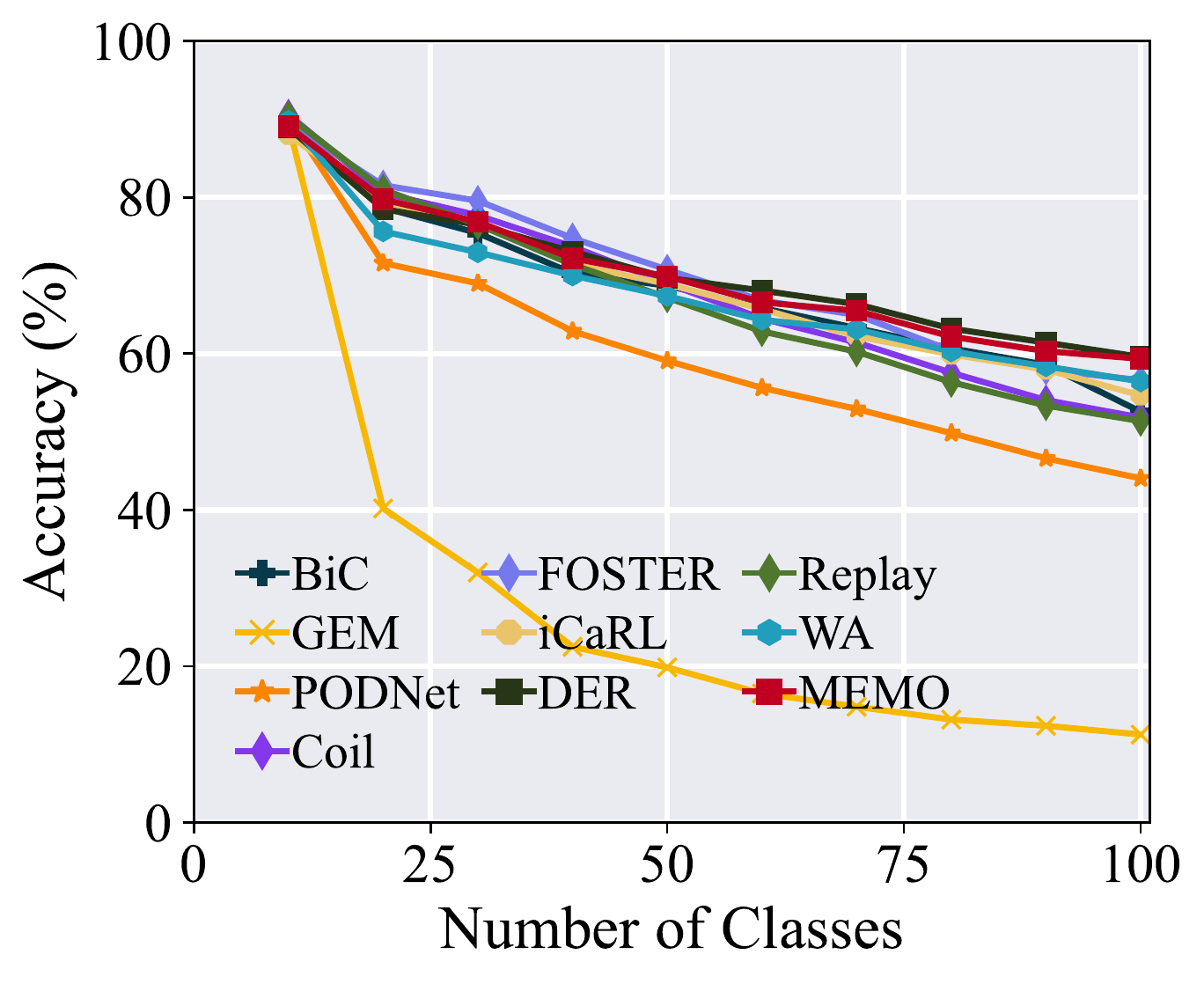}		
			
		}
	\end{center}
	\caption{Implementation details and performance curve of CIFAR100 when memory size=16.0 MB.
	} \label{fig:cif-16.0M}
\end{figure}
\begin{table}[h]
	\caption{Numerical details when memory size=16.0 MB.	}
	\centering
	\begin{tabular}{@{\;}l@{\;}|@{\;}c@{\;}c@{\;}c@{\;}
			c@{\;} c@{\;}}
		\addlinespace
		\toprule
		16.0MB  & \#$\mathcal{E}$ & $S(\mathcal{E})$ & Model Type & \#P & Model Size\\
		\midrule
		SingleNet     &4900  & 14.3MB & ResNet32& 0.46M & 1.76MB \\
		DER    &2000  & 5.85MB & ResNet20& 2.69M & 10.2MB \\
		MEMO &2768  & 8.10MB & ResNet20& 2.1M & 8.01MB\\
		\bottomrule
	\end{tabular}
\end{table}
\noindent\textbf{CIFAR100 with 16.0 MB Memory Size:} The implementations are shown in Figure~\ref{fig:cif-16.0M}.
By raising the total memory cost to $16.0$ MB, SingleNet can utilize the extra memory size to exchange $2900$ exemplars, and dynamic networks can switch to larger backbones to get better representation ability. We use ResNet20 for DER and MEMO in this setting.
The results are consistent with the former setting, where we can see that DER and MEMO show competitive results with stronger backbones and outperform other methods.

\begin{figure}[H]
	\begin{center}
		\subfigure[Memory Usage]
		{	\includegraphics[width=.47\columnwidth]{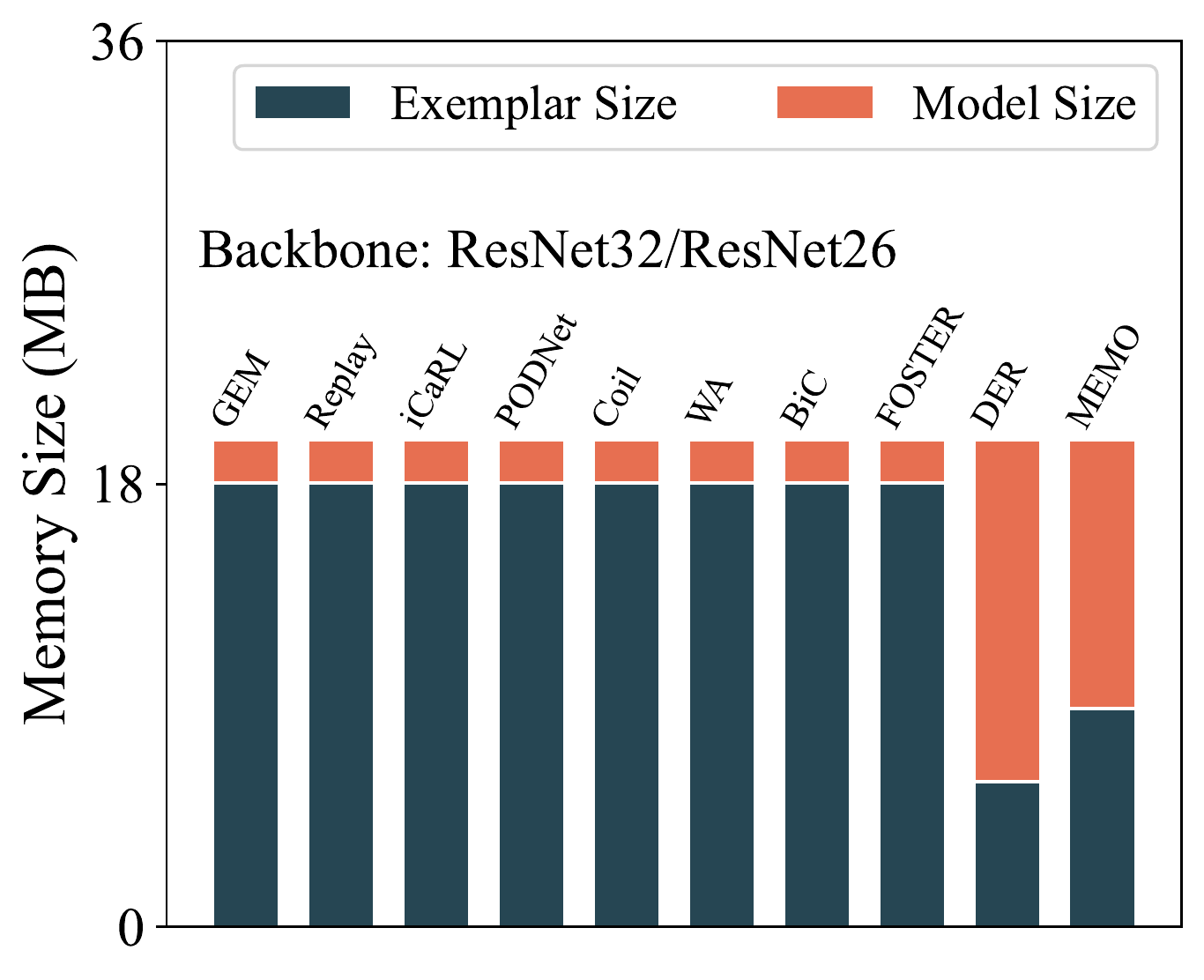}
		}
		\subfigure[Performance Curve]
		{	\includegraphics[width=.47\columnwidth]{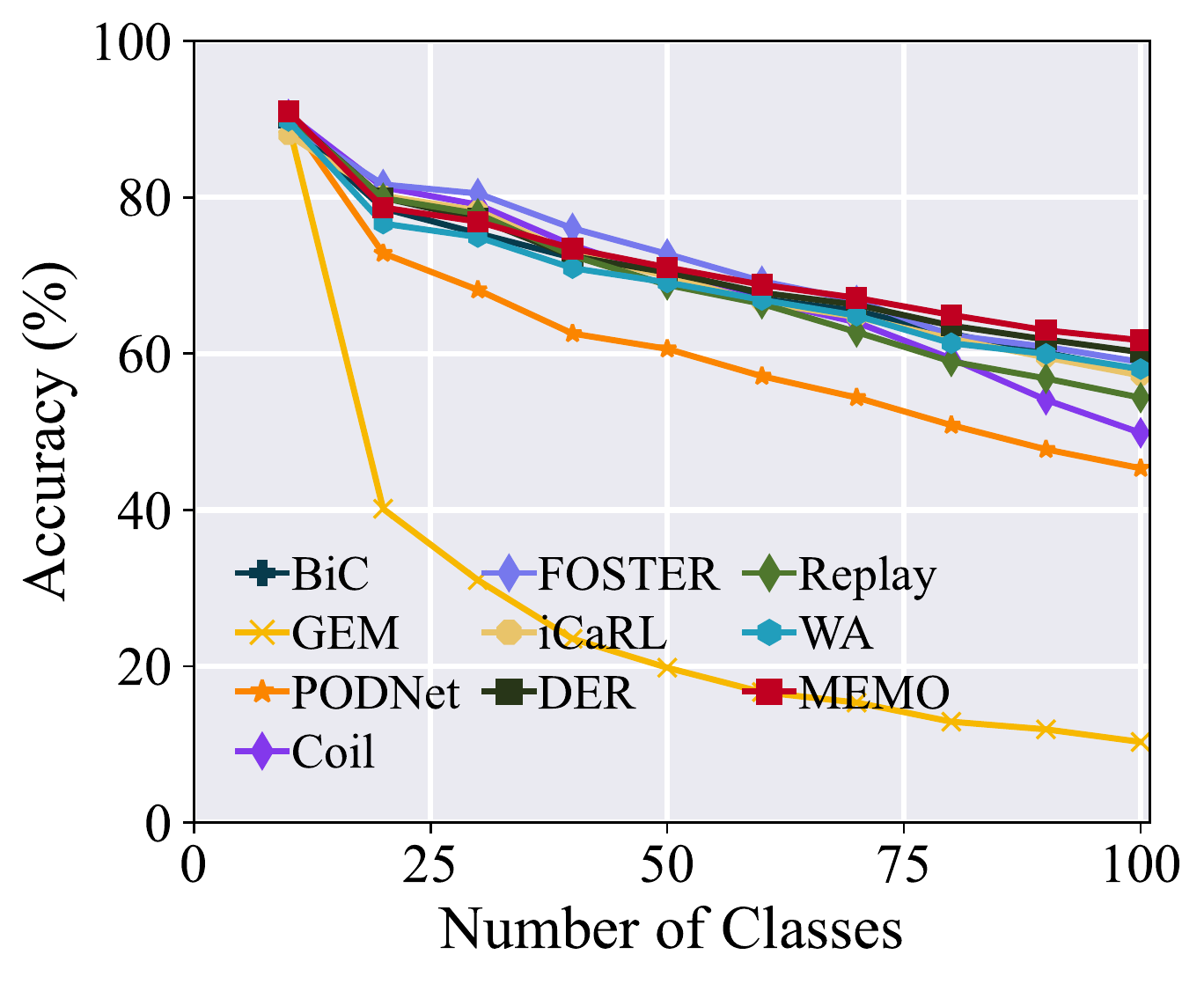}		
			
		}
	\end{center}
	\caption{Implementation details and performance curve of CIFAR100 when memory size=19.8 MB.
	} \label{fig:cif-19.8M}
\end{figure}
\begin{table}[h]
	\caption{Numerical details when memory size=19.8 MB.	}
	\centering
	\begin{tabular}{@{\;}l@{\;}|@{\;}c@{\;}c@{\;}c@{\;}
			c@{\;} c@{\;}}
		\addlinespace
		\toprule
		19.8MB  & \#$\mathcal{E}$ & $S(\mathcal{E})$ & Model Type & \#P & Model Size\\
		\midrule
		SingleNet     &6165  & 18.06MB & ResNet32& 0.46M & 1.76MB  \\
		DER    &2000  & 5.85MB & ResNet26& 3.60M  & 13.9MB \\
	MEMO &3040  & 8.91MB & ResNet26& 2.86M  & 10.92MB\\ 
		\bottomrule
	\end{tabular}
\end{table}
\noindent\textbf{CIFAR100 with 19.8 MB Memory Size:} The implementations are shown in Figure~\ref{fig:cif-19.8M}.
By raising the total memory cost to $19.8$ MB, SingleNet can utilize the extra memory size to exchange $4165$ exemplars, and dynamic networks can switch to larger backbones to get better representation ability. We use ResNet26 for DER and MEMO in this setting.
The results are consistent with the former setting, where we can infer that dynamic networks show competitive results with stronger backbones and outperform other methods.

\begin{figure}[H]
	\begin{center}
		\subfigure[Memory Usage]
		{	\includegraphics[width=.47\columnwidth]{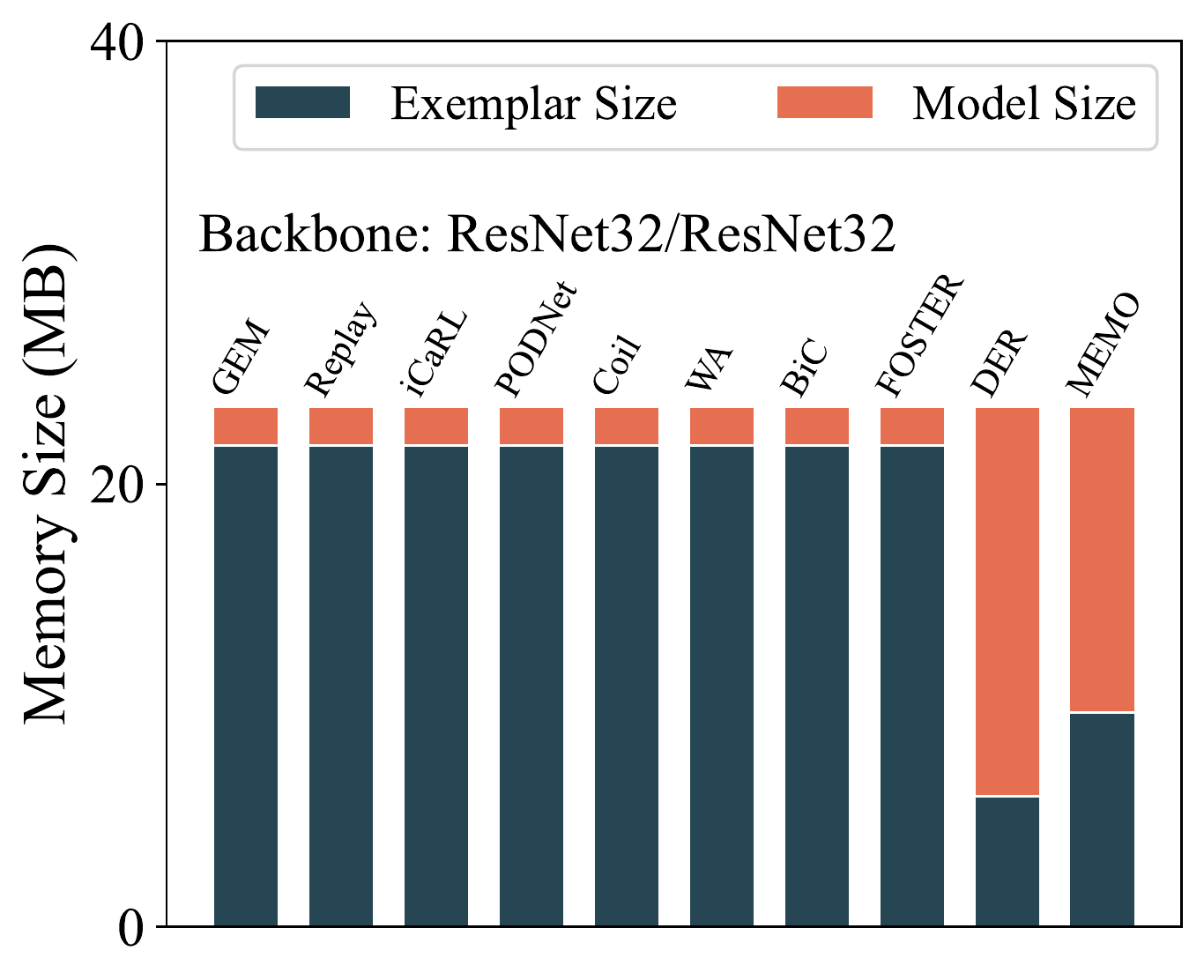}
		}
		\subfigure[Performance Curve]
		{	\includegraphics[width=.47\columnwidth]{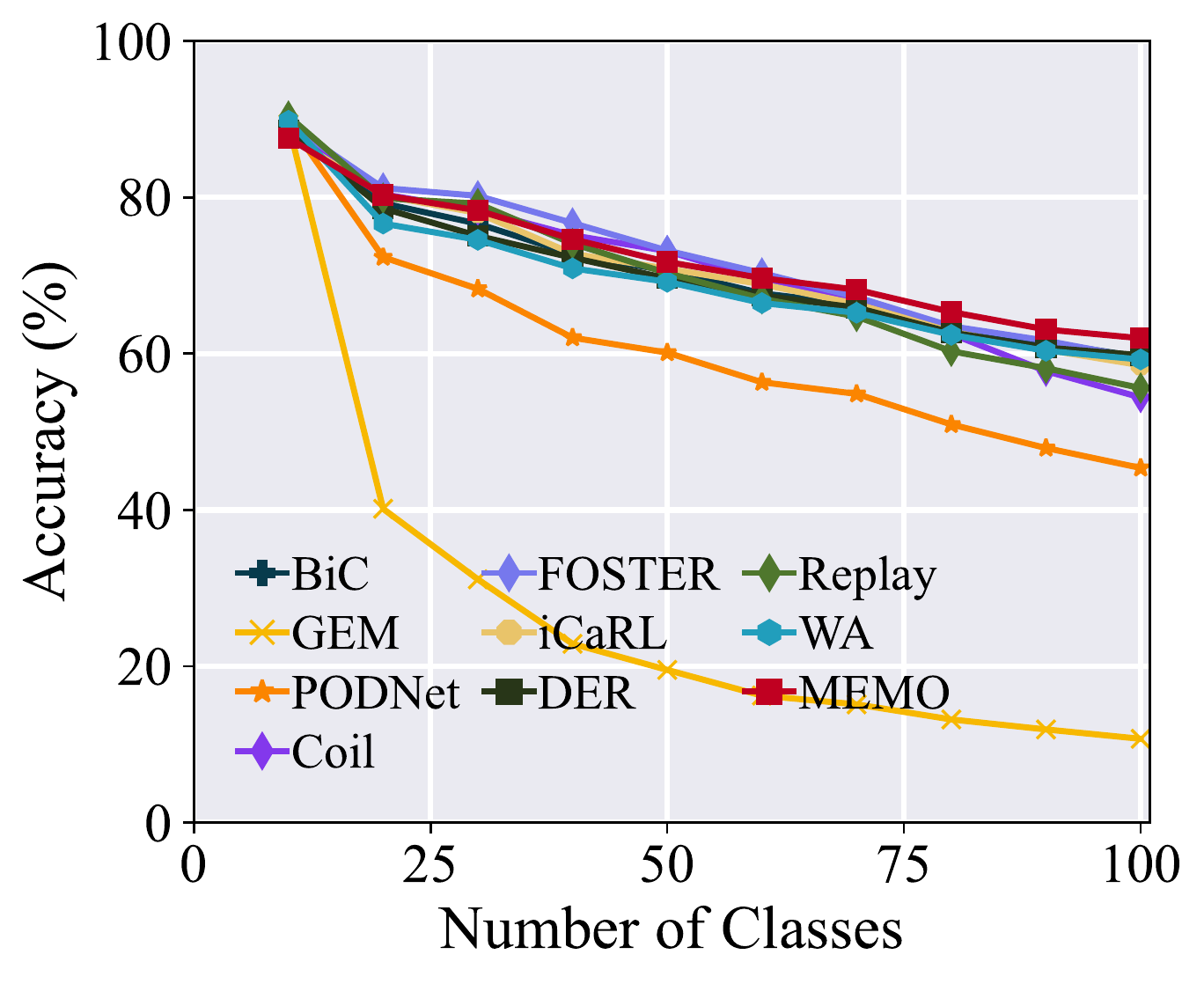}		
			
		}
	\end{center}
		\vspace{-4mm}
	\caption{Implementation details and performance curve of CIFAR100 when memory size=23.5 MB.
	} \label{fig:cif-23.5M}
\end{figure}
\begin{table}[h]
	\caption{Numerical details when memory size=23.5 MB.	}
	\vspace{-4mm}
	\centering
	\begin{tabular}{@{\;}l@{\;}|@{\;}c@{\;}c@{\;}c@{\;}
			c@{\;} c@{\;}}
		\addlinespace
		\toprule
		23.5MB  & \#$\mathcal{E}$ & $S(\mathcal{E})$ & Model Type & \#P & Model Size\\
		\midrule
		SingleNet    &7431  & 21.76MB & ResNet32& 0.46M & 1.75MB \\
		DER    &2000  & 5.86MB & ResNet32& 4.63M & 17.68MB \\
		MEMO &3312  & 9.7MB & ResNet32& 3.62M & 13.83MB \\
		\bottomrule
	\end{tabular}
\end{table}

\noindent\textbf{CIFAR100 with 23.5 MB Memory Size:} The implementations are shown in Figure~\ref{fig:cif-23.5M}.
By raising the total memory cost to $23.5$ MB, SingleNet can utilize the extra memory size to exchange $5431$ exemplars. In addition, dynamic networks can switch to larger backbones, \ie, ResNet32, to get better representation ability.

\subsubsection{ImageNet100 Implementations}

Similar to CIFAR100, we can conduct an exchange between the model and exemplars with the ImageNet100 dataset. 
For example, saving a ResNet18 model costs $11,176,512$ parameters (float), while saving an ImageNet image costs $3\times224\times224$ integer numbers (int). The budget for saving a backbone is equal to saving 
$11,176,512$ floats $\times4$ bytes/float $\div (3\times224\times224)$ bytes/image $\approx 297$ images for ImageNet. 
We conduct the experiment with the ImageNet100 Base50 Inc5 setting, as discussed in the main paper.

There are six X coordinates in the curve of ImageNet100, \eg, $329$, $493$, $755$, $872$, $1180$ and $1273$ MB. Following, we show the detailed implementation of different methods at these scales.

\begin{figure}[H]
	\begin{center}
		\subfigure[Memory Usage]
		{	\includegraphics[width=.47\columnwidth]{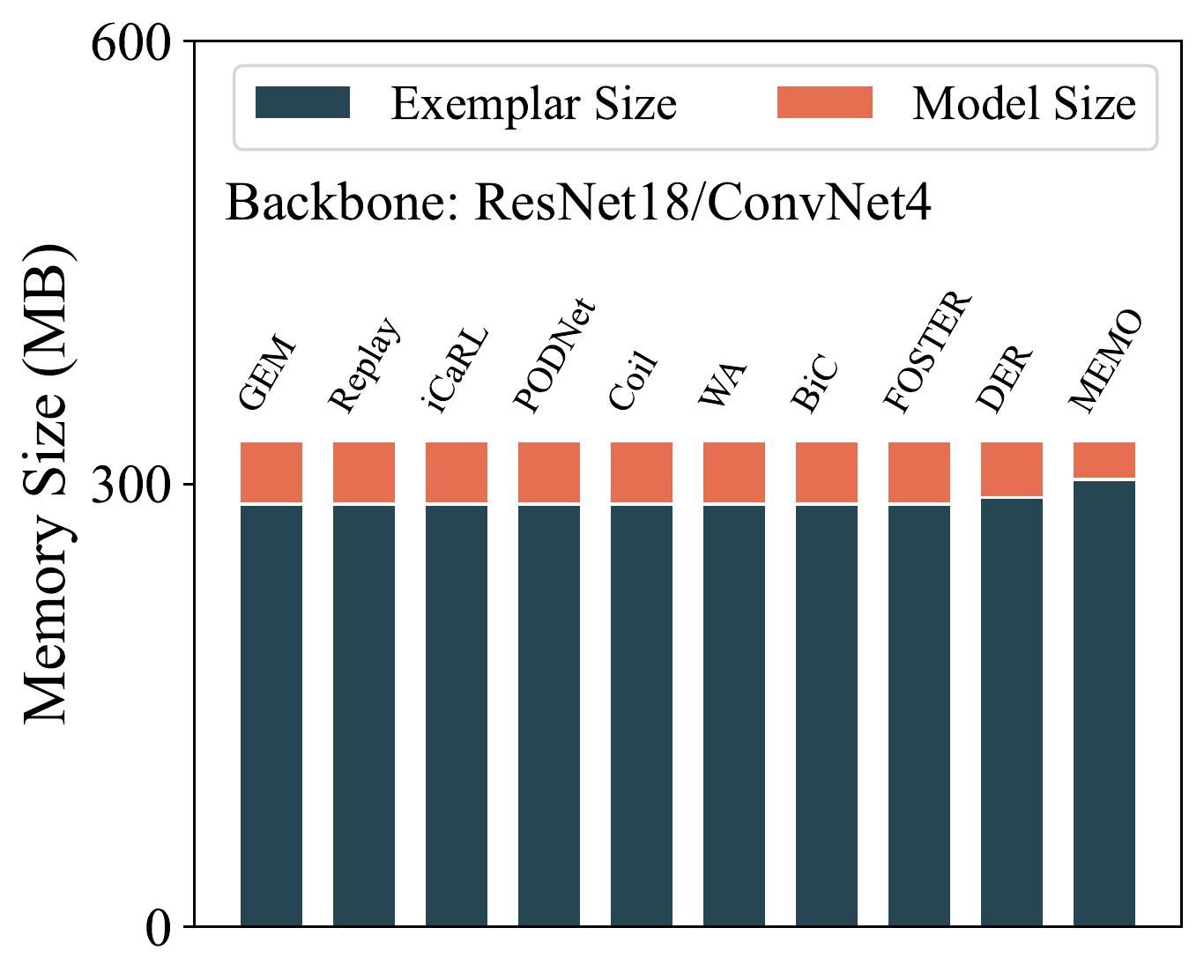}
		}
		\subfigure[Performance Curve]
		{	\includegraphics[width=.47\columnwidth]{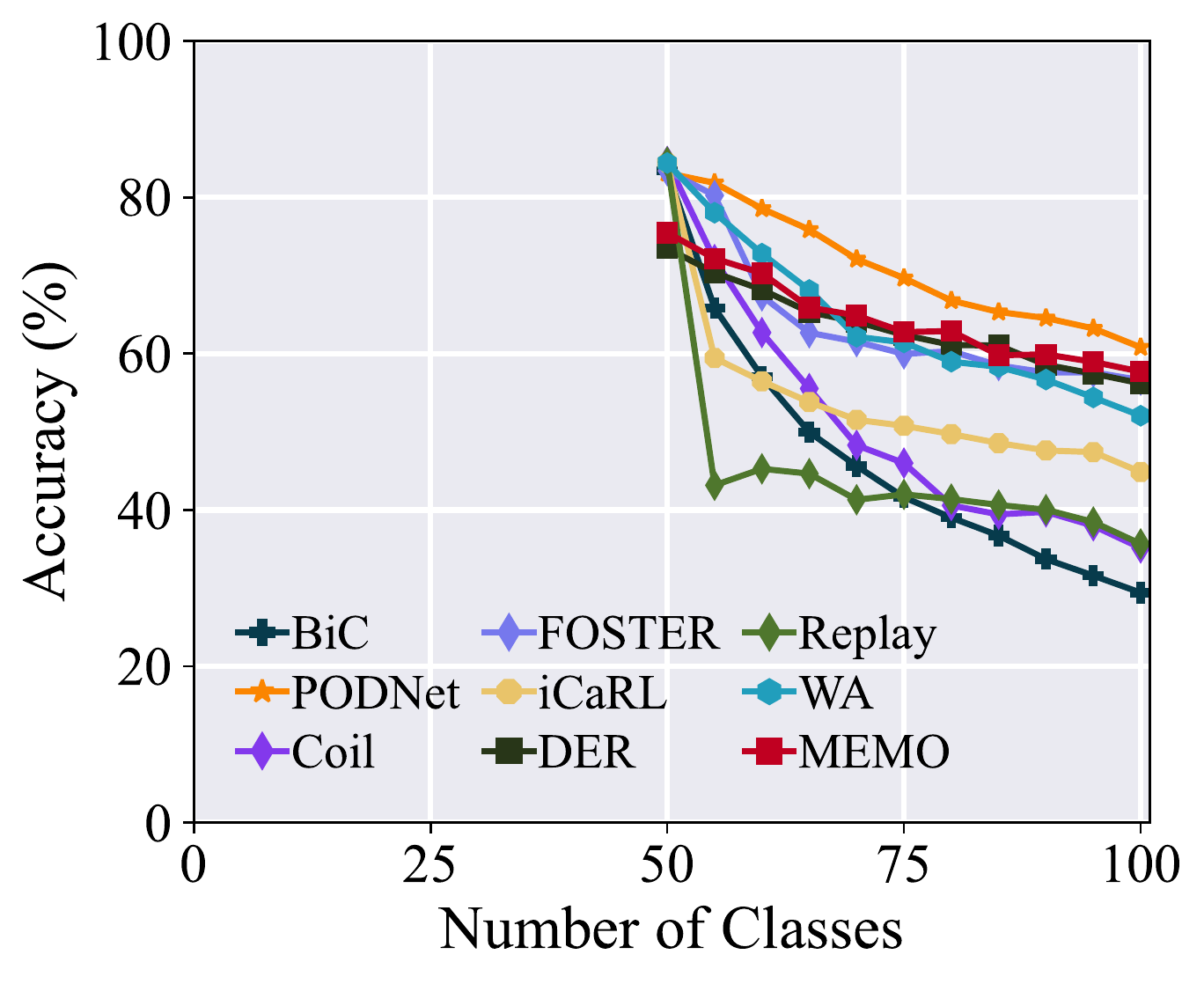}
		}
	\end{center}
	\caption{Implementation details and performance curve  of ImageNet100 when memory size=329 MB.
	} \label{fig:img-329M}
\end{figure}
\begin{table}[h]
	\caption{Numerical details when memory size=329 MB.	}
	\centering
	\begin{tabular}{@{\;}l@{\;}|@{\;}c@{\;}c@{\;}c@{\;}
			c@{\;} c@{\;}}
		\addlinespace
		\toprule
		329MB  & \#$\mathcal{E}$ & $S(\mathcal{E})$ & Model Type & \#P & Model Size\\
		\midrule
		SingleNet  	  &2000  & 287MB & ResNet18& 11.17M & 42.6MB \\
		DER    &2032  & 291MB & ConvNet4& 9.96M & 38.0MB \\
		MEMO &2115  & 303MB & ConvNet4& 6.81M & 26.0MB \\
		\bottomrule
	\end{tabular}
\end{table}
\noindent\textbf{ImageNet100 with 329 MB Memory Size:} The implementations are shown in  Figure~\ref{fig:img-329M}. $329$ MB is a relatively small memory size. Since we need to align the total budget of these methods, we are only able to use small backbones for DER and MEMO. These small backbones, \ie, {\em ConvNet with four convolutional layers}, have much fewer parameters than ResNet18, and saving 10 ConvNets matches the memory size of a single ResNet18.
We can infer from the figure that DER and MEMO are restricted by the learning ability of the inferior backbones, which perform poorly in the base task.

\begin{figure}[H]
	\begin{center}
		\subfigure[Memory Usage]
		{	\includegraphics[width=.47\columnwidth]{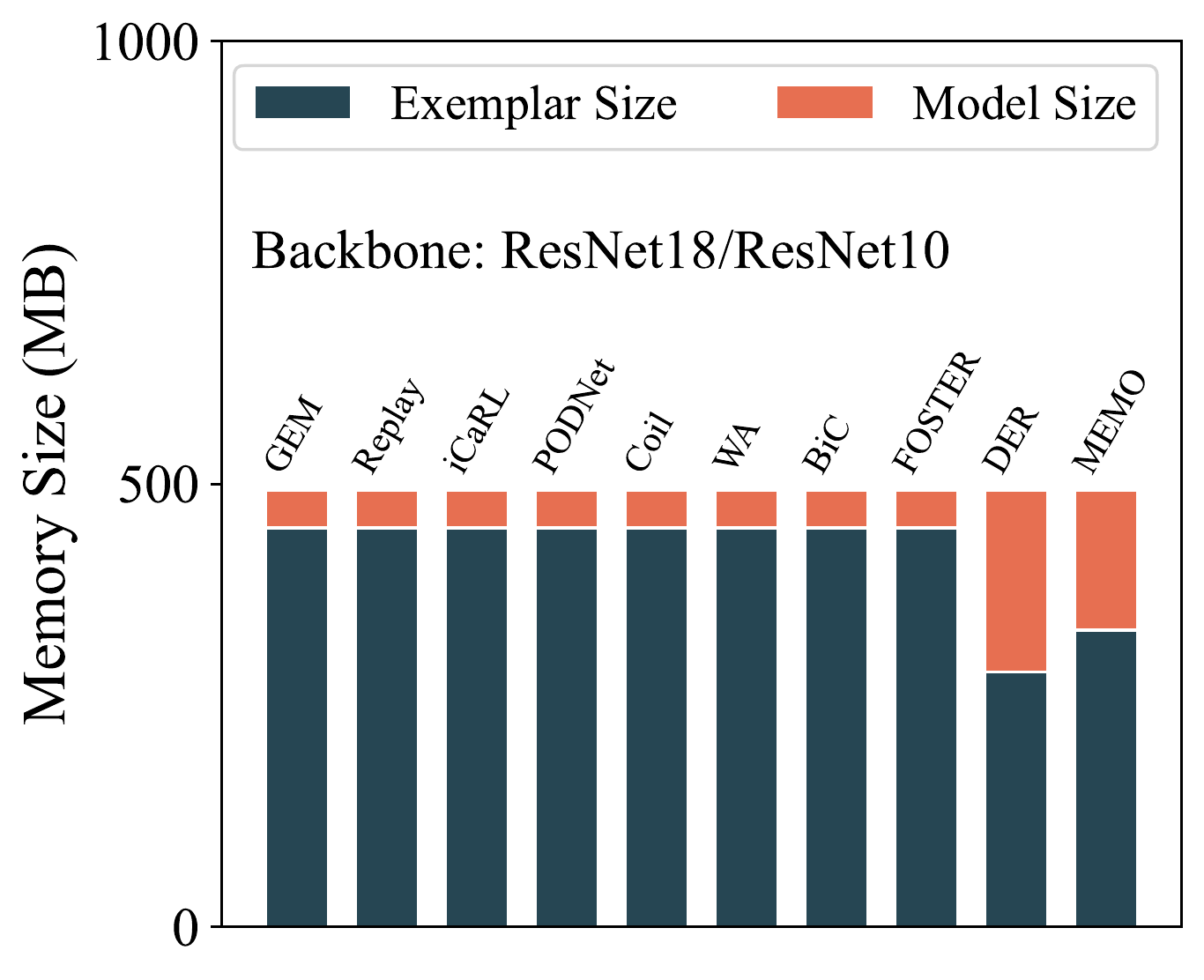}
		}
		\subfigure[Performance Curve]
		{	\includegraphics[width=.47\columnwidth]{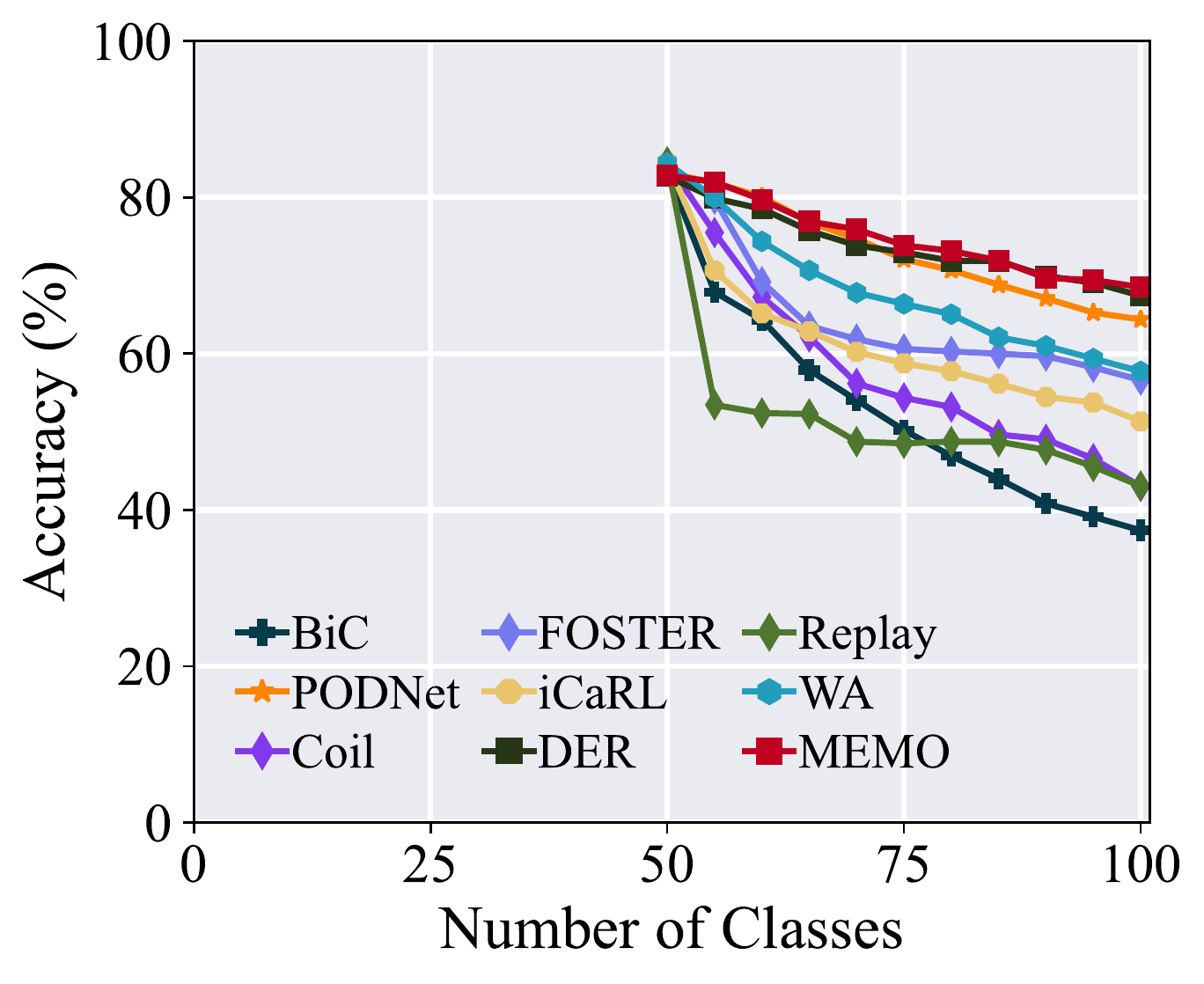}
		}
	\end{center}
	\caption{Implementation details and performance curve ImageNet100 when memory size=493 MB.
	} \label{fig:img-493M}
\end{figure}
\begin{table}[h]
	\caption{Numerical details when memory size=493 MB.	}
	\centering
	\begin{tabular}{@{\;}l@{\;}|@{\;}c@{\;}c@{\;}c@{\;}
			c@{\;} c@{\;}}
		\addlinespace
		\toprule
	493MB  & \#$\mathcal{E}$ & $S(\mathcal{E})$ & Model Type & \#P & Model Size\\
		\midrule
		SingleNet  	  &3136  & 450MB & ResNet18& 11.17M & 42.6MB \\
		DER    &2000  & 287MB & ResNet10& 53.96M & 205MB \\
		MEMO &2327  & 334MB & ResNet10& 41.63M & 158MB \\
		\bottomrule
	\end{tabular}
\end{table}
\noindent\textbf{ImageNet100 with 493 MB Memory Size:} The implementations are shown in  Figure~\ref{fig:img-493M}.
By raising the total memory cost to $493$ MB, SingleNet can utilize the extra memory size to exchange $1136$ exemplars. In addition, dynamic networks can switch to larger backbones, \ie, ResNet10, to get better representation ability.
We can infer from the figure that dynamic networks show competitive results with stronger backbones and outperform other methods in this setting.

\begin{figure}[H]
	\begin{center}
		\subfigure[Memory Usage]
		{	\includegraphics[width=.47\columnwidth]{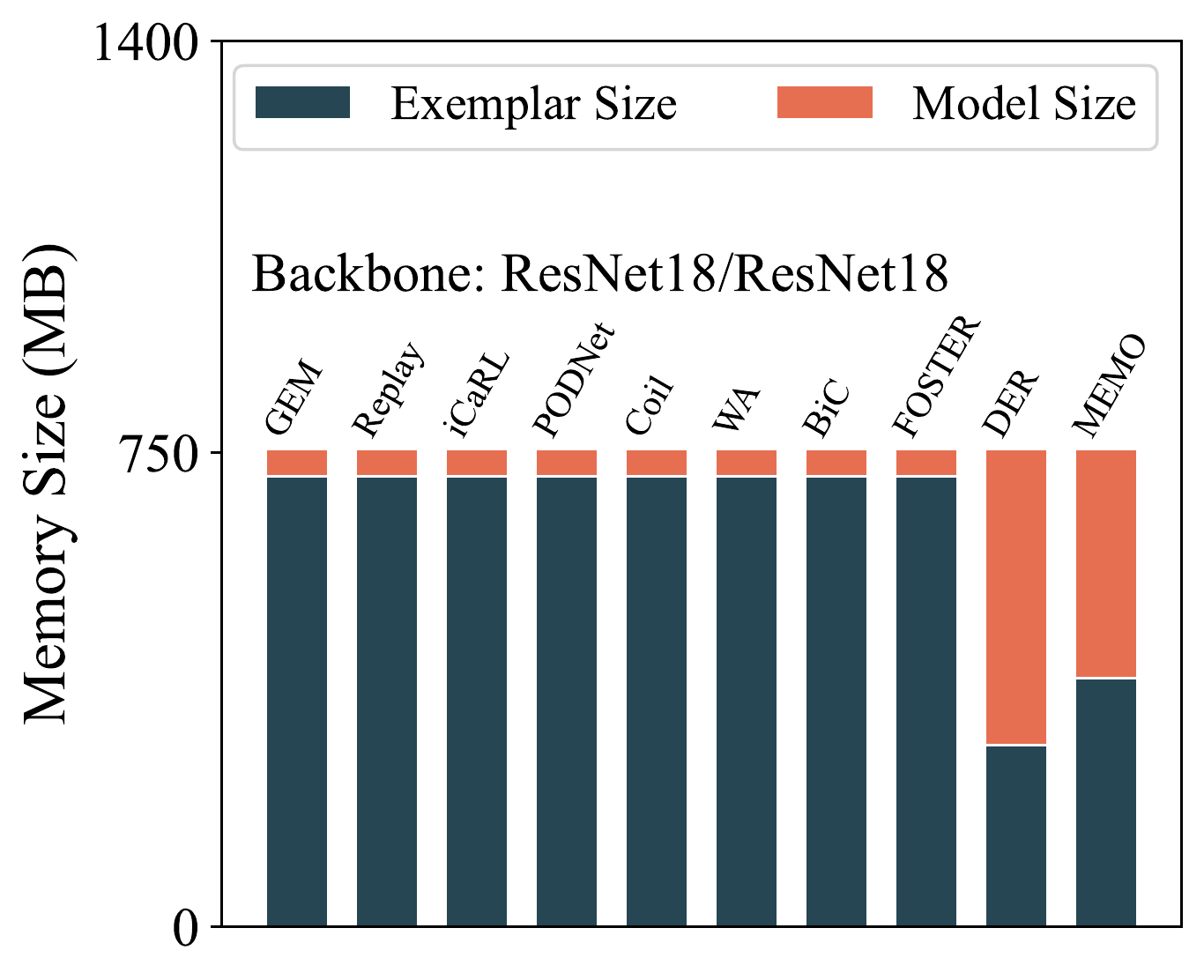}
		}
		\subfigure[Performance Curve]
		{	\includegraphics[width=.47\columnwidth]{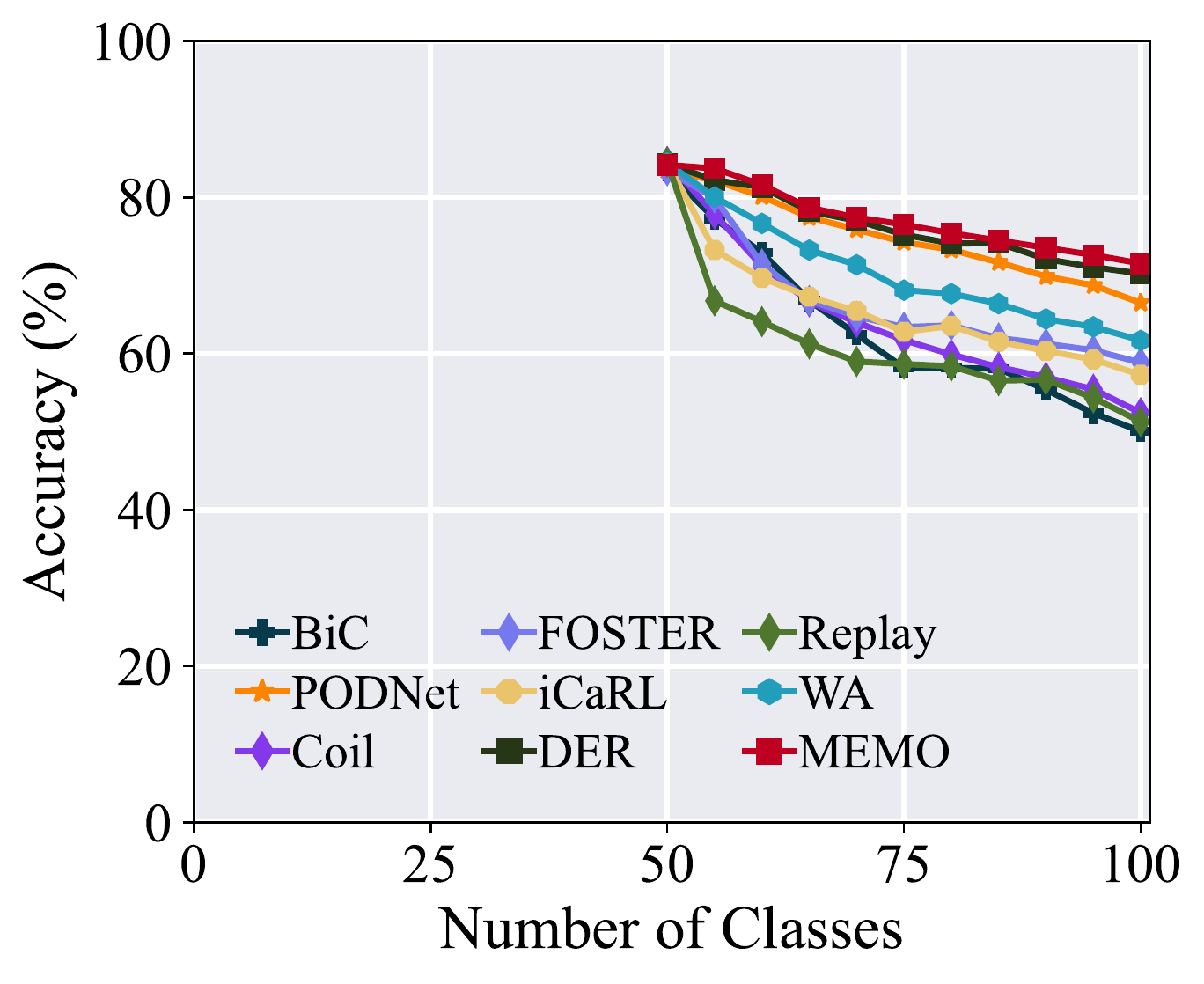}	}
	\end{center}
	\caption{Implementation details and performance curve ImageNet100 when memory size=755 MB.
	} \label{fig:img-775M}
\end{figure}
\begin{table}[h]
	\caption{Numerical details when memory size=755 MB.	}
	\centering
	\begin{tabular}{@{\;}l@{\;}|@{\;}c@{\;}c@{\;}c@{\;}
			c@{\;} c@{\;}}
		\addlinespace
		\toprule
		755MB  & \#$\mathcal{E}$ & $S(\mathcal{E})$ & Model Type & \#P & Model Size\\
		\midrule
		SingleNet    &4970  & 713MB & ResNet18& 11.17M & 42.6MB \\
		DER    &2000  & 287MB & ResNet18& 122.9M & 468MB \\
	MEMO&2739  & 393MB & ResNet18& 95.11M & 362MB \\
		\bottomrule
	\end{tabular}
\end{table}
\noindent\textbf{ImageNet100 with 755 MB Memory Size:} The implementations are shown in  Figure~\ref{fig:img-775M}.
By raising the total memory cost to $755$ MB, SingleNet can utilize the extra memory size to exchange $2970$ exemplars. At the same time, dynamic networks can switch to larger backbones, \ie, ResNet18, to get better representation ability.

\begin{figure}[H]
	\begin{center}
		\subfigure[Memory Usage]
		{	\includegraphics[width=.47\columnwidth]{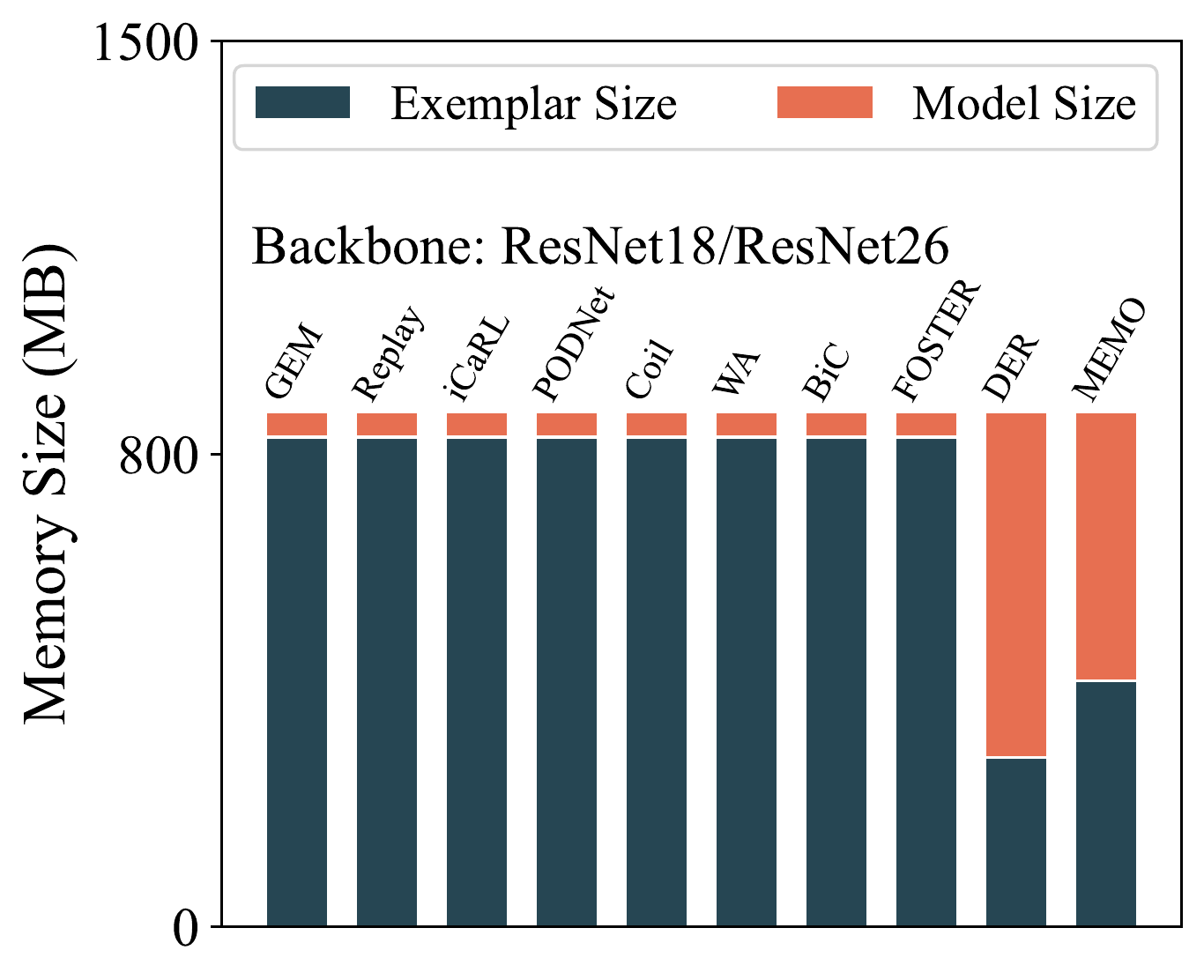}
		}
		\subfigure[Performance Curve]
		{	\includegraphics[width=.47\columnwidth]{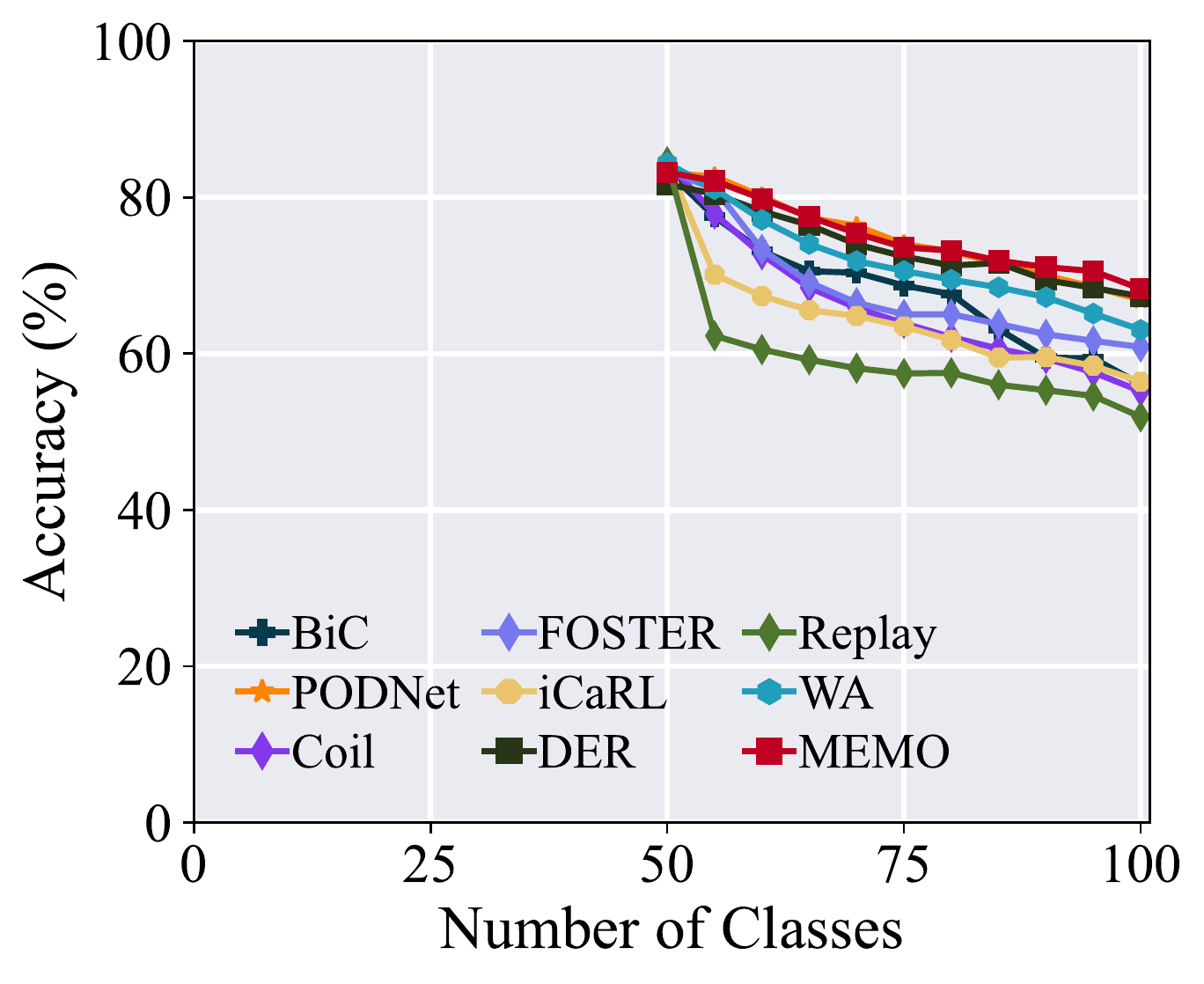}				}
	\end{center}
	\caption{Implementation details and performance curve ImageNet100 when memory size=872 MB.
	} \label{fig:img-872M}
\end{figure}
\begin{table}[h]
	\caption{Numerical details when memory size=872 MB.	}
	\centering
	\begin{tabular}{@{\;}l@{\;}|@{\;}c@{\;}c@{\;}c@{\;}
			c@{\;} c@{\;}}
		\addlinespace
		\toprule
		872MB  & \#$\mathcal{E}$ & $S(\mathcal{E})$ & Model Type & \#P & Model Size\\
		\midrule
		SingleNet   &5779  & 829MB & ResNet18& 11.17M & 42.6MB \\
		DER    &2000  & 287MB & ResNet26& 153.4M & 585.2MB \\
		MEMO&2915  & 417MB & ResNet26& 119.0M & 453MB \\
		\bottomrule
	\end{tabular}
\end{table}
\noindent\textbf{ImageNet100 with 872 MB Memory Size:} The implementations are shown in Figure~\ref{fig:img-872M}.
By raising the total memory cost to $872$ MB, SingleNet can utilize the extra memory size to exchange $3779$ exemplars. In addition, dynamic networks can switch to larger backbones, \ie, ResNet26 for better representation ability.

\begin{figure}[H]
	\begin{center}
		\subfigure[Memory Usage]
		{	\includegraphics[width=.47\columnwidth]{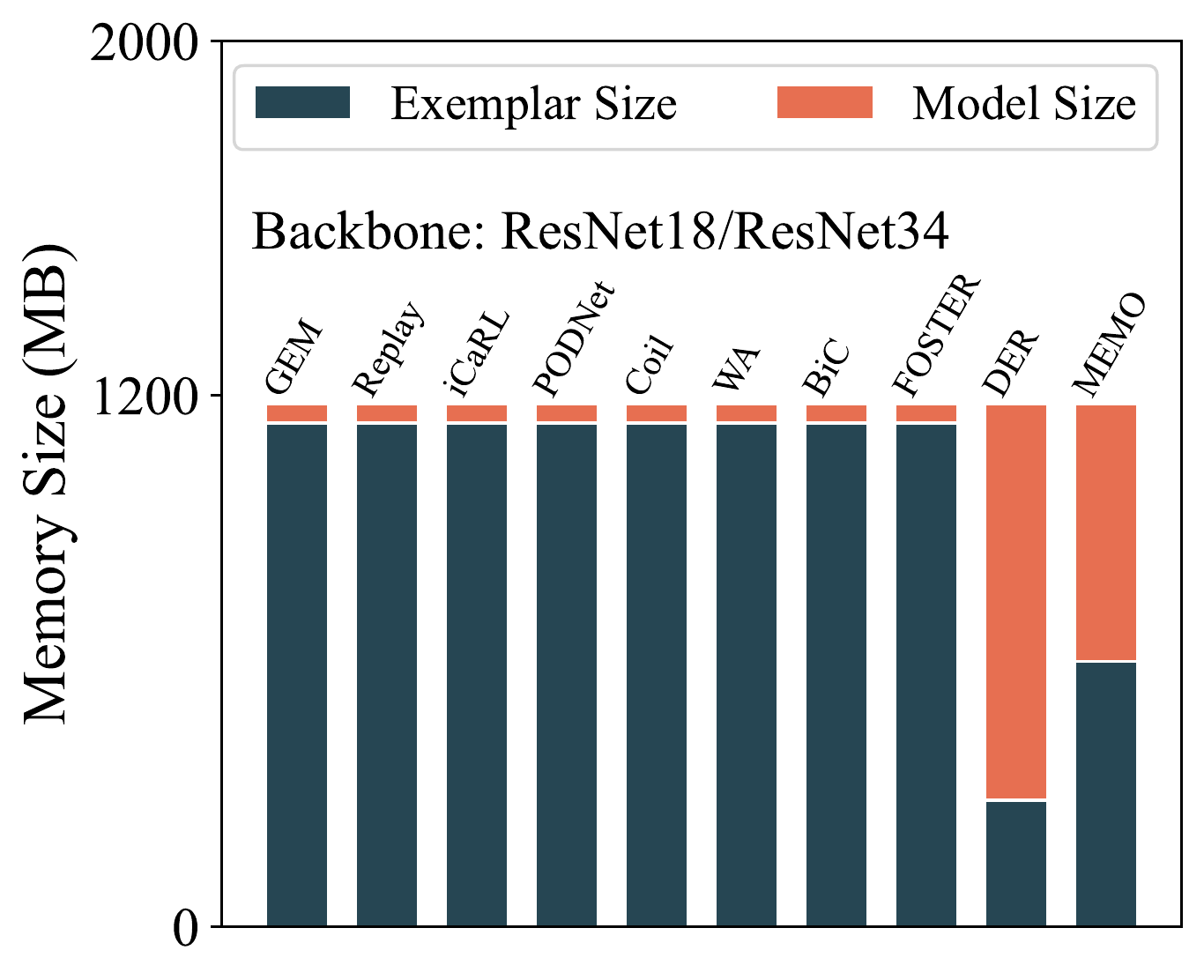}
		}
		\subfigure[Performance Curve]
		{	\includegraphics[width=.47\columnwidth]{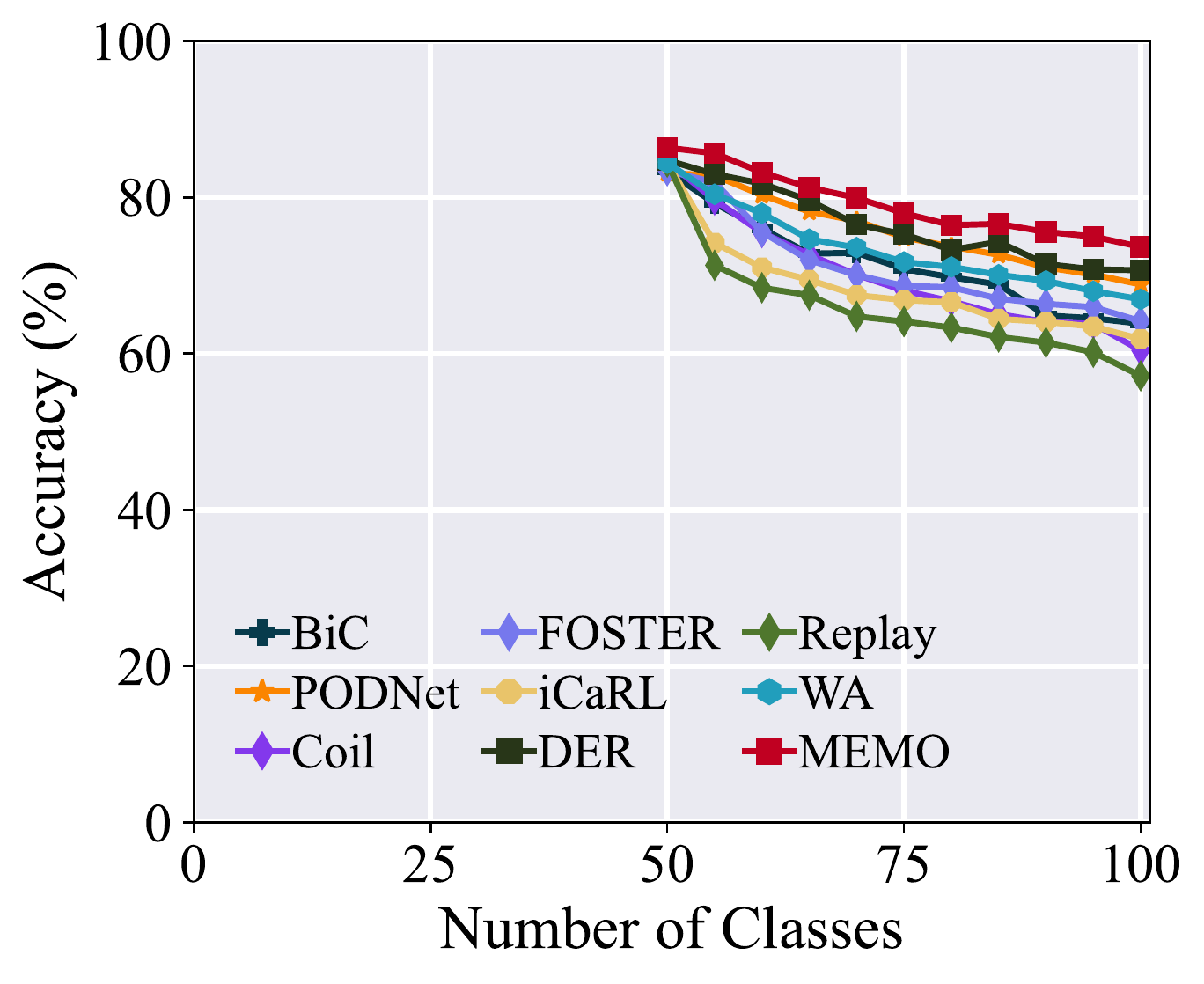}				}
	\end{center}
	\caption{Implementation details and performance curve ImageNet100 when memory size=1180 MB.
	} \label{fig:img-1180M}
\end{figure}
\begin{table}[h]
	\caption{Numerical details when memory size=1180 MB.	}
	\centering
	\begin{tabular}{@{\;}l@{\;}|@{\;}c@{\;}c@{\;}c@{\;}
			c@{\;} c@{\;}}
		\addlinespace
		\toprule
		1180MB  & \#$\mathcal{E}$ & $S(\mathcal{E})$ & Model Type & \#P & Model Size\\
		\midrule
		SingleNet   &7924  & 1137MB & ResNet18& 11.17M & 42.6MB \\
		DER    &2000  & 287MB & ResNet34& 234.1M & 893MB\\
		MEMO &4170  & 598MB & ResNet34& 152.4M & 581MB \\
		\bottomrule
	\end{tabular}
\end{table}
\noindent\textbf{ImageNet100 with 1180 MB Memory Size:} The implementations are shown in  Figure~\ref{fig:img-1180M}.
By raising the total memory cost to $1180$ MB, non-dynamic networks can utilize the extra memory size to exchange $5924$ exemplars. At the same time, dynamic networks can switch to larger backbones, \ie, ResNet34, to get better representation ability.

\begin{figure}[H]
	\begin{center}
		\subfigure[Memory Usage]
		{	\includegraphics[width=.47\columnwidth]{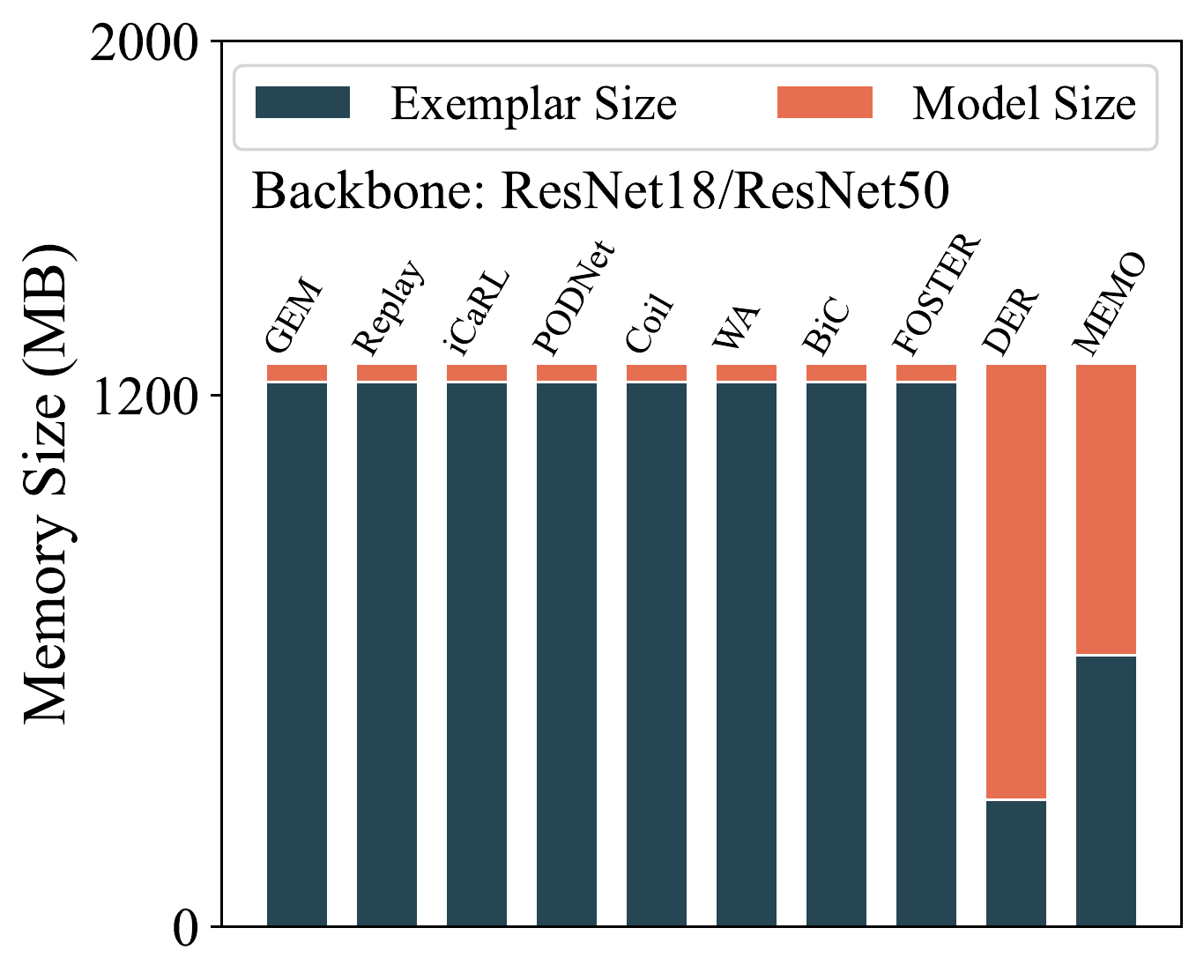}
		}
		\subfigure[Performance Curve]
		{	\includegraphics[width=.47\columnwidth]{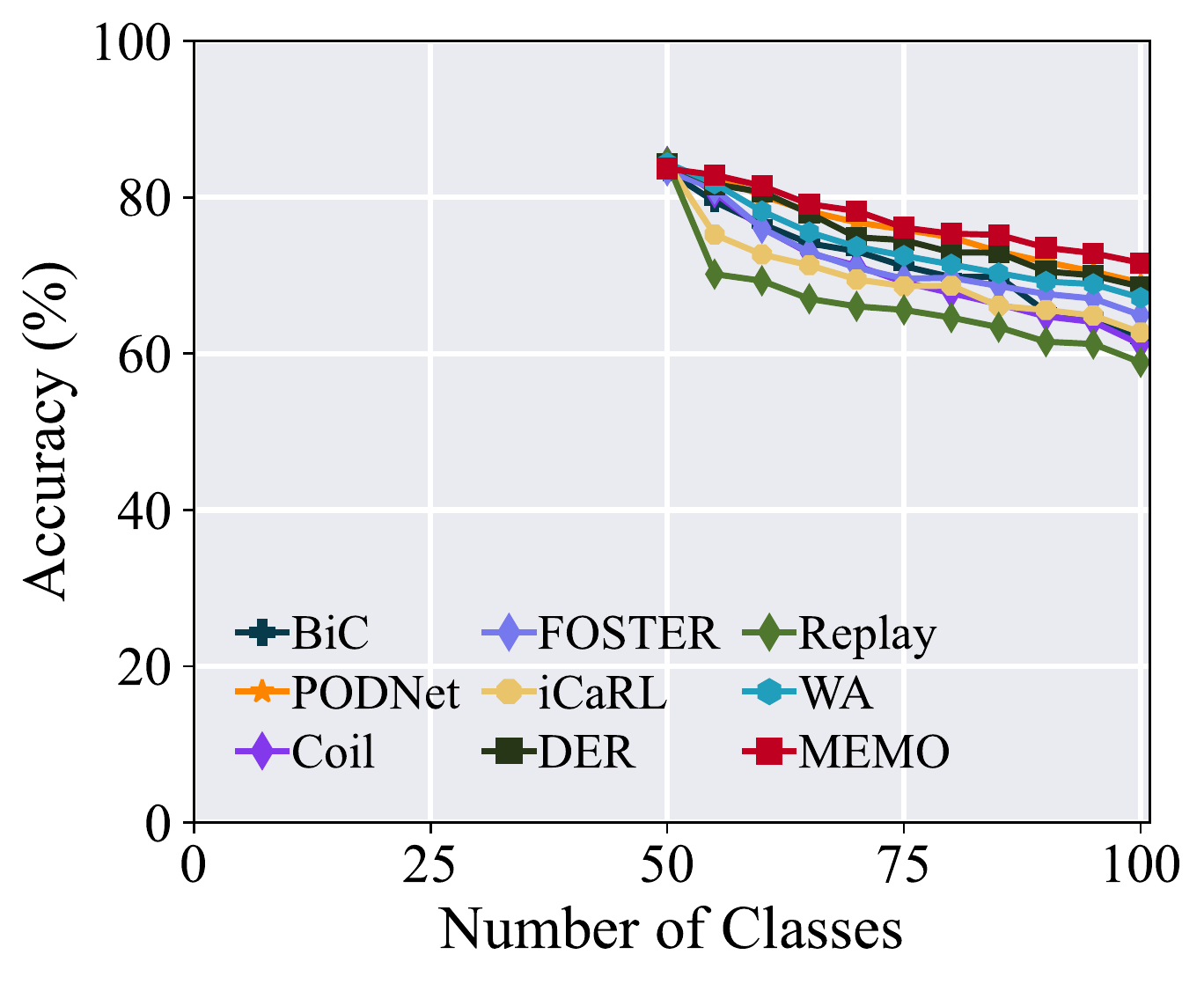}				}
	\end{center}
	\caption{Implementation details and performance curve ImageNet100 when memory size = 1273 MB.
	} \label{fig:img-1273M}
\end{figure}
\begin{table}[h]
	\caption{Numerical details when memory size=1273 MB.	}
	\centering
	\begin{tabular}{@{\;}l@{\;}|@{\;}c@{\;}c@{\;}c@{\;}
			c@{\;} c@{\;}}
		\addlinespace
		\toprule
		1273MB  & \#$\mathcal{E}$ & $S(\mathcal{E})$ & Model Type & \#P & Model Size\\
		\midrule
		SingleNet    &8574  & 1230MB & ResNet18& 11.17M & 42.6MB \\
		DER    &2000  & 287MB & ResNet50& 258.6M & 986MB \\
		MEMO &4270  & 612MB & ResNet50& 173.2M & 660MB \\
		\bottomrule
	\end{tabular}
\end{table}
\noindent\textbf{ImageNet100 with 1273 MB Memory Size:} The implementations are shown in  Figure~\ref{fig:img-1273M}.
By raising the total memory cost to $1273$ MB, SingleNet can utilize the extra memory size to exchange $6574$ exemplars, and dynamic networks can switch to larger backbones to get better representation ability. We use ResNet50 for DER and MEMO in this setting.

\noindent\textbf{Discussion about Backbones:} It should be noted that ResNet18 is the benchmark backbone for ImageNet,
and DER consumes about 755 MB memory budget under the benchmark setting. Hence, the last three points in the X-coordinate, \ie,  872, 1180, and 1273 MB, are larger than the benchmark setting.
There are two main reasons for the budget list design. First, handling large-scale image inputs requires more convolutional layers, and it is hard to find typical models with small memory budgets.
Second, we would like to investigate the performance when the model is large enough to see whether the improvement of stronger backbones will converge. The empirical results successfully verify our assumptions.

\section{Detailed Incremental Performance} \label{sec:supp_detailed_perf}

In this section, we report the detailed incremental performance of the main paper to facilitate a comparison of future works. Specifically, we report the performance on CIFAR100 and ImageNet100 in Table~\ref{table:cifar100b0inc5}, \ref{table:cifar100b0inc10}, \ref{table:cifar100b50inc10}, \ref{table:cifar100b50inc25}, \ref{table:imagenet100b0inc5}, \ref{table:Imagenet100b0inc10}, \ref{table:imagenet100b50inc10}, \ref{table:imagenet100b50inc25}.

\begin{table*}[h] 
	{
		\caption{ Incremental accuracy comparison of different methods under CIFAR100 Base0 Inc5 setting. }
		\centering	
		\resizebox{\textwidth}{!}{
			\begin{tabular}{l l l l l l l l l l l l l l l l l l l l l }
				\toprule
				\multicolumn{1}{c}{\multirow{2}{*}{Method}} & \multicolumn{20}{c}{Accuracy in each session (\%) $\uparrow$} \\ \cmidrule{2-21}
				\multicolumn{1}{c}{}    & 1      & 2      & 3    & 4     & 5  & 6     & 7      & 8     &   9  & 10 & 11      & 12      & 13    & 14     & 15  & 16     & 17      & 18     &   19  & 20      \\ \midrule
				
				Finetune & 97.00& 47.10& 31.80& 23.20& 19.72& 17.50& 13.80& 13.12& 11.09& 11.30& 9.27& 7.40& 7.66& 7.24& 6.35& 6.60& 5.96& 5.78& 5.07& 4.83 \\
				EWC & 97.00& 45.90& 32.07& 25.05& 20.68& 20.57& 14.00& 14.28& 12.02& 13.34& 11.78& 7.87& 8.62& 8.19& 6.75& 6.88& 7.07& 6.09& 4.57& 5.58 \\
				LwF & 97.00& 77.30& 58.33& 47.30& 40.76& 35.00& 29.17& 26.10& 25.13& 24.84& 21.11& 18.68& 17.85& 17.44& 15.36& 13.69& 14.32& 13.78& 12.85& 12.60 \\
				GEM & 97.00& 62.60& 46.00& 39.05& 36.92& 34.43& 28.54& 27.45& 26.69& 26.68& 25.45& 20.80& 20.97& 22.19& 21.32& 21.99& 21.38& 17.87& 17.40& 19.48 \\
				Replay & 97.00& 89.30& 79.93& 73.55& 71.56& 67.27& 63.03& 57.92& 57.58& 55.54& 52.60& 51.78& 48.65& 48.17& 45.83& 42.71& 42.80& 40.40& 39.61& 38.69 \\
				RMM & 97.00& 87.70& 82.20& 76.05& 74.12& 70.57& 69.43& 66.80& 65.27& 63.50& 62.35& 60.83& 58.58& 58.46& 56.80& 55.35& 54.07& 53.30& 51.57& 51.10 \\
				iCaRL & 97.00& 89.80& 81.93& 75.45& 73.80& 72.37& 68.63& 64.50& 63.04& 61.44& 58.98& 58.18& 56.49& 54.20& 52.55& 50.92& 49.25& 48.57& 47.57& 45.12 \\
				PODNet & 97.20& 84.90& 73.27& 62.40& 59.08& 54.60& 50.31& 46.40& 44.76& 43.28& 40.73& 39.35& 37.74& 36.73& 34.97& 32.99& 31.84& 30.62& 28.35& 27.99 \\
				Coil & 97.00& 89.90& 82.33& 75.35& 73.88& 69.40& 65.86& 60.65& 58.82& 56.38& 52.89& 50.57& 48.37& 45.51& 42.92& 39.25& 38.09& 36.57& 35.34& 34.33 \\
				WA & 97.60& 89.20& 80.47& 75.30& 73.24& 70.93& 68.77& 64.88& 64.07& 63.64& 61.36& 59.23& 57.82& 57.07& 55.24& 53.30& 51.55& 50.73& 50.09& 48.46 \\
				BiC & 97.40& 89.20& 79.60& 73.50& 72.24& 69.03& 66.63& 63.95& 63.93& 62.06& 60.05& 56.92& 55.18& 53.07& 51.59& 50.06& 48.38& 46.87& 44.83& 43.08 \\
				FOSTER & 96.40& 90.90& 82.67& 75.35& 72.68& 69.50& 66.06& 62.12& 60.73& 58.80& 57.58& 57.07& 55.40& 55.01& 53.72& 51.42& 51.85& 50.76& 50.11& 49.42 \\
				DER & 97.00& 89.40& 81.40& 76.85& 75.60& 74.20& 72.11& 68.72& 67.40& 66.04& 64.24& 63.37& 62.03& 61.69& 60.19& 58.59& 56.93& 55.48& 54.57& 53.95 \\
				MEMO & 97.00& 89.80& 82.53& 77.20& 75.48& 74.37& 71.43& 68.42& 68.00& 66.92& 64.49& 63.63& 62.51& 61.57& 60.56& 58.29& 57.16& 54.74& 54.84& 54.23 \\
				AANets & 96.40& 86.10& 78.47& 71.50& 68.84& 67.53& 64.37& 59.83& 58.62& 56.96& 54.09& 53.02& 50.95& 50.56& 48.57& 46.23& 45.22& 43.76& 43.35& 42.42 \\
				DyTox & 94.60& 90.20& 83.13& 79.40& 77.28& 75.73& 73.31& 71.53& 69.13& 67.80& 66.24& 63.63& 62.05& 60.83& 58.45& 55.23& 53.33& 54.06& 52.97& 52.23 \\
				L2P & 97.80& 96.90& 90.67& 88.75& 88.32& 86.67& 87.20& 85.12& 84.44& 83.76& 80.67& 80.28& 77.77& 79.40& 77.52& 78.72& 78.33& 79.27& 79.46& 78.96 \\
				
				\bottomrule
			\end{tabular}
		}
		\label{table:cifar100b0inc5}
	}
\end{table*}

\begin{table*}[h] 
	{
		\caption{ {Incremental  accuracy comparison of different methods under CIFAR100 Base0 Inc10 setting.} }
		\centering	
			\begin{center}
				
				\begin{tabular}{l l l l l l l l l l l}
					\toprule
					\multicolumn{1}{c}{\multirow{2}{*}{Method}} & \multicolumn{10}{c}{Accuracy in each session (\%) $\uparrow$}  \\ \cmidrule{2-11}
					\multicolumn{1}{c}{}    & 1      & 2      & 3    & 4     & 5  & 6     & 7      & 8     &   9  & 10      \\ \midrule
					Finetune & 90.80& 40.75& 30.63& 22.48& 18.56& 15.08& 13.47& 11.24& 10.41& 9.09 \\
					EWC & 90.80& 42.00& 33.77& 24.48& 24.24& 19.58& 18.90& 16.35& 15.17& 12.44 \\
					LwF & 90.80& 69.70& 57.17& 43.38& 39.30& 31.88& 30.09& 25.36& 24.70& 23.25 \\
					GEM & 90.80& 54.75& 47.93& 38.58& 34.96& 30.67& 28.40& 27.91& 25.37& 23.03 \\
					Replay & 90.80& 78.65& 71.40& 62.98& 57.86& 52.53& 50.83& 45.15& 41.89& 41.01 \\
					RMM & 89.40& 77.60& 75.50& 70.62& 68.00& 64.67& 63.49& 60.80& 58.67& 56.64 \\
					iCaRL & 90.80& 78.35& 73.97& 67.65& 63.94& 59.88& 57.77& 52.80& 51.04& 49.52 \\
					PODNet & 89.70& 72.10& 65.57& 57.45& 53.48& 49.57& 46.00& 42.10& 39.44& 36.78 \\
					Coil & 90.60& 79.25& 73.00& 65.00& 59.82& 55.40& 52.31& 45.76& 41.73& 39.85 \\
					WA & 90.80& 79.50& 75.10& 69.95& 67.50& 63.38& 61.44& 57.08& 53.81& 52.30 \\
					BiC & 88.80& 76.45& 72.77& 67.88& 64.36& 61.90& 59.17& 55.98& 52.70& 50.79 \\
					FOSTER & 89.40& 80.30& 75.63& 68.75& 64.48& 61.52& 59.63& 56.88& 55.09& 53.21 \\
					DER & 90.80& 78.80& 75.97& 71.72& 69.22& 66.58& 64.99& 61.75& 60.21& 58.59 \\
					MEMO & 89.60& 79.30& 77.17& 72.58& 69.76& 67.23& 65.47& 62.15& 60.28& 58.49 \\
					AANets & 91.00& 74.70& 71.30& 65.25& 60.36& 57.25& 53.77& 50.31& 47.83& 45.53 \\
					DyTox & 91.60& 80.00& 77.30& 74.35& 72.16& 67.68& 65.89& 62.63& 60.40& 58.72 \\
					L2P & 98.50& 95.15& 93.47& 91.00& 89.84& 86.32& 85.71& 85.17& 84.97& 83.39 \\
					
					\bottomrule
				\end{tabular}
		\end{center}
		\label{table:cifar100b0inc10}
	}
\end{table*}

\begin{table}[h] 
	{
		\caption{ {Incremental accuracy comparison of different methods under CIFAR100 Base50 Inc10 setting.} }
		
		\centering	
			\begin{center}
				\begin{tabular}{l l l l l l l l}
					\toprule
					\multicolumn{1}{c}{\multirow{2}{*}{Method}} & \multicolumn{6}{c}{Accuracy in each session (\%) $\uparrow$}  \\ \cmidrule{2-7}
					\multicolumn{1}{c}{}    & 1      & 2      & 3    & 4     & 5  & 6    \\ \midrule
					Finetune & 76.52& 15.18& 13.47& 11.62& 10.86& 9.09 \\
					EWC & 76.52& 20.53& 17.47& 14.51& 14.10& 11.47 \\
					LwF & 76.52& 48.50& 39.63& 30.45& 26.54& 25.06 \\
					GEM & 76.52& 27.43& 25.69& 25.92& 23.42& 21.33 \\
					Replay & 76.52& 54.70& 51.26& 46.41& 44.02& 41.26 \\
					RMM & 76.42& 73.32& 68.71& 64.39& 62.59& 59.75 \\
					iCaRL & 76.52& 66.05& 62.97& 56.90& 53.86& 52.04 \\
					PODNet & 76.60& 70.13& 66.00& 61.05& 57.73& 55.21 \\
					Coil & 76.52& 63.12& 57.10& 50.06& 45.53& 41.24 \\
					WA & 76.54& 68.65& 66.43& 60.74& 57.68& 55.85 \\
					BiC & 74.14& 68.62& 62.06& 58.29& 53.77& 49.19 \\
					FOSTER & 76.44& 72.02& 65.77& 62.25& 60.09& 57.82 \\
					DER & 76.52& 72.05& 69.79& 65.69& 63.33& 61.94 \\
					MEMO & 76.52& 73.28& 70.90& 67.58& 64.50& 62.83 \\
					AANets & 76.64& 72.38& 69.33& 64.66& 61.14& 59.36 \\
					DyTox & 79.96& 74.62& 71.16& 65.58& 62.73& 60.35 \\
					L2P & 92.56& 88.05& 85.96& 83.76& 81.60& 79.34 \\
					\bottomrule
				\end{tabular}
		\end{center}
		\label{table:cifar100b50inc10}
	}
\end{table}

\begin{table}[h] 
	{
		\caption{ {Incremental accuracy comparison of different methods under CIFAR100 Base50 Inc25 setting.} }
		\centering	
		\begin{center}
			
				\begin{tabular}{l l l l l}
					\toprule
					\multicolumn{1}{c}{\multirow{2}{*}{Method}} & \multicolumn{3}{c}{Accuracy in each session (\%) $\uparrow$} \\ \cmidrule{2-4}
					\multicolumn{1}{c}{}    & 1      & 2      & 3     \\ \midrule
					Finetune & 76.52& 27.29& 20.88 \\
					EWC & 76.52& 30.72& 23.85 \\
					LwF & 76.52& 48.48& 37.31 \\
					GEM & 76.52& 41.80& 37.29 \\
					Replay & 76.52& 52.85& 44.07 \\
					RMM & 76.42& 70.51& 64.63 \\
					iCaRL & 76.52& 64.88& 57.59 \\
					PODNet & 76.60& 66.83& 59.10 \\
					Coil & 76.52& 62.89& 51.08 \\
					WA & 76.54& 66.57& 61.49 \\
					BiC & 74.68& 65.91& 59.77 \\
					FOSTER & 76.44& 70.01& 62.38 \\
					DER & 76.52& 70.79& 65.10 \\
					MEMO & 76.52& 71.36& 66.06 \\
					AANets & 76.56& 69.79& 61.85 \\
					DyTox & 79.92& 71.89& 66.49 \\
					L2P & 92.56& 88.24& 86.29 \\
					\bottomrule
				\end{tabular}
		\end{center}
		\label{table:cifar100b50inc25}
	}
\end{table}

\begin{table*}[h] 
	{
		\caption{ {Incremental  accuracy comparison of different methods under ImageNet100 Base0 Inc5 setting.} }
		\centering	
		\resizebox{\textwidth}{!}{
			\begin{tabular}{l l l l l l l l l l l l l l l l l l l l l }
				\toprule
				\multicolumn{1}{c}{\multirow{2}{*}{Method}} & \multicolumn{20}{c}{Accuracy in each session (\%) $\uparrow$} \\ \cmidrule{2-21}
				\multicolumn{1}{c}{}    & 1      & 2      & 3    & 4     & 5  & 6     & 7      & 8     &   9  & 10 & 11      & 12      & 13    & 14     & 15  & 16     & 17      & 18     &   19  & 20      \\ \midrule
				
				Finetune & 94.40& 42.20& 30.13& 26.90& 19.76& 16.33& 13.83& 12.30& 10.40& 9.80& 9.42& 8.27& 7.51& 7.23& 6.72& 6.05& 5.79& 5.11& 5.20& 4.70 \\
				EWC & 94.40& 41.80& 32.93& 28.60& 23.84& 19.87& 15.89& 15.75& 12.49& 10.68& 10.00& 9.60& 8.80& 9.06& 8.00& 7.30& 7.25& 6.29& 6.19& 6.14 \\
				LwF & 94.40& 78.20& 63.87& 59.60& 56.56& 55.33& 51.20& 44.40& 42.00& 36.08& 31.45& 29.10& 29.20& 27.80& 27.95& 27.20& 22.99& 20.24& 20.32& 17.74 \\
				Replay & 94.80& 83.80& 73.33& 74.30& 66.16& 65.27& 61.71& 59.85& 54.27& 54.36& 52.84& 49.57& 46.95& 42.69& 43.73& 41.92& 43.13& 42.18& 39.31& 37.32 \\
				RMM & 94.40& 83.60& 80.67& 80.30& 76.72& 74.33& 72.86& 71.15& 69.73& 69.60& 68.87& 66.80& 64.22& 62.69& 61.36& 61.28& 60.19& 59.40& 58.74& 57.16 \\
				iCaRL & 94.40& 86.80& 80.67& 78.90& 73.44& 70.80& 66.06& 63.45& 62.76& 59.76& 58.40& 55.20& 53.94& 51.80& 50.61& 50.85& 48.38& 47.78& 47.98& 44.10 \\
				PODNet & 96.00& 84.20& 75.87& 72.20& 66.32& 63.47& 60.46& 56.15& 52.36& 50.04& 49.20& 44.97& 43.91& 39.03& 38.93& 37.55& 37.62& 37.24& 35.07& 33.34 \\
				Coil & 94.40& 86.60& 79.07& 76.30& 70.16& 67.40& 64.11& 60.00& 56.53& 54.96& 53.75& 48.60& 45.08& 40.46& 40.69& 37.50& 38.35& 37.82& 36.44& 34.00 \\
				WA & 94.80& 84.20& 78.27& 78.30& 73.12& 71.67& 69.14& 65.30& 64.27& 62.04& 61.42& 57.30& 54.31& 51.46& 49.79& 50.25& 49.95& 49.36& 48.17& 46.06 \\
				BiC & 94.80& 84.60& 78.67& 76.40& 72.08& 70.40& 66.51& 64.15& 59.60& 55.84& 54.95& 50.93& 49.94& 46.49& 44.24& 41.75& 40.02& 38.36& 36.38& 34.56 \\
				FOSTER & 94.40& 87.00& 77.60& 75.50& 71.36& 67.87& 65.54& 63.05& 61.42& 61.72& 61.71& 59.60& 56.25& 55.23& 56.05& 56.02& 55.91& 54.84& 54.84& 53.18 \\
				DER & 94.40& 87.20& 81.20& 80.30& 78.48& 77.47& 76.40& 75.65& 72.62& 73.16& 73.42& 72.53& 70.40& 68.60& 67.84& 66.82& 66.68& 64.02& 64.61& 63.66 \\
				MEMO & 94.40& 84.80& 77.47& 78.10& 72.96& 72.93& 71.20& 68.90& 67.64& 66.68& 67.38& 64.87& 63.20& 62.23& 60.37& 59.42& 59.67& 58.07& 57.41& 56.10 \\
				AANets & 96.80& 85.00& 76.93& 74.20& 65.76& 62.00& 59.66& 56.20& 55.60& 54.52& 54.04& 52.33& 49.94& 48.23& 46.61& 46.12& 46.00& 44.56& 42.48& 41.64 \\
				DyTox & 86.80& 85.60& 84.40& 80.80& 77.20& 77.67& 76.23& 73.95& 70.84& 68.36& 68.76& 66.40& 63.45& 62.91& 62.99& 61.48& 58.33& 56.56& 54.82& 53.82 \\
				
				\bottomrule
			\end{tabular}
		}
		\label{table:imagenet100b0inc5}
	}
\end{table*}

\begin{table*}[h] 
	{
		\caption{ {Incremental  accuracy comparison of different methods under ImageNet100 Base0 Inc10 setting.} }
		
		\centering	
			\begin{center}
				
				\begin{tabular}{l l l l l l l l l l l}
					\toprule
					\multicolumn{1}{c}{\multirow{2}{*}{Method}} & \multicolumn{10}{c}{Accuracy in each session (\%) $\uparrow$}  \\ \cmidrule{2-11}
					\multicolumn{1}{c}{}    & 1      & 2      & 3    & 4     & 5  & 6     & 7      & 8     &   9  & 10      \\ \midrule
					Finetune & 85.80& 43.80& 30.47& 23.25& 18.32& 15.57& 13.43& 11.80& 10.13& 9.30 \\
					EWC & 85.80& 45.00& 34.47& 25.20& 19.88& 16.63& 15.71& 12.92& 11.09& 11.10 \\
					LwF & 85.80& 72.90& 68.13& 62.15& 57.12& 48.87& 45.86& 43.02& 36.49& 33.10 \\
					Replay & 85.80& 78.50& 68.87& 63.50& 60.44& 54.10& 48.66& 47.00& 44.02& 41.00 \\
					RMM & 85.80& 81.20& 78.93& 76.95& 74.16& 71.90& 70.03& 69.25& 66.84& 65.66 \\
					iCaRL & 85.80& 82.10& 76.00& 71.95& 67.64& 63.77& 60.03& 56.55& 53.64& 50.98 \\
					PODNet & 91.00& 83.40& 74.80& 68.25& 63.28& 58.83& 55.14& 52.50& 47.69& 45.40 \\
					Coil & 85.80& 81.00& 73.73& 65.55& 62.72& 59.07& 53.00& 49.22& 45.13& 41.50 \\
					WA & 85.80& 82.00& 75.93& 72.95& 69.44& 65.97& 61.63& 59.65& 57.36& 55.04 \\
					BiC & 87.00& 80.50& 74.93& 72.90& 69.20& 64.17& 59.51& 53.52& 47.16& 42.40 \\
					FOSTER & 85.80& 79.80& 74.27& 70.05& 68.64& 65.97& 63.69& 63.12& 61.67& 60.58 \\
					DER & 88.40& 84.90& 80.87& 80.10& 78.32& 76.07& 73.09& 72.25& 69.91& 66.84 \\
					MEMO & 85.80& 79.60& 75.93& 73.35& 70.52& 69.77& 65.66& 65.05& 63.40& 60.96 \\
					AANets & 86.40& 78.80& 72.60& 66.35& 62.84& 59.23& 53.66& 52.38& 49.64& 46.60 \\
					DyTox & 87.20& 84.10& 80.60& 77.00& 72.40& 70.23& 68.66& 67.95& 64.11& 61.78 \\
					\bottomrule
				\end{tabular}
			
		\end{center}
		\label{table:Imagenet100b0inc10}
	}
\end{table*}

\begin{table}[h] 
	{
		\caption{ {Incremental accuracy comparison of different methods under ImageNet100 Base50 Inc10 setting.} }
		
		\centering	
			\begin{center}
				
				\begin{tabular}{l l l l l l l l}
					\toprule
					\multicolumn{1}{c}{\multirow{2}{*}{Method}} & \multicolumn{6}{c}{Accuracy in each session (\%) $\uparrow$}  \\ \cmidrule{2-7}
					\multicolumn{1}{c}{}    & 1      & 2      & 3    & 4     & 5  & 6    \\ \midrule
					Finetune & 84.40& 15.63& 13.57& 11.68& 10.20& 9.26 \\
					EWC & 84.40& 15.97& 17.51& 15.08& 12.98& 11.54 \\
					LwF & 84.40& 47.13& 41.60& 38.48& 34.42& 31.42 \\
					Replay & 84.40& 57.20& 51.49& 50.30& 47.58& 43.38 \\
					RMM & 83.40& 79.23& 70.97& 70.20& 67.78& 66.52 \\
					iCaRL & 84.40& 64.57& 59.94& 58.22& 54.89& 53.68 \\
					PODNet & 85.32& 80.83& 75.54& 71.82& 66.56& 62.94 \\
					Coil & 84.40& 69.13& 58.14& 54.00& 49.64& 43.40 \\
					WA & 84.40& 67.70& 63.66& 62.92& 59.56& 56.64 \\
					BiC & 84.60& 77.10& 67.17& 63.55& 55.82& 49.90 \\
					FOSTER & 84.40& 77.53& 66.43& 64.97& 63.62& 63.12 \\
					DER & 84.40& 80.67& 78.34& 76.18& 73.80& 71.10 \\
					MEMO & 84.40& 80.13& 77.71& 75.68& 72.82& 70.22 \\
					AANets & 85.08& 81.23& 76.51& 73.90& 69.78& 66.84 \\
					DyTox & 83.20& 79.13& 76.00& 74.10& 69.71& 65.76 \\
					\bottomrule
				\end{tabular}
		\end{center}
		\label{table:imagenet100b50inc10}
	}
\end{table}

\begin{table}[h] 
	{
		\caption{ {Incremental  accuracy comparison of different methods under ImageNet100 Base50 Inc25 setting.} }
		
		\centering	
			\begin{center}
				
				\begin{tabular}{l l l l l}
					\toprule
					\multicolumn{1}{c}{\multirow{2}{*}{Method}} & \multicolumn{3}{c}{Accuracy in each session (\%) $\uparrow$} \\ \cmidrule{2-4}
					\multicolumn{1}{c}{}    & 1      & 2      & 3     \\ \midrule
					Finetune & 84.40& 29.97& 22.14 \\
					EWC & 84.40& 30.88& 23.90 \\
					LwF & 84.40& 62.48& 51.36 \\
					Replay & 84.40& 56.85& 50.92 \\
					RMM & 83.40& 78.77& 76.38 \\
					iCaRL & 84.40& 69.68& 63.72 \\
					PODNet & 85.32& 77.65& 71.14 \\
					Coil & 84.40& 68.03& 58.64 \\
					WA & 84.40& 73.17& 68.88 \\
					BiC & 85.00& 77.20& 71.80 \\
					FOSTER & 84.40& 74.88& 68.50 \\
					DER & 84.40& 79.25& 74.88 \\
					MEMO & 84.40& 77.25& 72.96 \\
					AANets & 84.64& 76.80& 71.32 \\
					DyTox & 83.96& 77.07& 71.18 \\
					\bottomrule
				\end{tabular}
		\end{center}
		\label{table:imagenet100b50inc25}
	}
\end{table}

\begin{table*}[t]
	\caption{ Forgetting ($\mathcal{F}$) and intransigence ($\mathcal{I}$) measure on CIFAR100. 
	}
	\label{tab:forgetting-intransigence}
	\centering
			\begin{tabular}{@{}lcccccccccccc}
				\toprule
				\multicolumn{1}{c}{\multirow{2}{*}{Method}} & 
				\multicolumn{2}{c}{Base0 Inc5} & \multicolumn{2}{c}{Base0 Inc10} &
				\multicolumn{2}{c}{Base50 Inc10} & 
				\multicolumn{2}{c}{Base50 Inc25} &  \\
				& {$\mathcal{F}$} & $\mathcal{I}$ &  {$\mathcal{F}$} & $\mathcal{I}$ &  {$\mathcal{F}$} & $\mathcal{I}$  &  {$\mathcal{F}$} & $\mathcal{I}$ 
				\\
				\midrule
				Finetune&0.9114&\bf -0.1380&0.8834&\bf -0.1107&0.8653&\bf -0.1333&0.7921&\bf -0.0662&\\
				EWC&0.7483&0.0094&0.7287&-0.0049&0.7998&-0.1181&0.7038&-0.0452&\\
				LwF&0.6998&-0.0147&0.5978&0.0048&0.4443&-0.0280&0.4349&-0.0038&\\
				GEM&0.4892&0.1166&0.4990&0.0959&0.4992&0.0844&0.4407&0.0387&\\
				Replay&0.5174&-0.1023&0.5040&-0.0884&0.4645&-0.0995&0.4564&-0.0478&\\
				RMM&\bf 0.1719&0.1018&\bf 0.1459&0.0776&0.0981&0.0495&\bf 0.0673&0.0482&\\
				iCaRL&0.3539&-0.0113&0.3348&-0.0212&0.2243&-0.0031&0.1907&0.0152&\\
				PODNet&0.4271&0.0905&0.3400&0.1015&0.1218&0.1156&0.1284&0.0709&\\
				Coil&0.5482&-0.0880&0.4748&-0.0505&0.3880&-0.0143&0.3534&-0.0485&\\
				WA&0.2791&0.0264&0.2790&0.0012&0.1854&0.0230&0.1197&0.0422&\\
				BiC&0.1823&0.2388&0.1748&0.1401 &\bf 0.0980&0.4973&0.1006&0.0801&\\
				FOSTER&0.3325&-0.0340&0.2879&-0.0159&0.1478&-0.0022&0.1085&0.0225&\\
				DER&0.1894&0.0567&0.1840&0.0238&0.1098&0.0125&0.1065&0.0078&\\
				MEMO&0.2627&-0.0158&0.2081&0.0031&0.1226&-0.0201&0.0804&0.0180&\\
				AANets&0.3257&0.0425&0.2904&0.0586&0.1848&0.0507&0.1494&0.0405&\\
				\bottomrule
			\end{tabular}
	\end{table*}

	\begin{table*}[t]
		\caption{ Forgetting ($\mathcal{F}$) and intransigence ($\mathcal{I}$) measure on ImageNet100. 
		}\label{tab:forgetting-intransigence-img}
		\centering
			\begin{tabular}{@{}lcccccccccccc}
				\toprule
				\multicolumn{1}{c}{\multirow{2}{*}{Method}} & 
				\multicolumn{2}{c}{Base0 Inc5} & \multicolumn{2}{c}{Base0 Inc10} &
				\multicolumn{2}{c}{Base50 Inc10} & 
				\multicolumn{2}{c}{Base50 Inc25} &  \\
				& {$\mathcal{F}$} & $\mathcal{I}$ &  {$\mathcal{F}$} & $\mathcal{I}$ &  {$\mathcal{F}$} & $\mathcal{I}$  &  {$\mathcal{F}$} & $\mathcal{I}$ 
				\\
				\midrule
				Finetune&0.9404&\bf -0.0744&0.9138&\bf -0.0562&0.9156&\bf -0.0679&0.8716&-0.0333&\\
				EWC&0.8971&-0.0476&0.8771&-0.0412&0.8656&-0.0643&0.8506&\bf -0.0428&\\
				LwF&0.6524&0.0688&0.5513&0.0320&0.4392&-0.0069&0.3804&-0.0195&\\
				Replay&0.5800&-0.0582&0.5489&-0.0448&0.4989&-0.0656&0.4664&-0.0323&\\
				RMM&0.2535&0.0536&0.1807&0.0400&0.1457&0.0017&0.0678&0.0175&\\
				iCaRL&0.4589&-0.0110&0.4024&-0.0128&0.2967&-0.0080&0.2484&-0.0095&\\
				PODNet&0.5131&0.0452&0.4024&0.0430&0.2513&0.0115&0.1472&0.0319&\\
				Coil&0.5956&-0.0398&0.5318&-0.0344&0.4588&-0.0528&0.3444&-0.0348&\\
				WA&0.4139&0.0122&0.3169&0.0236&0.2430&0.0317&0.1722&0.0101&\\
				BiC&0.1400&0.4522&0.1589&0.3590&-0.0454&0.5754&0.1358&0.0204&\\
				FOSTER&0.3733&-0.0204&0.2971&-0.0140&0.1605&-0.0139&0.1644&-0.0041&\\
				DER&\bf 0.0299&0.2010&\bf 0.0633&0.1338&\bf -0.0894&0.2112&\bf 0.0320&0.0692&\\
				MEMO&0.2785&0.0404&0.2431&0.0308&0.0658&0.0527&0.0948&0.0308&\\
				AANets&0.3817&0.0870&0.3611&0.0682&0.1278&0.0889&0.1122&0.0592&\\
				\bottomrule
			\end{tabular}
	\end{table*}

	\begin{figure*}[t]
		\begin{center}
			\subfigure[CIFAR100 B0 Inc10 ]
			{	\includegraphics[width=.9\columnwidth]{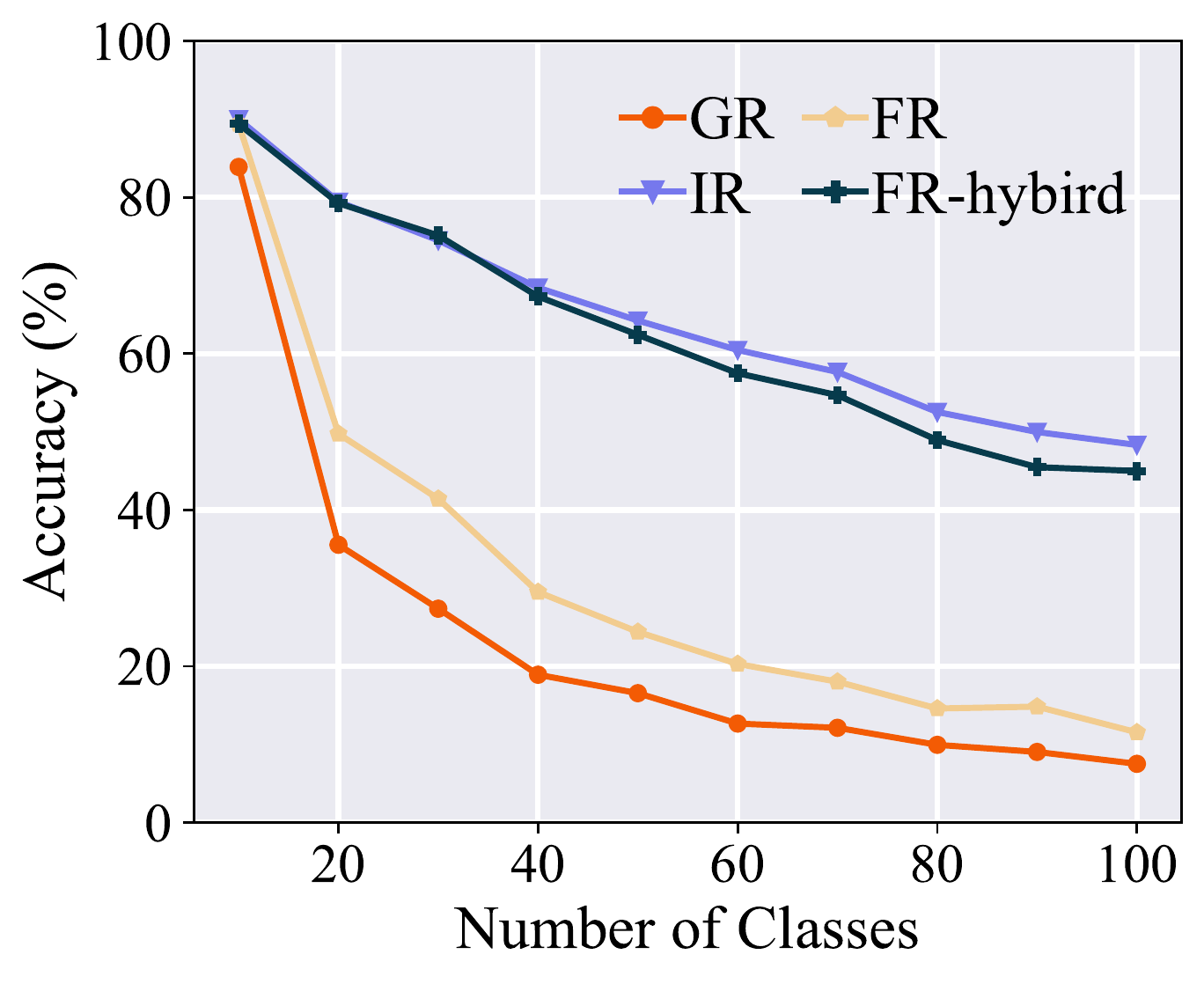}
			}
			\hfill
			\subfigure[CIFAR100 B50 Inc10]
			{	\includegraphics[width=.9\columnwidth]{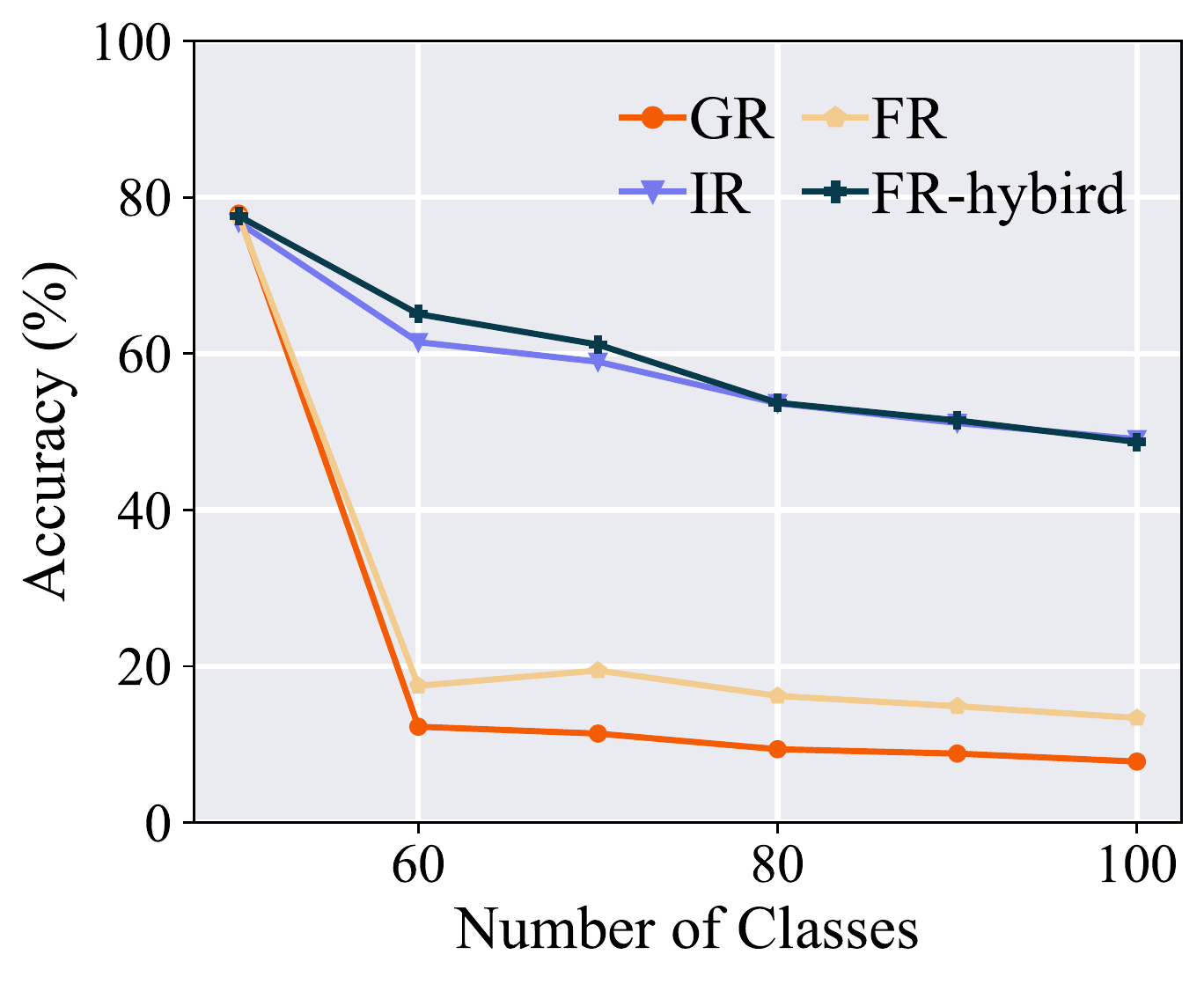}		
			}
		\end{center}
		\caption{	Incremental performance of different replay strategies. We align the total memory budget for fair comparison.
		} \label{figure:replay-comparison}
	\end{figure*}

\section{Forgetting and Intransigence Measure} \label{sec:supp_forgetting}

In the main paper, we mainly utilize accuracy as the performance measure. Moreover, \cite{chaudhry2018riemannian} proposes two metrics for evaluating continual learners, \ie, forgetting ($\mathcal{F}$) and intransigence ($\mathcal{I}$). Specifically, we use $a_{k,j}$ to represent model's performance on task $j$'s testing set after learning task $k$. Hence, the forgetting of a single task is defined as:
\begin{equation}
	f_{j,k}=\max _{l \in\{1, \cdots, k-1\}} a_{l, j}-a_{k, j}, \quad \forall j<k \,.
\end{equation}
$f_{j,k}$ denotes the gap between  the best performance ($ a_{l, j}$) and the final performance ($a_{k, j}$) on the $j$-th task.
Moreover, we can measure the gap between all tasks via the average forgetting of all tasks at the last stage:
\begin{equation} \label{eq:forgetting}
	\mathcal{F}= \frac{1}{B-1} \sum_{j=1}^{B-1} f_{j,B} \,,
\end{equation}
where $B$ is the number of all tasks in the learning process. Note that the last task will not suffer forgetting, and there are only $B-1$ terms in Eq.~\ref{eq:forgetting}.
Since the model often forgets former tasks during the learning process, the forgetting is usually larger than $0$. 

Apart from forgetting, we also have the intransigence measure to reflect the inability of a model to learn new tasks. The intransigence of task $k$ is defined as:
\begin{equation}
	I_k=a_k^*-a_{k, k} \,,
\end{equation}
where $a_k^*$ represents the performance on the $k$-th task of a joint training model. `Joint training' denotes a randomly initialized model optimized with all seen tasks ($\D^1\cup \cdots \D^k$). Hence, $I_k$ measures the gap between the best performance it can achieve ($a_k^*$) and the incrementally learned performance ($a_{k, k}$). In~\cite{chaudhry2018riemannian}, the intransigence is originally defined on a single task, and we follow the definition of forgetting and average it on all tasks:
\begin{equation} \label{eq:intransigence}
	\mathcal{I}= \frac{1}{B-1} \sum_{j=2}^{B} I_j \,.
\end{equation}
Note that there is no gap between joint and incremental training for the first task, and Eq.~\ref{eq:intransigence} averages $B-1$ following tasks. 

As defined in Eq.~\ref{eq:forgetting} and~\ref{eq:intransigence}, we  report these new performance measures during the learning process. Specifically, we calculate these performance measures for settings in Figure 6 in the main paper, and report the results in Table~\ref{tab:forgetting-intransigence}, \ref{tab:forgetting-intransigence-img}. We can summarize two conclusions from the table:

\begin{itemize}
	\item Generally, we find methods with higher last performance $\bar{\mathcal{A}}$ show better performance on the forgetting measure, \ie, lower $\mathcal{F}$. The main reason is that the last performance accounts for the second term in the forgetting calculation, indicating the inherent relationship between these metrics.
	\item 
	Additionally, we find methods with less regularization tend to perform better on
	the intransigence measure, \ie, finetune shows the best performance. Since intransigence represents the gap between the algorithm's performance and joint training's performance on new tasks, regularization terms will harm the model's plasticity and result in poor intransigence measure.
\end{itemize}

\section{Experiment on Replay Strategies} \label{sec:supp_replay}

		There are several kinds of data replay to help the model recover former knowledge, \ie, image~\cite{ratcliff1990connectionist,rebuffi2017icarl}, feature~\cite{iscen2020memory,petit2022fetril} and generative replay~\cite{shin2017continual,he2018exemplar}. Facing a new task $\D^b$, all of them utilize the concatenation of exemplars and the newest training set (\ie, $\mathcal{E}\cup \D^b$) to update the model. However, the difference lies in the configuration of exemplars. Specifically, image replay directly saves raw images to construct $\mathcal{E}$, \ie, $\mathcal{E}=\left\{\left(\x_{j}, y_{j}\right)\right\}_{j=1}^{M}$, $y_j \in \mathcal{Y}_{b-1}$. Using raw images enables the model to quickly recover former knowledge while having two main drawbacks --- raw images consume large memory budget and saving them may violate privacy issues. To tackle this problem, feature replay is proposed by saving a set of features instead of raw images, \ie, $\mathcal{E}=\left\{\left(\phi(\x_{j}), y_{j}\right)\right\}_{j=1}^{M}$, $y_j \in \mathcal{Y}_{b-1}$. Comparing to raw images with many pixels, features $\phi(\x)$ are encoded in the embedding space with fewer dimensions, which helps to reduce the cost for memory budget. For example, saving a raw image of CIFAR costs $3\times 32\times 32$ integer numbers (int), while saving a feature by ResNet32 only costs $64$ float numbers. Hence, the budget of saving a feature is equal to saving $64$ floats $\times 4$ bytes/float $\div (3\times 32\times 32)$ bytes/image $=\frac{1}{12}$ image of CIFAR.
		Additionally, since features are irreversible representations of raw images, privacy issues can be alleviated by saving these features. However, the fatal problem of feature replay lies in the format of $\phi(\x)$. Since the embedding function is changing throughout the learning process, the features extracted at former stages will not be compatible with latter stages. Hence, it requires further alignment or mapping stage to transform features into the same embedding space. Finally, there is also a typical line of work on generative replay, \ie, utilizing generative models to model the distribution of former classes and generate exemplars when needed. A typical line of work using GAN~\citep{goodfellow2014generative} to memorize the distribution of former tasks and then replay them when learning new tasks~\citep{he2018exemplar,shin2017continual}. Specifically, GR~\citep{shin2017continual} considers saving an extra GAN model as the generator and incrementally updates it when new data arrives. The classification model is optimized jointly with $\D^{b} \cup \mathcal{E}$, where $\mathcal{E}$ stands for the generated dataset from the former distribution. However, incrementally updating a single GAN model will also incur catastrophic forgetting. It also requires a larger budget to save the GAN model than saving exemplars.

		\begin{table}[t]
			\caption{ Implementation details for different kinds of replay.
				`\# I' stands for the number of images saved in exemplar set, and `\# F' stands for the number of features.
				`MS' stands for the total memory size (including exemplars and models).	
			}\label{tab:replay-comparison}
			\centering
				\begin{tabular}{@{\;}l@{\;}|@{\;}c@{\;}c@{\;}c@{\;}
						c@{\;} c@{\;}}
					\addlinespace
					\toprule
					Method  & \# I & \# F & Model Type & \# Parameters & MS\\
					\midrule
					GR   & 0  & 0 & ResNet32 + WGAN& 3.35 M & 12.78 MB \\
					IR &3759 & 0 & ResNet32 & 0.46 M &12.78 MB  \\
					FR & 0 & 45108 & ResNet32 & 0.46 M &12.78 MB \\
					FR-hybrid &2000  & 21108& ResNet32 & 0.46 M & 12.78 MB \\
					\bottomrule
				\end{tabular}
			\end{table}
			
		We supply an empirical study on these three kinds of replay strategies, and choose image replay~\cite{ratcliff1990connectionist} (IR), feature replay~\cite{iscen2020memory} (FR) and generative replay~\cite{shin2017continual} (GR) for comparison on CIFAR100 B0 Inc10 and CIFAR100 B50 Inc10 settings.
		We reimplement~\cite{iscen2020memory} and~\cite{shin2017continual} following the main paper.
		We use WGAN~\citep{arjovsky2017wasserstein} as the generative model and implement it with four transposed convolutional layers.
		The optimization details (rounds, learning rate, optimizer) are set according to the original paper. 
		Specifically, training an extra GAN model requires much more parameters than saving exemplars/features, and we follow the comparison protocol in the main paper to align the memory cost of all methods. The comparison details are shown in Table~\ref{tab:replay-comparison}. Note that \cite{iscen2020memory} also includes a variation namely FR-hybird that saves both images and features in the memory, and we also implement it by dividing the total budgets into two parts. We report the performance comparison in Figure~\ref{figure:replay-comparison}.
		
		As we can infer from these figures, there are three main conclusions.
		\begin{itemize}
			\item  Utilizing generative replay fails in this setting, indicating the generative model also suffers catastrophic forgetting.
			\item Utilizing feature replay alone cannot achieve competitive performance. However, using the hybrid replay memory can achieve competitive performance than image replay.
			\item  When the memory size is limited, using image replay shows the best performance and do not require other tuning techniques. Hence, when there is no restriction on privacy issues, using image replay is a simple yet effective solution.
		\end{itemize}

\section{Introduction about compared methods}\label{sec:compared_methods}

In this section, we give the detailed introduction about selected compared methods in the main paper.
In the comparison, we aim to contain all kinds of methods in our taxonomy. Hence, we systematically choose 17 methods, including: Replay~\cite{ratcliff1990connectionist}, RMM~\cite{liu2021rmm} ({\it data replay}), GEM~\cite{lopez2017gradient} ({\it data regularization}), EWC~\cite{kirkpatrick2017overcoming} ({\it parameter regularization}), AANets~\cite{liu2021adaptive}, FOSTER~\cite{wang2022foster}, MEMO~\cite{zhou2022model}, DER~\cite{yan2021dynamically}, DyTox~\cite{douillard2022dytox}, L2P~\cite{wang2022learning} ({\it dynamic networks}), 
LwF~\cite{li2016learning}, iCaRL~\cite{rebuffi2017icarl}, PODNET~\cite{douillard2020podnet}, Coil~\cite{zhou2021co} ({\it knowledge distillation}),
WA~\cite{zhao2020maintaining}, BiC~\cite{wu2019large} ({\it model rectify}).
The detailed introduction is as follows.

 \begin{itemize}
	 	\item {\bf Finetune} is a typical baseline in CIL, which only utilizes the cross-entropy loss in new tasks to update the model while ignoring former tasks. It suffers severe forgetting of former tasks.
	 	\item {\bf Replay~\cite{ratcliff1990connectionist}} ({\em data replay}) is the typical baseline of data replay, which concatenates the exemplar set with the current dataset to update the model.
	 	\item {\bf RMM~\cite{liu2021rmm}} ({\em data replay}) is a recent state-of-the-art algorithm in data replay, which meta-learns an exemplar set organization policy and utilizes it for future tasks.
	 	\item {\bf GEM~\cite{lopez2017gradient}} ({\em data regularization}) is a representative data regularization-based method, which utilizes exemplars as indicators during updating. 
	 	\item {\bf EWC~\cite{kirkpatrick2017overcoming}} ({\em parameter regularization}) is a representative parameter regularization-based method, which measures the parameter importance via Fisher information matrix and prevents important parameters from  drifting away.
	 	\item {\bf  AANets~\cite{liu2021adaptive}} ({\em dynamic networks})  is a representative dynamic network-based method, which builds a dual-branch network to capture task-specific information and adaptively aggregates them to balance stability and plasticity.
	 	\item {\bf DER~\cite{yan2021dynamically}} ({\em dynamic networks}) is a representative dynamic network-based method, which expands an individual network for a new task and aggregates all backbones for feature representation.
	 	\item {\bf FOSTER~\cite{wang2022foster}} ({\em dynamic networks})  is a state-of-the-art dynamic network-based method, which compresses the feature representation of multiple backbones into a single one to control the total budget. 
	 	\item {\bf MEMO~\cite{zhou2022model}} ({\em dynamic networks}) is a state-of-the-art dynamic network-based method, which decouples the network structure into specialized and generalized blocks and only expands specialized blocks based on the shared generalized blocks.
	 	\item {\bf DyTox~\cite{douillard2022dytox}} ({\em dynamic networks})  is a state-of-the-art dynamic network-based method for ViT. Instead of expanding backbones, it proposes to expand a task-specific token to enhance the model's representation ability.
	 	\item {\bf L2P~\cite{wang2022learning}} ({\em dynamic networks})  is a state-of-the-art dynamic network-based method with pre-trained ViT. It learns a prompt pool to encode task-specific information. During inference, it utilizes a key-value matching strategy to select instance-specific prompts for encoding.
	 	\item {\bf LwF~\cite{li2016learning}} ({\em knowledge distillation}) is a representative knowledge distillation-based method, which is the first to introduce knowledge distillation into CIL.
	 	\item {\bf iCaRL~\cite{rebuffi2017icarl}} ({\em knowledge distillation \& template-based classification}) is a popular method in CIL, which extends LwF with exemplar replay and template-based classification (NCM).
	 	\item {\bf PODNET~\cite{douillard2020podnet}} ({\em knowledge distillation})  is a state-of-the-art knowledge distillation-based method, which distills diverse pooled features among a set of models to resist forgetting.
	 	\item {\bf Coil~\cite{zhou2021co}} ({\em knowledge distillation}) is a representative knowledge distillation-based method, which addresses the cross-modal distillation to enhance bi-directional knowledge flow.
	 	\item {\bf BiC~\cite{wu2019large}} ({\em model rectify}) is a representative model rectification-based method, which learns an extra bias correction layer to rectify the biased output of fully-connected layers.
	 	\item {\bf WA~\cite{zhao2020maintaining}} ({\em model rectify}) is a representative model rectification-based method, which normalizes the fully-connected layers to reduce its inductive bias to resist forgetting.
	 \end{itemize}

The choice of these methods follows the development timeline of  CIL, which is also the way we introduce these works.  Additionally, it not only contains all seven aspects of CIL algorithms in our taxonomy but also includes early works (\eg, EWC, LwF, iCaRL) and recent state-of-the-art (DER, MEMO, L2P). The choice also gives consideration to CNN-based methods,  ViT-based methods (DyTox), and even pre-trained ViT-based methods (L2P).

\section{Discussions about Classification Criterion and Scopes of This Survey} \label{sec:supp_detection_seg}

In the main paper, we divide current class-incremental learning methods into seven groups based on the `key point' of these methods. 
For example, for the methods discussing how to save and utilize exemplars for better rehearsal, we divide them into  data replay-based methods. Similarly, we denote the methods of designing model expansion techniques to fit the network structure as data evolves as dynamic network-based methods. 
It must be noted that these methods also learn from each other, and there is no strict boundary between them. For example, iCaRL~\cite{rebuffi2017icarl} utilizes  knowledge distillation, template-based classification, and exemplar replay, but we assign it to the knowledge distillation-based methods. In fact, data replay is becoming a standard protocol in the class-incremental learning community, which is widely adopted in most methods.
As shown in main paper Figure 3, dynamic networks and knowledge distillation methods have dominated publications in recent years. However, it does not indicate that traditional methods like data replay are no longer popular. By contrast, most of these methods rely on the exemplar set to get better performance.

On the other hand, in the discussion of this paper, we mainly focus on the development of class-incremental learning in the image classification field. The main reason is that most influential works about CIL address the image classification problem. 
Furthermore, it has shown the potential to apply the techniques in classification into other tasks, \eg, semantic segmentation~\cite{yang2022uncertainty,ohalife,cermelli2022incremental,zhang2022representation,cermelli2020modeling}, object detection~\cite{perez2020incremental,shmelkov2017incremental,joseph2021towards,wang2021wanderlust,dong2021bridging,joseph2020incremental,feng2022overcoming}, 
video understanding~\cite{park2021class,villa2022vclimb}, medical surgery~\cite{vu2021data,xu2021class}, language model~\cite{wu2021pretrained,jang2021towards,qin2021lfpt5,korbak2022reinforcement}, language-vision model~\cite{yan2022generative,srinivasan2022climb}, etc. 
Specifically, object detection and semantic segmentation are two popular vision tasks that may also face the incremental learning scenario. Continually updating a detection/segmentation model will also suffer catastrophic forgetting, making incremental learning ability a core factor for these scenarios. We then discuss the application of former techniques in these two popular settings.

\noindent\textbf{Incremental Object Detection:} Object detection focuses on identifying and locating objects within an image or a video sequence. While typical works focus on training an object detection model with all training sets at once, real-world applications also require continually updating the detection model, \ie, incremental object detection (IOD)~\cite{shmelkov2017incremental}. In IOD, training samples for different object categories are observed in phases, restricting the ability of the trainer to access past data. The final target is to locate and identify the objects belonging to all seen classes, which is similar to CIL. However, since images can contain multiple objects in IOD, only the new categories are annotated in any given training phase, making it more challenging. 

Knowledge distillation and data replay have been found effective in IOD. \cite{shmelkov2017incremental} applies knowledge distillation to the output of Fast R-CNN~\cite{girshick2015fast}. Moreover, the distillation can also be applied to other detectors (\eg, Faster R-CNN~\cite{ren2015faster}, GFL~\cite{li2020generalized}, RetinaNet~\cite{lin2017focal}, DERT~\cite{carion2020end} and  CenterNet~\cite{zhou2019objects}) via intermediate features, which is similar to different distillation levels in CIL.
ERD~\cite{feng2022overcoming} focuses on elastically learning
responses from the classification head and the regression
head. MVCD~\cite{yang2022multi} designs correlation distillation losses from channel-wise, point-wise, and instance-wise views to regularize the learning of the incremental
model. Similarly, \cite{zhou2020lifelong}  applies knowledge distillation on both the region proposal network and the region classification network. Other works also apply knowledge distillation on the region proposal networks~\cite{chen2019new,hao2019end,peng2020faster}. Moreover, exemplars can also help resist forgetting in IOD.
\cite{joseph2021towards} utilizes a set of exemplars for replay after each incremental step.
\cite{liu2020multi} also designs an adaptive exemplar selection strategy for efficient exemplar selection in IOD.
Most of these works are based on conventional detectors, while  CL-DERT~\cite{liu2023continual} finds directly applying knowledge distillation or data replay works poorly on transformer-based detectors (\eg, Deformable DETR~\cite{zhu2020deformable} and UP-DETR~\cite{dai2021up}). It further designs detector knowledge distillation loss, focusing on
the most informative and reliable predictions from old versions of the model. Other recent works also address the specific  settings like few-shot IOD~\cite{perez2020incremental,yin2022sylph,cheng2021meta,li2021class}, open-world IOD~\cite{joseph2021towards} and 3D IOD~\cite{zhao2022static}.

\begin{figure*}[t]
	\begin{center}
		\subfigure[Average accuracy ]
		{	\includegraphics[width=.9\columnwidth]{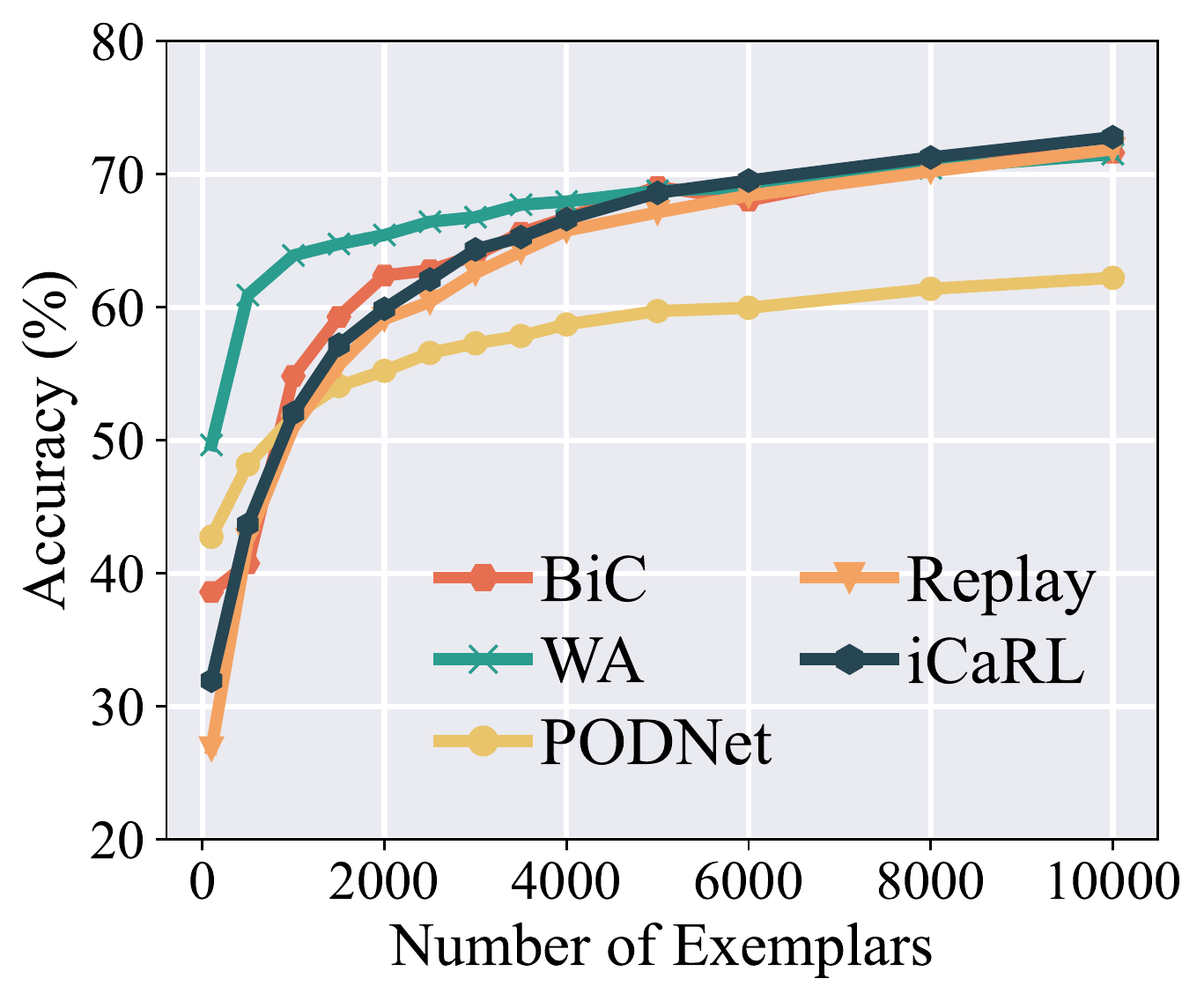}
		}
		\hfill
		\subfigure[Last accuracy]
		{	\includegraphics[width=.9\columnwidth]{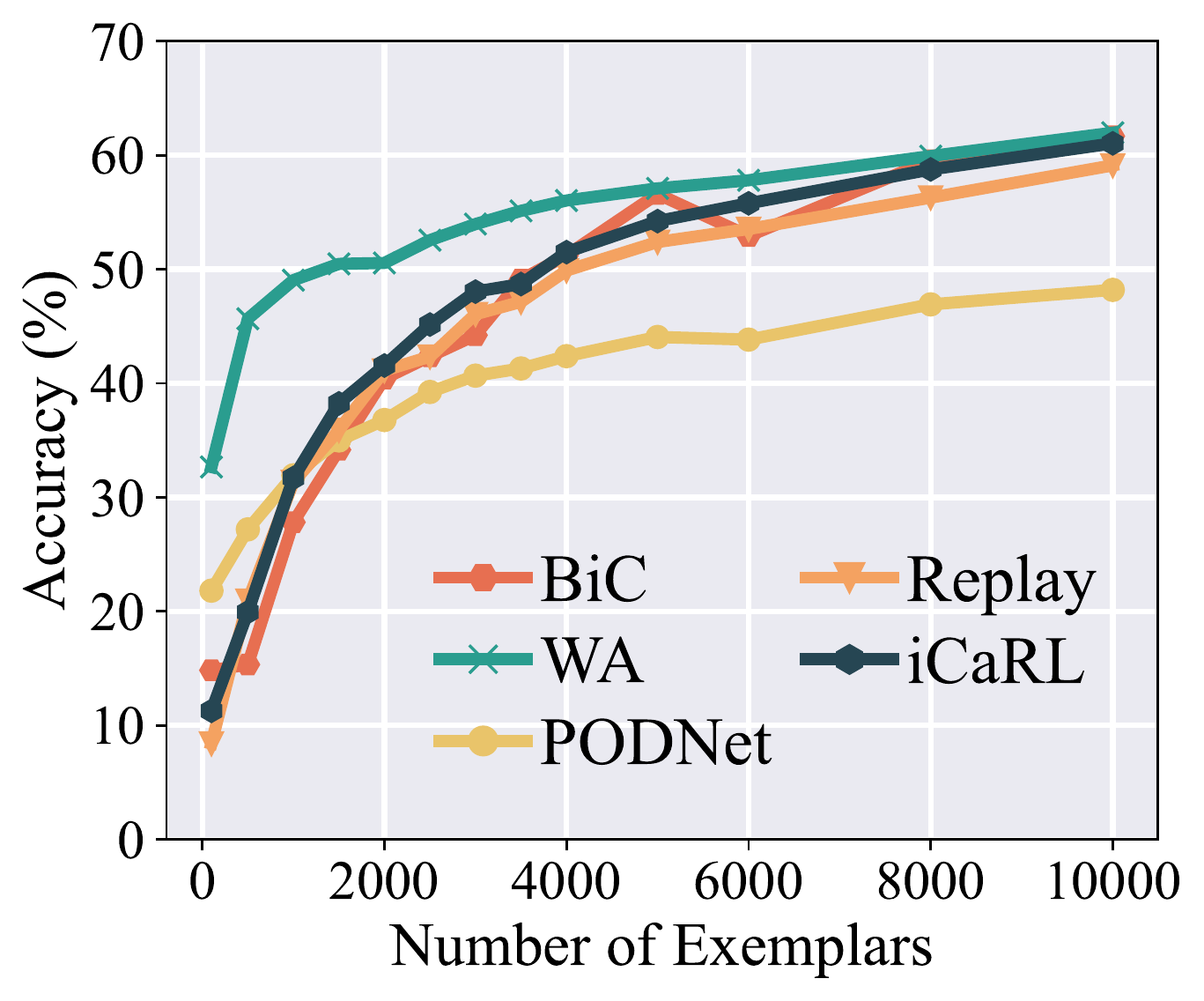}		
		}
	\end{center}
	\caption{	Average and last accuracy of different methods on CIFAR100 Base0 Inc10 with the change of exemplar number.
	} \label{figure:exemplar-change}
\end{figure*}

\begin{figure}[t]
	\centering
	\includegraphics[width=.9\columnwidth]{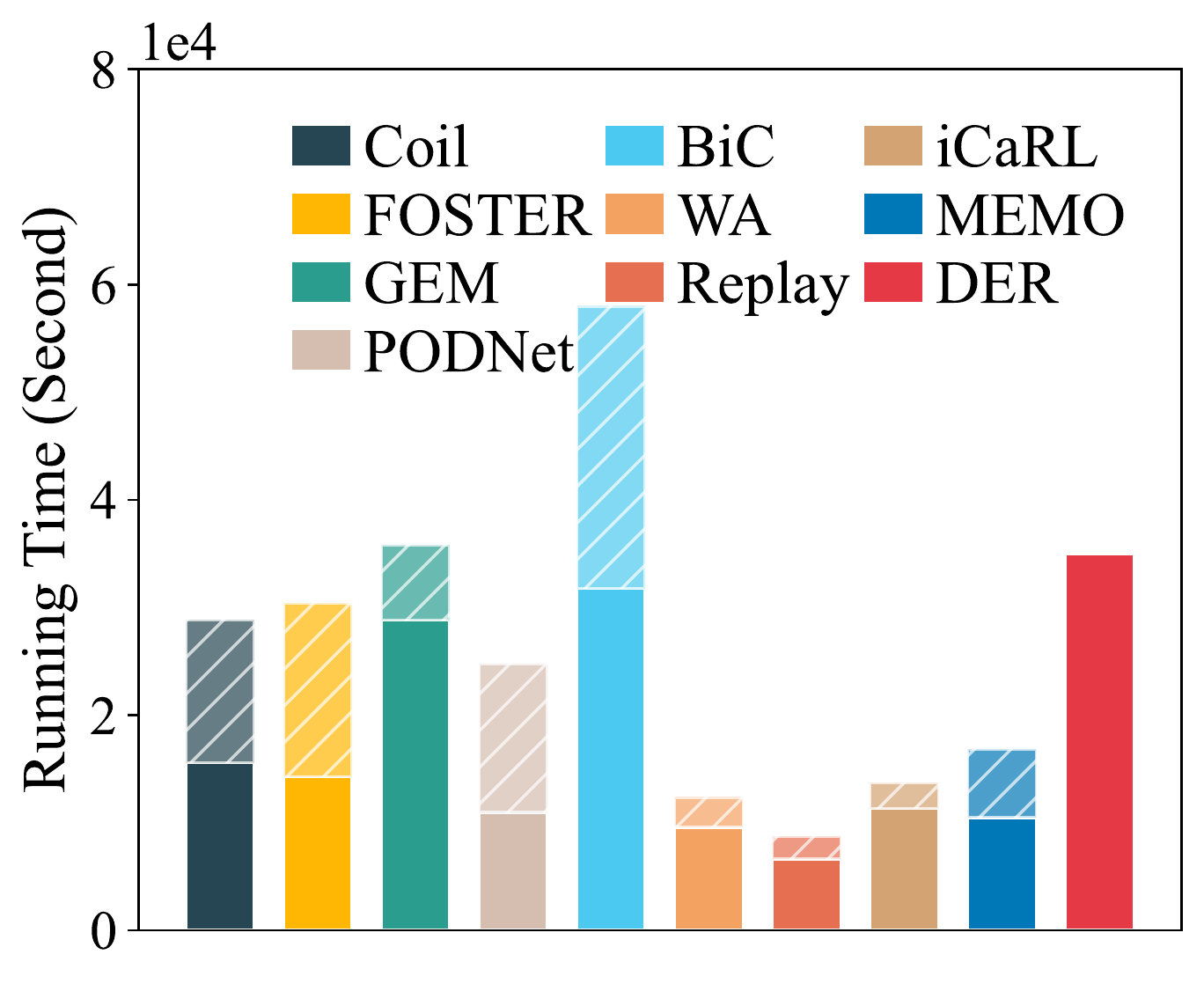} 
	\caption{Running time comparison of different methods on CIFAR100 B0 Inc10. The non-shaded area stands for running time with 2000 exemplars, while the shaded area stands for the extra time with aligned exemplars to DER.}
	\label{fig:running-time}
\end{figure}

\noindent\textbf{Continual Semantic Segmentation:} As another popular topic, semantic segmentation aims to simplify the representation of the image into something more meaningful  by assigning a label to each pixel of an image. When deploying semantic segmentation with incremental new classes, the setting is called continual semantic segmentation (CSS)~\cite{cermelli2020modeling,michieli2019incremental}. Similar to object detection, the application of typical CIL algorithms (\eg, knowledge distillation, data replay, and dynamic networks) can also be applied to CSS. 

MiB~\cite{cermelli2020modeling} points out a core problem in CSS, namely background shift. Since each training step provides annotation only for a subset of all possible classes, pixels of the background class (\ie, pixels that do not belong to any other classes) exhibit a semantic distribution shift. To tackle this problem, it combines the output space distillation with cross-entropy loss to enable model updating without forgetting.
To distill high-level information,
PLOT~\cite{douillard2021plop} proposes a multi-scale pooling distillation scheme that preserves long- and short-range spatial relationships at the feature level. 
SDR~\cite{michieli2021continual} combines knowledge distillation with contrastive loss, enabling the model to distill latent representations.
RCIL~\cite{zhang2022representation} designs a pooled cube knowledge distillation strategy on both spatial and channel dimensions to further enhance the plasticity and stability of the CSS model. 
The second group relies on data replay to revisit former knowledge when learning new. SSUL~\cite{cha2021ssul} and EM~\cite{yan2021framework} explore the application of exemplar replay in CSS.
AMSS~\cite{zhu2023continual} proposes a memory sample selection
mechanism that selects informative samples for effective replay in a fully automatic way.
Besides, works also investigate the application of generative replay~\cite{maracani2021recall} and auxiliary data~\cite{yu2022self,maracani2021recall} for data replay.
Additionally, RCIL~\cite{zhang2022representation} also explores the application of dynamic networks in CSS. It freezes the convolutional layers of the former stage and learns a representation compensation module to fit new tasks. During inference, the old and new modules are merged via structural reparameterization. 
EWF~\cite{xiao2023endpoints} utilizes model fusion among different stage backbones to strike a balance between old and new knowledge.
PFCSS~\cite{lin2023preparing} considers representation learning from the forward compatible perspective~\cite{zhou2022forward} and utilizes contrastive loss to enhance future knowledge.

Recent works in CSS also involve incremental learning in more challenging scenarios, \eg, learning with few-shot annotations~\cite{cermelli2020prototype,shi2022incremental}, different domains~\cite{kalb2023principles,stan2021unsupervised}, continual instance segmentation~\cite{gu2021class,ganea2021incremental} and weakly semantic segmentation~\cite{yu2023foundation}. Moreover, with the prosperity of foundation models, there are also works~\cite{yu2023foundation} transferring knowledge from the complementary foundation models for better performance.

\section{Further Analysis} \label{sec:supp:further}

\subsection{Influence of exemplar numbers
} 
Using exemplar replay to recover former knowledge is a common technique in CIL, whose capability is bounded by the size of the exemplar set. Specifically, if we can save all historical instances as exemplars and replay them during training, we can get an offline model that does not suffer forgetting. Hence, it is common to observe performance improvement with more saved exemplars. In this section, we change the number of exemplars for several typical methods, \eg, Replay, BiC, WA, PODNet, and iCaRL. We experiment on the CIFAR100 B0 Inc10 setting, and the benchmark protocol uses 2000 exemplars in total.  To explore the influence of exemplar number, we change it among $\{$100, 500, 1000, 1500, 2000, 2500, 3000, 3500, 4000, 5000, 6000, 8000, 10000$\}$ to investigate its influence on the final accuracy. We plot the results in Figure~\ref{figure:exemplar-change}. Firstly, with the increase of exemplars, all methods achieve better performance, indicating their benefit in the incremental learning process.  However, we also find the slope of all methods is becoming lower as exemplars increase. It indicates the trade-off between accuracy and memory consumption, \ie, we observe the accuracy increasing rate is almost saturated for exemplars $>4000$. 

\subsection{Running time comparison:}
 A recent work~\cite{prabhu2023computationally} considers another aspect of efficiency in CIL, \ie, computational efficiency. It focuses on realistic scenarios that
cope with high-throughput streams, where computational
bottlenecks impose implicit constraints on learning from past
samples that can be too many to be revisited during training. In this section, we also supply an empirical study on computational efficiency. We consider reporting the running time comparison of different methods, which reveals another aspect of computational efficiency. All the running time is evaluated on a single NVIDIA 3090 GPU. In Figure~\ref{fig:running-time}, we report the running time of DER and other compared methods in our memory-agnostic comparison. Specifically, we report two running times of other methods, one for 2000 exemplars (with the non-shaded area) and the other (with the shaded area) for memory-aligned exemplars to DER. As we can infer from this figure, equipping these methods with more exemplars leads to better performance, while the running time also faces a drastic increase. When the computational budget is strictly bounded, \ie, with limited iterations or updating time, we need to design more computationally-efficient algorithms.

\end{document}